\pdfoutput=1
\documentclass[a4paper,11pt]{book}

\usepackage[T1]{fontenc}
\usepackage[utf8]{inputenc}
\usepackage[authoryear]{natbib}

\usepackage[french,german,english]{babel}

\usepackage{lmodern} 

\usepackage{setspace} 

\usepackage{graphicx,xcolor}
\graphicspath{{figures/}}

\usepackage{booktabs}
\usepackage{lipsum}
\usepackage{microtype}
\usepackage{url}

\usepackage{fancyhdr}

\fancypagestyle{plain}{
	\fancyhf{}

	\fancyfoot[EL,OR]{\thepage}}
\fancypagestyle{addpagenumbersforpdfimports}{
	\fancyhead{}
	
	\fancyfoot{}
	\fancyfoot[RO,LE]{\thepage}
}

\usepackage{listings}
\usepackage{hyperref}
\hypersetup{pdfborder={0 0 0},
	colorlinks=true,
	linkcolor=black,
	citecolor=black,
	urlcolor=black}
\ifpdf
\usepackage[final]{pdfpages}
\else
\usepackage{calc}
\usepackage{breakurl}
\usepackage[nlwarning=false]{hypdvips}
\usepackage{backref}
\renewcommand*{\backref}[1]{}
\fi
\usepackage{bookmark}

\makeatletter
\renewcommand\@pnumwidth{20pt}
\makeatother

\makeatletter
\def\cleardoublepage{\clearpage\if@twoside \ifodd\c@page\else
    \hbox{}
    \thispagestyle{empty}
    \newpage
    \if@twocolumn\hbox{}\newpage\fi\fi\fi}
\makeatother \clearpage{\pagestyle{plain}\cleardoublepage}

\newcommand{\gbar}{\begin{tikzpicture}[
anchor=base, baseline, 
outer sep=0pt, 
]
\node[](A) {$\gamma$};
\draw[shorten >=2pt, shorten <=3pt] ([yshift=-0.5pt]A.west) to ([yshift=-0.5pt]A.east);
\end{tikzpicture}}

\usepackage{color}
\usepackage{tikz}
\usepackage[explicit]{titlesec}
\newcommand*\chapterlabel{}
\titleformat{\chapter}[display]  
	{\normalfont\bfseries\Huge} 
	{\gdef\chapterlabel{\thechapter\ }}     
 	{0pt} 
 	  {\begin{tikzpicture}[remember picture,overlay]
    \node[yshift=-8cm] at (current page.north west)
      {\begin{tikzpicture}[remember picture, overlay]
        \draw[fill=black] (0,0) rectangle(35.5mm,15mm);
        \node[anchor=north east,yshift=-7.2cm,xshift=34mm,minimum height=30mm,inner sep=0mm] at (current page.north west)
        {\parbox[top][30mm][t]{15mm}{\raggedleft \rule{0cm}{0.6cm}\color{white}\chapterlabel}};  
        \node[anchor=north west,yshift=-7.2cm,xshift=37mm,text width=\textwidth,minimum height=30mm,inner sep=0mm] at (current page.north west)
              {\parbox[top][30mm][t]{\textwidth}{\rule{0cm}{0.6cm}\color{black}#1}};
       \end{tikzpicture}
      };
   \end{tikzpicture}
   \gdef\chapterlabel{}
  } 
\titlespacing*{name=\chapter,numberless}{-3.7cm}{83.2pt-\parskip}{-3.2pt+\parskip}
\titlespacing*{\chapter}{-3.7cm}{50pt-\parskip-\parskip}{30pt+\parskip+\parskip}
\titlespacing*{\section}{0pt}{13.2pt}{1em-\parskip}  
\titlespacing*{\subsection}{0pt}{13.2pt}{1em-\parskip}
\titlespacing*{\subsubsection}{0pt}{13.2pt}{1em-\parskip}
\titlespacing*{\paragraph}{0pt}{13.2pt}{1em-\parskip}

\newcounter{myparts}
\newcommand*\partlabel{}
\titleformat{\part}[display]  
	{\normalfont\bfseries\Huge} 
	{\gdef\partlabel{\thepart\ }}     
 	{0pt} 
 	  {\ifpdf\setlength{\unitlength}{20mm}\else\setlength{\unitlength}{0mm}\fi
	  \addtocounter{myparts}{1}
	  \begin{tikzpicture}[remember picture,overlay]
    \node[anchor=north west,xshift=-65mm,yshift=-6.9cm-\value{myparts}*20mm] at (current page.north east) 
      {\begin{tikzpicture}[remember picture, overlay]
        \draw[fill=black] (0,0) rectangle(62mm,20mm);   
        \node[anchor=north west,yshift=-6.1cm-\value{myparts}*\unitlength,xshift=-60.5mm,minimum height=30mm,inner sep=0mm] at (current page.north east)
        {\parbox[top][30mm][t]{55mm}{\raggedright \color{white}Part \partlabel \rule{0cm}{0.6cm}}};  
        \node[anchor=north east,yshift=-6.1cm-\value{myparts}*\unitlength,xshift=-63.5mm,text width=\textwidth,minimum height=30mm,inner sep=0mm] at (current page.north east)
              {\parbox[top][30mm][t]{\textwidth}{\raggedleft \rule{0cm}{0.6cm}\color{black}#1}};
       \end{tikzpicture}
      };
   \end{tikzpicture}
   \gdef\partlabel{}
  } 
\titlespacing*{\part}{11.06cm}{26.4pt-\parskip-\parskip}{0pt}

\usepackage{amsmath}
\usepackage{amsfonts}
\usepackage{amssymb}
\usepackage{mathtools}
\makeatletter
\def\resetMathstrut@{%
  \setbox\z@\hbox{%
    \mathchardef\@tempa\mathcode`\(\relax
      \def\@tempb##1"##2##3{\the\textfont"##3\char"}%
      \expandafter\@tempb\meaning\@tempa \relax
  }%
  \ht\Mathstrutbox@1.2\ht\z@ \dp\Mathstrutbox@1.2\dp\z@
}
\makeatother
\usepackage{amsthm}
\usepackage{amsmath, bm}
\usepackage{amssymb}
\usepackage{mathrsfs}
\usepackage{mathtools}
\usepackage{graphicx}
\usepackage{multicol}
\usepackage{pgfplots}
\usepackage{changepage}
\usepackage[font={small, it}]{caption}
\usepackage{subcaption}
\usepackage{url}
\usepackage{siunitx}
\pgfplotsset{compat=newest}
\usetikzlibrary{plotmarks}
\usetikzlibrary{arrows.meta}
\usetikzlibrary{spy}
\usepgfplotslibrary{patchplots}
\usepackage{grffile}
\usepackage{nccmath}
\usepackage{array}
\usepackage[export]{adjustbox}
\usepackage{multirow}
\usepackage{setspace}
\usepackage{algorithm} 
\usepackage{algpseudocode}  
\usepackage{diagbox}
\usepackage{bbm} 
\usepackage[autostyle]{csquotes}  

\usepackage{xspace}
\usepackage{amsfonts}       
\usepackage{nicefrac}       
\usepackage{microtype}      
\usepackage[acronym]{glossaries}
\usepackage{mathrsfs,oldgerm}
\usepackage{wrapfig}
\usepackage{cleveref}
\usepackage{verbatim}
\usepackage[disable]{todonotes} 
\usepackage[export]{adjustbox}
\usepackage{pifont}
\usepackage{placeins}       
\newcommand{\cmark}{\ding{51}}%
\newcommand{\xmark}{\ding{55}}%

\newcommand{\todoi}[1]{\todo[inline]{#1}}

\usepackage{mdframed}
\usepackage{xcolor}
\theoremstyle{definition}
\newtheorem*{definition}{Definition}

\theoremstyle{definition}
\newtheorem*{remark}{Remark}
\surroundwithmdframed[linewidth = 0 ,%
leftmargin = 20,%
rightmargin = 20,%
backgroundcolor=black!10,%
innertopmargin = 10,%
splittopskip = 0,%
ntheorem]{remark}

\surroundwithmdframed[linewidth = 0 ,%
leftmargin = 20,%
rightmargin = 20,%
backgroundcolor=black!10,%
innertopmargin = 10,%
splittopskip = 0,%
ntheorem]{definition}

\newtheoremstyle{bfnoteonly}%
{}{}%
{}{}%
{\itshape}{.}%
{ }%
{\thmnote{#3}}
\theoremstyle{bfnoteonly}
\newtheorem*{extremark}{}
\surroundwithmdframed[linewidth = 0 ,%
leftmargin = 20,%
rightmargin = 20,%
backgroundcolor=black!10,%
innertopmargin = 10,%
skipabove=20,
ntheorem]{extremark}

\theoremstyle{theorem}
\newtheorem{proposition}{Proposition}
\surroundwithmdframed[linewidth = 0 ,%
leftmargin = 20,%
rightmargin = 20,%
backgroundcolor=black!10,%
innertopmargin = 10,%
ntheorem]{proposition}

\theoremstyle{theorem}
\newtheorem{theorem}{Theorem}
\surroundwithmdframed[linewidth = 0 ,%
leftmargin = 20,%
rightmargin = 20,%
backgroundcolor=black!10,%
innertopmargin = 10,%
ntheorem]{theorem}

\usepgfplotslibrary{fillbetween}
\pgfplotsset{every tick label/.append style={font=\scriptsize}}
\def\shortestskip{\setlength{\abovedisplayskip}{0pt}%
\setlength{\belowdisplayskip}{0pt}%
\setlength{\abovedisplayshortskip}{0pt}%
\setlength{\belowdisplayshortskip}{0pt}}

\shortestskip

\makeatletter
\newcounter{algorithmicH}
\let\oldalgorithmic\algorithmic
\renewcommand{\algorithmic}{%
  \stepcounter{algorithmicH}
  \oldalgorithmic}
\renewcommand{\theHALG@line}{ALG@line.\thealgorithmicH.\arabic{ALG@line}}
\makeatother

\pgfplotsset{compat=newest,
       colormap={parula}{
            rgb255=(53,42,135)
            rgb255=(15,92,221)
            rgb255=(18,125,216)
            rgb255=(7,156,207)
            rgb255=(21,177,180)
            rgb255=(89,189,140)
            rgb255=(165,190,107)
            rgb255=(225,185,82)
            rgb255=(252,206,46)
            rgb255=(249,251,14)
        },
    }

\usepackage{booktabs}
\usepackage{soul}
\makeatletter

\def\*#1{\mathbf{#1}}

\makeatletter
\newenvironment{tsubarray}[1]{%
  \vcenter\bgroup
  \Let@ \restore@math@cr \default@tag
  \baselineskip\fontdimen10 \scriptfont\tw@
  \advance\baselineskip\fontdimen12 \scriptfont\tw@
  \lineskip\thr@@\fontdimen8 \scriptfont\thr@@
  \lineskiplimit\lineskip
  \check@mathfonts
  \ialign\bgroup\ifx c#1\hfil\fi
    \normalfont\fontsize\sf@size\z@\selectfont\ignorespaces##\unskip\hfil\crcr
}{%
  \crcr\egroup\egroup
}
\makeatother

\input cyracc.def

\usepackage{amsmath,amsfonts,bm}

\def\eqref#1{equation~\ref{#1}}
\def\Eqref#1{Equation~\ref{#1}}

\def\1{\bm{1}}

\def\rx{{\textnormal{x}}}

\def\rvx{{\mathbf{x}}}
\def\rvy{{\mathbf{y}}}
\def\rvz{{\mathbf{z}}}

\def\va{{\bm{a}}}

\def\vd{{\bm{d}}}

\def\ve{{\bm{e}}}
\def\vf{{\bm{f}}}
\def\vg{{\bm{g}}}

\def\vm{{\bm{m}}}

\def\vs{{\bm{s}}}

\def\vv{{\bm{v}}}
\def\vw{{\bm{w}}}
\def\vx{{\bm{x}}}
\def\vy{{\bm{y}}}
\def\vz{{\bm{z}}}

\def\mA{{\bm{A}}}
\def \mAo{\mA_\omega}

\def\mD{{\bm{D}}}

\def\mF{{\bm{F}}}

\def\mI{{\bm{I}}}

\def\mP{{\bm{P}}}

\def\mS{{\bm{S}}}

\def\mW{{\bm{W}}}

\def\mZ{{\bm{Z}}}

\DeclareMathAlphabet{\mathsfit}{\encodingdefault}{\sfdefault}{m}{sl}
\SetMathAlphabet{\mathsfit}{bold}{\encodingdefault}{\sfdefault}{bx}{n}

\newcommand{\R}{\mathbb{R}}

\DeclareMathOperator*{\argmax}{arg\,max}
\DeclareMathOperator*{\argmin}{arg\,min}

\def \bepsilon{\boldsymbol{\epsilon}}

\def \R{\mathbb{R}}

\def \x{\vx}
\def \y{\vy}
\def \yo{\vy_\omega}

\def \z{\mathbf{z}}

\def \Po{\mP_\omega}
\def \mAo{\mA_\omega}

\def \px{p(\rvx)}

\def \ft{f_\theta}

\def \xh{\bm{\hat{x}}}

\makeatletter
\newcommand{\bianca}{\renewcommand\NAT@open{[}\renewcommand\NAT@close{]}}
\makeatother

\begin{document}
\setlength{\parindent}{0pt}
\setlength{\parskip}{0pt} 
\frontmatter
\begin{titlepage}
    \begin{otherlanguage}{french}
        \begin{center}
            \sffamily

            \null\vspace{2cm}
            {\huge Learning to sample in Cartesian MRI} \\[24pt]

            \vfill

            \begin{tabular} {cc}
                \parbox{0.3\textwidth}{\includegraphics[width=4cm]{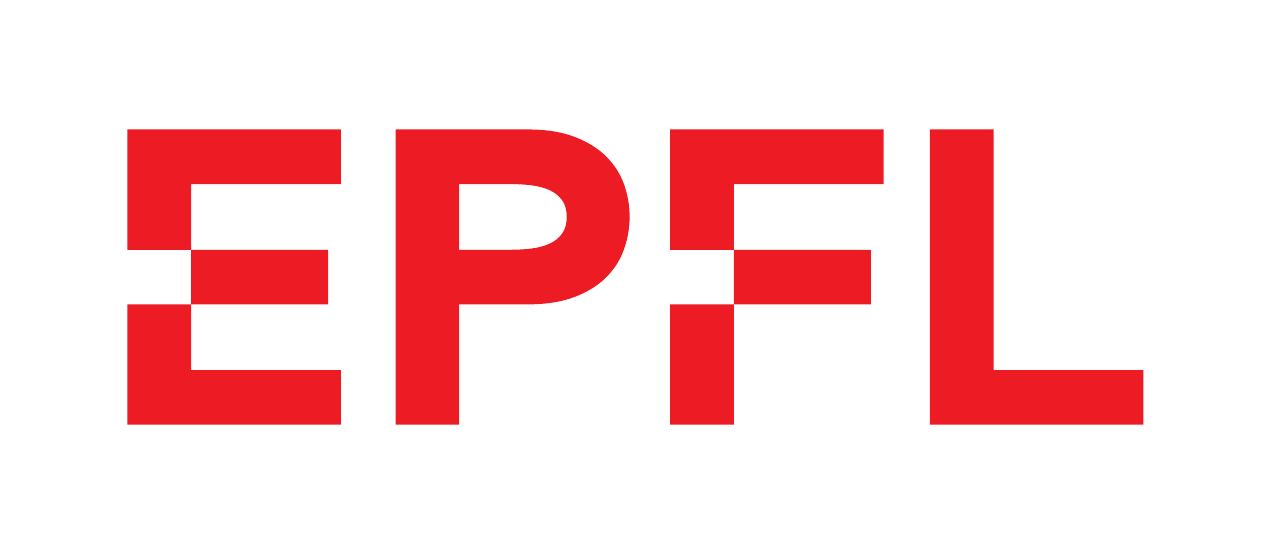}}
                 &
                \parbox{0.7\textwidth}{%
                Thèse n. 9981                                        \\
                présentée le 3 juin 2022                             \\
                Faculté des sciences et techniques de l'ingénieur    \\
                Laboratoire de systèmes d'information et d'inférence \\
                Programme doctoral en informatique et communications \\[6pt]
                pour l'obtention du grade de Docteur ès Sciences     \\
                par                                                  \\ [4pt]
                \null \hspace{3em} Thomas Sanchez                    \\[9pt]
                \small
                Acceptée sur proposition du jury :                   \\[4pt]
                Prof Alexandre Alahi, président du jury              \\
                Prof Volkan Cevher, directeur de thèse               \\
                Prof Dimitri van de Ville, rapporteur                \\
                Dr Ruud van Heeswijk, rapporteur                     \\
                Dr Philippe Ciuciu, rapporteur                       \\[12pt]
                }
            \end{tabular}
        \end{center}
        \vspace{2cm}
    \end{otherlanguage}
\end{titlepage}

\cleardoublepage
\thispagestyle{empty}

\vspace*{5cm}
\begin{raggedleft}
    On God rests my salvation and my glory;\\
    my mighty rock, my refuge is God.\\
    --- The Bible, \textit{Psalm 62:8}\\
\end{raggedleft}

\vspace{4cm}


\setcounter{page}{0}
\chapter*{Acknowledgements}
\markboth{Acknowledgements}{Acknowledgements}
\addcontentsline{toc}{chapter}{Acknowledgements}

\bigskip

This thesis was a journey that would have been impossible without the help and support of many people.
\vspace{0.5cm}

First, I would like to thank my advisor Volkan Cevher. He never gave up on me despite some tumultuous steps, and fought for me to reach the end of my thesis. Few would have done this. I am also thankful that he let me pursue freely my research interests, and I am grateful to him for having built an amazing group with colleagues who are both brilliant researchers and great friends. 
\vspace{0.5cm}

Second, I would to express my gratitude to Prof. Alexandre Alahi, Prof. Dimitri van de Ville, Dr. Ruud van Heeswijk and Dr. Philippe Ciuciu for being part of my thesis committee and for the interest that they took in reading and evaluating my thesis, as well as for their useful suggestions. I would like to thank in particular Dr. Ruud van Heeswijk and Dr. Philippe Ciuciu for all the insight that they gave me about MRI, and Dr. Philippe Ciuciu hosting me at Neurospin, despite the visit being shortened by the start of a pandemic. I have very good memories of my short time there.

\vspace{0.5cm}

My thanks then naturally go to my friends and colleagues, who supported me during these four years and made them unforgettable. I can never repay them enough for their kindness. Thank you Baran, Paul, Ya-Ping, Ahmet, Fatih, Ali K., Kamal, Thomas, Fabian, Leello, Pedro, Igor, Chaehwan, Maria, Arda, Armin, Yura, Nadav, Grigoris, Ali R., Andrej, Luca, Stratis, Kimon, Fanghui, and all the others for your friendship and the moments that we have shared. Of course, a couple of special mentions are in order.
I want to thank Baran for introducing me to the world of MRI, and teaching me much about research and life.
I want to thank Igor for the close collaboration that we had throughout my PhD, for putting up with me through manifold argumentative and extended meetings, pesky bugs and seemingly incomprehensible experimental results, and for being so patient and supportive. I am also thankful for his healthy German honesty and for the good friend he is. 
I want also to thank Leello for being one of the most fun people on this planet -- I don't think that someone made me laugh more than him --  for his optimistic
look on life, for his practical wisdom and overall for being such an amazing friend. Thank you for bringing joy and friendship to everyone around you.
Of course, I ought to thank Gosia, the secretary of our lab, for her amazing work and efficiency in every practical matter. I suspect that she is the reason that the lab has not yet collapsed into chaos.
\vspace{0.5cm}

I am also grateful to all my friends for their support, in particular to Clément and Nathan Dupertuis, Peter Lecomte, Marion Conus and Christian Cuendet, as well as everyone from church, from the house group and from GBU. They are too numerous to be listed individually, and I feel very fortunate to be surrounded by so many kind and generous people.
\vspace{0.5cm}

Of course, without the support from my family, this thesis would not have been possible. I have been blessed by the unconditional love and encouragement from my parents, Esther and David, from my sister Myriam and her fiancé Sam, and from my brother Nicolas. I also cannot thank enough my parents-in-law, Karin and Jean-Paul.

\vspace{0.5cm}
Finally, my words turn to my best friend and amazing wife, Aline. Thank you for who you are. Thank you for your love, your support and unfailing encouragement. Without you, I could not have completed this thesis. You make every day brighter and push me to become a better man.


\vspace{0.5cm}

\noindent\textit{Lausanne, May 12, 2022}
\hfill T.~S.


\cleardoublepage
\chapter*{Abstract}
\markboth{Abstract}{Abstract}
\addcontentsline{toc}{chapter}{Abstract (English/Français)} 

Magnetic Resonance Imaging (MRI) is a non-invasive, non-ionizing imaging modality with unmatched soft tissue contrast. However, compared to imaging methods like X-ray radiography, MRI suffers from long scanning times, due to its inherently sequential acquisition procedure. Shortening scanning times is crucial in clinical setting, as it increases patient comfort, decreases examination costs and improves throughput.\\

Recent developments thanks to compressed sensing (CS) and lately deep learning allow to reconstruct high quality images from undersampled images, and have the potential to greatly accelerate MRI. Many algorithms have been proposed in the context of reconstruction, but comparatively little work has been done to
find acquisition trajectories that optimize the quality of the reconstructed image downstream.\\ 

Although in recent years, this problem has gained attention, it is still unclear what is the best approach to design acquisition trajectories in Cartesian MRI. 
In this thesis, we aim at contributing to this problem along two complementary directions.\\

First, we provide novel \textit{methodological} contributions to this problem. We first propose two algorithms that improve drastically the greedy learning-based compressed sensing (LBCS) approach of \citet{gozcu2018learning}. These two algorithms, called lazy LBCS and stochastic LBCS scale to large, clinically relevant problems such as multi-coil 3D MR and dynamic MRI that were inaccessible to LBCS. 
We also show that generative adversarial networks (GANs), used to model the posterior distribution in inverse problems, provide a natural criterion for adaptive sampling by leveraging their variance in the measurement domain to guide the acquisition procedure.\\


Secondly, we aim at deepening the \textit{understanding} of the kind of structures or assumptions that enable mask design algorithms to perform well in practice. In particular, our
experiments show that state-of-the-art approaches based on deep reinforcement learning (RL), which have the ability to adapt trajectories on the fly to patient and perform long-horizon planning, bring at best a marginal improvement over stochastic LBCS, which is neither adaptive nor does long-term planning.\\ 

Overall, our results suggest that methods like stochastic LBCS offer promising alternatives to deep RL. They shine in particular by their scalability and computational efficiency and could be key in the deployment of optimized acquisition trajectories in Cartesian MRI.\\

\textbf{Keywords:} magnetic resonance imaging, experiment design, inverse problems, compressed sensing, reinforcement learning, deep learning, generative adversarial networks


\begin{otherlanguage}{french}
    \cleardoublepage
    \chapter*{Résumé}
    \markboth{Résumé}{Résumé}
    L'imagerie par résonance magnétique (IRM) est une modalité d'imagerie non invasive, non ionisante et offrant un contraste inégalé des tissus mous. Cependant, en raison de procédures d'acquisition inhéremment séquentielles, la vitesse d’acquisition en IRM est  lente comparé à des méthodes telles que la radiographie par rayons X. La réduction des temps d’acquisition est cruciale en milieu clinique, car elle profite au confort du patient, diminue les coûts d'examen et améliore le rendement.\\

    Les innovations récentes grâce à l’acquisition comprimée (CS) et dernièrement l'apprentissage profond permettent de reconstruire des images de haute qualité à partir d'images sous-échantillonnées et ont le potentiel de grandement accélérer l'IRM. De nombreux algorithmes ont été proposés dans le contexte de la reconstruction, mais comparativement peu de travaux ont été réalisés afin de trouver des trajectoires d'acquisition qui optimisent la qualité de l'image reconstruite en aval.\\

    Bien que ces dernières années, ce problème ait gagné en attention, il n'est toujours pas clair quelle est la meilleure approche pour concevoir des trajectoires d'acquisition en IRM cartésienne. Dans cette thèse, nous visons à contribuer à ce problème selon deux directions complémentaires.\\

    Premièrement, nous proposons de nouvelles contributions \textit{méthodologiques} à ce problème. Nous proposons tout d'abord deux algorithmes qui améliorent considérablement l'approche gloutonne d’acquisition comprimée basée sur l’apprentissage (LBCS) de \citep{gozcu2018learning}. Ces deux algorithmes, appelés \textit{LBCS paresseux} et \textit{LBCS stochastique}, s’étendent à des problèmes importants et cliniquement pertinents, tels que l'IRM parallèle 3D et dynamique, qui étaient inaccessibles à LBCS. Nous montrons également que les réseaux antagonistes génératifs (GAN), utilisés pour modéliser la distribution \textit{a posteriori} dans les problèmes inverses, fournissent un critère naturel pour l’acquisition adaptative en utilisant leur variance dans le domaine de la mesure pour guider la procédure d'acquisition.\\

    Deuxièmement, nous cherchons à approfondir la \textit{compréhension} du type de structures ou d'hypothèses qui permettent aux algorithmes de conception de masques d'être performants en pratique. En particulier, nos expériences montrent que les approches de pointe basées sur l'apprentissage par renforcement (RL) profond, qui ont la capacité d'adapter leurs trajectoires au patient à la volée et d'effectuer une planification à long terme, apportent au mieux une amélioration marginale par rapport à LBCS stochastique, qui n'est ni adaptatif ni ne fait pas de planification à long terme.\\

    Dans l'ensemble, nos résultats suggèrent que des méthodes comme LBCS stochastique offrent des alternatives prometteuses au RL profond. Elles brillent notamment par leur extensibilité et leur efficacité de calcul et pourraient être déterminantes dans le déploiement de trajectoires d'acquisition optimisées en IRM cartésienne.\\

    \textbf{Mots-clé~:} imagerie par résonance magnétique, planification d'expériences, problèmes inverses, acquisition comprimée, apprentissage par renforcement, apprentissage profond, réseaux antagonistes génératifs.

\end{otherlanguage}


\cleardoublepage
\pdfbookmark{\contentsname}{toc}
\tableofcontents



\setlength{\parskip}{1em}

\mainmatter
\cleardoublepage
\chapter{Introduction}
\markboth{Introduction}{Introduction}

Since its inception in the 1970s, Magnetic Resonance Imaging (MRI) has deeply impacted radiology and medicine. It allows for a non-invasive, non-ionizing imaging of the body, and enables among other things to visualize anatomical structures, physiological functions and metabolic information with unmatched precision \citep{wright1997magnetic,feng2016compressed}.  Magnetic Resonance Imaging, as its name indicates,  constructs an image of an object of interest by exploiting magnetic resonance. Namely, when submitted to a strong static magnetic field $B_0$, the protons in the object, mostly in water molecules, will resonate when excited through a radio frequency pulse. In order to encode spatial information in the observed signal instead of a receiving global response, additional time and spatially varying magnetic fields, referred to as gradient fields, are superimposed to the original magnetic field $B_0$.

\begin{figure}[!t]
    \centering
    \includegraphics[width=.75\linewidth]{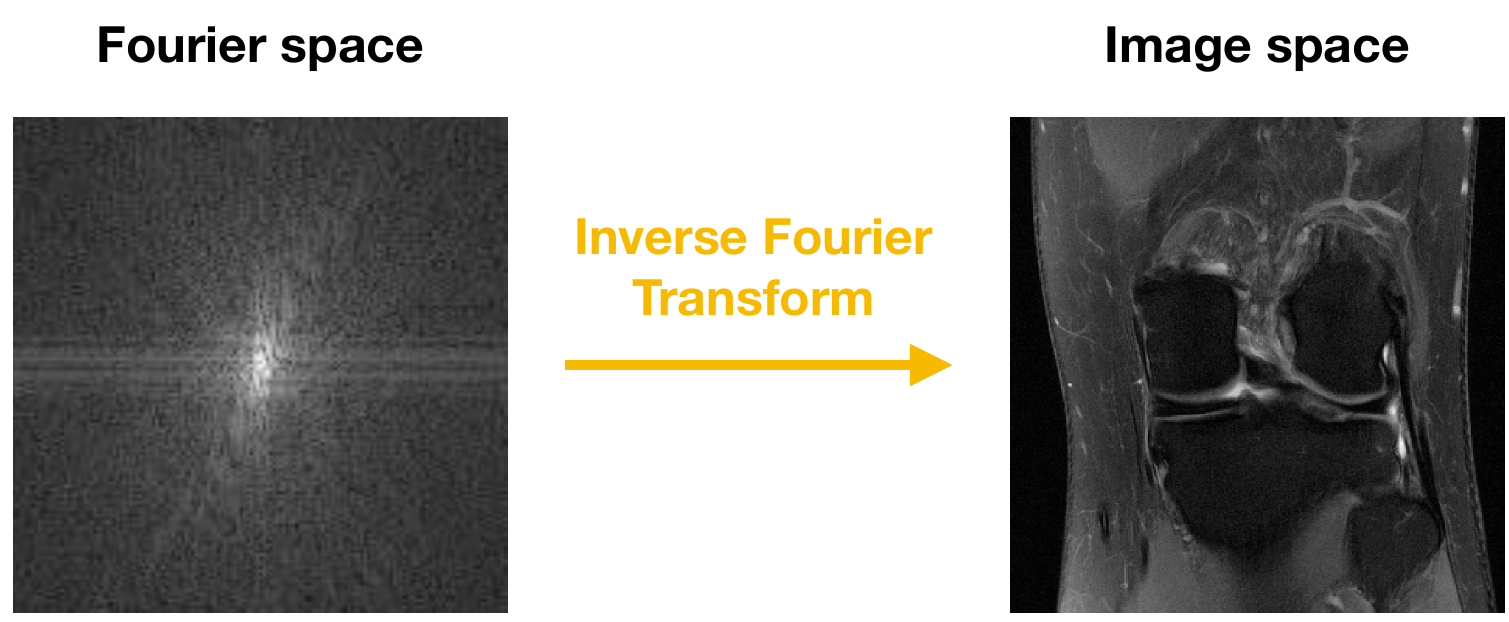}
    \caption{Fourier domain and image domain representation of a knee MRI scan.}\label{fig:fourier_representation}
\end{figure}

However, this process does not give direct access to an image, but rather informs us about the different \textit{frequencies} that construct it. By varying the gradient fields, it is possible to obtain frequency information relating to different locations and represent them in what is known as the Fourier space, or \textit{k-space}. Figure \ref{fig:fourier_representation} shows the difference between an image and its Fourier space representation. The object of interest is excited several times using different varying gradient fields, which defines a \textit{k-space trajectory} along which data are acquired. When sufficient information has been acquired, an image can be readily obtained by computing the inverse Fourier transform of the signal.

There is considerable flexibility in the design of k-space trajectories, but a particularly common approach is to cover the k-space in a grid-like fashion, known as \textit{Cartesian} sampling. A grid-like covering of k-space implies that the image can be directly obtained by applying a fast Fourier transform (FFT). Other approaches, known as non-Cartesian, do not cover k-space in a grid-like fashion, but can consist of radial spokes \citep{lauterbur1973image}, spirals \citep{meyer1992fast} or space-filling curves \citep{lazarus2019sparkling}. They can provide a faster acquisition than Cartesian sampling and benefits such as resistance to motion artifacts, but require a more involved and slower reconstruction through techniques such as gridding \citep{o1985fast,jackson1991selection} or non-uniform fast Fourier transform (NUFFT) \citep{fessler2003nonuniform}, as well as density compensation \citep{pipe1999sampling,pauly2007non}. In addition to this, non-Cartesian trajectories can be sensitive to errors in the magnetic field gradients, due to field imperfections of various sources \citep{vannesjo2013gradient}. As a result, Cartesian trajectories have been the most popular in practice \citep{lustig2008compressed,feng2016compressed}.



\begin{figure}[!ht]
    \centering
    \includegraphics[width=\linewidth]{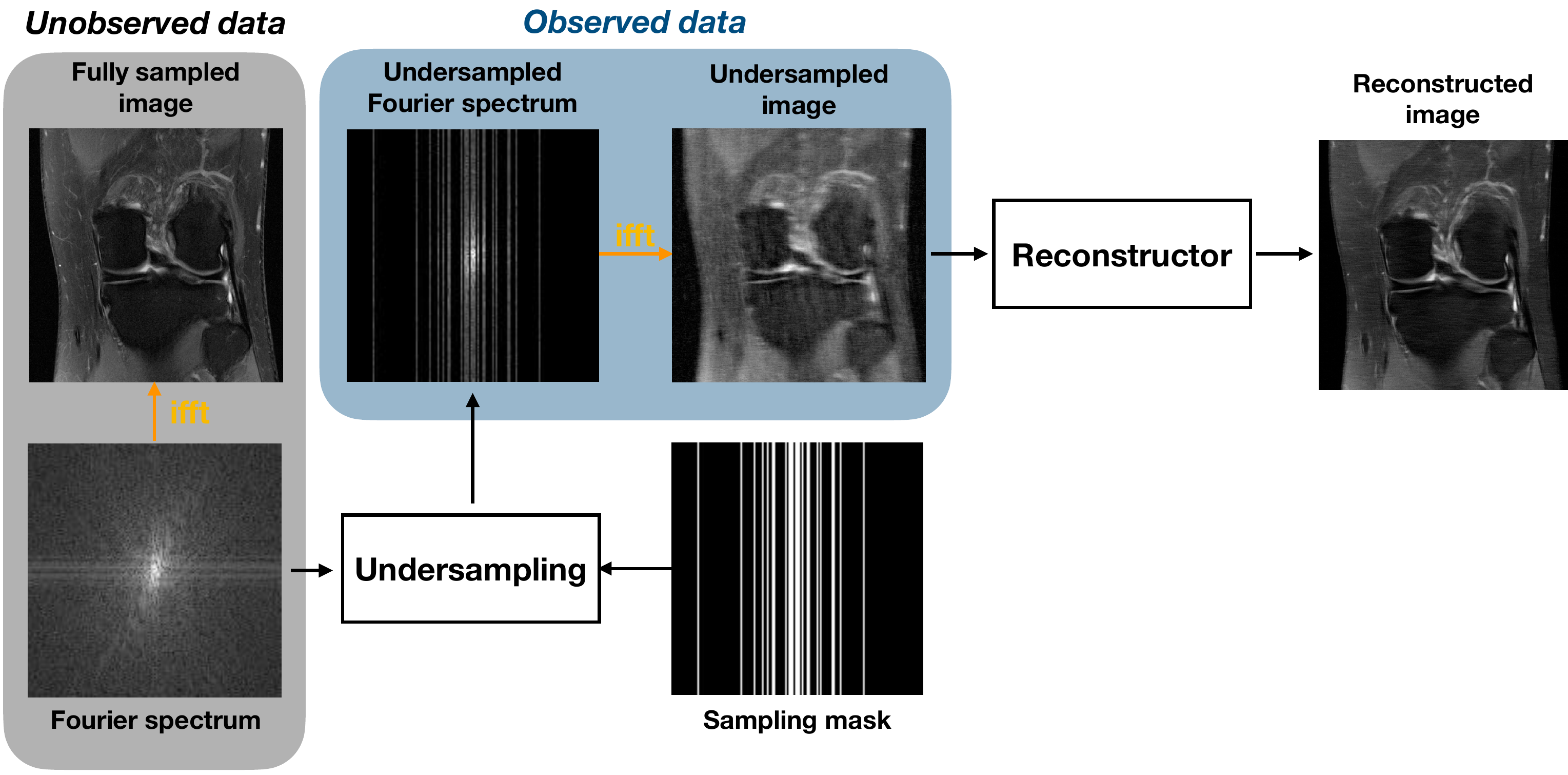}
    \caption{Accelerated MRI by undersampling. Undersampling is governed by a \textit{sampling mask} that tells what locations are acquired. The undersampled Fourier spectrum is then mapped back to image domain through an inverse Fourier transform, which gives an aliased image which needs to go through a reconstruction method in order to obtain a de-aliased estimation of the original, unobserved image. Here \texttt{ifft} stands for inverse fast Fourier transform.}\label{fig:fourier_undersampled}
\end{figure}

However, MRI examinations suffer from being relatively slow compared to other imaging modalities, due to the sequential nature of the acquisition. For instance, MRI acquisition can easily exceed 30 minutes \citep{zbontarFastMRIOpenDataset2019}. Improving imaging speed has been a major concern for researchers of the community and has led to many innovations along the years, such as rapid imaging sequences \citep{frahm1986rapid,hennig1986rare} or parallel/multi-coil imaging \citep{sodickson1997simultaneous,pruessmann1999sense,griswold2002generalized}. But there is a need to further accelerate imaging. As scanning time is directly related to the number of acquisitions, collecting fewer of them directly enables faster imaging. In Cartesian imaging,
this can be achieved by \textit{undersampling} the Fourier space, which is implemented by \textit{not} acquiring some lines in Fourier space. However, as we see on Figure \ref{fig:fourier_undersampled}, such undersampling degrades the quality of the image by introducing artifacts. This is called \textit{aliasing}. As a result, an additional step of \textit{reconstruction} is required in order to obtain a cleaner estimation of the original image.

The problem of reconstructing data from undersampled measurements is a typical instance of an \textit{ill-posed inverse problem}. It is called an \textit{inverse problem} because we aim at constructing an image of an object of interest from \textit{indirect measurements}, in this case frequency information about the image. It is \textit{ill-posed} because there is not sufficient information to directly retrieve the image of interest from the undersampled measurements. More precisely, there are infinitely many signals that could produce these partial measurements.

The typical way of solving this problem is by exploiting \textit{prior knowledge} about the data, by leveraging some \textit{structure} that is present in them. A particularly successful approach relies on \textit{sparsity}-based priors, where it is assumed that if the signal can be represented sparsely, or compactly in some transform domain, then, it can be reconstructed from undersampled measurements. This approach is known as Compressed Sensing (CS) \citep{donoho2006compressed,candes2006robust}, and has been successfully applied to MRI \citep{lustig2008compressed}, for instance by representing the signal sparsely in the Wavelets domain. The signal is retrieved by solving an optimization problem with two components: a \textit{data-fidelity} term that ensures that the reconstructed image is close to the measurements, and a \textit{regularization} term that encourages the reconstructed to be sparse in some domain.

In the latter half of 2010s, deep learning-based approaches allowed pushing even further the boundaries of the state-of-the-art in MRI reconstruction \citep{sun2016deep,schlemper2017deep,mardani2018neural}. By leveraging increasingly large amount of data, these methods are able to learn the structure directly from data, rather than imposing it through a mathematical model such as sparsity. These methods not only improved the quality of reconstruction, but also enabled fast reconstruction times, where CS methods often relied on slow iterative procedures.

However, in parallel to these developments, only few works tackled the problem of \textit{how} the k-space should be undersampled. Although CS prescribed random sampling, practitioners quickly turned to heuristics favoring low frequency sampling, where a lot of energy and information about the structure of the signal is located \citep{lustig2008compressed}. Nevertheless, there were only a few works in the early 2010s aiming at directly optimizing the sampling mask\footnote{We will refer to this problem interchangeably as the problem of mask design, sampling optimization, mask optimization, sampling mask optimization, sampling pattern optimization, or optimizing the acquisition trajectory.}, which can be attributed to the difficulty of the problem. Optimizing a sampling mask is a \textit{combinatorial} problem, for which commonly used gradient-based optimization techniques are not readily available. As a result, two main trends emerged: approximate solutions to the combinatorial problem \citep{ravishankar2011adaptive,liu2012under}, or the optimization of an underlying probability distribution from which a random mask is then sampled \citep{knoll2011adapted,chauffert2013variable}.

In more recent years, there has been a renewed interest in this problem, and a flurry of approaches have been proposed \citep{gozcu2018learning,jin2019self,sherry2019learning,bahadir2019learning,Huijben2020Deep,pineda2020active,bakker2020experimental,yin2021end}. This is mainly due to the wide availability of training data, the fast reconstruction times of deep learning-based methods and the wider use of auto-differentiation frameworks such as PyTorch and TensorFlow, which facilitate the routine computation of derivatives through complex models. While these methods are very diverse, they fit under two distinctive categories. Whereas previous approaches leveraged an underlying \textit{model} from which candidate sampling masks were drawn, modern approaches are \textit{learning-based} in the sense that they learn this model directly from data. Furthermore, earlier methods were \textit{model-driven}, designing masks to minimize a mathematical criterion such as coherence \citep{lustig2008compressed}. In contrast, recent works are \textit{data-driven}, as they learn policies that minimize the reconstruction error against a ground truth.

In this thesis, we aim at contributing to the problem of optimizing sampling masks along two complementary directions.

First, we will provide novel \textit{methodological} contributions to the problem of sampling optimization. More specifically, we aim at developing mask optimization algorithms that do not depend on a specific reconstruction algorithm, as this will allow the method to remain usable in a field where the state-of-the-art evolves very quickly. In addition, we are concerned with designing \textit{scalable} methods. Most clinically relevant settings involve high resolution, multi-coil images, and it is imperative for sampling optimization methods to scale up to such challenging settings in order to provide practical value.

Secondly, we aim at deepening the \textit{understanding} of the kind of structures or assumptions that enable mask design algorithms to perform well in practice. It is not clear from the current literature what components are fundamental in a mask design algorithm.
Should the mask design algorithm plan several steps of ahead? Should it adapt to each different patient?
What kind of feedback should be available to the mask design algorithm in order to achieve the best performance? Amidst the race to develop ever more complex models, we aim at identifying the key components that drive the performance of modern mask design algorithms, in order to \textit{efficiently} reach state-of-the-art performance.

\newpage

\begin{figure}[!t]
    \centering
    \includegraphics[width=\linewidth]{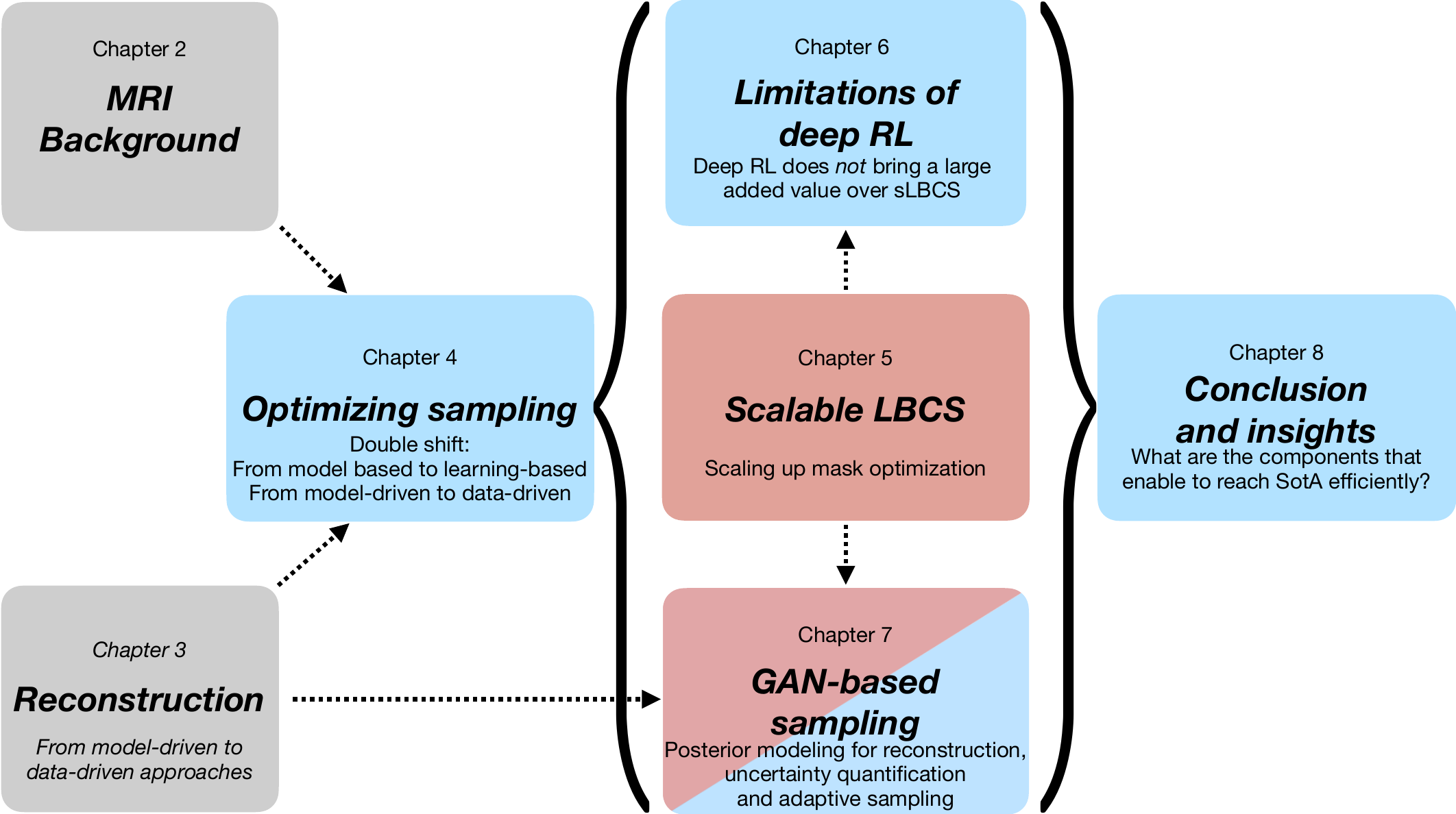}
    \caption{Outline of the different chapters in this thesis. \textbf{Grey} boxes refer to chapters dealing with background material. \textbf{Red} boxes denote chapters with \textit{methodological} contributions, and \textbf{blue} boxes denote chapters that contribute to a better \textit{understanding} of the problem of optimizing sampling. Arrows show particular, direct dependencies between chapters.}\label{fig:organization}
    \vspace{-.5cm}
\end{figure}

\section*{Organization and contributions}
Figure \ref{fig:organization} illustrates the organization of this thesis, and each chapter is briefly described below, along with their respective contributions.

\subsection*{Chapter \ref{ch:mri_background} --  MRI Background}
In this chapter, we provide a brief introduction to the physics underlying image acquisition in MRI, focusing on {Cartesian} sampling. We introduce {multi-coil} acquisition as well as dynamic imaging. We then provide the mathematical model that will be used to describe the acquisition in accelerated MRI.

\subsection*{Chapter \ref{chap:rec} -- Reconstruction methods}
In this chapter, we review the algorithms used to reconstruct images from undersampled measurements. We describe the shift that has taken place from {\textit{model-driven}} approaches based on Compressed Sensing \citep{lustig2007sparse} towards {\textit{data-driven}} approaches that construct a model from training data. We then focus on the different trends that have taken place within data-driven approaches, focusing on trained reconstruction models and generative approaches. Finally, we conclude by highlighting the trade-offs between the {speed of reconstruction} and the dependence on large training datasets.

\subsection*{Chapter \ref{ch:sampling} -- Optimizing sampling patterns}
In this last introductory chapter, we review the different approaches that have been taken to optimize {sampling masks}. We argue that, in a similar way that reconstruction methods moved from model-driven to data-driven approaches, a double shift has taken place in mask design. We show that mask design methods gradually moved from {\textit{model-based}, \textit{model-driven} to \textit{learning-based}, \textit{data-driven} approaches}. We discuss how this trend can be seen in the literature, and how it ties in with the empirical performance of these methods.
Then, we proceed to formalizing the problem of learning-based, data-driven mask design and show how this problem changes when one aims at designing {\textit{fixed policies}} -- where a mask is trained and re-used as is at test time -- or {\textit{adaptive policies}} -- that adapt to the current patient by integrating on-the-fly the information as it acquired. We conclude by proving the optimality of the discrete mask optimization problem, compared to the problem of finding the optimal {sampling \textit{distribution}}.

\textbf{Related publication.}
\begin{adjustwidth}{.5cm}{}
    Sanchez, T., G{\"o}zc{\"u}, B., van Heeswijk, R. B., Eftekhari, A., Il{\i}cak, E., \c{C}ukur, T., and Cevher, V. (2020a). Scalable learning-based sampling optimization for compressive dynamic MRI. In \textit{ICASSP 2020 - 2020 IEEE International Conference on Acoustics, Speech and Signal Processing (ICASSP)}, pages 8584–8588.
\end{adjustwidth}
\subsection*{Chapter \ref{chap:lbcs} -- Scalable learning-based sampling optimization}
In this chapter, we extend the Learning-Based Compressive Sensing (LBCS) approach of \citet{gozcu2018learning}. We primarily aim at \textit{drastically} improving their scalability of their method. We first discuss how their approach is rooted in submodular optimization \citep{krause2014submodular} and how algorithms proposed in this field could provide solutions to the limitation of LBCS. Their method scales as $O(P^2)$, where $P$ is the resolution of the image, such a complexity is prohibitive in high dimensional settings such as 3D MRI or dynamic MRI, and is not convenient when reconstruction becomes slow, such as in multi-coil MRI. To address this, we first propose \textit{lazy LBCS} (lLBCS), for the multi-coil and 3D settings. This algorithm leverages lazy evaluations \citep{minoux1978accelerated}, where a list of upper bounds is \textit{precomputed} and then traversed sequentially, drastically reducing the amount of computation over LBCS. Secondly, in the context of dynamic (multi-coil) MRI and large datasets, we propose\textit{ stochastic LBCS} (sLBCS) and show that it can achieve a very significant computational gain compared to LBCS, while retaining its performance.

\textbf{Related publications.}
\begin{adjustwidth}{.5cm}{}
    G{\"o}zc{\"u}, B., Sanchez, T., and Cevher, V. (2019). Rethinking sampling in parallel MRI: A data-driven approach. In \textit{27th European Signal Processing Conference (EUSIPCO)}.\\[0.4cm]
    Sanchez, T., G{\"o}zc{\"u}, B., van Heeswijk, R. B., Eftekhari, A., Il{\i}cak, E., \c{C}ukur, T., and Cevher, V. (2020a). Scalable learning-based sampling optimization for compressive dynamic MRI. In \textit{ICASSP 2020 - 2020 IEEE International Conference on Acoustics, Speech and Signal Processing (ICASSP)}, pages 8584–8588.
\end{adjustwidth}

\subsection*{Chapter \ref{ch:rl_mri} -- On the benefits of deep RL in accelerated MRI sampling}
In this chapter, we assess the performance of state-of-the-art (SotA) deep reinforcement learning (RL) methods for mask designs. These methods improve in principle over LBCS by designing policies that \textit{adapt} to the patient at hand and leverage \textit{long-horizon} planning in order to make better decisions. We mainly aim at making sense of seemingly contradictory conclusions. The work of \citet{pineda2020active} seems to indicate that long term planning could be the most important component in deep RL, as their results show that a non-adaptive, long term planning policy model trained on the dataset can perform as well as an adaptive, long-term planning policy. On the contrary, the contribution of \citet{bakker2020experimental} highlights the importance of adaptivity, as a greedy policy, that does not do long-term planning is found to closely match policies that do long term planning.

Our results, surprisingly show that sLBCS, a non-adaptive method that does not attempt long term planning, can perform as well as the state-of-the-art approaches of \citet{pineda2020active,bakker2020experimental}. We further show that the current SotA RL methods only bring at best a limited added value, and that other factors such as the architecture of the reconstruction algorithm or the masks used to train it have a much larger impact on the overall performance. This work  highlights the need for further discussions in the community about standardized metrics, strong baselines, and careful design of experimental pipelines to evaluate MRI sampling policies fairly.

\textbf{Related preprint.}
\begin{adjustwidth}{.5cm}{}
    Sanchez, T.$^{*}$, Krawczuk, I.$^{*}$ and Cevher, V. (2021). On the benefits of deep RL in accelerated MRI sampling. \textit{Available at \url{https://openreview.net/pdf?id=fRb9LBWUo56}}. $^{*}$ denotes equal contribution.
\end{adjustwidth}

\subsection*{Chapter \ref{ch:gans} -- Uncertainty driven adaptive sampling via GANs}
In this chapter, we take a step back from the question of designing the best sampling policy, and primarily aim at exploring how conditional Generative Adversarial Networks (GANs) \citep{goodfellow2014generative} can be used to model inverse problems in a Bayesian fashion.
GANs have been used mainly as reconstruction models in the context of inverse problems \citep{yang_dagan_2018,chen2022ai}, but a few works took an additional step to show the value of learning the full posterior distribution in order to have the ability to provide uncertainty quantification as well \citep{adler2018deep}.

We argue that they do more and that conditional GANs for MRI reconstruction can naturally provide a criterion for adaptive sampling: by sequentially observing the location with the largest posterior variance in the measurement domain, the GAN provides an adaptive sampling policy without ever being trained for it.

Such a criterion is not restricted to acquisition in Fourier domain, like in MRI, but can also be readily used for image domain sampling. In this context, we show that our GAN-based policy can strongly outperform the non-adaptive greedy policy from sLBCS. However, in Fourier domain, the GAN-based policy does not match the performance of sLBCS. We provide an explanation to this phenomenon rooted in the concept of the information horizon used to make a decision at each step. We show that our model uses less information than LBCS to design its policy, and argue that this is the reason leading to its inferior performance. However, when compared with models that use the same amount of information to inform their policy, our model largely outperforms the competition.

\textbf{Related preprint and workshop paper.}
\begin{adjustwidth}{.5cm}{}

    Sanchez, T., Krawczuk I., Sun, Z. and Cevher V. (2019). Closed loop deep Bayesian inversion: Uncertainty driven acquisition for fast MRI. \textit{Preprint available at} \url{https://openreview.net/pdf?id=BJlPOlBKDB}.\\[0.4cm]
    Sanchez, T., Krawczuk, I., Sun, Z. and Cevher V. (2020). Uncertainty-Driven Adaptive Sampling via GANs. In \textit{NeurIPS 2020 Workshop on Deep Learning and Inverse Problems}. Available at \url{https://openreview.net/pdf?id=lWLYCQmtvW}.
\end{adjustwidth}

\subsection*{Chapter \ref{ch:conclusion} -- Conclusion and future works}
In this last chapter, we review the contributions of the different chapters and summarize the insights that our contributions bring to the design of masks for Cartesian MRI sampling. We then provide an outlook of the opportunities for future works.

\begin{remark}[Bibliographic note]
    At the end of Chapters \ref{ch:sampling}, \ref{chap:lbcs}, \ref{ch:rl_mri} and \ref{ch:gans}, we include a ``bibliographic note'' section, where we distinguish the contributions of the co-authors of the abovementioned publications.
\end{remark}


\textbf{Additional publication not included in this thesis.}
\begin{adjustwidth}{.5cm}{}
    Sun, Z., Latorre, F., Sanchez, T., and Cevher, V. (2021). A plug-and-play deep image prior. In \textit{ICASSP 2021 - 2021 IEEE International Conference on Acoustics, Speech and Signal Processing (ICASSP)}, pp. 8103-8107.
\end{adjustwidth}

\cleardoublepage
\chapter{MRI Background}\label{ch:mri_background}
Before diving in greater detail in the problems of reconstruction from undersampled measurements and optimization of the sampling masks, we first present the physical underpinnings of MRI. Although this survey will be very short, it hints at the remarkable depth and flexibility of MRI. This introduction is based upon the well-known textbook of \citet{brown2014magnetic}\footnote{Part of the content was presented in \citet{sanchez2018master}}.

\section{MRI physics}

\subsection{Magnetization}

Let us then consider a proton under a strong external magnetic field $\*B_0$. In practice, such a field is obtained by running a large electric current through a coil around the scanner. High current is required because typical magnetic fields in MRI are very large, around $\SI{3}{\tesla}$, and the strength of the  magnetic field is directly proportional to the one of the electric current\footnote{For comparison, the strength of the magnetic field of the earth is typically between $40$ and $\SI{70}{\micro\tesla}$.}.

\begin{figure}[!ht]
    \centering
    \includegraphics[width=.28\textwidth]{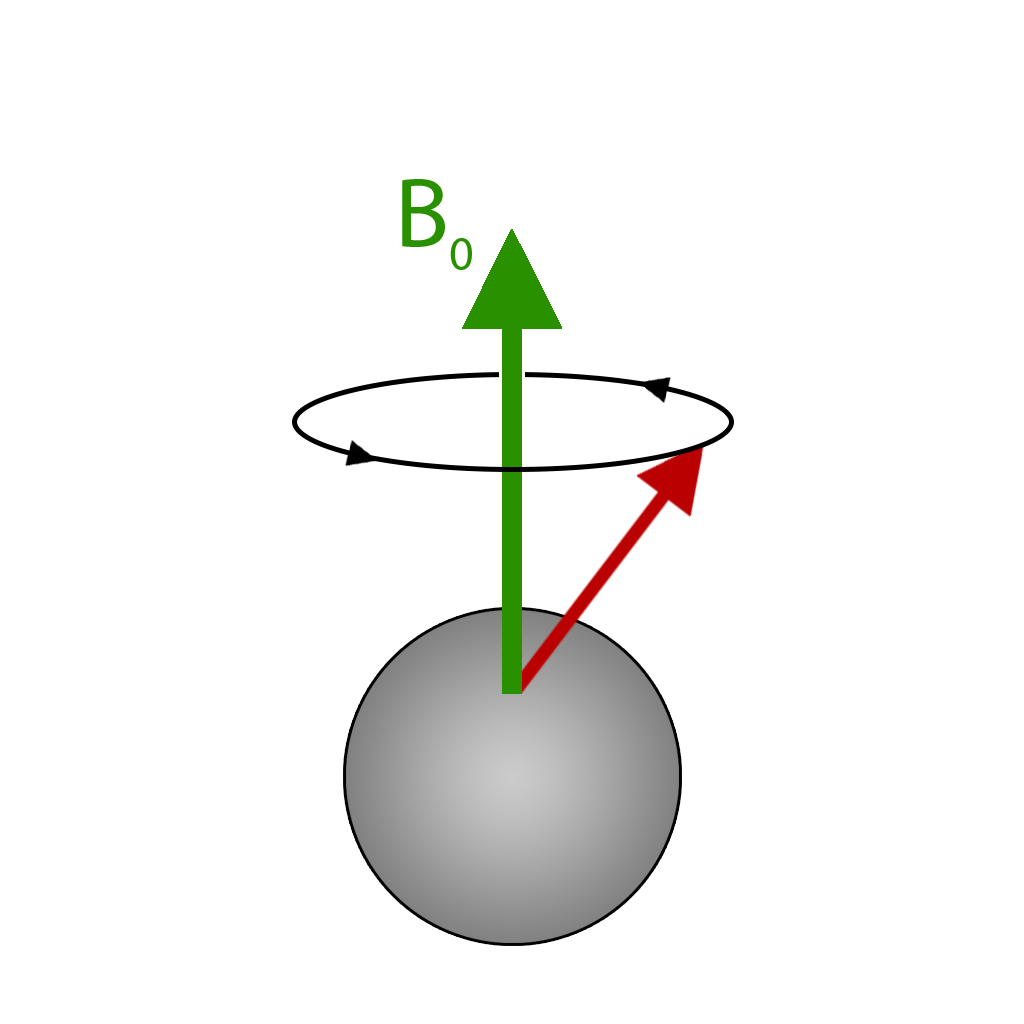}
    \hspace{0.5cm}
    \includegraphics[width=.65\textwidth]{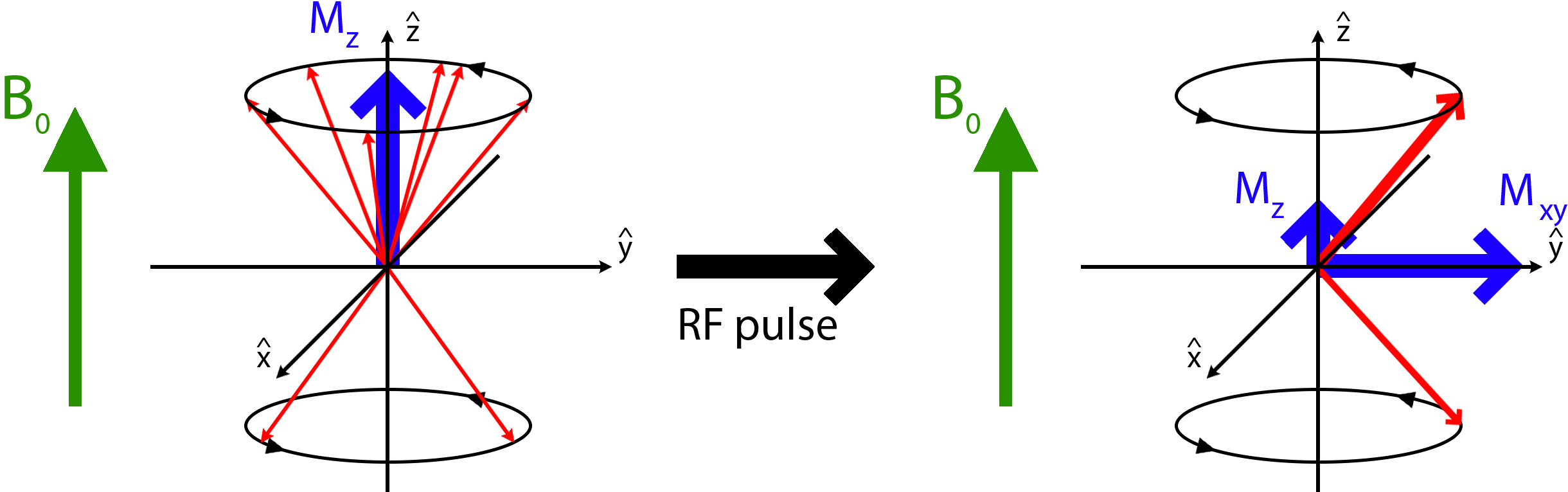}
    \caption{\textit{Left:} Illustration of precession of a particle when subject to an external field $\*B_0$ (large green arrow). The small red arrow corresponds to the spin of the particle, and its temporal evolution is illustrated by the black arrow. \textit{Right:} Illustration of the \textit{transverse} magnetization appearing when a body is subject to an RF pulse, due to the protons precessing synchronously, and reaching the higher energy level, which results in arrows pointing downwards. This Figure was first used in \citet{sanchez2018master}.}\label{fig:precession}
\end{figure}

The spin of a proton, which is positively charged, can be represented as a gyroscope precessing about the field direction, as illustrated on Figure \ref{fig:precession}. The precession angular frequency can be quantified as the Larmor frequency $\omega_0$, with
\begin{equation}
    \omega_0 = \gamma B_0\label{eq:larmor}
\end{equation}
where $\gamma$ is known as the gyromagnetic ratio, and is a particle dependent constant. For instance, in water, the hydrogen proton has $\gbar \triangleq \gamma/(2\pi) \approx \SI{42.6}{\mega\hertz\per\tesla}$. Note that deriving this formula is involved and requires tools from relativistic quantum mechanics. 

\begin{figure}
    \centering
    \includegraphics[width=0.8\linewidth]{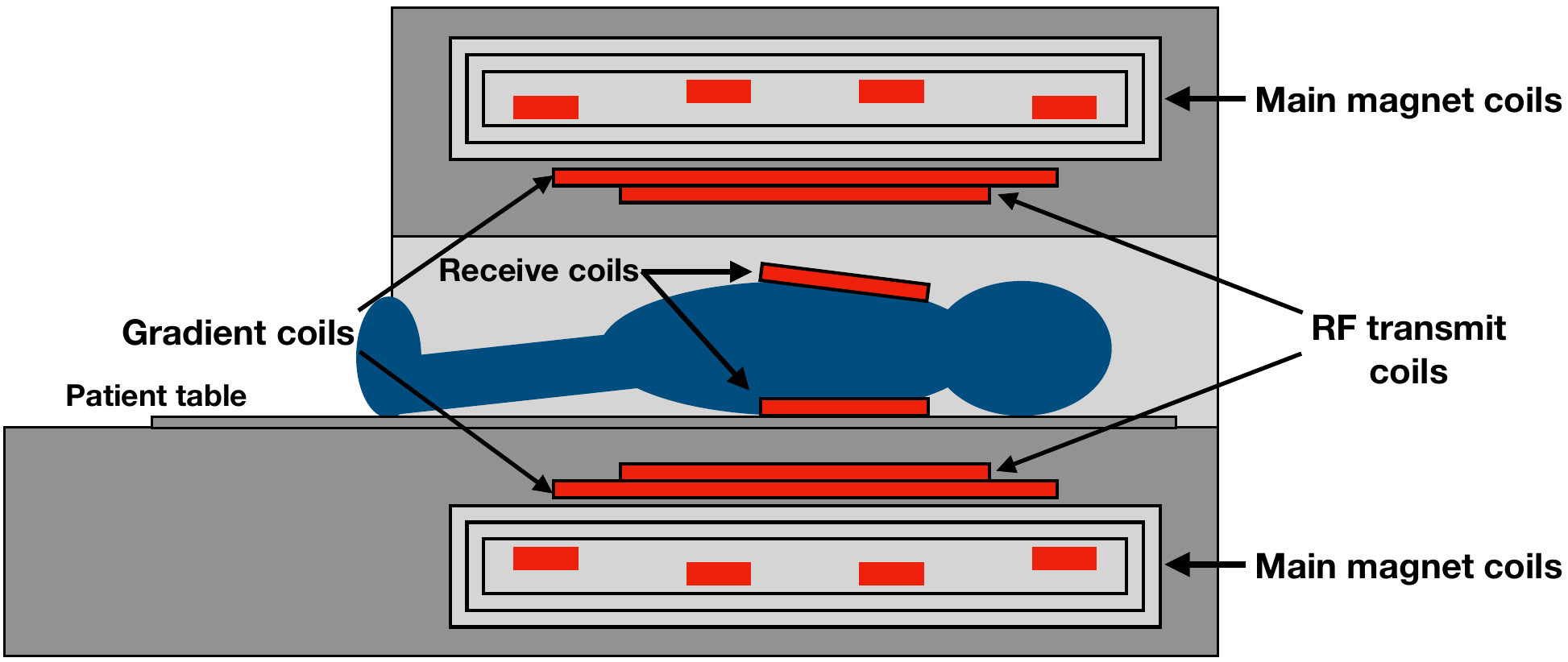}
    \caption{Illustration of an MRI scanner and the different coils and components required to produce an image.}\label{fig:MRI_scanner}
\end{figure}

This description considers a single hydrogen nucleus in a static magnetic field, but in practice, we are dealing with an entire population. Depending on their levels of energy, different protons of a population will either align in a parallel fashion against an external field $\*B_0$, while some will take an antiparallel position. Although there is only a very little amount of spins parallel to the magnetic field exceeding the number of spins antiparallel to it, this is sufficient to have a \textit{spin excess} that yields a noticeable net magnetization $\*M_0$ at a population level. $\*M_0$ will be parallel to $\*B_0$ because different protons will have incoherent phases, and will overall cancel each other out, leading to a net magnetization parallel to the static field. We call this resulting state the \textit{equilibrium} of the system, which is illustrated in the middle of Figure \ref{fig:precession}, where in this case $\*M_0 = \*M_z$. However, this net magnetization cannot be detected in the equilibrium, as we need to excite the protons in order to induce a change in the magnetization, which in turn will induce an electrical current in a \textit{receive coil}. This coil will allow to record a signal corresponding to the response of the protons to the excitation. The different components of the MRI that will be discussed throughout this chapter are illustrated on Figure \ref{fig:MRI_scanner}.

\subsection{Excitation}
Let us denote by $\*{\hat{z}}$ the direction of the external field $\*B_0$, and let us consider a body that we wish to image, for instance the knee of a patient. It is clear from the previous section that the hydrogen nuclei of the body will produce a net magnetization $\*M_0 = M_0 \*{\hat{z}}$, where $M_0$ denotes the intensity of magnetization. The atoms will precess around  $\*{\hat{z}}$ at the Larmor frequency $\omega_0$.

In the excitation step, the system is destabilized by sending a radiofrequency (RF) pulse. It is easier to describe it and consider its effect by placing ourselves in a frame rotating at an angular frequency $\omega_0$ around the axis $\*{\hat{z}}$. This gives the following rotating unit vectors $\*{\hat{x}'}$, $\*{\hat{y}'}$
\begin{align*}
    \*{\hat{x}'} & =\cos(\omega_0 t)\*{\hat{x}} -  \sin(\omega_0 t)\*{\hat{y}} \\
    \*{\hat{y}'} & = \sin(\omega_0 t)\*{\hat{x}} + \cos(\omega_0 t)\*{\hat{y}}
\end{align*}
and $\*{\hat{z}}$ remains unchanged. We define the RF pulse as a circularly rotating pulse of intensity $B_1$, namely $\*B_1 = B_1 \*{\hat{x}'}$. Applying this RF pulse will as a result cause the nuclei to resonate and flip the equilibrium magnetization $\*M_0$ into a new net magnetization $\*M$ in the $\*{\hat{x}'}-\*{\hat{y}'}$ plane. $\*M$ will now have both a \textit{transverse} component in the $\*{\hat{x}'}-\*{\hat{y}'}$ plane, and a \textit{longitudinal} component along the $\*{\hat{z}}$ axis. We refer to the angle between $\*M$ and the $\*{\hat{z}}$-axis as the \textit{flip angle} (FA), and it quantifies how strongly the magnetization $\*M$ has been moved away from its equilibrium state $\*M_0$. This process is illustrated in the right of Figure \ref{fig:precession}. In particular, we see that the transverse component arises due to the synchronization of the precessing phases of the hydrogen nuclei, which  cancelled each other out at equilibrium.

\subsection{Relaxation}
After applying the RF pulse, the system will gradually return to equilibrium, where $\*M_0$ is aligned with $\*B_0$. This will require the nuclei to re-emit the energy that they absorbed during the excitation phase. The relaxation is governed by two main components, namely a \textit{longitudinal} relaxation, characterized by a time constant $\mathbf{T_1}$, and a \textit{transverse} relaxation, characterized by  $\mathbf{T_2}$.


\paragraph{$T_1$ Relaxation.}~The $T_1$ relaxation describes the regrowth of the longitudinal component of the magnetization, i.e. how the $\*{\hat{z}}$ component of $\*M$ grows back over the relaxation. This is also known as \textit{spin-lattice} relaxation, as this describes a local process due to the interaction of the spins with their surroundings. The process is formally described as
\begin{equation}
    M_z(t) = M_0\big(1-\exp\left(-t/T_1\right)\big) + M_z\left(0\right)\exp\left(-t/T_1\right)
\end{equation}
where $M_z(t)$ is the longitudinal magnetization, $M_z(0)$ describes the amount of longitudinal magnetization left after the RF pulse, and $T_1$ is a constant that depends on the type of tissue considered. We see that the longitudinal magnetization will be recovered at an exponential speed, and the speed will be determined by $T_1$.

\paragraph{$T_2$ Relaxation.}~The $T_2$ relaxation process describes how clusters of spins gradually lose phase with each other after excitation. This dephasing will cause the transverse magnetization $M_{\hat{x}',\hat{y}'}$ or $M_{\perp}$ to progressively disappear, as the equilibrium state is reached (cf. middle of Figure \ref{fig:precession}). We have
\begin{equation}
    M_{\perp}(t) = M_0\exp\left(-t/T_2\right)
\end{equation}
where $T_2$ is a tissue-dependent time constant that describes how fast the transverse magnetization vanishes. In practice, $T_2 \ll T_1$, making the $T_2$ relaxation much faster. Also, in practice, the signal is likely to suffer additional suppression due to dephasing from inhomogeneities in the external field $\*B_0$.

As $T_1$ and $T_2$ are tissue dependent constants, constructing images by performing measurements at different steps of the relaxation allows to highlight different characteristics of tissues. They are one of the key components of the flexibility of MR imaging.


\subsection{Image acquisition}\label{sec:imacquisition}
Until now, we have discussed the global response of a sample to an RF pulse, but we have not addressed how to encode spatial information into the signal, in order to be able to construct an image. As previously discussed, the signal will be recorded by a receive coil, where an electric current is induced depending on the relaxation of the spins.

But the question of localizing the signal remains whole so far. To solve this issue, we will try to give to each location in the body a unique combination of phase and frequency, in order to be able to retrieve its contribution from the signal in the receive coil. This is achieved by introducing three additional coils to the scanner, known as gradient coils, whose role will be to produce some spatially varying magnetic fields in the ${\hat{x}}$,  $\*{\hat{y}}$ and  $\*{\hat{z}}$ directions respectively. As a result, the tissues of specific areas of the object will be excited, and this will allow to observe their local responses. Then, by repeating a measurement with different gradients, we will be able to sequentially construct a map of how the tissue responds to the RF pulses depending on its location. This is why an acquisition is often referred to as a \textit{pulse sequence}, as we gather a sequence of measurements in order to obtain a full image.

\begin{figure}[!ht]
    \vspace{1cm}
    \centering
    \includegraphics[width=0.8\linewidth]{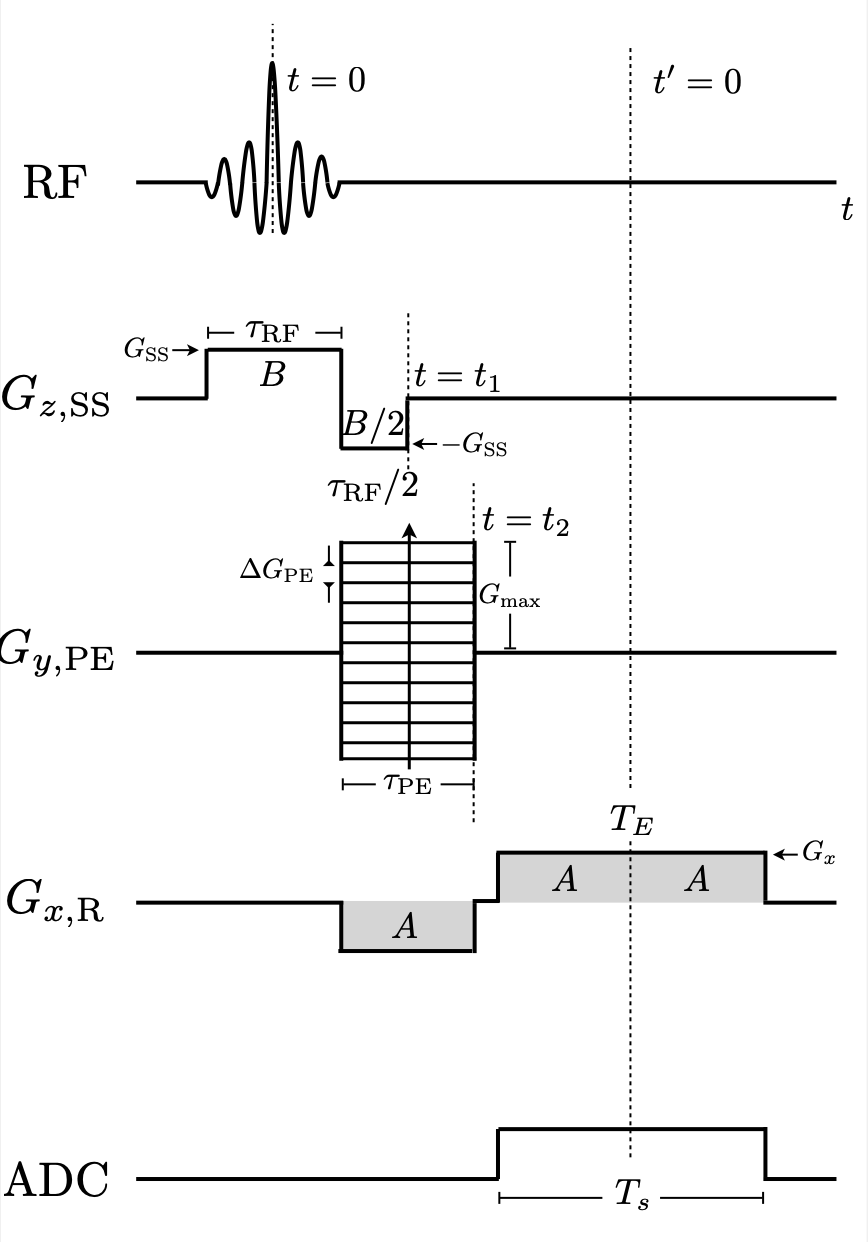}
    \caption{A 2D gradient echo sequence diagram. Each line describes how different components interact during the time of a single excitation. On the first line, we see the RF pulse enabled at $t=0$. On the second line, we have the slice selecting gradient $G_{z,\text{SS}}$, on the third line we have the phase encoding gradient $G_{y,\text{PE}}$, on the fourth we have the readout gradient $G_{x,\text{R}}$ and finally the analog-to-digital converter (ADC).}\label{fig:gre_diagram}
    \vspace{1cm}
\end{figure}

There exists several ways to construct an image with MRI, and we will illustrate how it works by considering a \textit{2D gradient echo} sequence, and we will explain its different components. Sequences are frequently represented as diagrams, such as the one of Figure \ref{fig:gre_diagram}, where each line represents how different component vary in intensity through time. If we simply excite a body with an RF pulse, neglecting the relaxation effects, we can write the signal observed in the receive coil as
\begin{equation}
    S(t) = \int \rho(\*x) e^{i(\omega_0 t + \phi(\*x,t))} d\*x\label{eq:signal_og}
\end{equation}
where $\rho(\*x)$ is the \textit{effective spin density}, and is the quantity that we wish to measure, and $\phi(\*x,t)$ is the \textit{accumulated phase}
\begin{equation}
    \phi(\*x,t) = - \int_0^t \omega(\*x,t')dt'. \label{eq:accumulated_phase}
\end{equation}
We see that in Equation \ref{eq:signal_og}, without additional considerations, the signal that we observe is integrated over the whole volume being imaged. In Equation \ref{eq:accumulated_phase}, the main reason that $\omega$ is now a function of space and time is that we will add spatially varying gradient fields that in turn make $\omega$ dependent on space and time. One can also see that \Eqref{eq:signal_og} shows that measurements will not be provided directly in image space, but rather in Fourier space, often referred to as \textit{k-space}: MRI allows us to obtain \textit{frequency} information about our object of interest, and then an image can be obtained by mapping back these measurements into image space.


\paragraph{Slice Selecting (SS) Gradient.}~As we see on Figure \ref{fig:gre_diagram}, during the RF pulse, a constant $\*z$-gradient is applied during a time $\tau_{\text{RF}}$ with intensity $G_{\text{SS}}$, and reversed afterwards. We will not detail here why this gradient is reversed, but this has the effect of making all components in the slice in phase, with a common accumulated phase $\phi = 0$. This enables to select the slice of interest and make sure that the signal will be strong for the acquisition. As a result, for a slice at location $\z_0$ with thickness $\Delta z$, the signal will be integrated out on the slice only, i.e. we will have the spin density $$ S(t_1) = \int \int \left[\int_{z_0 -\frac{\Delta z}{2}}^{z_0 +\frac{\Delta z}{2}} \rho(x,y,z) dz\right]dydx  \approx \int \int\rho(x,y,z_0) dydx.$$

\textbf{Phase Encoding (PE) Gradient.} Let us now add the phase encoding $\*y$-gradient to the picture. Assume that it is applied during a time $\tau_{\text{PE}}$ at intensity $G_{\min} \leq G_y \leq G_{\max}$. This will result in gradients acquiring a $\*y$-dependent phase, and as a result
$$S(t_2) = \int \left[\int\rho(x,y,z_0) e^{-i2\pi \gbar G_y \tau_{\text{PE}}y}dy \right]dx$$


\textbf{Readout (R) Gradient.} Finally, the $\*x$-gradient is first turned on negatively, dephasing the phases, followed by a rephasing stage where the observation through the ADC is carried out. It is enabled at the strength $G_x$ and for a duration $T_S$. Given the shifted variable $t' = t-T_E$, we have as a result
\begin{equation}
    S(t') = \int \left[\int\rho(x,y,z_0) e^{-i2\pi \gbar G_y \tau_{\text{PE}}y}dy \right] e^{-i2\pi \gbar G_xt'x}dx, \text{~~} -T_S/2 \leq t' \leq T_S/2. \label{eq:s_full}
\end{equation}

\begin{figure}[!ht]
    \centering
    \includegraphics[width=0.6\linewidth]{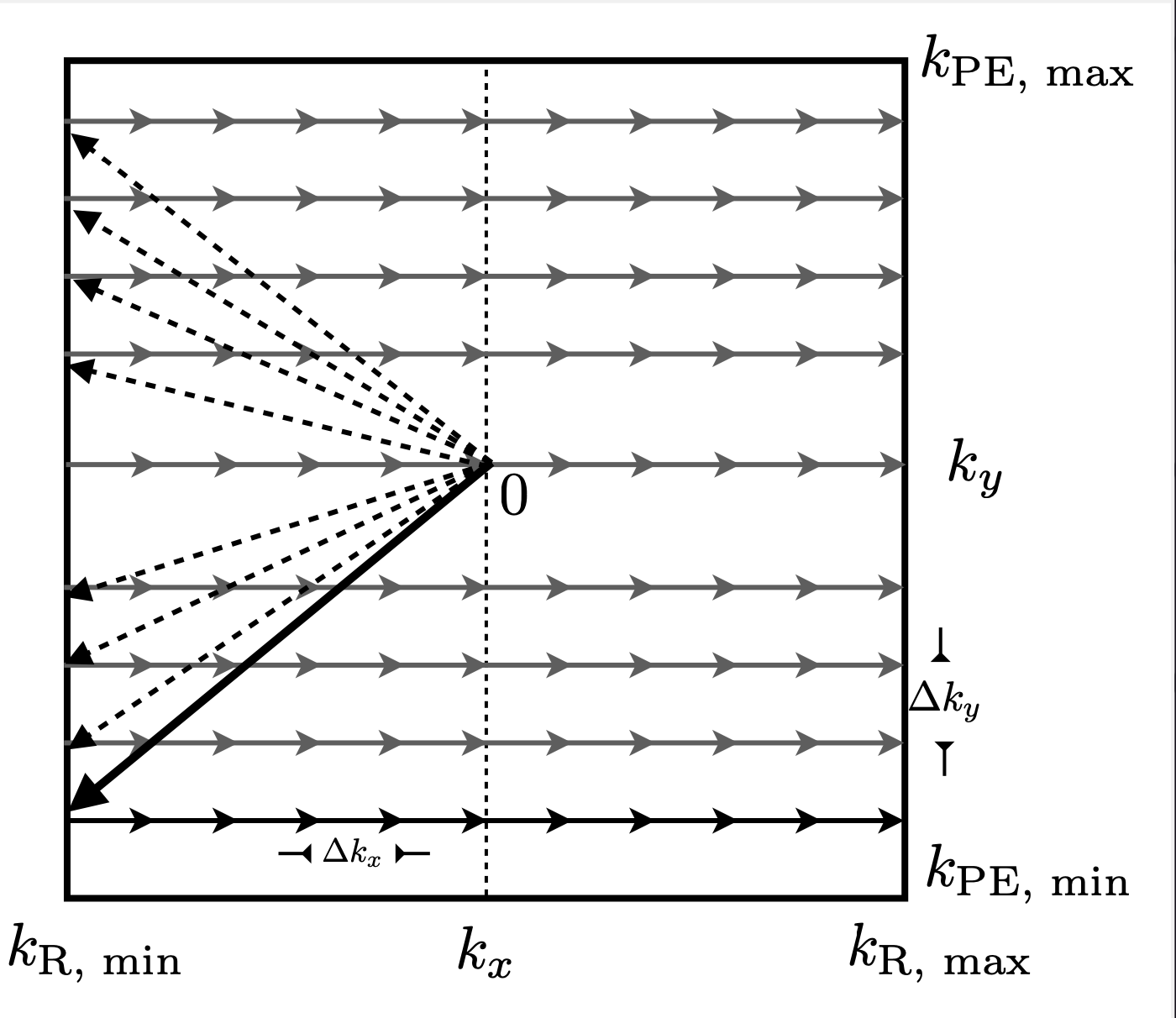}
    \caption{Traversal of k-space for a gradient echo pulse sequence. The combined application of the phase encoding gradient $G_{y,\text{PE}}$ and readout gradient $G_{x,\text{R}}$ males the signal move in k-space to the desired location in a diagonal trajectory. The phase encoding gradient is then stopped, and the readout gradient is reversed, bringing the spins back through a complete set of $k_x$ value while keeping $k_y$ constant. The experiment is then repeated by returning to a different $k_y$ value until the whole of k-space has been covered in a grid-like fashion.}\label{fig:kspace_coverage}
\end{figure}

\Eqref{eq:s_full} is very interesting as one can rewrite the $(\*x,\*y)$-directions as spatial frequencies depending on $t$ and $\tau_{y}$, namely $$(k_x,k_y) = (\gbar G_xt', \gbar G_y \tau_{\text{PE}}).$$
This in turns enables us to rewrite \Eqref{eq:s_full} as
\begin{equation}
    S(k_x,k_y) =  \int \int\rho(x,y,z_0) e^{-i2\pi (k_x x + k_y y)} dxdy\label{eq:skspace}
\end{equation}
which turns out to be the Fourier transform of the spin density. More precisely, the signal obtained in the presence of a gradient echo structure in the readout direction after its encoding by a fixed $G_y$ describes a \textit{line} in the k-space of the Fourier transform of $\rho(\*x)$. This implies that by acquiring several lines at different $k_y$ by varying $G_y$, we can effectively cover the k-space, as illustrated in Figure \ref{fig:kspace_coverage}. As a result, when the k-space has been sufficiently densely covered, the spin-density can be recovered by taking the inverse Fourier transform of \Eqref{eq:skspace}, i.e.
\begin{equation}
    \rho(x,y,z_0) =  \int \int S(k_x,k_y) e^{+i2\pi (k_x x + k_y y)} dk_xdk_y\label{eq:inversion}
\end{equation}

This approach is referred to as \textit{Cartesian sampling} because it covers k-space in a grid-like fashion, as illustrated on Figure \ref{fig:kspace_coverage}. Note that there exist different sequences that can perform Cartesian sampling, and even more sequences that are \textit{not} Cartesian. We refer to such approaches as non-Cartesian sampling approaches, and they can include radial sampling \citep{lauterbur1973image}, spiral sampling \citep{meyer1992fast} or even some more flexible coverages of k-space \citep{lazarus2019sparkling}.

Cartesian MRI has the advantage over these methods of enabling the use of the fast Fourier transform to compute the inversion of Equation \ref{eq:inversion}, which is \textit{not} the case for non-Cartesian methods. These approaches require some more sophisticated inversion, involving gridding \citep{o1985fast} and non-uniform Fourier transforms \citep{fessler2003nonuniform}. The differences are however deeper than merely the inversion technique, as radial trajectories are known to be more robust to motion artifacts, while spiral acquisitions enable fast imaging but are prone to other artifacts, and \citet{lustig2008compressed} discuss these issues in greater depth.

\subsection{Parallel imaging}

The need to cover k-space through sequential readouts makes MR imaging a slow imaging technique, and the quality of the images is greatly dependent on the absence of motion during the acquisition. This also implies that for body MRI, the patient should not breathe during the whole acquisition process, as this motion would yield a reconstruction image with many undesired artifacts \citep{glockner2005parallel}. This has motivated further research into ways to accelerate the acquisition.

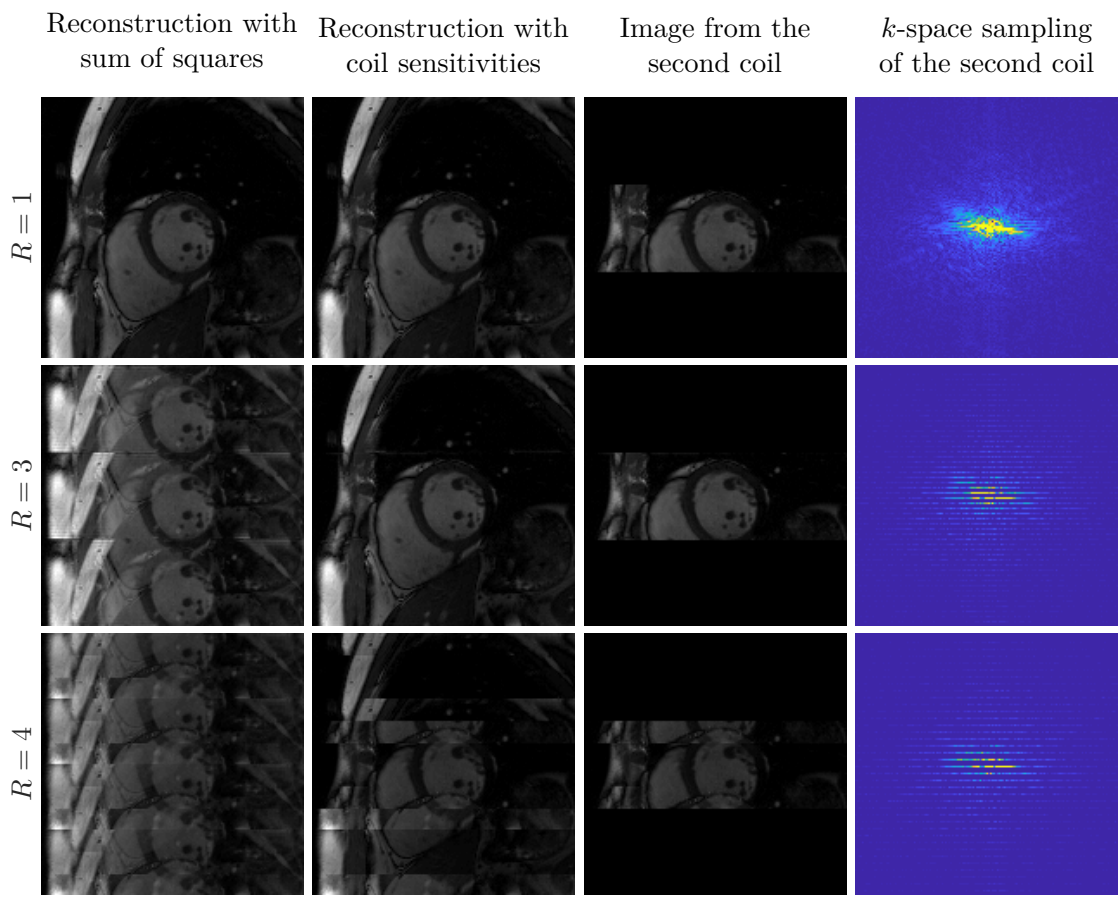
\begin{figure}[!t]
    \centering
    \resizebox*{\linewidth}{!}{
%
%
\begin{tikzpicture}

\begin{axis}[%
width=1.4in,
height=1.4in,
at={(2.179in,2.931in)},
scale only axis,
point meta min=0,
point meta max=1,
axis on top,
xmin=0.5,
xmax=152.5,
tick align=outside,
y dir=reverse,
ymin=0.5,
ymax=152.5,
title style={align=center},
title={Reconstruction with\\coil sensitivities},
axis line style={draw=none},
ticks=none
]
\addplot [forget plot] graphics [xmin=0.5, xmax=152.5, ymin=0.5, ymax=152.5] {pdm/parallel_sampling/parallel_sampling-1.png};
\end{axis}

\begin{axis}[%
width=1.4in,
height=1.4in,
at={(0.729in,2.931in)},
scale only axis,
point meta min=0,
point meta max=1,
axis on top,
xmin=0.5,
xmax=152.5,
tick align=outside,
y dir=reverse,
ymin=0.5,
ymax=152.5,
title style={align=center},
title={Reconstruction with\\sum of squares},
ylabel style={font=\color{white!15!black}},
ylabel={$R = 1$},
axis line style={draw=none},
ticks=none
]
\addplot [forget plot] graphics [xmin=0.5, xmax=152.5, ymin=0.5, ymax=152.5] {pdm/parallel_sampling/parallel_sampling-2.png};
\end{axis}

\begin{axis}[%
width=1.4in,
height=1.4in,
at={(3.629in,2.931in)},
scale only axis,
point meta min=0,
point meta max=1,
axis on top,
xmin=0.5,
xmax=152.5,
tick align=outside,
y dir=reverse,
ymin=0.5,
ymax=152.5,
title style={align=center},
title={Image from the\\second coil},
axis line style={draw=none},
ticks=none
]
\addplot [forget plot] graphics [xmin=0.5, xmax=152.5, ymin=0.5, ymax=152.5] {pdm/parallel_sampling/parallel_sampling-3.png};
\end{axis}

\begin{axis}[%
width=1.4in,
height=1.4in,
at={(2.179in,1.487in)},
scale only axis,
point meta min=0,
point meta max=0.334620332940932,
axis on top,
xmin=0.5,
xmax=152.5,
tick align=outside,
y dir=reverse,
ymin=0.5,
ymax=152.5,
axis line style={draw=none},
ticks=none
]
\addplot [forget plot] graphics [xmin=0.5, xmax=152.5, ymin=0.5, ymax=152.5] {pdm/parallel_sampling/parallel_sampling-4.png};
\end{axis}

\begin{axis}[%
width=1.4in,
height=1.4in,
at={(0.729in,1.487in)},
scale only axis,
point meta min=0,
point meta max=0.334620332940932,
axis on top,
xmin=0.5,
xmax=152.5,
tick align=outside,
y dir=reverse,
ymin=0.5,
ymax=152.5,
ylabel style={font=\color{white!15!black}},
ylabel={$R=3$},
axis line style={draw=none},
ticks=none
]
\addplot [forget plot] graphics [xmin=0.5, xmax=152.5, ymin=0.5, ymax=152.5] {pdm/parallel_sampling/parallel_sampling-5.png};
\end{axis}

\begin{axis}[%
width=1.4in,
height=1.4in,
at={(3.629in,1.487in)},
scale only axis,
point meta min=0,
point meta max=0.334620332940932,
axis on top,
xmin=0.5,
xmax=152.5,
tick align=outside,
y dir=reverse,
ymin=0.5,
ymax=152.5,
axis line style={draw=none},
ticks=none
]
\addplot [forget plot] graphics [xmin=0.5, xmax=152.5, ymin=0.5, ymax=152.5] {pdm/parallel_sampling/parallel_sampling-6.png};
\end{axis}

\begin{axis}[%
width=1.4in,
height=1.4in,
at={(2.179in,0.044in)},
scale only axis,
point meta min=0,
point meta max=0.35396953352875,
axis on top,
xmin=0.5,
xmax=152.5,
tick align=outside,
y dir=reverse,
ymin=0.5,
ymax=152.5,
axis line style={draw=none},
ticks=none
]
\addplot [forget plot] graphics [xmin=0.5, xmax=152.5, ymin=0.5, ymax=152.5] {pdm/parallel_sampling/parallel_sampling-7.png};
\end{axis}

\begin{axis}[%
width=1.4in,
height=1.4in,
at={(0.729in,0.044in)},
scale only axis,
point meta min=0,
point meta max=0.35396953352875,
axis on top,
xmin=0.5,
xmax=152.5,
tick align=outside,
y dir=reverse,
ymin=0.5,
ymax=152.5,
ylabel style={font=\color{white!15!black}},
ylabel={$R=4$},
axis line style={draw=none},
ticks=none
]
\addplot [forget plot] graphics [xmin=0.5, xmax=152.5, ymin=0.5, ymax=152.5] {pdm/parallel_sampling/parallel_sampling-8.png};
\end{axis}

\begin{axis}[%
width=1.4in,
height=1.4in,
at={(3.629in,0.044in)},
scale only axis,
point meta min=0,
point meta max=0.35396953352875,
axis on top,
xmin=0.5,
xmax=152.5,
tick align=outside,
y dir=reverse,
ymin=0.5,
ymax=152.5,
axis line style={draw=none},
ticks=none
]
\addplot [forget plot] graphics [xmin=0.5, xmax=152.5, ymin=0.5, ymax=152.5] {pdm/parallel_sampling/parallel_sampling-9.png};
\end{axis}

\begin{axis}[%
width=1.4in,
height=1.4in,
at={(5.079in,2.931in)},
scale only axis,
point meta min=0,
point meta max=1,
axis on top,
xmin=0.5,
xmax=152.5,
tick align=outside,
y dir=reverse,
ymin=0.5,
ymax=152.5,
title style={align=center},
title={$k$-space sampling\\of the second coil},
axis line style={draw=none},
ticks=none
]
\addplot [forget plot] graphics [xmin=0.5, xmax=152.5, ymin=0.5, ymax=152.5] {pdm/parallel_sampling/kspace-1.png};
\end{axis}

\begin{axis}[%
width=1.4in,
height=1.4in,
at={(5.079in,1.487in)},
scale only axis,
point meta min=0,
point meta max=1,
axis on top,
xmin=0.5,
xmax=152.5,
tick align=outside,
y dir=reverse,
ymin=0.5,
ymax=152.5,
axis line style={draw=none},
ticks=none
]
\addplot [forget plot] graphics [xmin=0.5, xmax=152.5, ymin=0.5, ymax=152.5] {pdm/parallel_sampling/kspace-2.png};
\end{axis}

\begin{axis}[%
width=1.4in,
height=1.4in,
at={(5.079in,0.044in)},
scale only axis,
point meta min=0,
point meta max=1,
axis on top,
xmin=0.5,
xmax=152.5,
tick align=outside,
y dir=reverse,
ymin=0.5,
ymax=152.5,
axis line style={draw=none},
ticks=none
]
\addplot [forget plot] graphics [xmin=0.5, xmax=152.5, ymin=0.5, ymax=152.5] {pdm/parallel_sampling/kspace-3.png};
\end{axis}
\end{tikzpicture}%
}

    \caption{Example of parallel imaging on a cardiac image, simulated with three coils, whose sensitivities are three superimposed rectangles, each covering one third of the image. \textit{First column:} combination via the sum of squares (SoS) technique, where the images from each coil are simply summed weighted by their modulus. \textit{Second column:} Reconstruction taking into account the sensitivity maps, allowing for precise reconstruction for an acceleration rate $R$ up to $3$. \textit{Third column:} Image reconstructed for the second coil when taking into account its sensitivity. \textit{Fourth column:} $k$-space of the second channel at different undersampling rates. } \label{fig:parallel}
\end{figure}

A first step in this direction is the use of \textit{parallel imaging}. This technique relies on a \textit{multicoil} MRI acquisition, which was introduced at the beginning of the 1990s by \cite{roemer1990nmr}. The idea was to use several receive coils positioned at different locations instead of a single large one. The coils record different signal intensities depending on their spatial location, which enables to achieve a larger signal to noise ratio (SNR) compared to a single coil acquisition.
It was subsequently observed that instead of increasing the SNR compared to a single coil acquisition, exploiting the geometry of the coil array could help accelerate MRI by resolving the aliasing due to spacing out successive k-space phase encodes by a factor $R$ known as the \textit{acceleration factor} \citep{sodickson1997simultaneous}. This is what is known as \textit{parallel imaging}. However, this acceleration introduces a tradeoff between acceleration and SNR as increasing $R$ will reduce the SNR by a factor $\sqrt{R}$. Reconstructing the image becomes also non-trivial, as the images from each coil have a decreased field of view (FoV), resulting in aliasing when extended to the full FoV, as illustrated on Figure \ref{fig:parallel}.



Various approaches have tackled this challenge, but they generally rely on some patient-dependent calibration. For instance, SENSE\footnote{Sensitivity encoding} \citep{pruessmann1999sense} considered \textbf{coil sensitivity maps}, which describe how each coil responds to a signal at each location of the FoV. They are typically estimated from data from a prescan \citep{blaimer2004smash}. GRAPPA\footnote{Generalized Autocalibrating Partially Parallel Acquisitions} \citep{griswold2002generalized} uses a small part of the signal near the center of k-space, called autocalibration signal, in order to compute how the data from different \textit{channels} (coils) should be combined. Several reviews \citep{blaimer2004smash, glockner2005parallel,deshmane2012parallel} examine the differences and similarities of these techniques in greater detail and detail various clinical applications. Calibrationless methods have also been explored in more recent years \citep{trzasko2011calibrationless,majumdar2012calibration,el2019calibrationless}.

\textbf{An example of parallel imaging with coil sensitivities.} We illustrate the role of sensitivity maps to provide additional information to resolve the aliasing due to the acceleration in Figure \ref{fig:parallel}. This example assumes three coils with box-like sensitivities covering each a third of the image.
We see that in the case $R=1$, there is no aliasing, and reconstruction is simply achieved by an inverse Fourier transform. In the case $R=3$. We see that adding the information of each coil allows to still resolve the aliasing, but this is not sufficient for $R=4$, where the three coils do not provide enough information in order to resolve aliasing. We can see also on the last column how increasing $R$ results in acquisition lines getting more spread out in Fourier space.

Note that this is an idealized example, as in this case the coil-sensitivities do not overlap, which means that summing up individual images multiplied by their respective sensitivities suffices to reconstruct the image. The SENSE algorithm is more involved and can resolve images with overlapping coil sensitivities.

\subsection{Dynamic MRI}
Before terminating this chapter, we will also cover the case of dynamic MRI, which will be relevant to Chapter \ref{chap:lbcs} of this thesis.  Dynamic MRI, as its name suggests aims at imaging dynamic, moving objects. It can be used for instance for cardiac imaging \citep{tsao2003k} or vocal tract imaging \citep{echternach2010vocal}. The data are acquired in k-space at different times, resulting in k-t raw data, and the direct reconstruction is done with a frame by frame inverse Fourier transform.

In this setting, imaging speed is most critical for several reasons: if one wants to image a fast-moving object precisely, then a high temporal resolution will be required, which presents a challenge for MRI \citep{tsao2003k}. Accelerating the acquisition in this case will not only improve the comfort of the patient, but might also provide higher temporal resolution.

However, the motion between two frames should not be too large, and in practice, which would require breath-held examinations for Cartesian cardiac imaging \citep{gamper2008compressed, jung2009k}, which is often impractical for patients. In this case, techniques such as radial sampling can be greatly beneficial, as radial spokes continually sample the center of the k-space, and then motion can be retrospectively corrected in free-breathing samples \citep{winkelmann2007optimal, uecker2010real, feng2014golden, feng2016compressed}.

If one wishes to perform Cartesian sampling in this case, higher acceleration rates would be required, and parallel imaging might not be sufficient to resolve aliasing. As a result, one might use a priori structure in the images to constrain the reconstruction of the image and obtain a good quality outcome. This is known as Compressed Sensing \citep{candes2006robust,donoho2006compressed}, and will be discussed in Chapter \ref{chap:rec}.

\section{Mathematical description}
We close this chapter by introducing the mathematical framework that will be used in the sequel. So far, we have described our problem in a continuous fashion, where the signal is a function of a continuous time $t$ or k-space location $(k_x, k_y)$. However, in a Cartesian acquisition, the data can be represented on a finite grid as a discrete set of values, depending on the total number of readouts. Then, within each readout, samples will be recorded at a time interval $\Delta t$, and so the resolution of k-space can be described as
\begin{align*}
    \Delta k_x = \gbar G_x \Delta t \\
    \Delta k_y = \gbar \Delta G_{\text{PE}} \tau_{\text{PE}}
\end{align*}
Note that in order to guarantee an image without aliasing, $\Delta t $ and $\Delta G_{\text{PE}}$ must be sufficiently small, and their precise value is dictated by the famous Nyquist-Shannon sampling theorem. Assuming a total of $N_y$ phase encoding lines with each continuing $N_x$ samples, we can represent the signal $S(k_x,k_y)$ as a matrix $\mS \in \mathbb{C}^{N_x \times N_y}$, which can be written in a vectorized form $\vs \in \mathbb{C}^P$, where $P=N_x \cdot N_y$.

In the Cartesian case, the spin density $\rho(x,y)$ will be obtained by computing an inverse Fourier transform on $\vs$. It is common for the resulting image to be described as $\boldsymbol{\rho} \in \mathbb{C}^P$, with the same dimensions as $\vs$. The inversion operation is performed by the inverse Fourier transform $\mathcal{F}^{-1}$, applied to $\vs$, namely
\begin{equation}
    \boldsymbol{\rho} = \mathcal{F}^{-1}(\vs).
\end{equation}
This operation is often represented as matrix vector multiplication, where the operator $\mathcal{F}^{-1}$ is explicited as the FFT matrix $\mF^{-1}$, and in addition, one can model the imperfections in the measurement as additive Gaussian noise $\bepsilon \in \mathbb{C}^P$, which yields
\begin{equation}
    \boldsymbol{\rho} = \mF^{-1} \vs + \bepsilon. \label{eq:ifft}
\end{equation}
Note that in practice, Equation \ref{eq:ifft} is not evaluated explicitly, but rather the inverse Fourier transform is directly computed using the Fast Fourier Transform (FFT) algorithm.

\subsection{Accelerated MRI}\label{s:acc_MRI}
In the case of accelerated MRI, the full k-space is not observed, and in practice, entire phase encoding lines are skipped at once, resulting in an acceleration of the scanning time proportional to the number of lines removed. We represent this operation as masking our unobserved coefficients, which is achieved by defining a diagonal masking matrix $\mP_\omega$ that masks out coefficients indexed by the set $\omega$, namely $\mP_{\omega,ii} = 1$ if $i\in \omega$ and $0$ otherwise. $\omega$ is defined as the sampling mask and has cardinality $|\omega| = N$. $\omega$ is typically structured so that entire lines are skipped at once, and a formal presentation of this structure will be discussed in Chapter \ref{chap:lbcs}. We refer to the ratio $N/P$ as the \textit{sampling} rate. This masking results in observing the partial image
\begin{equation}
    \boldsymbol{\rho} = \mF^{-1} \mP_\omega \vs + \bepsilon \label{eq:acceleratedMRI1}
\end{equation}

\begin{remark}[Sampling mask and trajectories]
    There is a difference between a sampling mask and a sampling trajectory. The former only provides a fixed view of all locations acquired, while the latter comprises a \textit{dynamic} component. The sampling trajectory tells not only what locations were acquired, but also the temporal ordering of the acquisition, detailing the order in which the samples are acquired throughout the multiple acquisition steps.
\end{remark}

In the sequel, we will consider problems where we aim at learning how to reconstruct data from partial Fourier measurements, and how to optimize sampling masks to obtain the best reconstruction quality. In order to be aligned with the commonly used terminology, we will need to redefine some of the notation used until here. We assume that there exists some underlying ground truth image $\vx \in \mathbb{C}^P$ from which we acquire partial Fourier measurements (or observations) $\vy_\omega$ that can be noisy, i.e.
\begin{equation}
    \yo = \mAo \vx + \bepsilon = \Po \mF \vx + \bepsilon.\label{eq:acquisition}
\end{equation}
This will be our basic acquisition model for the whole of this thesis. $\mAo$ is defined as a shorthand for $\Po \mF$. Note that Equations \ref{eq:acceleratedMRI1} and \ref{eq:acquisition} can be related by setting $\vs = \mF\vx$ and $\boldsymbol{\rho} = \mF^{-1} \yo$. This reformulation allows to describe the image $\vx$ as the fundamental unknown, and the partial measurements to be described in the Fourier domain.

\begin{remark}
    In the case of parallel or multicoil acquisition, Equation \ref{eq:acquisition} is slightly adapted to account for acquisition on $j=1,\ldots,C$ coils with different spatial sensitivities
    \begin{equation}
        \vy_{\omega,j} = \mAo \mS_j \vx + \bepsilon_j = \mP_{\omega} \mF \mS_j \x+ \bepsilon_j, \label{eq:acquisition_parallel}
    \end{equation}
    where the coil sensitivities can be represented as a $P\times P$ diagonal matrix where $S_{j,ii}$ describes the spatial sensitivity for the $i$-th pixel. We will also sometimes  use the compact notation
    \begin{equation}
        \yo = \mAo \mS \vx + \bepsilon\label{eq:acquisition_parallel_compact}
    \end{equation}
    for multicoil MRI, and in this case, $\yo \in \mathbb{C}^{C P}$ will represent the stacked observation, $\mAo$ will be the coil-wise Fourier transform, $\mS \in \mathbb{R}^{CP\times P}$ will be the stacked sensitivities.
\end{remark}

Recovering the ground truth image $\vx$ from partial measurement $\yo$ is challenging, as the problem of recovering an image from partial measurements is fundamentally ill-posed: there is mathematically infinitely many solutions that are consistent with $\yo$, but they are not physically meaningful. Such problems are often referred to as ill-posed \textit{inverse} problems, where we aim at recovering a reference image from partial information that entail a loss of information. In this context, \Eqref{eq:acquisition} is referred to as a \textit{forward} model. In the next chapter, we will discuss the approaches that have been studied in the literature to construct an estimate $\hat{\vx}$ of $\vx$ from partial measurements $\vy.$



\cleardoublepage
\chapter{Reconstruction methods}\label{chap:rec}

In this chapter, we consider the solutions that have been explored to the problem of \textit{accelerated MRI} in the literature. We have defined this problem formally in Section \ref{s:acc_MRI}: we consider partial measurements $\yo$ obtained through the forward model of Equation \ref{eq:acquisition}, and aim at constructing an estimate $\hat{\vx}$ of the underlying, unknown ground truth $\vx$.

The difficulty of this problem resides in its ill-posedness: there exist infinitely many solutions consistent with the observations that are not physically meaningful. To counter it, it is necessary to impart additional information to the problem in order to reach a sensible solution. This was the approach taken in  Compressed Sensing (CS) applied to MRI \citep{lustig2008compressed}. The ill-posedness of the problem is alleviated by imposing additional \textit{structure} on it. All subsequent approaches rely on imposing some form of structure to estimate $\x$ from $\y$, but have relied on increasingly complex, data-driven approaches.

We will trace back the evolution from CS approaches, that used simple mathematical models to obtain structure, to recent deep learning-based approaches that construct their models from training data: we moved from model-driven approaches to data-driven approaches. A more exhaustive presentation of these approaches can be found in \citet{ravishankar2019image, doneva2020mathematical}.

\section{Model-driven approaches}

\subsection{Compressed Sensing}
It has been long known that a band-limited signal can be exactly reconstructed from a set of uniformly sampled frequencies, provided that their density if at least twice the highest frequency of the original signal. This result is famously known as the Nyquist-Shannon sampling theorem. However, sampling at the Nyquist rate can remain expensive, and there has been extensive research to achieve sub-Nyquist rate sampling and reconstruction. In their seminal works, \citet{donoho2006compressed,candes2006robust} introduced the formalism of Compressed Sensing (CS), where they proved that by leveraging the structure present in natural images, such as sparsity or low-rankness, one can perfectly reconstruct a signal sampled much below the Nyquist rate. Achieving this required using a non-linear reconstruction method, framed as the following prototypical problem
\begin{equation}
    \xh = \argmin_\x \frac{1}{2}\|\mAo \x - \yo\|_2^2 + \lambda \mathcal{R}(\x)\label{eq:cs}
\end{equation}
where the first term, known as \textit{data-fidelity}, enforces consistency with the measurement, and the second term, called \textit{regularization}, defines a statistical model, capturing the desired structure of the signal. In the literature, $\mathcal{R}(\x)$ has taken various forms, but typical models include sparsity in the wavelet domain, $\mathcal{R}(\x) = \|\mW \x\|_1$ \citep{candes2006robust,donoho2006compressed,lustig2007sparse}, or sparsity with Total Variation (TV) kind of constraints \citep{block2007undersampled,knoll2011second}. An alternative approach would rather impose a low-rank structure on the resulting image, typically using $\mathcal{R}(\x) = \|\x\|_*$ \citep{lingala2011accelerated}, or more sophisticated structures such as the Hankel matrix \citep{jin2016general}. The common trend in all of these methods is the idea that \textit{the structure of the signal implies a redundancy that can be exploited to represent the signal in a compact fashion} when transforming it appropriately. As expressing the structure in a mathematical fashion is not a straightforward task, these methods are the expression of various attempts to capture this idea.

Note that all these approaches typically introduce a trade-off between acquisition speed (by lowering the number of k-space lines acquired) and reconstruction speed. Solving \Eqref{eq:cs} requires using an iterative method, and while the computation is carried out offline, the reconstruction can last up to hours for specific settings with large accelerations \citep{feng2016xd}.

\begin{figure}[ht!]
    \centering
    \begin{subfigure}[b]{0.2\textwidth}
        \centering
        \includegraphics[width=\textwidth]{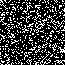}
        \caption{}\label{fig:maska}
    \end{subfigure}
    \begin{subfigure}[b]{0.2\textwidth}
        \centering
        \includegraphics[width=\textwidth]{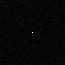}
        \caption{}\label{fig:conva}
    \end{subfigure}
    \begin{subfigure}[b]{0.2\textwidth}
        \centering
        \includegraphics[width=\textwidth]{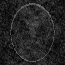}
        \caption{}\label{fig:phantoma}
    \end{subfigure}

    \begin{subfigure}[b]{0.2\textwidth}
        \centering
        \includegraphics[width=\textwidth]{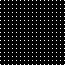}
        \caption{}\label{fig:maskb}
    \end{subfigure}
    \begin{subfigure}[b]{0.2\textwidth}
        \centering
        \includegraphics[width=\textwidth]{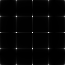}
        \caption{}\label{fig:convb}
    \end{subfigure}
    \begin{subfigure}[b]{0.2\textwidth}
        \centering
        \includegraphics[width=\textwidth]{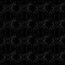}
        \caption{} \label{fig:phantomb}
    \end{subfigure}

    \caption{Random \textit{(a)} and regular \textit{(d)} sampling masks, acquiring $25\%$ of the total locations, and their corresponding point spread functions (PSF) in \textit{(b)} and \textit{(e)}. The convolution of the PSF with the Shepp–Logan phantom is shown respectively in figures \textit{(c)} and \textit{(f)}. We see that the random undersampling produces incoherent or noise-like aliasing, which still allows us to distinguish the original image. On the other hand, the aliasing produced by regular, grid-based sampling (known as fold-over artifacts) yields a low-intensity result where the original image cannot be distinguished.} \label{fig:aliasing}
\end{figure}

We note also for completeness that CS relies not only on leveraging additional structure in the image to be reconstructed, but also on incoherent sampling \citep{lustig2008compressed}. In practice, this often involves random-like sampling strategies, which enable the aliasing to be incoherent, i.e. to feature a noise-like structure, as illustrated on Figure \ref{fig:aliasing}. We will defer a more complete discussion of sampling to the next chapter.

Overall, it is clear that such approaches rely on \textit{statistical} models, constructed to implement the designer's assumptions, rather than being learned from data. While such methods enabled great progress in the last decade, the statistical models restrict the expressivity of the approach: they express abstract mathematical structures encoding our assumptions rather than focusing on the real data distribution expressed through the training set. This is why we will refer such approaches as \textit{model-driven}.

\section{Data-driven approaches}
Many approaches tried to learn their model directly from data, and in this Section, we survey a couple of the most significant representatives of data-driven methods.

\subsection{Dictionary learning}
Dictionary learning approaches rely on the assumption that an image can be represented as a sparse combination of atoms from a dictionary, i.e. $\vx =\mD\vz$ where $\mD$ is a dictionary and $\z$ the sparse coefficient vector. This is sometimes referred to as a synthesis formulation, as opposed to \ref{eq:cs}, which shows an analysis formulation. Synthesis models are often applied patch-wise to images, and they aim at simultaneously learning a reconstruction $\vx$, a dictionary $\mD$ and an encoding $\mZ$ \citep{ravishankar2011mr,caballero2014dictionary}
\begin{equation}
    \begin{gathered}
        \min _{\vx, \mD, \mZ} \frac{1}{2}\|\mAo\vx-\yo\|_{2}^{2}+\beta \sum_{j=1}^{N}\left\|\mP_j \vx-\mD \vz_{j}\right\|_{2}^{2} \\
        \text { s.t. }\left\|\vz_{j}\right\|_{0} \leq s,\left\|\vd_{i}\right\|_{2}=1, \forall i, j
    \end{gathered}
\end{equation}
\subsection{Unrolled neural networks}
Unrolled neural networks are a data-driven, deep learning-based approach inspired by the iterative algorithms used to solve \Eqref{eq:cs}. This problem is known as a composite optimization problem, and can be solved by various approaches. A famous approach is known as the Iterative Soft-Thresholding Algorithm (ISTA) \citep{daubechies2004iterative}, or Proximal Gradient method \citep{combettes2011proximal}, solves the problem by iteratively computing
\begin{equation}
    \vx^{t+1}=\text{prox}_{\lambda\mathcal{R}}\left(\vx^t-\alpha \mA^\dagger(\mA \x^t-\yo)\right).\label{eq:prox}
\end{equation}
The proximal operator is defined as $\text{prox}_{g}(\vx) \triangleq \argmin_\vv g(\vv) + \frac{1}{2}\|\vv-\vx\|_2^2$, and for some functions, can have a closed-form solution. This is typically the case when $\mathcal{R}(\vx) = \|\mW\vx\|_1$, where the proximal takes the form of a soft-thresholding operator, i.e. $\text{prox}_{
        \lambda\|\mW\cdot\|_1}(\vx) = \text{sign}(\mW\x)\odot \max(\mW\x-\lambda,0)$.

Motivated by the formulation of \Eqref{eq:prox}, unrolled neural networks parameterize the proximal operator with a neural network $\vf(\yo,\omega; \theta)$, and then train a model of the form
\begin{equation}
    \vx^{t+1} = \vf\left(\vx^t-\alpha \mA^\dagger(\mAo \x^t-\yo); \theta_{(t)}\right)
\end{equation}
for a fixed, small number of iterations $t=1,\ldots,T$. The output of the model is thus $\vf(\yo,\omega; \theta)=\vx^T$ where $\theta = \{\theta_1,\ldots,\theta_T\}$. As the proximal term capture the statistical properties of the signal that wish to recover, parametrizing it with a neural network enables to learn a model directly from data. The model being more flexible, it allows for improved reconstruction performance over methods using statistical models.

A flurry of approaches has been motivated by unrolling iterative algorithms, and we highlight a few examples hereafter. A more complete discussion of unrolled neural networks can be found in \citet{liang2020deep}. In \citet{mardani2018neural}, the parameters of the proximal mapping can be shared across iterations, or different at each iteration \citep{schlemper2017deep}. The unrolling can leverage the structure of a different optimization algorithm, like the Alternating Direction Method of Multipliers (ADMM) in \citet{sun2016deep}, or the Primal-Dual Hybrid Gradient in \citet{adler2018learned}. In the noiseless case, it is also common to replace the inner step $\vx^t-\alpha \mA^\dagger(\mAo \x^t-\yo)$ with a step called \textit{Data Consistency} (DC) \citep{schlemper2017deep,zhang2019reducing}, that replaces the reconstruction with the exact values at observed locations. Essentially, the inner step becomes
\begin{equation}
    \text{DC}(\vx^t,\yo)=\mF^\dagger (\mP_{\omega^C}\mF \vx^t + \yo) = \mF^\dagger (\mP_{\omega^C}\mF \vx^t + \mP_{\omega}\mF\x).\label{eq:DC}
\end{equation}
We see that at the observed locations $\omega$, we simply keep the observation, while at the non-observed locations $\omega^C$. Note that here, $\mP_{\omega^c}=\mI-\Po$.

These unrolled approaches require training the models $f_\theta$ on a dataset of reference images and observations $\{\vx_i,\vy_i\}_{i=1}^{m}$ by minimizing a loss $\ell(\cdot,\cdot)$:
\begin{equation}
    \theta^* = \argmin_\theta \frac{1}{m}\sum_{i=1}^m \ell(\vx_i,\vf(\y_i;\theta)).
\end{equation}
Various losses have been used in the literature. Initial works used $\ell_2$ loss \citep{schlemper2017deep,hammernik2018learning}, but lately practitioners have turned to using $\ell_1$ loss, SSIM or compounds of these \citet{muckleyStateoftheArtMachineLearning2020}. $\ell_1$ was indeed found to outperform $\ell_2$ on vision tasks \citet{zhao2016loss}. The structural similarity (SSIM) \citep{wang2004image} is a metric developed to match more closely the human perception, and recent work have proposed that a weighted sum between $\ell_1$ loss and the multiscale SSIM (MS-SSIM) as a compromise \citep{zhao2016loss}.


Deep learning-based reconstruction methods further shift the burden of computation, compared to compressed sensing approaches. While these enabled faster acquisition at the cost of a slower, iterative reconstruction procedures, deep learning-based methods retain the benefit of enabling sub-Nyquist sampling, but also enable fast reconstruction, in the order of $10^{-1}$ second for a slice \citep{jin2017deep,hammernik2018learning}. However, the computational burden is only shifted an additional step: these methods need to be trained with large, fully-sampled\footnote{In their recent work, \citet{yaman2020self} proposed approaches to train only using undersampled data, but having access to a dataset of undersampled images is still required.} datasets \citep{zbontarFastMRIOpenDataset2019}.

\subsection{Deep image prior}
In a spirit closer to dictionary learning and classical CS, \citet{ulyanov2018deep} have shown that untrained CNN can provide a sufficiently strong prior to enable image reconstruction, even without training data. Given an untrained CNN $\vf_\theta$ a single measurement instance $\yo$, the idea in deep image prior (DIP) is simply to minimize
\begin{equation}
    \min_\theta \|\yo - \mAo \vf_\theta(\vz)\|_2^2
\end{equation}
where $\vz$ is a fixed, randomly sampled vector, and where the parameters $\theta$ are randomly initialized. Then, using an iterative optimization algorithm, fitting the weights to the observations enables the network to output a high-quality reconstructed image without supervision or training data. This method was successfully applied to MRI in  \citep{darestani2021accelerated,yoo2021time}. A common pitfall of DIP is that this flexibility comes at the price of an increase inference time: the network weights must be retrained for each individual image. In addition, DIP can suffer from overfitting and so heuristics such as early stopping are used to counteract this \citep{ulyanov2018deep,sun2021plug}.

\subsection{A Bayesian interpretation of reconstruction}\label{ss:bayesian_recon}
So far, the discussion has been framed as moving from mathematically-defined to data-driven regularization in order to improve the recovery of Problem \ref{eq:cs}. However, this approach can also be framed from a Bayesian perspective, where the data consistency term $ \frac{1}{2}\|\mAo \x - \yo\|_2^2$ is viewed as a likelihood and the regularization $ \lambda \mathcal{R}(\x)$ as a prior. This was initially discussed by \citet{ji2008bayesian} for compressed sensing. The idea revolves around Bayes rule, where the posterior $p(\rvx|\rvy)$ can be expressed as
\begin{equation}
    p(\rvx|\rvy) = \frac{\overbrace{p(\rvy|\rvx)}^{\text{likelihood}}\overbrace{p(\rvx)}^{\text{prior}}}{\underbrace{p(\rvy)}_{\text{marginal}}} \propto p(\rvy|\rvx) p(\rvx).\label{eq:bayes}
\end{equation}
Under Gaussian noise, we can derive an explicit likelihood model for \Eqref{eq:acquisition}, where, assuming $\bepsilon \sim \mathcal{N}(0,\sigma^2\mI)$, i.e assuming additive Gaussian noise, we get $p(\rvy|\rvx) = \frac{1}{(2\pi\sigma^2)^{P/2}}\exp\left(-\frac{1}{2\sigma^2}\|\rvy - \mAo \x\|_2^2\right) \sim \mathcal{N}(\rvy;\Po \mF \x,\sigma^2\mI)$. The regularization term can then be viewed as a prior from which the ground truth $\rvx$ is sampled, in the case of classical CS, the prior can we specified as the Laplace density function $p(\rvx) = \left(\frac{\lambda}{2}\right)^P \exp\left(-\lambda\|\mW\rvx\|_1\right)$ \citep{ji2008bayesian}. Such an approach enables to view Problem \ref{eq:cs} as a maximum a posteriori (MAP) estimation, where one aims to find the reconstruction $\xh$ that maximizes posterior probability
\begin{align*}
    \xh & = \argmax_\rvx p(\rvx|\rvy) = \argmax_\rvx \frac{p(\rvy|\rvx) p(\rvx)}{p(\rvy)}  &  & \text{by definition}                                          \\
        & =  \argmax_\rvx p(\rvy|\rvx) p(\rvx)                                             &  & \text{as the optimization is indep. from } \rvy               \\
        & = \argmin_\rvx -\log\left(p(\rvy|\rvx) p(\rvx)\right)                            &  & \text{by taking the} \log \text{ and negative of the problem} \\
        & = \argmin_\rvx \frac{1}{2\sigma^2}\|\rvy - \mAo \x\|_2^2 +  \lambda\|\mW\rvx\|_1 &  & \text{explicit the functions and discard the constants}       \\
\end{align*}
The Bayesian perspective gives a different interpretation of reconstruction and of the role of regularization as a prior, and lends itself naturally to an extension to data-driven models, which act as learned prior models. The Bayesian framing of Equation \ref{eq:cs} has also the advantage of naturally defining a criterion to acquire observations that have large uncertainty according to the model \citep{ji2008bayesian}. We will discuss these implications in greater depth in the next Chapter as well as in Chapter \ref{ch:gans}. For now, we turn to our last category of deep learning-based approaches which make use of this Bayesian insight and attempt to explicitly learn meaningful, data-driven priors.

\subsection{Generative approaches}\label{ss:generative}
The role of the reconstruction methods can be seen as learning a data-driven prior which has proven to enable higher quality of reconstruction than mathematically motivated models. While reconstruction methods embed the full iterative resolution of Problem \ref{eq:cs} into a deep architecture, other approaches, relying on deep generative models, propose to explicitly parametrize the prior with a neural network \citep{bora2017compressed, jalal2021robust, patel2021gan}. This has the advantage of keeping the optimization procedure more transparent, as it will be about explicitly finding a sample from a deep generative prior that is consistent with the observation. The approach, proposed initially by \citet{bora2017compressed} relies on training a generative adversarial network (GAN) to capture the prior distribution. We first briefly introduce GANs.

\textbf{Generative adversarial networks.} GANs \citep{goodfellow2014generative} are a specific type of deep learning models where a generator $\vg_\theta(\vz)$ and a discriminator $d_\phi(\vx)$ are trained in a competitive fashion: starting from random noise $\rvz \sim p(\rvz)$, the generator tries to construct samples that imitate a distribution of references images $p(\rvx)$ from which we are given a large set of samples $\{\vx_i\}_{i=1}^m$. The discriminator, on the other hand, tries to classify a sample based on how likely it is to originate from the reference distribution. The problem can formally be described as a game between the generator and the discriminator, where we solve
\begin{equation}
    \min_{\theta} \max_{\phi} \mathbb{E}_{p(\rvx)}\left[d_\phi(\rvx)\right] - \mathbb{E}_{p(\rvz)}\left[d_\phi(\vg_\theta(\rvz))\right].\label{eq:gan}
\end{equation}

\textbf{Generative priors.} Given a trained generator $\vg_{\theta^*}(\rvz)$ approximating the image distribution of interest $p(\rvx)$, the approach of \citet{bora2017compressed}, applied also to MRI in \citet{patel2021gan}, aims at finding a sample from the generator consistent with the observations by solving
\begin{equation}
    \min_\vz \|\mAo \vg_{\theta^*}(\vz)-\yo\|_2 \label{eq:generative_priors}
\end{equation}
While this approach does not enable to directly construct a maximum a posteriori (MAP) estimate, due to the GAN only yielding samples and no probabilities, the low-dimensionality of the latent space $\rvz$ enables to efficiently carry out Markov Chain Monte Carlo (MCMC) computations \citep{patel2021gan} to construct posterior samples.


\section{Trade-offs}
While all the methods presented in this chapter enable reconstruction from sub-Nyquist sampling rates, they feature different advantages and drawbacks. 
The state-of-the-art in terms of reconstruction performance is occupied by deep learning-based reconstruction methods, such as unrolled algorithms. They allow for the best quality and fastest inference among all methods presented here, but require large training datasets and are not robust to distribution shifts \citep{jalal2021robust,darestani2021measuring}. Generative priors also require large amounts of data to train the generative model, and generally perform less well than trained reconstruction networks, while also having a slower inference time \citep{darestani2021measuring}. They have the advantage however of being robust to some distribution shifts like change in the distribution of masks $\omega$ and their inference is more interpretable than deep learning methods, as the conditioning on the observations is explicitly computed from the prior, rather than through a black box like in an end-to-end reconstruction algorithm.

Compressed sensing (CS), dictionary learning (DIL) and deep image prior (DIP) are generally slower at inference, as they all rely on iterative reconstruction, and require the optimization of increasingly large amounts of variable: from the image in CS to the image and dictionary in DIL to the parameters of a deep network in DIP. However, DIP is found to generally outperform CS-based approaches. The trade-offs between DIL, DIP and Generative priors is however less clear, as these methods have not been extensively compared in the literature.

Overall, this survey highlights a couple of the general trends observed in the literature and is not exhaustive. It shows however a clear trend where different methods try to exploit the redundancy due to the structure in the data. Earlier works attempted to explicitly embed this structure by compactly representing the signal in a variety of different bases, such as Wavelet domain \citep{lustig2007sparse}, or through representing the signal as low-rank  \citep{lingala2011accelerated}. Later, approaches aimed rather at \textit{learning} how to leverage this structure from data, and use the powerful tools of deep learning to represent the signal in complex ways, following a data-driven paradigm. All these methods rely however on the idea that the structure of the signal allows to alleviate the ill-posedness of the inverse problem, and retrieve a signal from undersampled observations, because of its underlying structure.

In the next chapter, we will detail the evolution in mask designs $\omega$ used to generate the observation and how they have been optimized in order to provide the most informative observations to enable the best reconstruction downstream.


\cleardoublepage
\chapter{Optimizing sampling patterns}\label{ch:sampling}
In the previous chapter, we discussed the evolution of reconstruction methods for accelerated MRI, where a shift from model-driven approaches to data-driven approaches was observed. A similar shift can be observed in the design of sampling patterns.

\section{A taxonomy of mask design methods}
While the evolution of reconstruction methods can generally be described along the axis of moving from model-based approaches towards learning-based ones, the picture is less clear for the mask design methods. It is nonetheless to consider two axes of discussion that run in parallel to the evolution witnessed in reconstruction methods.

These two axes pertain to the two main components needed to optimize sampling patterns (or masks or trajectories). One needs \textbf{i)} a prior distribution from which candidates patterns are drawn, and \textbf{ii)} a criterion, a metric to quantify its performance. We will see that at the onset of compressed sensing, sampling patters were drawn from heuristically designed parametric distributions, a setting widely known as \textit{variable density sampling}, and gradually moved towards distributions tailored for a specific dataset. Similarly, the criteria used to establish the quality of a mask initially were mathematical quantities such as \textit{coherence}, but the criteria became increasingly related to the reconstruction performance enabled by subsampling data using a given mask. One can clearly see a shift from \textbf{i)} \textit{model-based} approaches towards \textit{learning-based} ones, and from \textbf{ii)} \textit{model-driven} criteria towards \textit{data-driven} ones.

\begin{remark}
    We chose to denote the distribution from which candidate sampling patterns are drawn as the \textit{base} for the sampling optimization procedure, while the criterion used as what \textit{drives} it.
\end{remark}

In the rest of this chapter, we will survey the literature from the first application of compressed sensing to MRI onwards to show how mask design evolved from model-based, model-driven approaches towards learning-based, data-driven approaches. We will conclude by formalizing the problems linked with optimizing mask design in this context.


\section{Compressed sensing and MRI: Model-based, model-driven approaches}
\subsection{The contribution of \texorpdfstring{\citet{lustig2007sparse}}{Lustig et al. (2007)}}
With the onset of Compressed Sensing (CS), random sampling patterns were used, as they allowed to obtain noise-like, incoherent aliasing, as illustrated on Figure \ref{fig:aliasing}. This was indeed theoretically motivated, as random sampling easily enables to achieve incoherent sampling, which in turn meant that a small number of measurements were sufficient to exactly solve the problem \ref{eq:cs} for $\mathcal{R}(\vx) = \|\mW\vx\|_1$ with high probability.

\begin{definition}
    Let $(\mA, \mW)$ be two orthonormal bases of $\mathbb{C}^p$, where $\mA$ is the observation domain, and $\mW$ is the sparsifying (representation) basis. Then, the \textbf{coherence} between these bases is defined as \citep{candes2008introduction}
    \begin{equation}
        \mu(\mA,\mW) = \sqrt{p} \cdot \max_{1\leq i,j \leq p} |\langle \va_i, \vw_j \rangle|.
    \end{equation}
    The coherence measures the maximal correlation between any element of $\mA$ and $\mW$. $\mu(\mA,\mW) \in [1, \sqrt{p}]$, where coherence $1$ implies maximal incoherence between the bases, and $\sqrt{p}$ maximal coherence.
\end{definition}

Coherence then allows to bound the minimal number of measurements $N$ required for exactly solving \ref{eq:cs} with high probability. Indeed, for a given signal $\vx \in \mathbb{C}^p$ observed in domain $\mA$, assuming that $\vx$ is $S$-sparse in the transformed domain $\mW$, then a typical compressive sensing result \citep{candes2007sparsity} dictates that, given $N$ \textit{random} measurements, exact reconstruction will occur with high probability when
\begin{equation}
    N \geq C \cdot S\cdot \mu^2(\mA, \mW) \cdot \log P \label{eq:n_measurement}
\end{equation}
for some positive constant $C$.

While the theory of CS obtains results for \textit{random} measurements, practitioners have quickly sought to optimize the measurements, and a natural initial criterion has been to minimize the coherence between measurements, i.e. to select a sampling pattern $\omega$ with $|\omega| = N$ that enables to minimize $\mu(\mAo, \mW)$ \citep{lustig2007sparse}. However, when applying compressed sensing to MRI, it was quickly observed that purely random undersampling of the Fourier space is not practical, for several reasons \citep{lustig2007sparse,lustig2008compressed}: \textbf{i)} it does not satisfy hardware and physical constraints that require smooth trajectories in Fourier space and the use of trajectories robust to system imperfections, \textbf{ii)} it does not consider the energy distribution of the signal in k-space, which is very concentrated around the center 

For this reason, the seminal work of \citet{lustig2007sparse} proposed to use a variable-density sampling approach, where the k-space is less undersampled near the origin (low frequency, high energy region) and more towards the periphery (high frequency but low energy)\footnote{The name \textit{variable-density sampling} originates from the very fact that k-space is not uniformly subsampled, but rather with a variable density \citep{tsai2000reduced}.}.  They used a distribution centered at the origin of k-space, where the probability of acquiring a location decayed at a polynomial rate as a distance of the origin, namely
\begin{equation}
    P(k_x,k_y) = \left(1-\frac{2}{\sqrt{p}}\sqrt{k_x+k_y}\right)^d \label{eq:vds_distribution}
\end{equation}
with decay $d>1$, and $-p/2\leq k_x,k_y\leq p/2$. This heuristic was proposed to imitate the decay in the spectrum observed on real data. The mask to be used is then estimated by a Monte-Carlo procedure: a set of masks are randomly sampled from Equation \ref{eq:vds_distribution}, and the mask with minimal coherence is retained.


The idea proposed by \citet{lustig2007sparse} of using variable-density sampling for MRI was not totally novel, as variable-density sampling had previously been used in the MRI community to design undersampling trajectories with incoherent aliasing when using linear reconstructions \citep{marseille1996nonuniform,tsai2000reduced}. However, \citet{lustig2007sparse} were the first to integrate this heuristic within the framework of CS.

It should be clear from the presentation that the approach of \citet{lustig2007sparse} is a \textit{model-based approach}, as the masks are sampled from a heuristic probability distribution (eq. \ref{eq:vds_distribution}) and also a \textit{model-driven} approach, as the criterion used to select the masks is \textit{coherence}, a mathematical structure prescribed by the theory of compressed sensing.

\subsection{Model-based, model-driven sampling in the sequel}
A similarly model-based, model-driven was followed by \citet{gamper2008compressed}, which considers dynamic MRI, although from a different initial mask: they attempted at randomizing a deterministic mask design proposed by \citet{tsao2003k} to achieve incoherent sampling. The \textit{prior} in this case is given by a structured mask.

Some refinements to variable-density sampling (VDS) were developed for instance in \citep{wang2009pseudo}, but in the sequel, two broad approaches were followed to improve the design of sampling patterns. The first approach leveraged learning-based approaches, and will be discussed in the next section, while the second attempted at developing theoretical models that more closely captured the structure of real-world signals.

The work of \citet{wang2009variable} is a first step in this direction, where improved VDS functions are proposed by exploiting a priori statistical information of real-world images in the wavelets domain. One can notice that this approach chose to tie the sampling pattern specifically to a type of regularization, namely sparsity in the wavelet domain, and obtains better sampling density for this problem but lose in generality over competing approaches.

The work of \citet{puy2011variable} proposes a coherence-driven optimization of the variable density sampling procedure. Instead of relying on a heuristic distribution such as \eqref{eq:vds_distribution}, the procedure proposes an objective that yields a VDS distribution that minimizes the coherence between the measurement and representation bases.

A large body of subsequent works \citep{krahmer2013stable,roman2014asymptotic,adcock2015quest,adcock2017breaking} rely on some additional theoretical tools such as \textit{local coherence}, \textit{asymptotic sparsity}, and \textit{multilevel sampling} to give a solid ground to heuristics such as VDS that show good performance but lack theoretical justification.

\section{Towards learning-based, data-driven approaches}
In parallel to these works, there was a motivation to move towards learning-based approaches, due to some weaknesses in the VDS approach of \citet{lustig2007sparse}. One of them is that the VDS distribution of \Eqref{eq:vds_distribution} is a heuristic that requires tuning to work well on different types of data, and that heuristically tuned variable density distributions generally performed better than the ones solely prescribed by CS theory \citep{chauffert2013variable}.

\textbf{Learning-based, model-driven sampling.} A first work of \citet{knoll2011adapted} aimed at solving this issue by constructing the VDS distribution from data rather than from a tuned heuristic. They simply average the spectra of several similar images, normalized it and sampled their mask from the resulting distribution, minimizing coherence to select the final mask, as in \citet{lustig2007sparse}. This approach was then improved by \citet{zhang2014energy,vellagoundar2015robust}.

\textbf{Early learning-based, data-driven approaches.} \citet{ravishankar2011adaptive,liu2012under} proposed similar methods where they aimed at designing a sampling pattern that minimized the \textit{reconstruction error} in Fourier space, rather than coherence. Both algorithms partition k-space in blocks, and iteratively design a mask by removing samples from low-error regions in order to re-assign them in high-error regions. \citet{liu2012under} made the important point that the design of a mask should reconstruction-based and not observation-based. Until then, approaches did not include the reconstruction algorithm into the design of the mask, but relied on model-based criteria relying on the observations, such as coherence. This insight proved to be impactful in the performance of the methods, and while no quantitative comparison was available in the paper, further works confirmed the impact of including the reconstruction algorithm in the mask design \citep{zijlstra2016evaluation}.

While these methods capture the paradigm that will be overwhelmingly dominant in the deep learning era, there was little follow-up work until much later, around 2018.

\textbf{The case of \citet{seeger2010optimization}.} The work of \citet{seeger2010optimization} lies in its own category. This approach builds on a sequential Bayesian experiment design perspective for sparse linear models \citep{seeger2007bayesian, seeger2008compressed}. First a model of the posterior distribution $p(\rvx|\yo)$ is built, and then, a sampling pattern is iteratively built by successively acquiring the readouts that have the highest entropy. This approach is able to sequentially tailor the sampling pattern to the need of an unseen image, given the previously acquired measurements. However, the approach features a quite high computational cost for a single image. \citet{liu2012under} raise that is not practical, as the matrix inversion required to compute the next measurement that must be acquired is too slow to match the readout time of a scanner.

While methods to optimize sequentially the sampling mask are common in the deep learning era, the approach of \citet{seeger2010optimization} is unique in its time, as all other works focused on optimizing the sampling pattern for a fixed sampling rate, and not in a sequential fashion. While more principled than other approaches, and using maximum entropy as the criterion to optimize the acquisition trajectory, this method remains model-based (using a Laplace prior) and model-driven (maximum entropy).

\section{A matter of performance: how do these methods compare to each other?}
Very few of the works discussed until then feature an extensive comparison of performance of their proposed approaches. Most works rely on visual comparisons, or state the weaknesses of previous methods, but do not compare against them. In this regard, the contributions that bring a comparison of various approaches to mask design are particularly valuable, and in this section, we will discuss two of them, namely \citet{chauffert2013variable} and \citet{zijlstra2016evaluation}.

The work of \citet{chauffert2013variable} aims at designing a better VDS density, while also providing a comparison between theoretically informed sampling patterns, and the ones obtained using the polynomial of \Eqref{eq:vds_distribution} tuned at various degrees $d \in \{1,2,\ldots, 6\}$, and a hybrid distribution mixing some theoretically prescribed components and heuristics. It is interesting to observe that in their data, the tuned hybrid distribution performs best, followed by the tuned polynomial and the optimal theoretically predicted distribution. This theoretically optimal distribution is based on the arguments from \citet{rauhut2010compressive} uses rejection of already acquired locations \citep{puy2011variable} and minimizes coherence, while the polynomial was tuned on the data using directly the final evaluation metric, namely PSNR. The authors conclude that the theory would need to capture additional structure of real-world images for sampling patterns obtaining better practical performance.

The article of \citet{zijlstra2016evaluation} compares the performance of coherence-based VDS using \Eqref{eq:vds_distribution}, of \citet{knoll2011adapted}, of \citet{liu2012under} and of VDS optimized to minimize normalized root mean squared error (NRMSE). The results show that coherence-based VDS is generally outperformed by \citet{knoll2011adapted}, which is matched by the optimized VDS. Both are outperformed by the method of \citet{liu2012under}. The conclusion is however careful: \blockquote{\textit{When putting the presented results in perspective, it is important, to remember that data-driven optimization of an undersampling pattern for CS is just one aspect of the CS implementation. Data-driven approaches should be taken to optimize other aspects of the CS implementation, such as the sparse domain and the reconstruction algorithm.}}

Overall, while these works do not exhaustively compare the existing methods in the literature, they give us reasons to view learning-based, data-driven methods as the most direct avenue towards improvement in mask design optimization.



\section{The advent of deep learning: Learning-based, data-driven approaches}

Interest for learning-based, data-driven approaches to mask optimization was renewed around 2018, with the advent of the first deep learning-based reconstruction methods applied to MRI \citep{sun2016deep,schlemper2017deep,mardani2018neural}. This enabled the possibility to scale up the complexity of mask design methods, as deep learning methods enable fast reconstruction times, which was not the case for CS methods.

The first work that we discuss in this section comes slightly before this time and is the one of \citet{baldassarre2016learning}, where the authors propose a first principled, learning-based approach to optimizing a deterministic sampling pattern given a set of training signals. Note that in our taxonomy, their approach is both learning-based \textit{and} data-driven. While the optimization only considers the energy of the signal in the case of Fourier sampling, and while the paper on \textit{linear} reconstruction, the authors show improvement about the sampling patterns prescribed in \citet{lustig2007sparse,roman2014asymptotic}. However, as they optimize only the sampling pattern to capture the \textit{energy} of the signal and do not include the reconstruction algorithm in their optimization procedure, they fail to obtain significant improvement by turning to non-linear reconstruction methods.

However, the authors published a sequel to this work with \citet{gozcu2018learning}, that introduced a sequential optimization of the sampling pattern in a fully data-driven fashion. The principle consists, at each step, to evaluate the improvement for a given performance metric of including a candidate location in the final sampling pattern. Once all candidate locations have been evaluated, the algorithms picks the one that brings the largest improvement and moves to the next step. This method was shown to outperform existing approaches, such as the one of \citet{lustig2007sparse} and \citet{knoll2011adapted} by a large margin, due to its approach being both learning-based and data-driven. Part of the improvement is due also to not solely adding sampling locations based on the current error, but computing how adding a location would \textit{actually} decrease the error in the reconstruction. The distinction is important as the method of \citet{gozcu2018learning} is able to take into account how the newly acquired location affects the behavior of the non-linear reconstruction algorithm, rather than acquiring locations based on the current reconstruction error. We will describe the setting of \citet{gozcu2018learning} in greater depth in Chapter \ref{chap:lbcs} as this is the basis upon which this thesis was built.

At roughly the same time as \citet{gozcu2018learning}, \citet{haldar2019oedipus} published another work aiming at optimizing sampling, based on an experiment design approach to minimize the Cramér-Rao bound (CRB). The CRB provides a lower-bound to the covariance matrix that is based on the measurement matrix $\mAo$. This approach is independent on the reconstruction algorithm, and provides a different take on a learning-based, model-driven approach, aiming at minimizing the covariance of an estimator. In the context of MRI sampling, this amounts to finding the measurement matrix that will encode the most information about the original signal $\vx$.

The last non-fully deep learning-based work that we will discuss is the one of \citet{sherry2019learning}, where a bilevel optimization procedure is proposed: the first level solves an empirical risk minimization problem for the mask, while the second level consists in an iterative algorithm to solve a Problem similar to \ref{eq:cs}. The work takes an original approach whereby relaxing the mask optimization problem into a continuous optimization problem, they are able to tackle the mask optimization and the reconstruction under a unified framework. While the reconstruction remains model-based, the mask stage is clearly learning-based and data-driven, as training data are used, and aim at minimizing some loss between the original image and the reconstruction.

Due to the overwhelming performance of deep learning methods, most subsequent works focus exclusively on deep learning-based reconstruction methods, and all follow  a learning-based, data-driven paradigm. All methods initialize the prior from which masks are drawn randomly, and aim at refining it throughout training in order to maximize some performance metric.

There are however some trends that can be highlighted within these methods. Some of them aim at \textit{jointly} train a reconstructor and a sampling pattern for some fixed sampling budget. Others tackle a problem of designing a mask \textit{sequentially}, by adding new sampling locations based on the currently acquired information. Finally, some methods aim at training a model that is able to propose a \textit{patient-adaptive} mask design. These methods often leverage tools from reinforcement learning (RL) and this is why we refer to such models as \textit{policy} models.

The works of \citet{bahadir2019learning,aggarwal2020j,weiss2020joint,huijben2020learning} relax the problem of optimizing the mask to a continuous optimization problem, and train the sampling pattern along with the neural network using backpropagation. After training, the mask is then mapped back as a boolean vector. Other approaches use a pre-trained reconstruction model and train a policy model that performs patient-adaptive sequential mask optimization using reinforcement learning \citep{pineda2020active,bakker2020experimental}. The approach of \citet{zhang2019reducing} is unique, as they aim at training an evaluator that tries to estimate the current reconstruction error for each location in k-space, while using this evaluator to inform the training of the reconstruction model. Finally, a fourth category of approaches tackles the challenging problem jointly learning a reconstruction model and a policy model for patient-adaptive sampling \citep{jin2019self,van2021active,yin2021end}.

\begin{remark}[A short discussion of non-Cartesian sampling methods]
    The problem of optimizing sampling trajectories becomes more complex when moving to the non-Cartesian realm. The problem is inherently continuous, and there is a lot of freedom in designing non-Cartesian trajectories as long as hardware constraints such as bounded first and second derivatives of the trajectory \citep{boyer2016generation}.

    As a result, there is great diversity in the methods proposed. Traditional approaches such as radial spokes \citep{lauterbur1973image} and spirals \citep{meyer1992fast} are typically \textit{model-based, model-driven}, but more recent space filling trajectories like SPARKLING \citep{lazarus2019sparkling} fall under this category. \citet{weiss2019pilot,chaithya2022hybrid} attempted to train non-Cartesian sampling and reconstruction jointly, in an end-to-end fashion, yielding \textit{learning-based, data-driven} approaches, while \citet{wang2021b} proposed to parametrize the trajectories with B-spline to facilitate learning, resulting in a \textit{model-based data-driven} method. Finally, \citep{chaithya2021learning} also proposed to refine the SPARKLING algorithm by learning from data the underlying variable-density distribution, yielding a \textit{learning-based, model-driven} approach.

    These approaches were compared in \citet{chaithya2022benchmarking}, where additional challenges due to non-Cartesian sampling are highlighted. In this setting, it appears that the algorithm used to optimize the sampling trajectory greatly matters, as it is possible for trajectories to remain stuck close to initialization. As a result, merely being a learning-based, data-driven approach does not guarantee an improved performance over model-based model-driven alternatives. Nonetheless, the best performing method was found to be the hybrid learning approach of \citet{chaithya2022hybrid}.
\end{remark}

\section{Formalizing learning-based, data-driven mask design}
Recall the observation setting described in Section \ref{s:acc_MRI}. We consider the problem of accelerated MRI, where we obtain partial information $\yo \in \mathbb{C}^P$ about a ground truth signal $\x \in \mathbb{C}^P$
\begin{equation*}
    \yo = \mAo\x + \bepsilon =\Po \mF \x + \bepsilon
\end{equation*}
where $\bepsilon$ a complex Gaussian noise vector. $\mF$  is the Fourier transform operator, and $\Po$ is our masking operator, implemented as a diagonal matrix with diagonal entries $\mP_{\omega,ii} = 1 \text{ if } i \in \omega$ and $0$ otherwise. $\omega$ denotes the sampling mask and $|\omega| = N$ denotes is cardinality.

As described in Chapter \ref{chap:rec}, we construct an estimate $\xh$ of the ground truth $\vx$ using a reconstruction algorithm $f_\theta$ such that $\xh = f_\theta(\y, \omega)$, where $\theta$ denote the parameters of the reconstruction algorithm. This can correspond to the regularization parameter of a compressed sensing method, or denote all the parameters of a neural network.

\begin{figure}[!t]
    \centering
    \includegraphics[width=\linewidth]{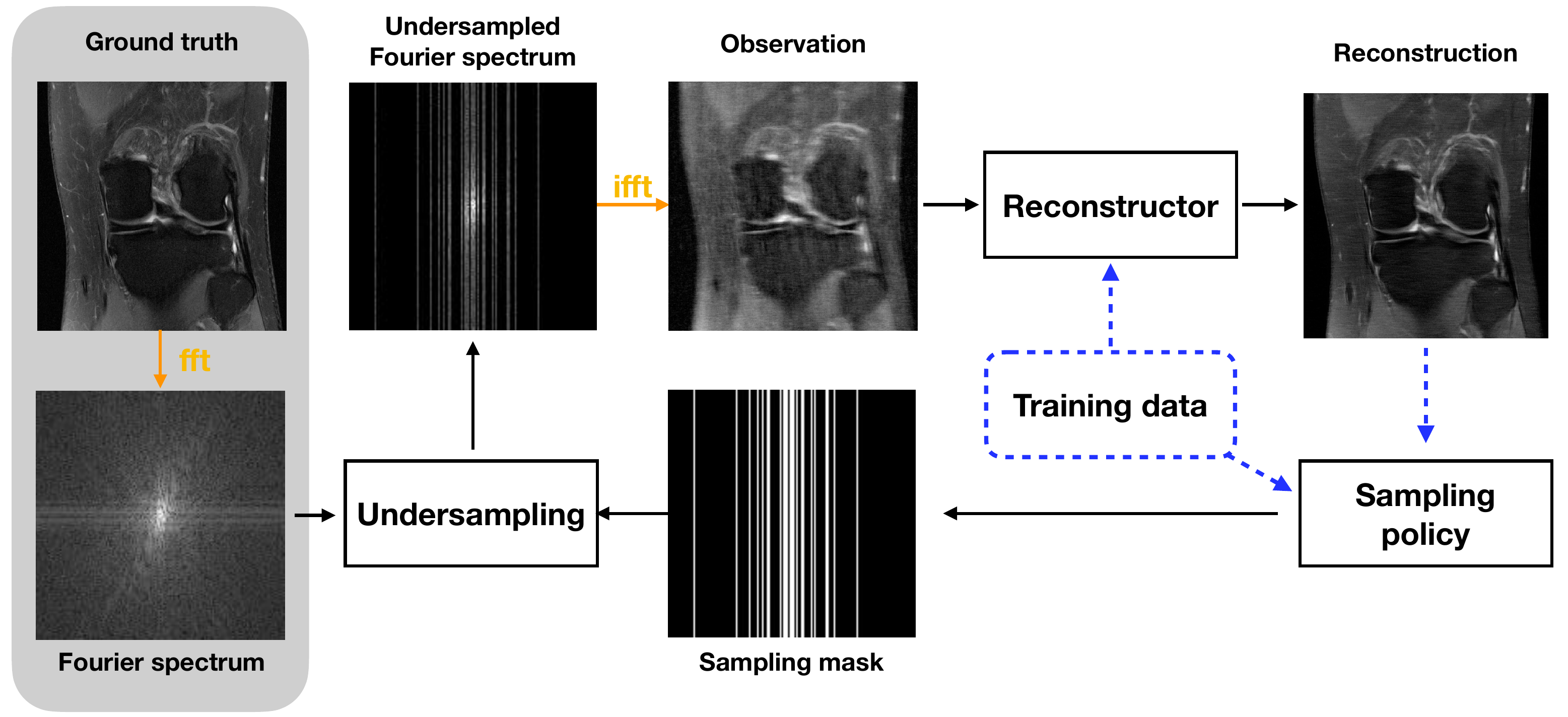}
    \caption{Overview of sampling in accelerated MRI, using a knee image. Acquisition physically happens sequentially in Fourier space (full readout lines acquired at once), and a \textit{sampling policy} can decide ahead of time what sampling mask to use (non-adaptive sampling), or integrate information from the current reconstruction to decide on the next location to acquire (adaptive sampling).\\
        Blue dashed lines indicate optional relations: not all policies and reconstruction methods rely on training data, and not all policies are patient-adaptive. The light grayed area indicates that we do not have access to the full ground truth and spectrum at evaluation time, but that we rather directly acquire undersampled data.}
    \label{fig:acc_model}
\end{figure}

A learning-based approach requires access to a data distribution $p(\rvx)$ that can give us access to reference images, while a data-driven approach requires a way to quantify how close an estimation $\xh$ is from the ground truth $\vx$. This is done by using a performance metric $\eta(\cdot, \cdot): \mathbb{C}^P \times \mathbb{C}^P \to \mathbb{R}$, which we will aim to maximize\footnote{A similar reasoning would hold if one would consider a loss $\ell(\cdot, \cdot): \mathbb{C}^P \times \mathbb{C}^p \to \mathbb{R}$ instead of a performance metric. In this case, one would then need to find the sampling mask that minimizes it.}. The process is illustrated on Figure \ref{fig:acc_model}.

An ideal sampling algorithm would tailor the mask to each instance of $\vx \sim \px$, solving
\begin{equation}
    \max_{\omega: |\omega|\leq N} \eta(\vx,\hat\vx_\theta(\vy_{\omega}=\mathbf{P}_{\omega}\mA\vx)), \label{eq:adaptive_long}
\end{equation}
However, it is evident that such an approach is infeasible in practice, because solving \Eqref{eq:adaptive_long} would require access to the ground truth image $\vx$ at the time of evaluation. This is not the case in reality, where we only have access to fully sampled training signals for training. We turn now to two main approaches to this problem that have been discussed before. They rely on either \textbf{i)} using \textit{non-adaptive} masks, designed on a training set offline and deployed at testing time or \textbf{ii)} using \textit{adaptive} masks, where a sampling \textit{policy} or \textit{heuristic} $\pi_\phi$ is trained offline and then takes patient-adaptive decision during the acquisition at evaluation time.

\subsection{Non-adaptive sampling masks}\label{sec:non-adapt}
In this first setting, given access to a set of training data $\{\vx_1, \ldots, \vx_m\} \sim p(\rvx)$, we aim at finding the mask that exhibits the best average performance, namely
\begin{equation}
    \max_{\omega: |\omega| \leq N} \frac{1}{m} \sum_{i=1}^m \eta(\vx_i, \vf_\theta(\vy_{\omega,i}; \omega)) \text{~s.t.~}\vy_{\omega,i} = \mAo\vx_i + \bepsilon.\label{eq:emp_perf}
\end{equation}
However, solving \Eqref{eq:emp_perf} remains challenging, due to the combinatorial nature of the problem. It also brings questions of \textit{generalization}, of whether a mask $\omega$ that shows good performance on the training set would also show good performance on a testing set of unseen samples from the same distribution $p(\rvx)$. While standard learning-theoretic can give answer on the generalization problem \citep{gozcu2018learning}, the combinatorial nature of the problem dooms to failure any naive approach, as the total number of different masks grows exponentially, with a complexity $O(2^N)$.

Two main types of approaches have been proposed for Problem \ref{eq:emp_perf}, relying either on tackling the combinatorial problem in an approximate fashion \citep{gozcu2018learning,zibetti2020fast} or relaxing the mask to a continuous variable that is then solved using gradient-based optimization \citep{bahadir2019learning,aggarwal2020j,weiss2020joint,huijben2020learning,sherry2019learning}. Combinatorial approaches typically tackle Problem \ref{eq:emp_perf} as a \textit{sequential} problem, starting from some initial mask $\omega_0$, and gradually growing the mask until the sampling budget $N$ is exhausted. On the contrary, \textit{continuous} approaches choose to directly optimize the mask at the maximal sampling budget $N$, a setting which we refer as \textit{fixed} sampling.

\subsection{Adaptive sampling masks}
In contrast to Equation \ref{eq:emp_perf}, adaptive sampling generally relies on a two-step procedure, where one must first train a \textit{policy} model $\pi_\phi$, and subsequently evaluate it on testing data. The problem of training the policy reads
\begin{equation}
    \max_{\phi} \frac{1}{m} \sum_{i=1}^m \eta(\vx_i, \vf_\theta(\vy_{\omega_T,i})), \text{~where~}
    \begin{cases}
        \omega_t = \omega_{t-1}\cup v_t                  \\
        v_t\sim \pi_\phi(\vf_\theta(\vy_{\omega_{t-1}})) \\
        |\omega_t| \leq N
    \end{cases} t=1,\ldots,T. \label{eq:emp_adapt}
\end{equation}
Here, $T$ describes the number of acquisition rounds that the policy does. While the problem may look similar to Problem \ref{eq:emp_perf}, there are \textit{two} fundamental differences to be noted. First, the optimization here is on the \textit{parameters} $\phi$ of the policy model, rather than on the mask itself. Secondly, the problem is inherently sequential, as the mask is gradually built from atoms $v_t$, $t=1,\ldots,T$ that are all given from the policy $\pi_\phi$ at different stages in the acquisition.

Solving the problem remains challenging, as gradient-based optimization of $\phi$ remains challenging, due to the sampling operation $v_t \sim \pi_\phi((\vf_\theta(\vy_{\omega_{t-1}}))$ preventing a direct computation of the gradients. As a result, two main approaches can be found to tackle this problem, relying either again on relaxing the mask to a continuous variable \citep{van2021active,yin2021end}, or leveraging reinforcement learning \citep{bakker2020experimental,pineda2020active, jin2019self}, which is naturally suited to consider the problem of searching the policy that yields the best performance given a changing environment. In our case, the environment is described by the unknown underlying ground truth.

The second stage of adaptive sampling is the \textit{inference} step, where the trained policy $\pi_\phi$ is then used to guide the acquisition of an unseen image $\vx_{\text{test}}$:
\begin{equation}
    v_{t+1} \in \argmax_{v \in [P]} [\pi_\phi(\vf_\theta(\vy_{\omega_t, \text{test}}))]_v \label{eq:inference_adaptive}
\end{equation}
where $[\cdot]_v$ denotes the $v$-th entry of the vector, $\omega_{t}=\omega_{t-1}\cup v_t$, for $t=1,\ldots,T$. At inference time, we see that the test image is gradually sampled at the locations $v_t$ provided by the policy and based on the reconstruction at the previous step.

\begin{remark}[Using heuristics] The inference of \Eqref{eq:inference_adaptive} is not restricted to policies trained following \Eqref{eq:emp_adapt}, but can be paired with different heuristics. For instance, \citet{zhang2019reducing} trained an \textit{evaluator} that estimates the error made by the reconstruction model at the different locations to be acquired. They then use this heuristic for sampling, by acquiring at each step the location that is estimated to have the highest reconstruction error.

    We refer to this approach as a \textit{heuristic} rather than a \textit{policy} because it is not specifically trained to estimate what would be the best location to answer next; it only trains a model that estimates the current reconstruction error, and then uses it in a greedy fashion.

    We will expand upon the use of heuristics for adaptive sampling in Chapter \ref{ch:gans}, where we propose to use the variance of a generative adversarial network \citep{goodfellow2014generative} to perform adaptive sampling.
\end{remark}

\begin{remark}[Non-sequential adaptive sampling] Although a sequential acquisition is a natural choice for adaptive sampling, it is possible to perform patient-adaptive sampling with a fixed (non-sequential) mask.

    This was done for instance by \citet{bakker2021learning}, where in the case of multi-coil MRI, the authors used the auto-calibration signal (ACS), a small set of measurements that is systematically acquired, to design a patient-adaptive fixed mask in a one-shot fashion. They inputted the measurements into a policy model that outputs a single mask for the rest of the acquisition as a result. However, their results surprisingly show that their best performing sampling policies explicitly learn to be non-adaptive, and their results provide then a state-of-the-art \textit{non-adaptive} sampling mask.
\end{remark}

\subsection{On the optimality of the discrete mask optimization problem}
In the previous sections, we defined Problems \ref{eq:adaptive_long} and \ref{eq:emp_perf} as optimization problems over masks. However, compressed sensing approaches \citet{puy2011variable, chauffert2013variable} rather optimized the probability \textit{distribution} from which masks are subsequently drawn. These arguments were also constructed in the context of model-based, model-driven methods, aiming at minimizing coherence. Can we relate the optimization of a probability distribution to the optimization of a discrete mask such as Problem \ref{eq:emp_perf}? How does the problem of finding an optimal probability distribution change when moving from a model-based, model-driven context to a learning-based, data-driven context? These are the questions that we will answer in this section.


While in \citet{chauffert2013variable}, the optimization was done on the distribution $\pi$ from which masks are subsequently sampled, in our case, the performance metric will be optimized directly for a fixed mask rather than a distribution. We will formally show below that in a data-driven context, we can construct an optimal sampling distribution $\pi$ from an optimal mask. The optimal mask describes the support of an optimal probability distribution.

We model the mask designing process as finding a probability mass function (PMF) $\pi \in S^{P-1}$, where $S^{P-1} := \{\pi \in [0,1]^P : \sum_{i=1}^P \pi_i =1\}$ is the standard simplex in $\mathbb{R}^P$. $\pi$ assigns to each location $i$ in the $k$-space a probability $\pi_i$ to be acquired. The mask is then constructed by drawing without replacement from $\pi$ until the cardinality constraint $|\omega|=N$ is met. We denote this dependency as $\omega(\pi,N)$. The problem of finding the optimal sampling distribution is subsequently formulated as \useshortskip
\begin{equation}
    \max_{\pi\in S^{P-1}} \eta(\pi),
    \qquad \eta(\pi) :=  \mathbb{E}_{\substack{\omega(\pi,N)\\ \vx \sim p(\rvx)}}\left[\eta\left(\vx, \vf_\theta\left(\vy,\omega\right)\right)\right],
    \label{eq:main}
\end{equation}
where the index set $\omega\subset [P] $ is generated from $\pi$ and $[P] := \{1,\ldots,P\}$. This problem corresponds to finding the probability distribution $\pi$ that maximizes the expected performance metric with respect to the data $p(\vx)$ and the masks drawn from this distribution. To ease the notation, we will use $\eta\left(\vx, \vf_\theta\left(\vy,\omega\right)\right) \equiv \eta\left(\vx; \omega\right)$.

In practice, we do not have access to $\mathbb{E}_{p(\vx)} \left[\eta(\vx; \omega)\right]$ and instead have at hand the training images $\{\vx_i\}_{i=1}^m$ drawn independently from $p(\vx)$. We therefore maximize the \textit{empirical} performance by solving \useshortskip
\begin{equation}
    \label{eq:emp}
    \max_{\pi\in S^{P-1}}  \eta_m(\pi), \text{~~~~} \eta_m(\pi) :=\frac{1}{m} \sum_{i=1}^m \mathbb{E}_{\omega(\pi,N)}\left[\eta(\omega,\vx_i)\right].
\end{equation}\useshortskip
Note that Problem \ref{eq:emp}, while being an empirical maximization problem, is distinct from Problem \ref{eq:emp_perf} above, as it seeks to optimize the distribution from which masks are drawn, and as a result, performs an expectation over all the masks drawn from it.

Given that Problem \ref{eq:emp} looks for  masks that are constructed by sampling $N$ times without replacement from $\pi$, the following holds.
\begin{proposition}
    There exists a maximizer of Problem \ref{eq:emp} that is supported on an index set of size at most $N$.\label{prop:1}

    {\normalfont \noindent \textit{Proof.}  Let the distribution $\widehat{\pi}_n$ be a maximizer of Problem~\ref{eq:emp}. We are interested in finding the support of $\widehat{\pi}_n$. Because $\sum_{|\omega|=N}\Pr[\omega]=1$, note that

        \begin{align}
            \max_{\pi\in S^{P-1}}  \eta_m(\pi) & := \max_{\pi\in S^{P-1}}  \sum_{|\omega|=N}  \frac{1}{m}\sum\nolimits_{i=1}^m\eta(\vx_{i};\omega) \cdot \Pr[\omega|\pi]\nonumber \\
                                               & \le \max_{\pi\in S^{P-1}} \max_{|\omega|=N} \frac{1}{m}\sum\nolimits_{i=1}^m\eta(\vx_{i}; \omega) \nonumber                      \\
                                               & = \max_{|\omega|=N}   \frac{1}{m}\sum\nolimits_{i=1}^m\eta(\vx_{i};\omega).  \nonumber
        \end{align}
        Let $\widehat{\omega}_N$ be an index set of size $N$ that maximizes the last line above.
        The above holds with equality when $\Pr[\widehat{\omega}_N]=1$ and $\Pr[\omega]=0$ for $\omega\ne \widehat{\omega}_N$ and $\pi=\widehat{\pi}_N$. This in turn happens when $\widehat{\pi}_N$ is  supported on $\widehat{\omega}$. That is, there exists a maximizer of Problem \ref{eq:emp} that is supported on an index set of size $N$. \hfill $\square$}
\end{proposition}

While this observation does not indicate how to find this maximizer, it nonetheless allows us to simplify Problem \ref{eq:emp}. More specifically, the observation that a distribution $\widehat{\pi}_N$ has a compact support of size $N$ implies the following:
\begin{proposition}
    \begin{equation}
        \text{Problem \ref{eq:emp}} \equiv  \max_{|\omega|=N}  \frac{1}{m}\sum_{i=1}^m \eta(\vx_{i}; \omega) \label{eq:greedy}
    \end{equation}\label{prop:2}

    {\normalfont \noindent \textit{Proof.} Proposition~\ref{prop:1} tells us hat a solution of Problem~\ref{eq:emp} is supported on a set of size at most $n$, which implies
        \begin{equation}
            \text{Problem \ref{eq:emp}} \equiv
            \max_{\pi\in S^{P-1}: |\text{supp}(\pi)| = N }\eta_m(\pi)\label{eq:main emp equiv 1}.
        \end{equation}
        That is, we only need to search over compactly supported distributions $\pi$. Let $S_\Gamma$ denote the standard simplex on a support $\Gamma\subset [P]$. It holds that
        \begin{align}
            \text{Problem \ref{eq:main emp equiv 1}} & \equiv
            \max_{|\Gamma|=N} \max_{\pi\in S_\Gamma}\eta_m(\pi) \nonumber                                                                                           \\
            = \max_{|\Gamma|=N}                      & \max_{\pi \in S_\Gamma} \frac{1}{m}\sum\nolimits_{i=1}^m\eta(\vx_{i};\Gamma) \cdot \Pr[\Gamma|\pi] \nonumber \\
            = \max_{|\Gamma|=N}                      & \max_{\pi \in S_\Gamma} \frac{1}{m}\sum\nolimits_{i=1}^m\eta(\vx_{i};\Gamma) \nonumber                       \\
            =  \max_{|\Gamma|=N}                     & \frac{1}{m}\sum\nolimits_{i=1}^m\eta(\vx_{i};\Gamma).
            \label{eq:main emp equiv 2}
        \end{align}
        To obtain the second and third equalities, one observes that all masks have a common support $\Gamma$ with $N$ elements, i.e. $\pi\in S_\Gamma$ allows only for a single mask $\omega$ with  $N$ elements, namely $\omega=\Gamma$. \hfill $\square$}

\end{proposition}
The framework of Problem~\ref{eq:emp} captures most variable-density based approaches of the literature that are defined in a learning-based fashion \citep{knoll2011adapted,ravishankar2011adaptive,vellagoundar2015robust,haldar2019oedipus,bahadir2019learning,sherry2019learning}, and Proposition \ref{prop:2} shows that Problem~\ref{eq:main emp equiv 2}, that we tackled in \citet{gozcu2018learning,gozcu2019rethinking,sanchez2019scalable} and develop in Chapter \ref{chap:lbcs}, also aims at solving the \textit{same} problem as these probabilistic approaches.
\todoi{Think again to what it is applicable. Does it capture most VD approaches? They are non data-driven in most cases. This might need to be revised.}

Note that while the present argument considers sampling \textit{points} in the Fourier space, it is readily applicable to the Cartesian case, where full lines are added to the mask at once.

Several additional observations and remarks can be made. The argument presented here is thus mainly a justification of the objective of \Eqref{eq:emp_perf} that we use for a greedy mask optimization: we do not aim at solving a fundamentally different problem than the one of finding the optimal sampling density in a data-driven approach when trying to find the best performing mask. Indeed, Proposition \ref{prop:2} shows that we do not need to find a density, but that finding the support of cardinality $N$ of an optimal distribution is sufficient. And proposition \ref{prop:1} shows the existence of an optimal distribution with such a support.

The argument does not formally prove the suboptimality of variable-density sampling, although one can argue for this from the intuition that these results bring. Indeed, any distribution not compactly supported on a set of $N$ optimal location will open the possibility to a suboptimal mask being sampled by doing sampling without replacement. The heuristic of picking sampling locations at random without replacement could be a cause of the worse practical performance of VDS compared to LBCS, as we will see in Section \ref{s:exp_slbcs} of next Chapter.

Finally, these propositions do not prescribe any particular algorithm that should be used to solve Problem \ref{eq:emp_perf}. Rather, they are limited to establishing an equivalence between \textit{problems}, and do not speak either about the expected \textit{generalization} of these solutions. Generalization is however addressed by standard learning theoretic arguments \citep{gozcu2018learning}.

\section*{Bibliographic note}
Propositions \ref{prop:1} and \ref{prop:2} are due to Armin Eftekhari.

\cleardoublepage
\chapter{Scalable learning-based sampling optimization}\label{chap:lbcs}
In this chapter, we present our extensions to the framework of \citet{gozcu2018learning}. We primarily aim at drastically improving the scalability of their method. We begin by presenting in detail their learning-based CS (LBCS) framework in Section \ref{sec:LBCS_baran}. In Section \ref{sec:submodularity}, we take a short detour in the field of submodular function maximization, from which \citet{gozcu2018learning} took inspiration, and discuss how algorithms that have been proposed in this context could help improve the LBCS method. In Section \ref{sec:improved_LBCS}, we present our two improvements over LBCS, namely \textit{lazy} LBCS and \textit{stochastic} LBCS. We then carry out extensive validation of these methods in Sections \ref{s:exp_slbcs} and \ref{s:exp_llbcs}, and discuss the results in Section \ref{s:lbcs_discussion}\footnote{The work in this chapter is based on the following publications:\\
G{\"o}zc{\"u}, B., Sanchez, T., and Cevher, V. (2019). Rethinking sampling in parallel MRI: A data-driven approach. In \textit{27th European Signal Processing Conference (EUSIPCO)}.\\
Sanchez, T., G{\"o}zc{\"u}, B., van Heeswijk, R. B., Eftekhari, A., Il{\i}cak, E., \c{C}ukur, T., and Cevher, V. (2020a). Scalable learning-based sampling optimization for compressive dynamic MRI. In \textit{ICASSP 2020 - 2020 IEEE International Conference on Acoustics, Speech and Signal Processing (ICASSP)}, pages 8584–8588.}.


\section{Learning-based Compressive Sensing (LBCS)}\label{sec:LBCS_baran}
\begin{figure}[!ht]
    \centering
    \includegraphics[width=\linewidth]{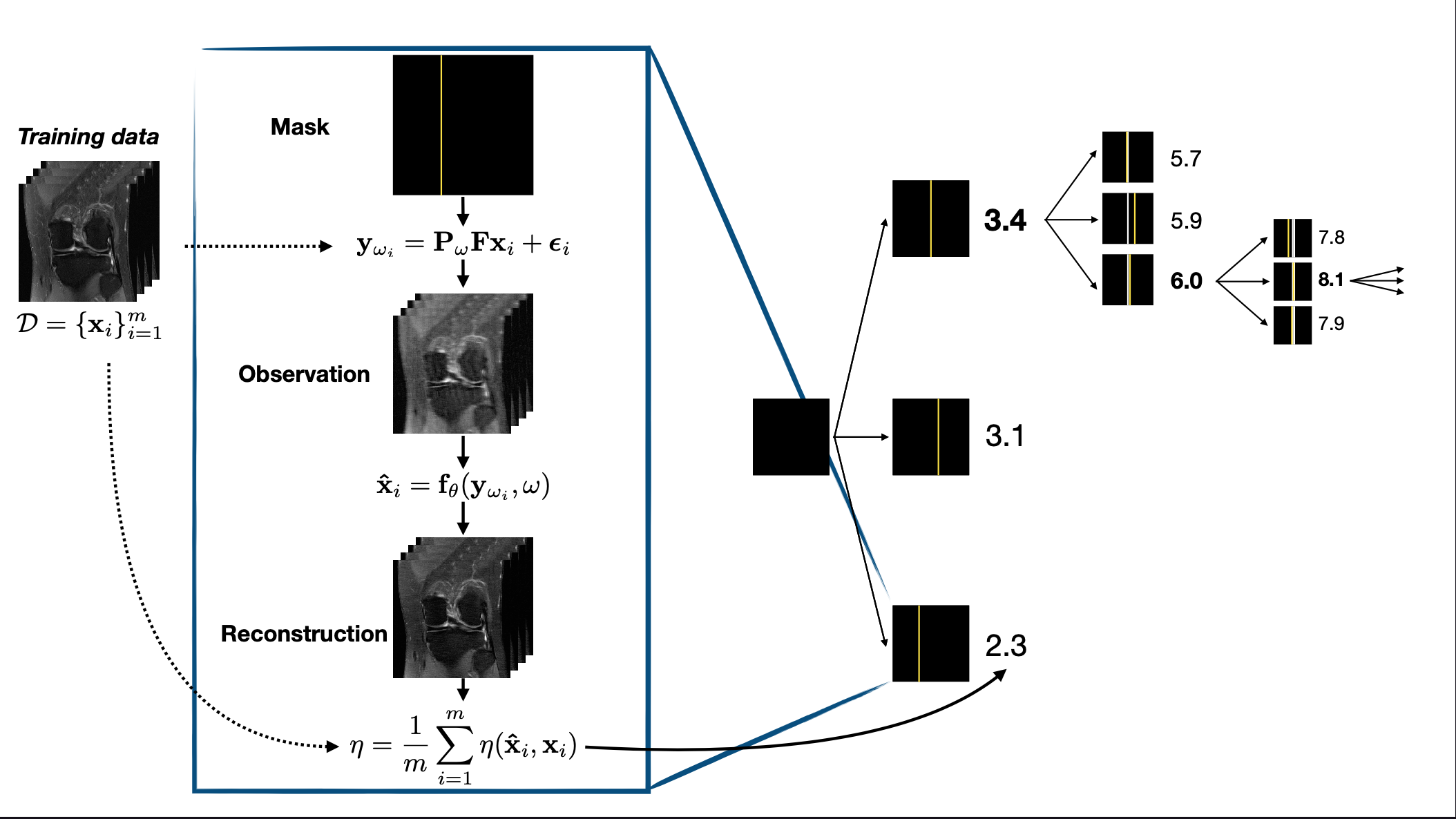}
    \caption{Illustration of the LBCS greedy approach. On the right, we start from an empty mask, and select the line that yields to the best performance. Once it is acquired, we keep this mask, try adding to it the line that yields the best performance given what has been acquired, and continue until the final sampling budget is reached. On the left, we see how the performance is assessed: we use a set $\mathcal{D}$ of training data, that are subsampled according to the current mask, then reconstructed and evaluated using the performance metric $\eta$. The computation is repeated for each different mask.}\label{fig:lbcs}
\end{figure}

LBCS aims at designing a non-adaptive mask in a sequential fashion, and tackles Problem \ref{eq:emp_perf}. Indeed, solving this problem in a naive, brute-force approaches would only be possible in very simple cases, as the total number of masks grows exponentially, with a rate $O(2^P)$. Therefore, one must use \textit{approximate} solutions to this problem. \citet{gozcu2018learning} proposed a greedy, parameter-free approach, where a mask is grown sequentially by adding elements to the mask in an iterative fashion. This approach, called Learning-Based Compressed Sensing (LBCS) allows to reduce the number of configurations evaluated to a complexity $O(P^2)$. The simplified procedure is illustrated on Figure \ref{fig:lbcs}.

A couple of elements must be introduced in order to formally define their algorithm. To implement the constraints of sampling in MRI, typically acquiring full lines in the Fourier space at once, we define a \textit{set of subsets} $\mathcal{S}$ of $\{1, \ldots, P\}$ and assume that the final mask will be constructed as
\begin{equation}
    \omega = \bigcup_{j=1}^t S_j, \text{~~s.t.~} S_j \in \mathcal{S}
\end{equation}
for some $t>0$. \citet{gozcu2018learning} also introduce a \textit{cost function} $c(\omega)\geq 0$ and a \textit{maximal cost} $\Gamma$ that the final mask $\omega$ must not exceed. In the case of Problem \ref{eq:emp_perf}, we have $c(\omega) = |\omega|$ and $\Gamma = N$. With these two additions, the LBCS method \citep{gozcu2018learning} is displayed in Algorithm \ref{alg:lbcs}. The procedure is very simple: while budget is available (l.2), the algorithm tests all feasible candidates $S\in \mathcal{S}$ (l.3-7) and adds permanently to the sampling mask the candidate that yields the largest performance gain (l.8).

\begin{figure}[!ht]
    \centering
    \begin{minipage}[b]{.7\textwidth}
        \begin{algorithm}[H]
            \caption{LBCS}\label{alg:lbcs}
            \textbf{Input}: Training data $\x_1, \dotsc, \x_m$, decoder $\vf_\theta$,\\ sampling subsets $\mathcal{S}$, cost function $c$, maximum cost $\Gamma$ \\
            \textbf{Output}: Sampling pattern $\omega$
            \begin{algorithmic}[1]
                \State $\omega \leftarrow \emptyset$
                \While{$ c(\omega) \leq  \Gamma$}

                \For{$S \in \mathcal{S}$ such that  $c(\omega \cup S) \le \Gamma$}
                \State $\omega' = \omega \cup S$
                \State For each $j$, set  $\vy_{j} \leftarrow \mP_{\omega'}\mF\x_j$, $\hat{\vx}_j \leftarrow \vf_\theta(\vy_j,\omega')$
                \State $\eta(\omega') \leftarrow \frac{1}{m}\sum_{j=1}^m \eta(\vx_j,\hat{\vx}_j)$
                \EndFor
                \State $\displaystyle  \omega \leftarrow \omega \cup S^*, \mbox{ ~~~}
                    S^* = \argmax_{S\,:\,c(\omega \cup S) \le \Gamma} \eta(\omega \cup S) - \eta(\omega)$
                \EndWhile
                \State {\bf return} $\omega$
            \end{algorithmic}
        \end{algorithm}
    \end{minipage}
\end{figure}

The greedy approach of LBCS brings multiple advantages over previous heuristics such as variable-density sampling. First, the mask $\omega$ has a \textit{nested} structure: by recording the order in which elements were added to the mask, it is possible to immediately adapt to different costs $\Gamma$. This feature is not possible for many approaches that directly optimize the sampling mask for a given sampling rate. In addition, this approach does not have any parameter that requires to be tuned, making it particular easy to implement in practice. It also does not rely on any probability distribution from which candidate masks are drawn, which makes it fully learning-based. Finally, the algorithm is not tied to a specific reconstruction method or performance metric, which makes it easy to integrate in novel approaches.


\section{LBCS motivation: Submodularity}\label{sec:submodularity}

The idea to apply a greedy approach to the problem of mask design was initially motivated by techniques used in the field of submodular function maximization \citep{baldassarre2016learning,krause2014submodular}. We discuss in greater depth these concepts here as the motivation for scaling up LBCS are also motivated by algorithms proposed in the context of submodular function maximization.

\begin{figure}[!ht]
    \centering
    \includegraphics[width=.7\linewidth]{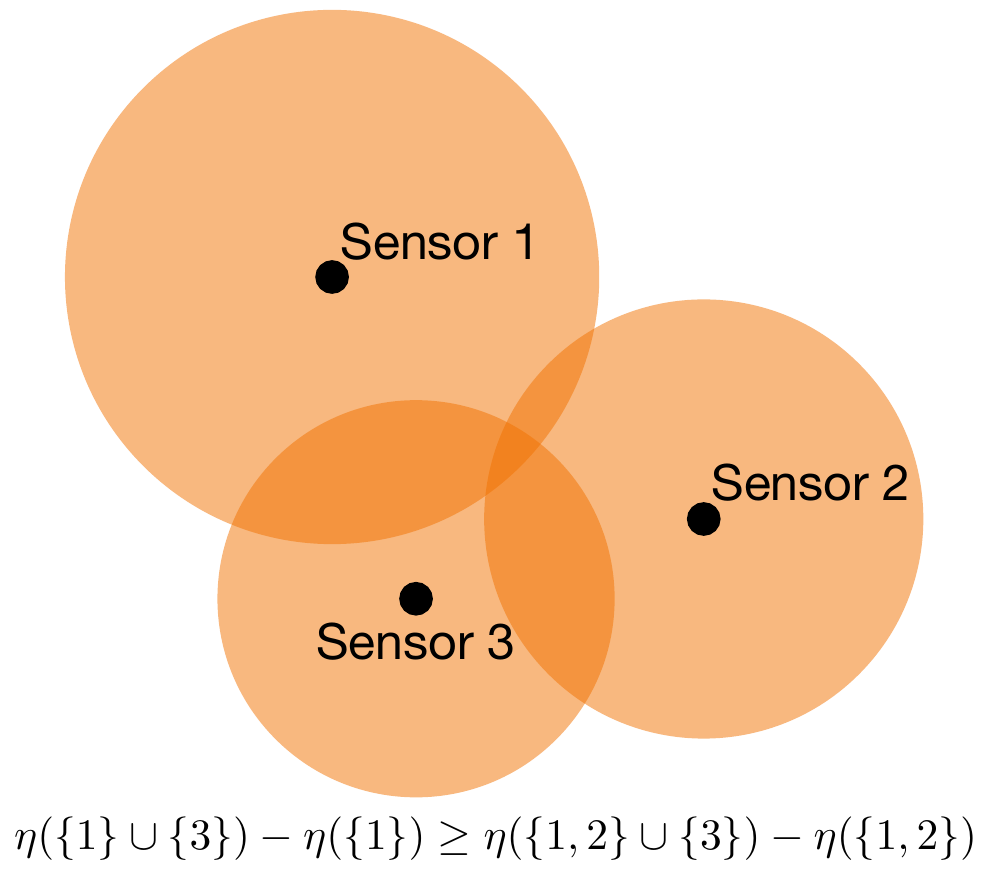}
    \caption{Illustration of the diminishing return property of submodular functions. If $h$ quantifies the total coverage of sensors, we see that adding sensor $3$ to sensor $1$ increases the total coverage more than when adding it to the set $\{1,2\}$, as there is more surface coverage surface overlapping in this case.} \label{fig:submodularity}
\end{figure}

Submodular functions have been studied for a long time \citep{nemhauser1978analysis, minoux1978accelerated}, and formalize among other things the idea of \textit{diminishing returns}. This is the idea that adding an element to a smaller set increases the objective more than when it is added to a larger set. Formally
\begin{definition}[Submodularity]
    A set function $\eta(\omega)$ mapping subsets $\omega \in [P] $ to real numbers is said to be \textbf{submodular} if, for all $\omega_1,\omega_2 \subset [P]$ with $\omega_1 \subseteq \omega_2$ and all $i \in [P]$, it holds that
    \begin{equation}
        \eta(\omega_1 \cup \{i\}) - \eta(\omega_1) \geq \eta(\omega_2 \cup \{i\}) - \eta(\omega_2).
    \end{equation}
    The function is said to be \textbf{modular} if the same holds true with equality in place of inequality
\end{definition}
An example of this is sensor placement. As one places more sensors to cover an area more fully, often, the area covered by sensors will be overlapping, and so adding a new sensor will bring less information if a large set of sensors is already available. This is illustrated on Figure \ref{fig:submodularity}.

This typically translates into problems of submodular function maximization, where a common constraint lies on the cardinality of the solution:
\begin{equation}
    \max_{\omega \subseteq [P]} \eta(\omega) \text{~subject to~}|\omega| \leq N.
\end{equation}

In the case of linear reconstruction in MRI, it can be shown that the problem of finding the sampling mask that gives minimal mean squared error is a \textit{modular} optimization problem: there is no diminishing returns, but each component can be optimized individually \citep{baldassarre2016learning}. In the case of a non-linear reconstruction in MRI, there is no formal proof that shows that submodularity exactly holds, but empirical results show a diminishing return throughout acquisition. This is what motivated \citet{gozcu2018learning} to leverage tools from submodular optimization for mask design in MRI.

Indeed, greedy algorithms have a special place in submodular optimization because of the celebrated result by \citet{nemhauser1978analysis}, who proved that the greedy algorithm provides a good approximation to the optimal solution of NP-hard instances of submodular function maximization \citep{krause2014submodular}.

\begin{theorem}[\citet{nemhauser1978analysis}]
    Let $\eta:2^P \to \mathbb{R}_+$ be a nonnegative monotone submodular function. Let ${\omega_t}$, ${t\geq 0}$ be the sequence obtained by the greedy algorithm. Then, for all positive $t, N$,
    \begin{equation}
        \eta(\omega_t) \geq \left(1- e^{-t/N}\right) \max_{\omega:|\omega|\leq N} \eta(\omega).
    \end{equation}
    In particular, for $t=N$, $\eta(\omega_N) \geq  \left(1- 1/e\right) \max_{\omega:|\omega|\leq N} \eta(\omega)$.
\end{theorem}
For many classes of submodular functions $h$, this result is the best that can be achieved  with any efficient algorithm \citep{krause2014submodular}. In particular, \citet{nemhauser1978best} proved that any algorithm that can only evaluate $h$ on a polynomial number of set will not be able to obtain an approximation guarantee that is better than $(1-1/e)$. This property could be key to explaining the strong empirical performance of the LBCS algorithm of \citet{gozcu2018learning} in the context of MRI.

Although the approximation guarantee cannot be generally improved, many works have proposed more \textit{efficient} approaches than the simple greedy algorithm. Two works are of particular interest to us, as they serve as the motivation to the algorithms that we will describe in the next section. These algorithms are the \textit{lazy greedy} \citep{minoux1978accelerated} and \textit{stochastic greedy}, called lazier than lazy greedy \citep{mirzasoleiman2015lazier}.

The lazy greedy algorithm relies on the fact that if at the $i$-th iteration of the greedy algorithm, an element $S$ is expected to bring a marginal benefit $ \Delta(S|\omega_{i}) = \eta(\omega_{i}\cup S) - \eta(\omega_{i})$, then it holds that $\Delta(S|\omega_{j}) \le \Delta(S|\omega_{i})$  for all $j \ge i$ for a submodular function $\eta$. Exploiting this fact, one can keep a list of upper bounds on the marginal benefits of each element $S$, called $\rho(S)$, initialized at $+\infty$.  Then, at each iteration, the element S with largest value $\rho(S)$ is picked and updated as $\rho(S) = \Delta(S|\omega)$.  If $\rho(S) \ge \rho(S')$ $\forall S' \in \mathcal S$, the marginal benefit of this element $S$ is larger than  the upper bounds of each other marginal contribution, and is consequently added permanently to $\omega$. This can speed up the algorithm by several orders of magnitude, since $\rho(S) \ge \rho(S')$ can be satisfied after a few trials, instead of trying each available $S$ at every iteration \citep{krause2014submodular}.

The stochastic greedy algorithm \citep{mirzasoleiman2015lazier} samples at each step of the greedy algorithm a subset of candidates $\mathcal{S}_{iter} \subseteq \mathcal{S}$, with $|\mathcal{S}_{iter}|=k$ at line 3 of Algorithm \ref{alg:lbcs}. As a result, given a nonnegative monotone submodular function $\eta:2^P \to \mathbb{R}_+$ and taking $k=\frac{P}{N} \log\left(1/\epsilon\right)$,  the stochastic greedy algorithm achieves a $(1-1/e-\epsilon)$ approximation guarantee in expectation to the optimum, using only $O\left(P\log(1/\epsilon)\right)$ evaluations.

Both of these algorithms inspired the methods that we proposed to scale up LBCS, and their implementation in the context of MRI will be respectively covered in Sections \ref{ss:llbcs} and \ref{ss:slbcs}.

\begin{remark}[Submodular optimization under noise]
    Although the cases so far mostly discussed optimizing submodular functions without measurement noise, such a setting is not realistic in the case of prospective MRI acquisitions, where repeated lines will have a different noise, and resampling can improve the average signal. Nonetheless, submodular optimization under noise has been studied, and algorithms that adapt to such settings have been proposed \citep{singla2016noisy,hassidim2017submodular}.
\end{remark}

\section{Scaling up LBCS}\label{sec:improved_LBCS}
\subsection{Limitations of LBCS}
As we have hinted above, in many practical scenarios, evaluating $O(P^2)$ configurations remains very expensive. This is also due to the fact that each configuration requires reconstructing $m$ data points. In particular, the LBCS approach becomes prohibitively expensive when either of the following occurs:
\begin{enumerate}
    \item When the number of candidate locations grows significantly. This is particularly the case, in 3D MRI, where one needs to undersample entire volumes, or dynamic MRI, where one undersamples multiple frames for a single 2D image.
    \item When the size of the dataset grows, it becomes impractical to evaluate LBCS on the whole testing set at once.
    \item When the running time of the reconstruction method becomes large. This is particularly the case for multi-coil MRI, which is the setting most commonly used in practice. This issue also occurs when dealing with larger data, from 3D or dynamic MRI.
\end{enumerate}

As a result, we wish to develop variants of LBCS that allow to tackle these different settings, while retaining the performance of LBCS as much as possible. We will discuss two variants of LBCS that were respectively proposed in \citet{gozcu2019rethinking} and \citet{sanchez2019scalable}, and that aim at tackling mostly the issues of multi-coil and 3D MRI as well as dynamic MRI. We start by explaining how they differ from LBCS.

\subsection{Stochastic LBCS}\label{ss:slbcs}
With stochastic LBCS (sLBCS), we aim at addressing two fundamental drawbacks of the approach of \citet{gozcu2018learning}, namely that \textbf{1.} its scaling is  quadratic with respect to the total number of readout lines, \textbf{2.} its scaling is linearly with respect to the size of the dataset. With this approach, we also aim at integrating a specific constraint of dynamic MRI (dMRI), namely that the sampling mask should have a fixed number of readouts by temporal frame.

Scalability with respect to the size of the dataset is essential for the applicability of sLBCS to modern datasets, which can contain several thousands of images, making it impractical to go through the entire dataset at each inner loop of the algorithm.

Before describing the algorithm, let us briefly recall the dynamic MRI setting. The main difference with static MRI is that the signal will be a vectorized video, i.e. $\vx \in \mathbb{C}^{P}$, where $P = H \times W \times T$ instead of $H \times W$. As a result, the operator $\mA_\omega$ will be constructed differently, and in particular, the Fourier transform will be applied frame-wise. In addition, the mask $\omega$ will be constructed as a union of elements from a set $\mathcal{S}$ that will represent lines at different time frames, and will be described as $\mathcal{S}= \mathcal{S}_1 \cup \ldots \cup \mathcal{S}_T$. We also define the set $\mathcal{L} \subseteq \{1,\ldots,m\}$ as a set of indices where the training data $\{\vx\}_{i=1}^m$ will be subsampled.

Let us now detail \textit{how} our proposed stochastic greedy method solves the limitations \textbf{1.} and \textbf{2.} of LBCS (Algo. \ref{alg:lbcs}). The issue \textbf{1.} is solved by picking uniformly at random at each iteration a batch possible readout lines $\mathcal{S}_{iter}$ of constant size $k$ from a given frame $\mathcal{S}_t$, instead of considering the full set of possible lines $\mathcal{S}$ (line 3 in Alg. \ref{alg:dslbcs}), and is inspired from the stochastic greedy of \citet{mirzasoleiman2015lazier};
the number of candidate locations is further reduced by iterating through the lines to be added from each frame $\mathcal{S}_t$ sequentially (lines 1, 3 and 11 in Algorithm \ref{alg:dslbcs});
\textbf{2.} is addressed by considering a fixed batch of training data $\mathcal{L}$ of size $l$ instead of the whole training set of size $m$ at each iteration (line 4 in Algorithm \ref{alg:dslbcs});
\begin{remark}
    We will discuss the impact of introducing the batch size $k$ for the readout lines and the subset $\mathcal{L}$ to subsample training data in the results. However, we want to mention here that one has to be careful to pick the subset $\mathcal{L}$ before the \texttt{for} loop of line 5 in Algorithm \ref{alg:dslbcs}. Indeed, failing to do this would result in the different candidates $S$ would be evaluated on different subsets of the data. As there is some variation within the data, for instance in terms of SNR, such an approach would yield performances that are \textit{not} comparable among different candidates. In turn, this would skew the results towards picking the candidate location $S$ with an associated subset $\mathcal{L}_S$ with high signal-to-noise ratio, at the expense of selecting a readout location for the performance improvement that it actually brings. This would be especially the case if $|\mathcal{L}|$ were small. Picking $\mathcal{L}$ before this \texttt{for} loop ensures that the readout line selected will indeed be the one that brings the largest performance gain among the candidates for the subset of data $\mathcal{L}$.
\end{remark}

The improvements of sLBCS allow to reduce the computational complexity from\linebreak $\Theta\left(mr(HT)^2\right)$ to $\Theta\left(lrkHT\right)$, effectively speeding up the computation by a factor $\mathbf{\Theta(\frac{m}{l}\frac{HT}{k})}$, where $r$ denotes the cost of a single reconstruction. Our results show that this is achieved without sacrificing any reconstruction quality. We will see that the improvement is particularly significant for large datasets, as the batch size $l$ will typically remain small, typically up to 100 samples, whereas MRI datasets can contain more than 15\,000 data points \citep{zbontarFastMRIOpenDataset2019}.


\begin{figure}[!ht]
    \centering
    \begin{minipage}[b]{.8\textwidth}
        \begin{algorithm}[H]
            \caption{sLBCS: Greedy mask optimization algorithm for (d)MRI}\label{alg:dslbcs}
            \textbf{Input}: Training data $\{\vx\}_{i=1}^m$, recon. rule $g$, sampling set $\mathcal{S}$, max. cardinality $n$, samp. batch size $k$, train. batch size $l$ \\
            \textbf{Output}: Sampling pattern $\omega$
            \begin{algorithmic}[1]
                \State \textcolor{red}{\textbf{(dMRI)}  Initialize $t=1$}
                \While{$|\omega| \leq n$}
                \State \textcolor{red}{Pick $ \left \{\begin{tabular}{l}
                            $\mathcal{S}_{iter}\subseteq \mathcal{S}$ \textbf{(MRI)} \\
                            $\mathcal{S}_{iter}\subseteq \mathcal{S}_t$ \textbf{(dMRI)}\end{tabular}\right.$  at random, with $|\mathcal{S}_{iter}| = k$}

                \State \textcolor{red}{Pick $\mathcal{L} \subseteq \{1,\ldots,m\} $, with $|\mathcal{L}| = l$}
                \For{$S \in \mathcal{S}_{iter} \text{ such that } |\omega \cup S| \leq \Gamma$}

                \State  $\omega' = \omega \cup S$
                \State \textcolor{red}{For each $\ell \in \mathcal{L}$} set ${\hat{\vx}}_\ell\leftarrow \vf_\theta(\mP_{\omega'}\mF\vx_\ell,\omega')$
                \State \textcolor{red}{$\eta(\omega') \leftarrow \frac{1}{|\mathcal{L}|} \sum_{\ell\in\mathcal{L}} \eta(\vx_\ell, {\hat{\vx}}_\ell)$}
                \EndFor
                \State $\displaystyle\omega \leftarrow \omega \cup S^*,  \text{ where }$   $\displaystyle S^* = \argmax_{\substack{S:|\omega \cup S| \leq n\\ S \in \mathcal{S}_{iter}}} \eta(\omega\cup S)$
                \State   \textcolor{red}{\textbf{ (dMRI) }   $t= (t \bmod T)+1$}
                \EndWhile
                \State {\bf return} $\omega$
            \end{algorithmic}
        \end{algorithm}
    \end{minipage}
\end{figure}

\subsection{Lazy LBCS}\label{ss:llbcs}
In cases where the cardinality of the candidate elements $\mathcal{S}$ grows extremely large, even Algorithm \ref{alg:dslbcs} can become impractical. In 2D MRI, LBCS deals with images of size $H \times W$ while having a candidate set of size $H$, with $H$ different lines. However, in the case of 3D MRI, the number of candidate locations explodes, as we move to images of size $P= H \times W \times S$, where $S$ is the number of slices, with a candidate set of size $H \times W$, i.e., one is allowed to sample in both phase and frequency encoding directions. While in 2D dynamic MRI, the candidate set is of size $H \times T$ with $T$ typically much smaller than $H$, in the case of 3D MRI, we often have $H \approx W$. In addition, we have less structure than in dMRI, as there is no option to cycle through frames. This implies that the batch of randomly sampled locations must be much larger to enable a good performance, which results in the cost remaining overall prohibitive.
An additional reason for this increased cost is the reconstruction time of 3D methods. The number of slices can be much larger than the number of frames in dMRI, slowing down reconstruction considerably.
Moving from single coil to multi-coil data further increases this cost, although this change does \textit{not} increase the complexity of the sampling optimization problem.

Lazy LBCS proposes a solution to this limitation by leveraging another approach to speed up LBCS. It uses \textit{lazy} evaluations \citep{minoux1978accelerated} which is equivalent to but faster than the greedy algorithm for submodular functions, and possess optimization guarantees for such functions \citep{nemhauser1978analysis}. While in our present setting, verifying submodularity is difficult because writing the performance of reconstructions in a closed form is generally not possible, we were motivated by the fact that Algorithm \ref{alg:llbcs} works well in practice even if the objective function is not exactly submodular. Indeed, as we will observe in the experiments, lazy evaluations also perform well for mask optimization.

The implementation of lazy evaluations for LBCS is described in Algorithm \ref{alg:llbcs}. We see that a list of upper bounds is initially computed (l.2), and that we sequentially traverse the list of upper bounds until we find a candidate sample $S$ that brings a larger improvement than any of the upper bounds in $\rho$ (ll.8-9).


\begin{algorithm}
    \begin{minipage}[b]{.8\textwidth}
        \caption{Lazy Learning-based Compressive Sensing (lLBCS)}
        \label{alg:llbcs}
        \textbf{Input}: Training data $\x_1, \dotsc, \x_m$, decoder $\vf$, sampling subsets $\mathcal{S}$, cost function $c$, maximum cost $\Gamma$ \\
        \textbf{Output}: Sampling pattern $\omega$
        \begin{algorithmic}[1]
            \State $\omega \leftarrow \emptyset$
            \State $\rho(S) \leftarrow +\infty \text{~}\forall S \in \mathcal S \text{ s.t. } c(\omega \cup S) \le \Gamma$

            \While{$ c(\omega) \leq  \Gamma$}
            \State $\displaystyle\omega' \leftarrow \omega \cup S, \text{where } S = \argmax_{S' \in \mathcal S\,:\,c(\omega \cup S') \le \Gamma} \rho(S')$
            \State For each $j$, set  $\vy_{j} \leftarrow \mP_{\omega'}\mF\x_j$, $\hat{\x}_j \leftarrow \vf_\theta(\vy_j,\omega')$
            \State $\displaystyle\eta(\omega') \leftarrow \frac{1}{m}\sum_{j=1}^m \eta(\x_j,\hat{\x}_j)$
            \State $\displaystyle\rho(S) \leftarrow \eta(\omega') - \eta(\omega)$
            \If{$\rho(S) \ge \rho(S') \text{~}\forall S' \in \mathcal S \text{ s.t. }c(\omega \cup S') \le \Gamma$}
            \State $\omega = \omega \cup S$
            \EndIf
            \EndWhile
            \State {\bf return} $\omega$
        \end{algorithmic}
    \end{minipage}
\end{algorithm}

\section{Experiments - Stochastic LBCS}\label{s:exp_slbcs}
In this Section, we validate the performance of sLBCS, and will discuss lLBCS in Section \ref{s:exp_llbcs}. We carry out extensive experiments to validate our stochastic LBCS method. We first compare sLBCS against the vanilla LBCS methods of \citet{gozcu2018learning}, paying particular attention to how it compares to the original method in terms of performance and computational complexity. We study the influence on the batch sizes $k$ and $l$ on performance and on the design of the mask. Then, we carry out experiments on both single coil and multicoil data, as well as noise free and noisy data. Finally, we show the benefit of our method on large scale datasets, for which LBCS is prohibitively expensive.

\subsection{Experimental setting}

\textbf{Dataset.} Our experiments were carried out on three different datasets.
\begin{itemize}
    \item \textit{Cardiac dataset.} The data set was acquired in seven healthy adult volunteers with a balanced steady-state free precession (bSSFP) pulse sequence on a whole-body Siemens 3T scanner using a 34-element matrix coil array. Several short-axis cine images were acquired during a breath-hold scan. Fully sampled Cartesian data were acquired using a $256\times 256$ grid, with relevant imaging parameters including $320 \times 320$ mm field of view (FoV), \SI{6}{\milli\meter} slice thickness, $1.37 \times 1.37$ mm spatial resolution, \SI{42.38}{\milli\second} temporal resolution, $1.63/3.26$ ms TE/TR, \ang{36} flip angle, \SI{1395}{\hertz}\mbox{/px} readout bandwidth. There were $13$ phase encodes acquired for a frame during one heartbeat, for a total of $25$ frames after the scan.

          The Cartesian cardiac scans were then combined to single coil data from the initial $256\times 256 \times 25 \times 34$ size, using adaptive coil combination \citep{walsh2000adaptive, griswold2002use}. This single coil image was then cropped to a $152 \times 152 \times 17$ image. This was done because a large portion of the periphery of the images are static or void, and also to enable a greater computational efficiency. In the experiments, we used three volumes for training and four for testing.

    \item \textit{Vocal dataset.} The vocal dataset that we used in the experiments comprised $4$ vocaltract scans with a 2D HASTE sequence (T2 weighted single-shot turbo spin-echo) on a 3T Siemens Tim Trio using a 4-channel body matrix coil array. The study  was  approved  by  the  local  institutional  review  board,  and  informed  consent  was  obtained  from  all  subjects  prior  to  imaging.  Fully sampled Cartesian data were acquired using a $256 \times 256$ grid, with $256 \times 256$ mm field of view (FoV), \SI{5}{\milli\meter} slice thickness, $1 \times 1$ mm spatial resolution, $98/1000$ ms TE/TR, \ang{150} flip angle,  \SI{391}{\hertz}\mbox{/px} readout bandwidth, \SI{5.44}{\milli\second} echo spacing ($256$ turbo factor). There was a total of $10$ frames acquired, which were recombined to single coil data using adaptive coil combination as well \citep{walsh2000adaptive, griswold2002use}.

    \item \textit{fastMRI.} The fastMRI dataset was obtained from the NYU fastMRI initiative \citep{zbontarFastMRIOpenDataset2019}. The anonymized dataset comprises raw k-space data from more than 1,500 fully sampled knee MRIs obtained on 3 and 1.5 Tesla magnets. The dataset includes coronal proton density-weighted images with and without fat suppression.
\end{itemize}

\textbf{Reconstruction algorithms.}  We consider two reconstruction algorithms, namely \textit{k-t FOCUSS} (KTF) \citep{jung2009k},  and \textit{ALOHA} \citep{jin2016general}.  Their parameters were selected to maintain a good empirical performance across all sampling rates considered.

KTF aims at reconstructing a dynamic sequence by enforcing its sparsity in the x-f domain. Formally, it aims at solving
$$
    \min_\vz \|\vz\|_1 \text{~subject to~} \|\yo - \Po \mF \mF_t^{-1} \vz  \|_2 \leq \epsilon
$$
where $\mF_t^{-1}$ denotes the inverse temporal Fourier transform. It tackles the problem by using a reweighted quadratic approach \citep{jung2007generalized}.

ALOHA \citep{jin2016general} is an annihilating filter based low-rank Hankel matrix approach. ALOHA exploits the duality between sparse signals and their transform domain representations, cast as a low-rank Hankel matrix, and effectively transforms CS reconstruction into a low-rank matrix completion problem.  We refer the reader to their detailed derivation for further details.

\textbf{Mask selection baselines:}

\begin{itemize}
    \item\textit{Coherence-VD} \citep{lustig2007sparse}: We consider a random \textit{variable-density} sampling mask with Gaussian density and optimize its parameters to minimize coherence. This is an instance of a \textit{model-based}, \textit{model-driven} approach.

    \item \textit{Golden} \citep{li2018dynamic}: This approach designs a Cartesian mask by adding progressively lines according to the golden ratio. This is a \textit{model-based}, \textit{model-driven} approach.

          \item\textit{LB-VD} \citep{gozcu2018learning,gozcu2019rethinking}: Instead of minimizing the coherence as in \textit{Coherence-VD}, we perform a grid search on the parameters using the training set to optimize reconstruction according to the same performance metric as our method. This makes this approach \textit{model-based}, \textit{data-driven}.

\end{itemize}

We also consider several variants of LBCS and sLBCS that we will use throughout the experiments. We define abbreviations for conciseness:
\begin{itemize}
    \item \textbf{G-v1} is the original LBCS algorithm.
    \item \textbf{G-v2} is a version of sLBCS that uses a batch of training data $l$ and cycles through the frames.
    \item \textbf{SG-v1} is a version of the stochastic LBCS algorithm that considers a batch of sampling locations $k$ but uses all training data, i.e. $l = |m|$.
    \item \textbf{SG-v2} is sLBCS, using a batch of sampling locations $k$, a batch of training data $l$ and cycling through the frames.
\end{itemize}

\begin{remark}
    In this work, we mainly focused on small scale datasets with a large set of candidate sampling locations. Our experiments therefore are carried out using SG-v1 and SG-v2. We did not observe any significant performance difference between these variants of sLBCS.

    However, we note that G-v2 might be of interest in deep-learning based settings, where one deals with large datasets, such as fastMRI \citep{zbontarFastMRIOpenDataset2019}. This will be further discussed in Section \ref{sec:fastMRI_sLBCS}.
\end{remark}

\subsection{Comparison of greedy algorithms}
We first evaluate SG-v1 and SG-v2 against G-v1 to establish how sensitive our proposed method is to its parameters $k$ and $l$. We carry out this experiment on the cardiac data, and report only the results on KTF, although similar trends can be observed with other reconstruction algorithms.

\begin{wrapfigure}{L}{0.45\linewidth}
    \centering
    \includegraphics[width=\linewidth]{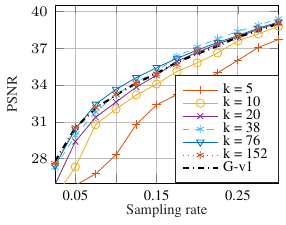}
    \caption{PSNR as a function of the rate for KTF, comparing the effect of the batch size on the quality of the reconstruction for SG-v1. The result is averaged on $4$ testing images of size $152\times 152\times 17$.}\label{fig:psnr_batch2}
\end{wrapfigure}

\begin{figure}[!ht]
    \centering
    \begin{minipage}[c]{.6\linewidth}
        \includegraphics[width=\linewidth]{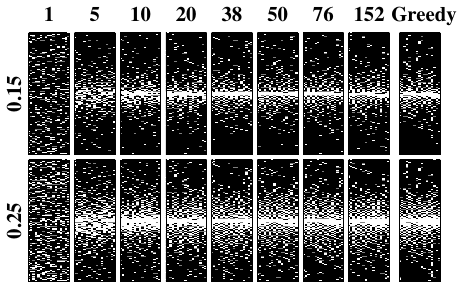}
    \end{minipage}\hfill
    \begin{minipage}[c]{.38\linewidth}
        \caption{Learning-based masks obtained with SG-v1 for different batch sizes $k$ using KTF as a reconstruction algorithm, shown in the title of each column, for $15\%$ and $25\%$ sampling rate. The optimization used data of size $152\times 152 \times 17$, with a total of $2584$ possible phase encoding lines for the masks to pick from.}\label{fig:mask_sto_cycling}
    \end{minipage}
\end{figure}

\textbf{Influence of the batch size $k$ on the mask design.}  We first discuss the tuning of the batch size used in SG-v1, to specifically study the effect of different batch sizes. Figure \ref{fig:psnr_batch2} provides quantitative results of the performance, while Figure \ref{fig:mask_sto_cycling} shows the shapes of the masks obtained with various batch sizes. Unsurprisingly, small batch sizes yield poor results, as this algorithm has to choose from a very small set of randomly sampled candidates, being likely to miss out on important frequencies, such as the center ones. It is interesting to see however that the PSNR reaches the result from G-v1 with as few as 38 samples (out of the $152$ candidates of a frame, and $152 \times 17 = 2584$ candidates overall).

In some cases, such as $k=38$, SG-v1 can even yield an \textit{improved} performance over G-v1, with $60$ times less computation that G-v1. We hypothesize that it is due to the noise introduced by sLBCS allowing to avoid suboptimal configurations encountered by G-v1. Note that this might not be expected if the problem was to be exactly submodular, but this is \textit{not} the case in practice.

Focusing on Figure \ref{fig:mask_sto_cycling}, one can make several observations. First of all, as expected, taking a batch size of $1$ yields a totally random mask, and taking a batch size of $5$ yields a mask that is more centered towards low frequency than the one with $k=1$ but it still has a large variance. Then, as the batch size increases, resulting masks seem to converge to very similar designs, but those are slightly different from the ones obtained with G-v1. Overall, our initial experiments suggest, as a rule of thumb, that roughly sampling $25\%$ of a frame enables sLBCS to match the performance of G-v1.

\begin{wrapfigure}{L}{0.45\linewidth}
    \centering
    \includegraphics[width=\linewidth]{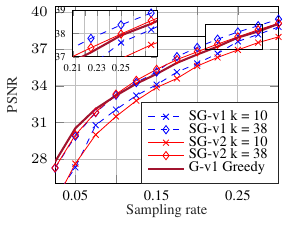}
    \caption{PSNR as a function of the sampling rate for KTF, comparing the different reconstruction methods as well as the effect of the batch size on the quality of the reconstruction for SG. }\label{fig:psnr_batch}
\end{wrapfigure}

\textbf{Influence of the batch size $l$.} We then include the effect of using a smaller data batch size $l$ into the equation, comparing the performances of G-v1 with SG-v1 and SG-v2. The results are presented on Figure \ref{fig:psnr_batch}. Both SG-v1 and SG-v2 behave similarly, and for $k=38$, both methods end up outperforming G-v1, with respectively $60$ times less and $180$ times less computation. As seen before, using a \textit{too} small batch size $k$ (e.g. $10$) yields a drop in performance. Using a batch of training images $l$ (SG-v2) does not seem to significantly reduce the performance compared to SG-v1, while substantially reducing computations. As a result, in the sequel, we use $k=38$ and $l=1$ for SG-v2.

\textbf{Computational costs.} We discuss now the computational costs for the different variations of the greedy methods used in the single coil experiments. Table \ref{tab:runtime}  provides the running times and empirically measured speedup for the greedy variation, and Table \ref{tab:LB-VD} provides the computational times required to obtain the learning-based variable density (LB-VD) parameters through an extensive grid-search. The empirical speedup is computed as
\begin{equation}
    \text{Speedup} = \frac{t_{\text{G-v1}}\cdot n_{\text{procs, G-v1}}}{t_{\text{SG-v2}}\cdot n_{\text{procs, SG-v2}}}
\end{equation}
where $t_{\text{G-v1}}$, $t_{\text{SG-v2}}$ refer to the measured running times of the algorithms, and $n_{\text{procs, G-v1}}$, $n_{\text{procs, SG-v2}}$ to the number of CPU processes used to carry out the computations.

The main point of these tables is to show that the computational improvement is very significant in terms of resources, and that our approach improves greatly the efficiency of the method of \citet{gozcu2018learning}. This ratio might differ from the predicted speedup factor of $\mathbf{\Theta\left(\frac{m}{l}\frac{NT}{k}\right)}$ due to computational considerations. Table \ref{tab:runtime} shows that we have roughly a factor $1.2$ between the predicted and the measured speedup, mainly due to the communication between the multiple processes as well as I/O operations.

Moving to the grid search computations for LB-VD, displayed in Table \ref{tab:LB-VD}, we ought to do several remarks. The number of parameters was chosen in order to ensure the same order of magnitude of computations to be carried out for both LB-VD and sLBCS. Note also that in opposition to SG-v2, the grid search cannot be sped up by using a batch of training data: all the evaluation should be done on the same data. In the case of SG-v2 however, one can draw a new batch of training data at each sampling round of the algorithm.

\begin{table}[!t]
    \centering

    \resizebox{\textwidth}{!}{
        \begin{tabular}{>{\bfseries}c crccrcccc}
            \toprule
            \multirow{2}{*}{\textbf{Algorithm}} & \multirow{2}{*}{\textbf{Setting}} & \multicolumn{2}{c}{\textbf{G-v1} (LBCS)} & \multicolumn{3}{c}{\textbf{SG-v1} (sLBCS-v1)} & \multicolumn{3}{c}{\textbf{SG-v2}  (sLBCS-v2)}                                                                                                             \\
            \cmidrule(r){3-4} \cmidrule(l){5-7} \cmidrule(l){8-10}
                                                &                                   & \multicolumn{1}{c}{Time}                 & $n_{\text{procs}}$                            & \multicolumn{1}{c}{Time}                       & $n_{\text{procs}}$ & Speedup     & \multicolumn{1}{c}{Time} & $n_{\text{procs}}$ & Speedup                \\
            \midrule

            \multirow{2}{*}{KTF}                & 152$\times$152$\times$17$\times$3 & $6\text{d }23\text{h~}$                  & $152$                                         & $11\text{h } 40$                               & $38$               & $58$ $(68)$ & $3\text{h } 25$          & $38$               & $\mathbf{170}$ $(204)$ \\
                                                & 256$\times$256$\times$10$\times$2 & $\sim 7\text{d~~}8\text{h}^*$            & $256$                                         & $12\text{h } 20$                               & $64$               & $57$ $(68)$ & $5\text{h } 20$          & $64$               & $\mathbf{173}$ $(204)$ \\ \cmidrule(l){2-10}
            IST                                 & 152$\times$152$\times$17$\times$3 & $3\text{d~}11\text{h~}$                  & $152$                                         & $5\text{h } 30$                                & $38$               & $60$ $(68)$ & $1\text{h } 37$          & $38$               & $\mathbf{184}$ $(204)$ \\ \cmidrule(l){2-10}
            ALOHA                               & 152$\times$152$\times$17$\times$3 & $\sim 25\text{d~~}1\text{h}^*$           & $152$                                         & $1\text{d }14\text{h } 25$                     & $38$               & $62$ $(68)$ & $18\text{h } 13$         & $38$               & $\mathbf{133}$ $(204)$ \\
            \bottomrule
        \end{tabular}}
    \caption{Running time of the greedy algorithms for different decoders and training data sizes. The setting corresponds to  $n_{x}$, $n_{y}$, $n_{\text{frames}}$, $n_{\text{train}}$. The number $n_{\text{procs}}$ denotes how many parallel processes are used by each simulation. $^*$ means that the runtime was extrapolated from a few iterations. We used $k=n_{\text{procs}}$ for SG-v1 and SG-v2 and $l=1$ for SG-v2. The speedup column contains the measured speedup and the theoretical speedup in parentheses.}\label{tab:runtime}
\end{table}

\begin{table}[!t]
    \centering

    \begin{tabular}{>{\bfseries}c cccr}
        \toprule
        Algo.                & Setting                               & $n_{\text{pars}}$ & $n_{\text{procs}}$ & Time      \\
        \midrule
        \multirow{2}{*}{KTF} & 152$\times$152$\times$17$\times$3~~   & 1200              & 38                 & $6$h~$30$ \\
                             & 256$\times$256$\times$10$\times$2$^*$ & 2400              & 64                 & $6$h~$45$ \\ \cmidrule(l){2-5}
        IST                  & 152$\times$152$\times$17$\times$3~~   & 1200              & 38                 & $3$h~$20$ \\ \cmidrule(l){2-5}
        ALOHA                & 152$\times$152$\times$17$\times$3~~   & 1200              & 38                 & $1$d~$8$h \\
        \bottomrule
    \end{tabular}
    \caption{Comparison of the learning-based random variable-density Gaussian sampling optimization for different settings. $n_{\text{pars}}$ denotes the size of the grid used to optimize the parameters. For each set of parameters, the results were averaged on $20$ masks drawn at random from the distribution considered. The $n_{\text{pars}}$ include a grid made of $12$ sampling rates (uniformly spread in $[0.025,0.3]$), $10$ different low frequency phase encodes (from $2$ to $18$ lines), and different widths of the Gaussian density (uniformly spread in $[0.05, 0.3]$) -- $10$ for the images of size $152\times 152$,
        $20$ in the other case. }\label{tab:LB-VD}
\end{table}

\FloatBarrier
\subsection{Single coil results}\label{sec:baseline_main}

\begin{figure}[!ht]
    \centering
    \includegraphics[width=.75\linewidth]{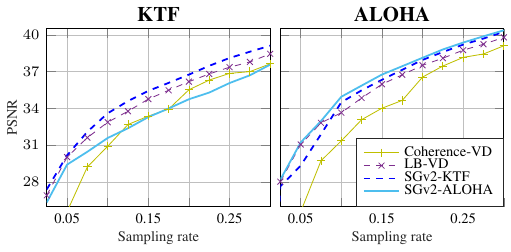}
    \caption{PSNR as a function of sampling rate, evaluating the different masks (Coherence-VD, LB-VD, SG-v2-KTF, SG-v2-ALOHA) on both reconstruction methods (KTF and ALOHA). The results are averaged over $4$ images.}\label{fig:psnr_cross}
\end{figure}

\begin{figure}[!ht]
    \centering
    \includegraphics[width=.6\linewidth]{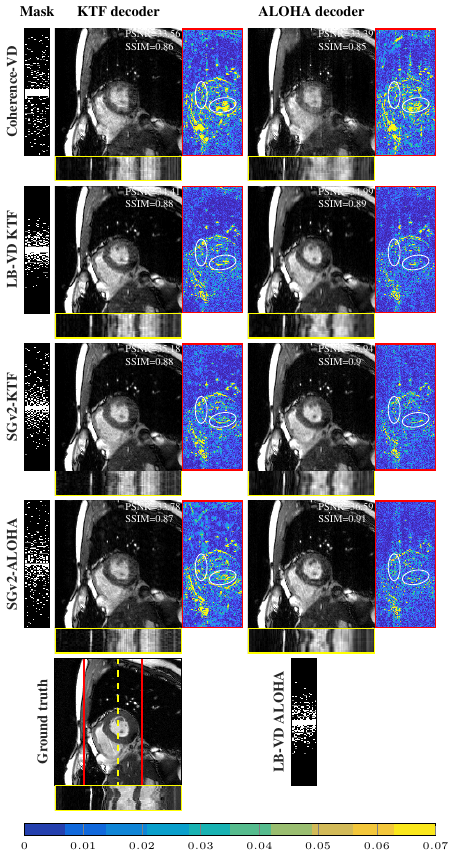}
    \caption{Comparison of the different reconstruction masks and decoders, for a sampling rate of $15\%$ on a single sample with its PSNR/SSIM performances.}\label{fig:plot_cross}
\end{figure}

\textbf{Main results.} We see on Figures \ref{fig:psnr_cross} and \ref{fig:plot_cross} that the SG-v2 brings a consistent improvement over all baselines considered, even though some variable-density techniques are able to provide good results for some sampling rates and algorithms. Compared to Coherence-VD, there is always at least $1$ dB improvement at any sampling rate, and it can be as much as $6.7$ dB at $5\%$ sampling rate for ALOHA. For the rest, the improvement is over $0.5$ dB. Figure \ref{fig:psnr_cross} also clearly indicates that the benefits of our learning-based framework become more apparent towards higher sampling rates, where the performance improvement over LB-VD reaches up to $1$ dB. Towards lower sampling rates, with much fewer degrees of freedom to achieve good mask designs, the greedy method and LB-VD yield similar performance, which is expected\footnote{At low sampling rates, learning-based approaches will prioritize low-frequency phase encoding lines, as those contain most of the energy at a given frame.}.

Focusing on Figure \ref{fig:plot_cross}, we observe that comparing reconstruction algorithms using the model-based, model-driven approach Coherence-VD does not allow for a faithful comparison of their performance. In this case, the performance difference between KTH and ALOHA is marginal. However, when moving to the model-based, data-driven approach LB-VD, the difference becomes quantitatively clearer. Moving to the learning-based, data-driven approach SG-v2 makes the difference even more noticeable: ALOHA with its corresponding mask clearly outperforms KTF, and this conclusion could not be made by looking solely at reconstructions with VD-based masks. This trend is also reflected on Figure \ref{fig:psnr_cross}.

This is an important observation, as most works have developed new reconstruction algorithms that were then tested using random masks designed with criteria such as minimal coherence. However, our results suggest that this approach does not reflect the \textit{true} performance of a reconstruction method, and that one should look to optimize the mask and the reconstruction algorithm jointly.

The results on Figure \ref{fig:psnr_cross} highlight also an interesting asymmetry between the performance of the SG-v2 mask optimized on KTF (SG-v2-KTF) or ALOHA (SG-v2-ALOHA): the one from KTF is more centered towards low frequencies than the one of ALOHA, and also performs well, and better than LB-VD on ALOHA. However, the reverse is not true: the mask from ALOHA is very scattered throughout k-space and leads to a very poor performance when applied to KTF. This trend is also observed on multi-coil data, as highlighted on Table \ref{tab:cross_recon}. This suggests that ALOHA, not relying explicitly on $\ell_1$ sparsity, might succeed in exploiting varied information, from both low-frequency emphasizing masks from models promoting sparsity, but also higher frequency information.

\begin{remark}
    This point does not hold directly for deep learning methods, that will be discussed in later chapters. Indeed, the reconstruction methods that we use in these experiments are all based on an underlying \textit{model}, such as sparsity or low-rank. They are \textit{not} data-driven approaches to reconstruction.

    The question of what truly reflects the performance of a reconstruction method becomes even more complex in the case of deep learning-based reconstruction method, where the model is trained on a distribution of different sampling masks, making reconstruction and sampling even more intricate. We will come back to this question in Chapter \ref{ch:rl_mri}.
\end{remark}


\begin{remark}
    The results in the sequel are obtained with the SG-v1 algorithm, rather than SG-v2. Recall that SG-v1 already accelerated G-v1 by a factor $60$, and that we did not observe, in our experiments, a significant performance difference between SG-v1 and SG-v2\footnotemark.
\end{remark}
\footnotetext{{The main reason of having these experiments on SG-v1 is that they were obtained from an earlier set of results. However, we believe that they remain insightful on the behavior of sLBCS, highlighting not only application on different settings, but we expect that such insights would also carry out to SG-v2.}}
\FloatBarrier
\textbf{Cross-performances of performance measures.} Up to here, we used PSNR as the performance measure, and we now compare it with the results of the greedy algorithm paired with SSIM, a metric that more closely reflects perceptual similarity \citep{wang2004image}. For brevity, we only consider ALOHA in this section. In the case where we optimized for SSIM, we noticed that unless a low-frequency initial mask is given, the reconstruction quality would mostly stagnate. This is why we chose to start the greedy algorithm with $4$ low-frequency phase encodes at each frame in the SSIM case.

\begin{figure}[!t]
    \centering
    \includegraphics[width=.75\linewidth]{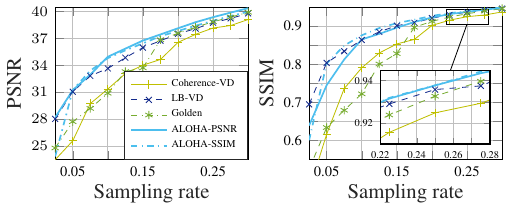}
    \caption{PSNR and SSIM as a function of sampling rate for ALOHA, comparing the SG-v1 results optimized for PSNR and SSIM with the three baselines, averaged on $4$ testing images of size 152$\times$152$\times$17.}\label{fig:plot_ssim}
\end{figure}

The reconstructions for PSNR and SSIM are shown on Figure \ref{fig:plot_ssim}, where we see that the learning-based masks outperform the baselines across all sampling rates except at $2.5\%$ in the SSIM case. The quality of the results is very close for both masks, but each tends to perform slightly better with the performance metric for which it was trained. The fact that the ALOHA-SSIM result at $2.5\%$ has a very low SSIM is due to the fact that we impose $4$ phase encodes across all frames, and the resulting sampling mask at $2.5\%$ is a low pass mask in this case.

\begin{figure}[!t]
    \centering
    \includegraphics[width=.7\linewidth]{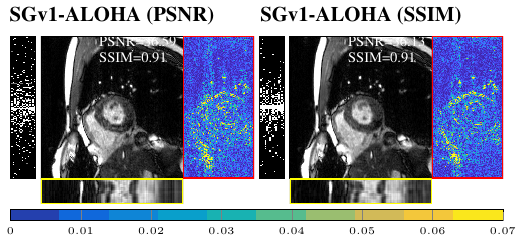}
    \caption{Comparison of the sampling masks optimized for PSNR and SSIM with ALOHA, at $15\%$ sampling. The images and masks can be compared to those of Figure \ref{fig:plot_cross}, as the settings are the same.} \label{fig:ssim}
\end{figure}

A visual reconstruction is provided in Figure \ref{fig:ssim}, we see that there is almost no difference in reconstruction quality, and that the masks remain very similar. Overall, we observe in this case that the performance metric selection does not have a dramatic effect on the quality of reconstruction, and our greedy framework is still able to produce masks that outperform the baselines when optimizing SSIM instead of PSNR.


\begin{figure}[!t]
    \centering
    \includegraphics[width=.7\linewidth]{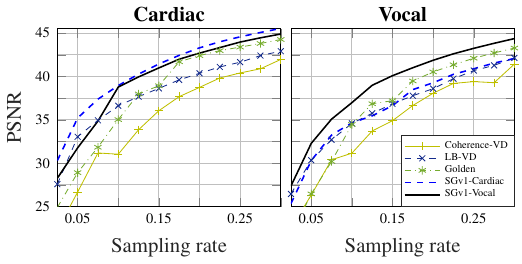}
    \caption{PSNR as a function of sampling rate for KTF, comparing SG-v1 with both baselines, averaged on $2$ testing images for both cardiac and vocal data sets of size 256$\times$256$\times$10.}\label{fig:psnr_anat}
\end{figure}

\textbf{Experiments with different anatomies.} In these experiments, we consider both the single coil cardiac dataset and the vocal imaging dataset, both of size $256\times 256 \times 10$. The cardiac dataset was trained on $5$ samples and tested on $2$, using only the first ten frames of each scan, whereas the vocal one used $2$ training samples and $2$ testing samples. In this setup, the k-space of the cardiac dataset tends to vary more from one sample to another than the vocal one, making the generalization of the mask more complicated. This issue would require more training samples, but imposing SG-v1 algorithm to start with $4$ central phase encoding lines on each frame was found to be sufficient to acquire the peaks in the k-space across the whole dataset. \textit{SGv1-Cardiac} refers to the stochastic greedy algorithm using cardiac data, and \textit{SGv1-Vocal} is its vocal counterpart. The algorithm used a batch of size $k = 64$ at each iteration\footnote{This follows the rule of thumb of using $25\%$ of the number of lines in a given frame for $k$.}, and the results were obtained using only KTF.

\begin{figure}[!ht]
    \centering
    \vspace{1.5cm}
    \includegraphics[width=.6\linewidth]{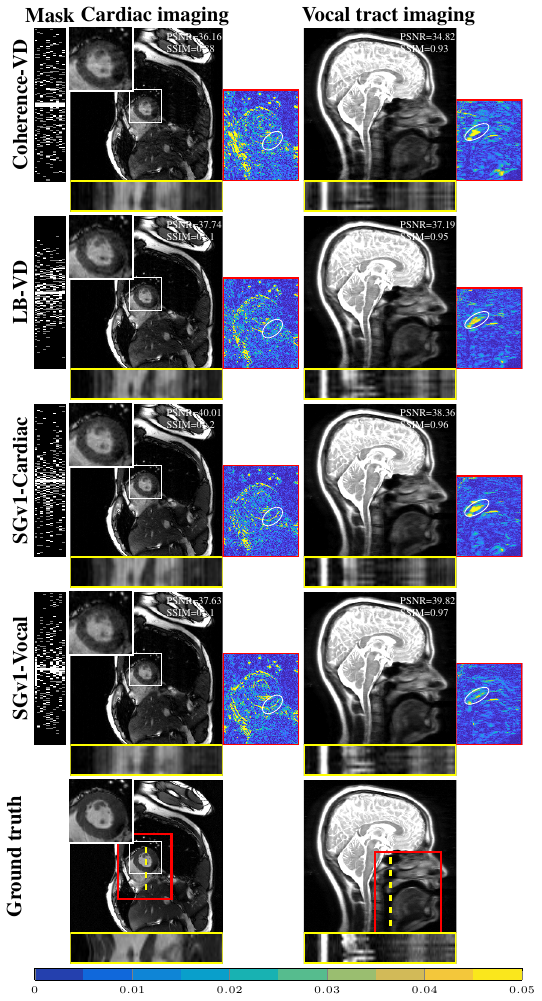}
    \caption{Reconstruction for KTF at $15\%$ sampling for the cardiac and vocal anatomies of size 256$\times$256$\times$10. Figures showing different frames for the vocal and cardiac images are available in Figures \ref{fig:cardiac} and \ref{fig:vocal}.}\label{fig:plot_cross_anat}
    \vspace{1.5cm}
\end{figure}

The results are reported on the Figures \ref{fig:psnr_anat} and \ref{fig:plot_cross_anat}, and we see that, for the both datasets, the greedy approach provides superior results against VD sampling methods across all sampling rates. It is striking that, in this setting, the SG-v1 approach outperforms even more convincingly all the baselines, and the LB-VD approach, in this case, is outperformed by more than $2$dB by SG-v1, where it remained very competitive in the other settings. This difference is clear in the temporal fidelity of both reconstructions on Figure \ref{fig:plot_cross_anat}, where we see that the LB-VD approach loses sharpness and accuracy compared to SG-v1.

\textbf{Comparison across anatomies.}\label{sec:anatomies} The main complication coming from applying the masks across anatomies is that the form of the k-space might vary heavily across datasets: the vocal spectrum is very sharply peaked, while the cardiac one is much broader.  Comparing the cross-performances on Figures \ref{fig:plot_cross_anat}, we see that the and \textit{SGv1-vocal} masks generalizes much better on the cardiac datasets than the other way around. This can be explained from the differences in the spectra: the cardiac one being more spread out, the cardiac mask less faithfully captures the very low frequencies of the k-space, which are absolutely crucial to a successful reconstruction on the vocal dataset, thus hindering the reconstruction quality. Also, we see that it is important for the trained mask to be paired with its anatomy to obtain the best performance.

\begin{figure}[!ht]
    \centering
    \includegraphics[width=.95\linewidth]{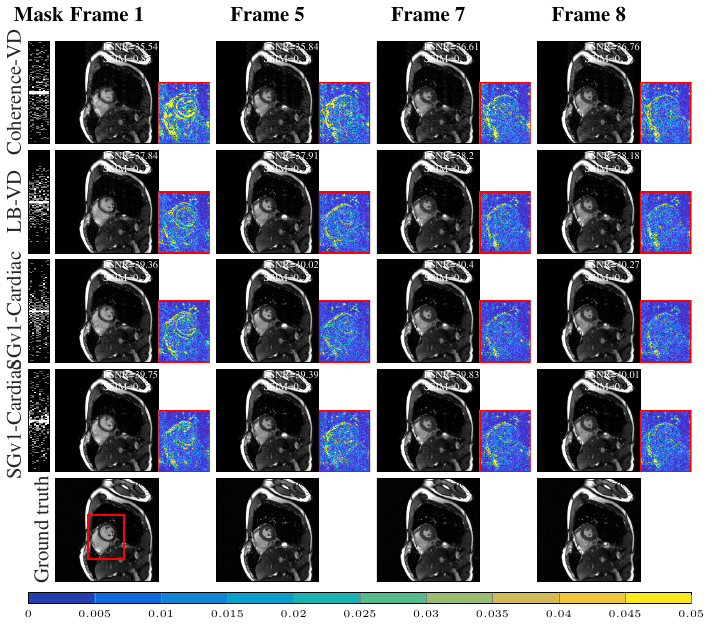}
    \caption{Reconstruction with KTF \citep{jung2009k} at $15\%$ sampling rate for the cardiac anatomy of size 256$\times$256$\times$10. It unfolds the temporal profile of Figure \ref{fig:plot_cross_anat}. The PSNR and SSIM displayed are computed for each image individually, and the overall PSNR for each image is the one of Figure \ref{fig:plot_cross_anat}. The ground truth is added at the end of each line for comparison.}\label{fig:cardiac}
\end{figure}

\begin{figure}[!ht]
    \centering
    \includegraphics[width=.95\linewidth]{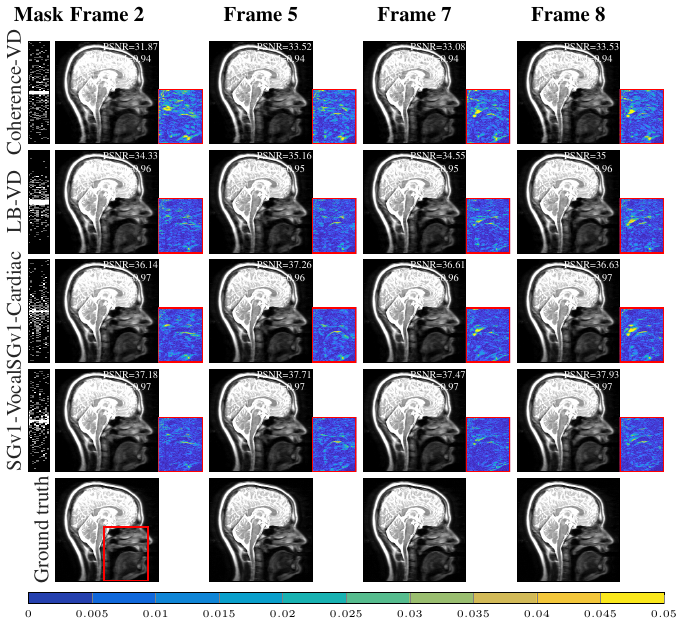}
    \caption{Reconstruction with KTF at $15\%$  \citep{jung2009k} sampling rate for the vocal anatomy of size 256$\times$256$\times$10. It unfolds the temporal profile of Figure \ref{fig:plot_cross_anat}. The PSNR and SSIM displayed are computed for each image individually, and the overall PSNR for each image is the one of Figure \ref{fig:plot_cross_anat}. The ground truth is added at the end of each line for comparison. }\label{fig:vocal}

\end{figure}

\textbf{Additional visual reconstructions for cardiac and vocal dataset.} We show in Figures \ref{fig:cardiac} and \ref{fig:vocal} reconstruction at different frames which provide clearer visual information to the quality of reconstruction compared to the temporal profiles.

For these images, the PSNR and SSIM are computed with respect to each individual frame, showing the quality of the reconstruction in a much more detailed fashion than before, where we considered each dynamic scan as a whole. Generally, we as previously observed, the mask trained for a specific anatomy will most faithfully capture the sharp contrast transitions in the dynamic regions of the images. For the vocal images, we see that sampling the first frame more heavily is important in order to avoid having a very large PSNR discrepancy, as observed for the other masks.  The PSNR remains quite stable across the frames otherwise.

\newpage
\textbf{Noisy experiments.}
In order to test the robustness of our framework to noise, we artificially added bivariate circularly symmetric complex random Gaussian noise to the normalized complex images, with a standard deviation $\sigma = 0.05$ for both the real and imaginary components. We then tested to see whether the greedy framework is able to adapt to the level of noise by prescribing a different sampling pattern than in the previous experiments.

\begin{figure}[!t]
    \centering
    \includegraphics[width=.7\linewidth]{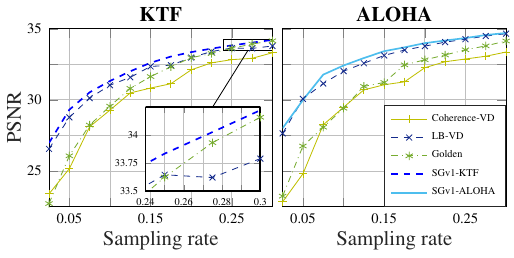}
    \caption{PSNR as a function of sampling rate for both reconstruction algorithms considered, comparing SG-v1 with the three baselines, averaged on $4$ noisy testing images of size 152$\times$152$\times$17. The PSNR is computed between the denoised reconstructed image and the original (not noisy) ground truth.}\label{fig:psnr_noisy}
\end{figure}

\begin{figure}[!t]
    \begin{minipage}[t]{.48\linewidth}
        \centering
        \vspace{1.4cm}
        \includegraphics[width=\linewidth]{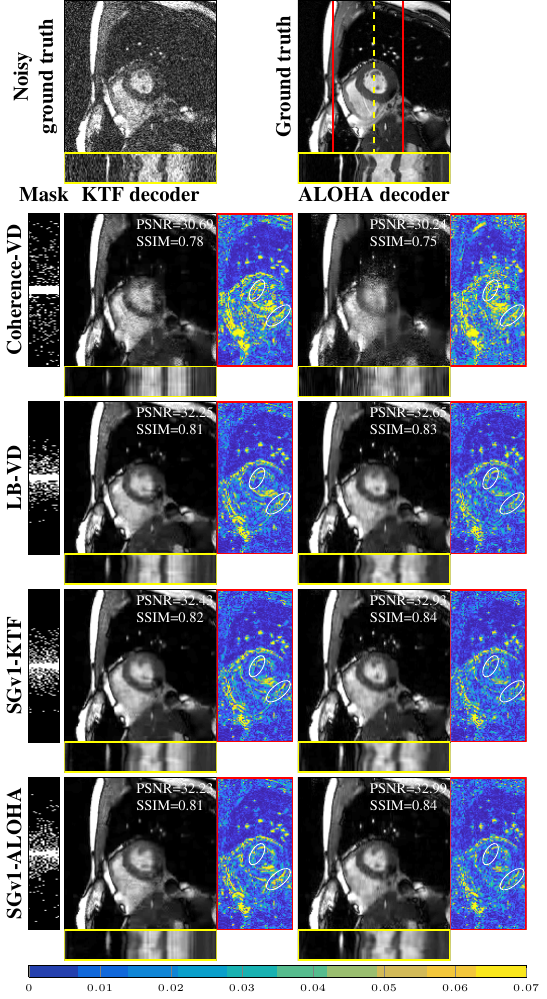}
        \caption{Reconstructed denoised version from the noisy ground truth on the first line, at $15\%$ sampling. The PSNR is computed with respect to the original ground truth on the top right.}\label{fig:plot_noisy}
    \end{minipage}
    \begin{minipage}[t]{.48\linewidth}
        \centering
        \vspace{1cm}
        \includegraphics[width=\linewidth]{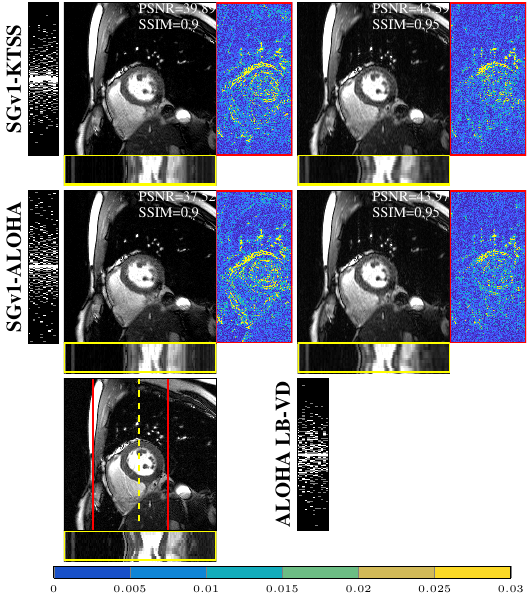}
        \caption{Reconstruction with KTSS \protect\citep{otazo2010combination} and ALOHA \citep{jin2016general} at $15\%$  sampling rate for a 4 coil parallel acquisition of cardiac cine size 256$\times$256$\times$12. The setting is otherwise similar as the one presented in Figure 5 of \citep{sanchez2019scalable}.}\label{fig:multi}
    \end{minipage}

    \vspace{1.3cm}
\end{figure}

We chose to use V-BM4D \citep{maggioni2012video} as denoiser with its default suggested mode using Wiener filtering and low-complexity profile, and provided the algorithm the standard deviation of the noise as the denoising parameter. The comparison between the fully sampled denoised images and the original ones yields an average PSNR of $24.95$ dB across the whole dataset. Due to the fact that none of the reconstruction algorithms that we used have a denoising parameter incorporated, we simply apply the V-BM4D respectively to the real and the imaginary parts of the result of the reconstruction. The results that we obtain are presented on the Figures \ref{fig:psnr_noisy} and \ref{fig:plot_noisy}.

It is interesting to notice on Figure \ref{fig:plot_noisy} that the sLBCS framework outperforms the baselines that are not data-driven by a larger margin than in the noiseless case, and this is again especially true at low sampling rates. In this case however, the difference between SG-v1 and LB-VD methods is much smaller, and this might be explained by the fact that noise corrupts the high frequency samples, and thus the masks concentrate more around low-frequencies, leaving less room for designs that substantially differ.

We see a clear adaptation of the resulting learning-based mask, as shown by comparing Figures \ref{fig:plot_cross} and \ref{fig:plot_noisy}: the masks SGv1-KTF and SGv1-ALOHA, which are trained on the noisy data, are closer to low-pass masks, due to the high-frequency details being lost to noise, and hence, no very high frequency samples are added to the mask.

Also, notice than even if the discrepancy in PSNR is only around $0.8-1$ dB between the golden ratio sampling and the optimized one, the temporal details are much more faithfully preserved by the learning-based approach, which is crucial in dynamic applications. The inadequacy of coherence-based sampling is highlighted in this case, as very little temporal information is captured in the reconstruction with both decoders. Also, for both decoders, there is a clear improvement on the preservation of the temporal profile when using learning-based masks compared to the baselines; the improvement of the SGv1-ALOHA mask of around $3$dB also shows how well our framework is able to adapt to this noisy situation, whereas Coherence-VD yields results of unacceptable quality.

\subsection{Multicoil experiment}
We now illustrate the applicability of our SG-v1 algorithm to multicoil data, a setting much more representative of routine clinical practice than single coil.

For the multicoil experiment, we used the previously described cardiac dataset, but we did not crop the images. We took the first $12$ frames for all subjects, and selected $4$ coils that cover the region of interest. Each image was then normalized in order for the resulting sum-of-squares image to have at most unit intensity.  When required, the coil sensitivities were self-calibrated according to the idea proposed by \citet{feng2013highly}, which averages the signal acquired over time in the k-space and subsequently performs adaptive coil combination \citep{walsh2000adaptive,griswold2002use}.

The advantage of using self-calibration is that the greedy optimization procedure can simultaneously take into account the need for accurate coil estimation as well as accurate reconstruction, thus potentially eliminating the need for a calibration scan prior to the acquisition. A more complete discussion of the accuracy of self-calibrated coil sensitivities is presented in \citet{feng2013highly}.

We used k-t SPARSE-SENSE \citep{otazo2010combination} and ALOHA \citep{jin2016general} for reconstruction. While the first requires coil sensitivities, the second reconstructs the images directly in k-space before combining the reconstructed data.
\begin{figure}[!t]
    \centering
    \includegraphics[width=.75\linewidth]{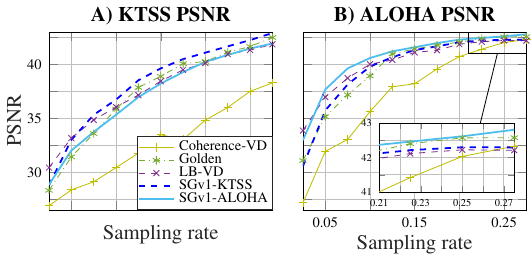}
    \caption{PSNR as a function of sampling rate for KTSS \protect\citep{otazo2010combination} and ALOHA \citep{jin2016general} in the multicoil setting, comparing SG-v1 with the coherence-VD \citep{lustig2007sparse}, LB-VD  and golden ratio Cartesian sampling \protect\citep{li2018dynamic}, averaged on 4 testing images of size 256$\times$256$\times$12 with 4 coils. }\label{fig:psnr_multi}
\end{figure}

\subsection{Large scale static results}\label{sec:fastMRI_sLBCS}
\begin{figure}[!t]
    \centering
    \begin{minipage}[c]{.55\linewidth}
        \includegraphics[width=\linewidth]{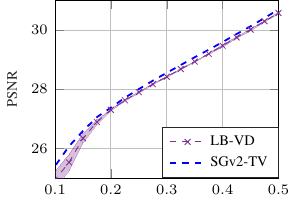}
    \end{minipage}\hfill
    \begin{minipage}[c]{.4\linewidth}
        \caption{PSNR as a function of the sampling rate for TV, averaged on the 13 most central slices of the fastMRI validation set \citep{zbontarFastMRIOpenDataset2019} (2587 slices). SGv2 outperforms LB-VD over all sampling rates.}\label{fig:psnr_fastmri}
    \end{minipage}
\end{figure}

This last experiment shows the scalability of our method to very large datasets. We used the fastMRI dataset \citep{zbontarFastMRIOpenDataset2019} consisting of knee volumes  and trained the mask for reconstructing the $13$ most central slices of size $320\times 320$, which yielded a training set containing 12\,649 slices. For the sake of brevity, we only report computations performed using total variation (TV) minimization with NESTA \citep{becker2011nesta}. For mask design, we used the SG-v2 method with $k=80$ and $l=20$ (2\,500 fewer computations compared to G-v1). The LB-VD method was trained using 80 representative slices and optimizing the parameters with a similar computational budget as SG-v2. The result on Figure \ref{fig:psnr_fastmri} shows a uniform improvement of our method over the LB-VD approach.

\section{Experiments - Lazy LBCS}\label{s:exp_llbcs}
\textbf{Dataset.} Our experiments were carried out on two datasets.
\begin{itemize}
    \item \textit{Brain data.} The data were acquired on a 3T MRI system (Siemens), and the protocols were approved by the local ethics committee. The dataset consists of 2D T1-weighted brain scans of seven healthy subjects, which were scanned with a FLASH pulse sequence and a 12-channel receive-only head coil.  In our experiments, we use 4 different slices {of sizes 256$\times$256} from two subjects as training data and 100 slices from 5 subjects as test data. To speed up the computations, we compressed our $12$-channel data into $4$-channel data.
    \item \textit{Knee data.} The data were acquired by a 3T system (GE Healthcare) using FSE CUBE sequence with the proton density weighting. The number of channels is 8 and the matrix size is 320 $\times$ 320. We used the data of 8 subjects as training data and remaining 12 subjects as test data. This dataset is publicly available at \textit{http://mridata.org}.
\end{itemize}
We used PSNR as the image quality metric $\eta$, but also reported the results evaluated on SSIM \citep{wang2004image}.

\textbf{Reconstruction algorithms.} We consider the following decoders: Basis Pursuit (BP) \citep{donoho2006compressed}, Total Variation (TV), SENSE \citep{pruessmann1999sense}, and ALOHA \citep{jin2016general}. For BP, we let the sparsifying transform to be the Shearlet transform \citep{kutyniok2016shearlab}. For SENSE, we used ESPIRiT \citep{uecker2014espirit} method to estimate coil sensitivities.

Basis pursuit is a method that directly minimizes the $\ell_1$ norm of the image in a sparse domain. A similar type of convex optimization formulation is {\em total variation} (TV) minimization.
SENSE is a method that requires the knowledge of coil sensitivities to retrieve an unaliased image from parallel acquisitions \citep{pruessmann1999sense}. We use its implementation in the BART toolbox \citep{uecker2015berkeley} to iteratively solve 
\begin{equation}
    \min_\z \|\vy_\omega - \Po \mF \mS \z\|_2^2 + \lambda \|\mW\z\|_1 \label{eq:opti}
\end{equation}
where $\vy_\omega \in \mathbb{C}^{CP}$ is the coil-wise subsampled Fourier measurements, $\mS =  \begin{bmatrix} \mS_1 \ldots \mS_C \end{bmatrix}^T \in \R^{CP\times P}$ are the stacked coil sensitivities,  $\mP_\omega \mF$ is the subsampled Fourier transform applied coil-wise to the product of the image with the stacked coil sensitivities and $\mW$ is a sparsifying operator. Recall that the multicoil formulation was discussed in greater depth in Equation \ref{eq:acquisition_parallel_compact}.
Other compressed sensing multicoil algoritms include \citep{uecker2014espirit,chaari2011wavelet}. Recall that ALOHA exploits the duality between sparse signals and their transform domain representations cast as a low-rank Hankel matrix, and transforms the problem into a low-rank matrix completion.

\subsection{2D multicoil setting}
\begin{figure}[!ht]
    \hspace{4mm}
    \begin{minipage}[c]{.55\linewidth}
        \centering
        \includegraphics[width=\linewidth]{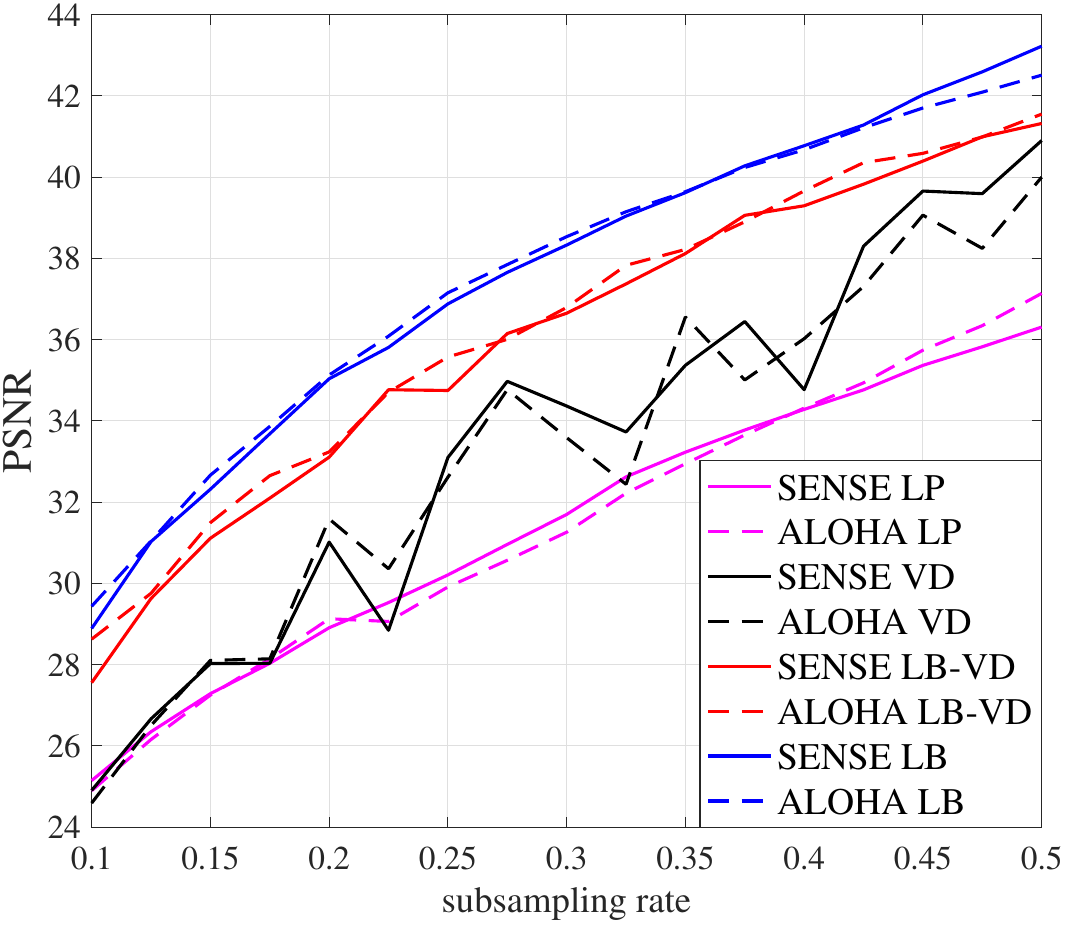}
    \end{minipage}
    \hfill
    \begin{minipage}[c]{0.3\linewidth}
        \centering
        \caption{PSNR performances of various masks averaged over 100 test images: LP (low-pass mask), VD (variable density), LB-VD and lLBCS (LB).}\label{fig:figu1}
    \end{minipage}
\end{figure}

We first compare the performances of lLBCS with various baselines.  As it can be seen in Figure \ref{fig:figu1} for a range of subsampling rates, the coherence-based VD masks can perform better than low-pass (LP) mask. However, they have weaker performance compared to the parameter sweep with PSNR as the objective function (LB-VD). This is expected as the LB-VD masks exploit the knowledge of the training set, whereas the coherence-based VD masks do not. The best performance is provided by our proposed approach, lLBCS (LB in the results). Indeed, the greedy LB algorithm is free from the constraint of a parametric distribution, which allows the data to decide which mask is best suited to a given decoder. Also, the VD-based masks result in artifacts especially visible in the ALOHA reconstructions that are suppressed in the LB-case.

\textbf{Cross-performances of single and multi-coil masks.}

\begin{figure}[!ht]
    \begin{minipage}[c]{.6\linewidth}
        \centering
        \includegraphics[width=\linewidth]{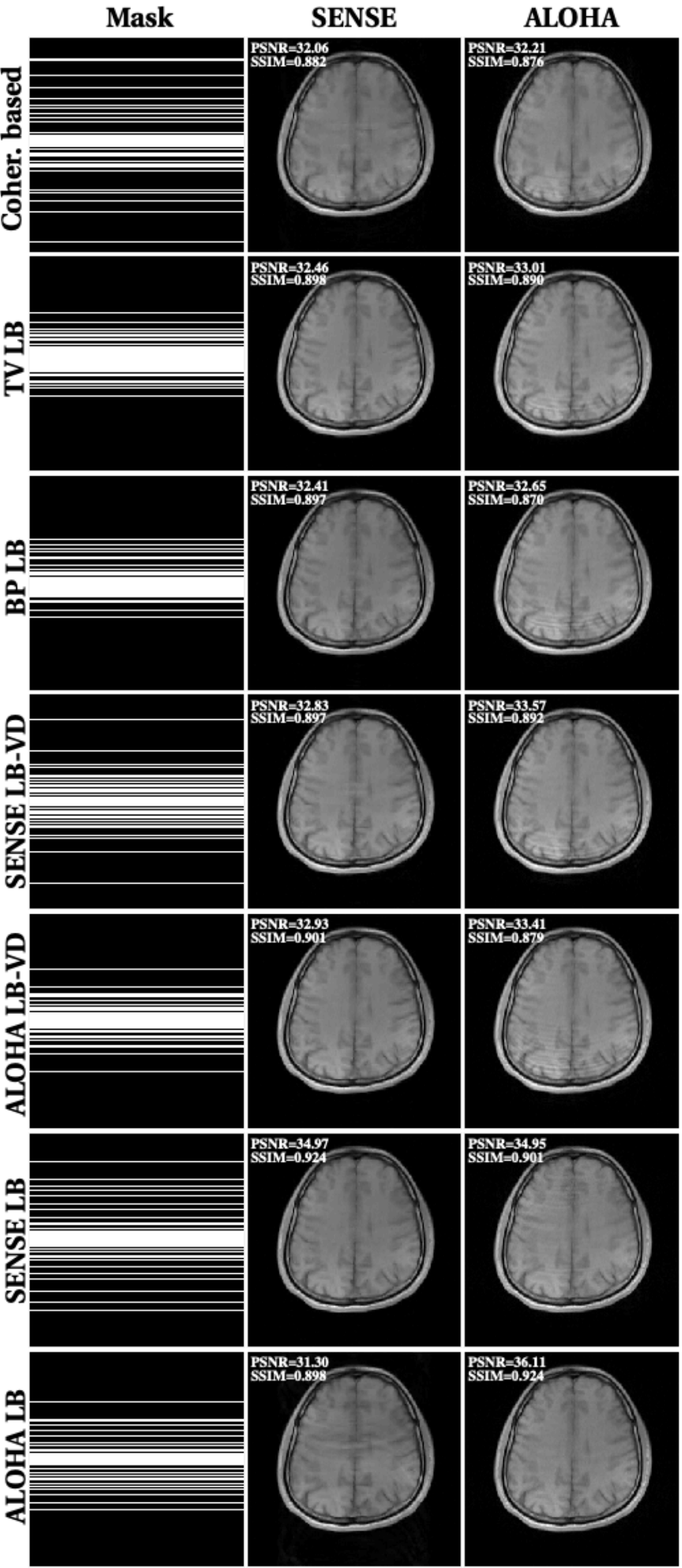}
        (a)
    \end{minipage}
    \hfill
    \begin{minipage}[c]{0.3\linewidth}
        \centering
        \includegraphics[width=0.7\linewidth]{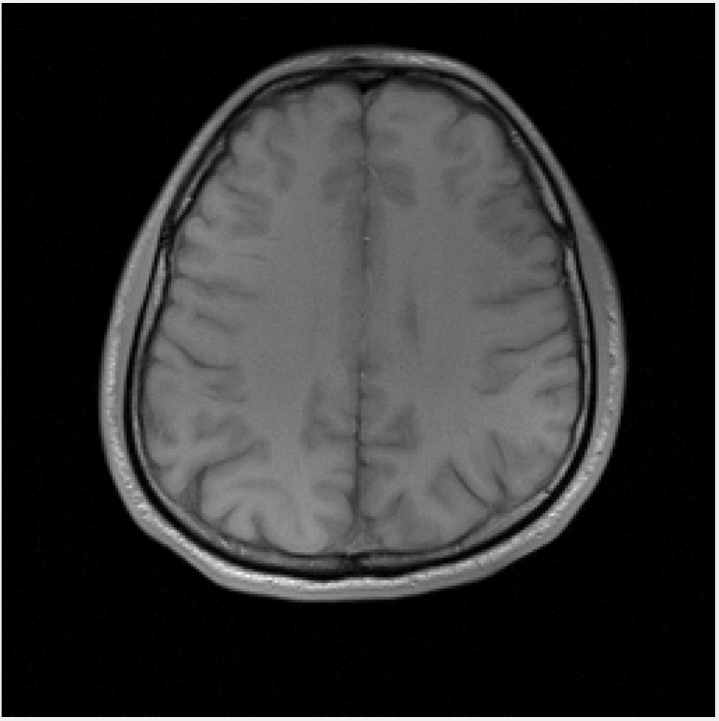}

        (b)

        \vspace{\fill}
        \caption{Optimized masks and example reconstructions under SENSE and ALOHA decoding at 20\% sampling rate. See Figure \ref{fig:figu1} for ground truth.} \label{fig:recons}
    \end{minipage}
\end{figure}

In Figure \ref{fig:recons}, we have also included single coil masks (TV and BP) obtained from using LBCS on single coil data. They result in inferior performance when used with the multi-coil SENSE and ALOHA reconstruction algorithms. Compared to these single coil masks, the optimal multi-coils masks are found to be more spread in k-space. As expected, one obtains the best performances in terms of visual quality and performance metrics when the LB-mask is paired with the reconstruction algorithm for which it was trained.  In Table \ref{tab:cross_recon}, we provide an average cross-performance of single/multi-coil masks and reconstruction algorithms. We see that it is important to optimize single and multi-coil masks separately as they are quite different from each other, which was also observed by \citet{haldar2019oedipus}.  

\begin{table}
    \centering

    \begin{tabular}{>{\bfseries}lcccc}\hline
        \toprule
        \multirow{2}{*}{Mask} & \multicolumn{2}{c}{\textit{Single-coil}} & \multicolumn{2}{c}{\textit{Multicoil}}                           \\
        \cmidrule(l){2-3}\cmidrule(l){4-5}
                              & TV                                       & BP                                     & SENSE      & ALOHA      \\
        \midrule
        Coherence             & 29.04                                    & 29.60                                  & 33.10      & 32.61      \\
        Low pass              & 30.59                                    & 30.78                                  & 30.21      & 29.91      \\
        \midrule
        TV LB                 & {\bf32.82}                               & 34.23                                  & 34.56      & 35.17      \\
        BP LB                 & 32.90                                    & {\bf34.27}                             & 35.02      & 35.07      \\
        SENSE LB              & 31.51                                    & 31.82                                  & {\bf36.88} & 36.61      \\
        ALOHA LB              & 29.00                                    & 26.95                                  & 30.73      & {\bf37.15} \\
        \bottomrule
    \end{tabular}
    \caption{\label{tab:cross_recon} Cross performance of single and multi-coil masks on various reconstruction algorithms, at 20\% subsampling rate, averaged over 100 test slices.  }
\end{table}

\subsection{3D multicoil setting}
\label{ssec:subhead}

\begin{figure}[!ht]
    \centering
    \includegraphics[width=0.5\textwidth]{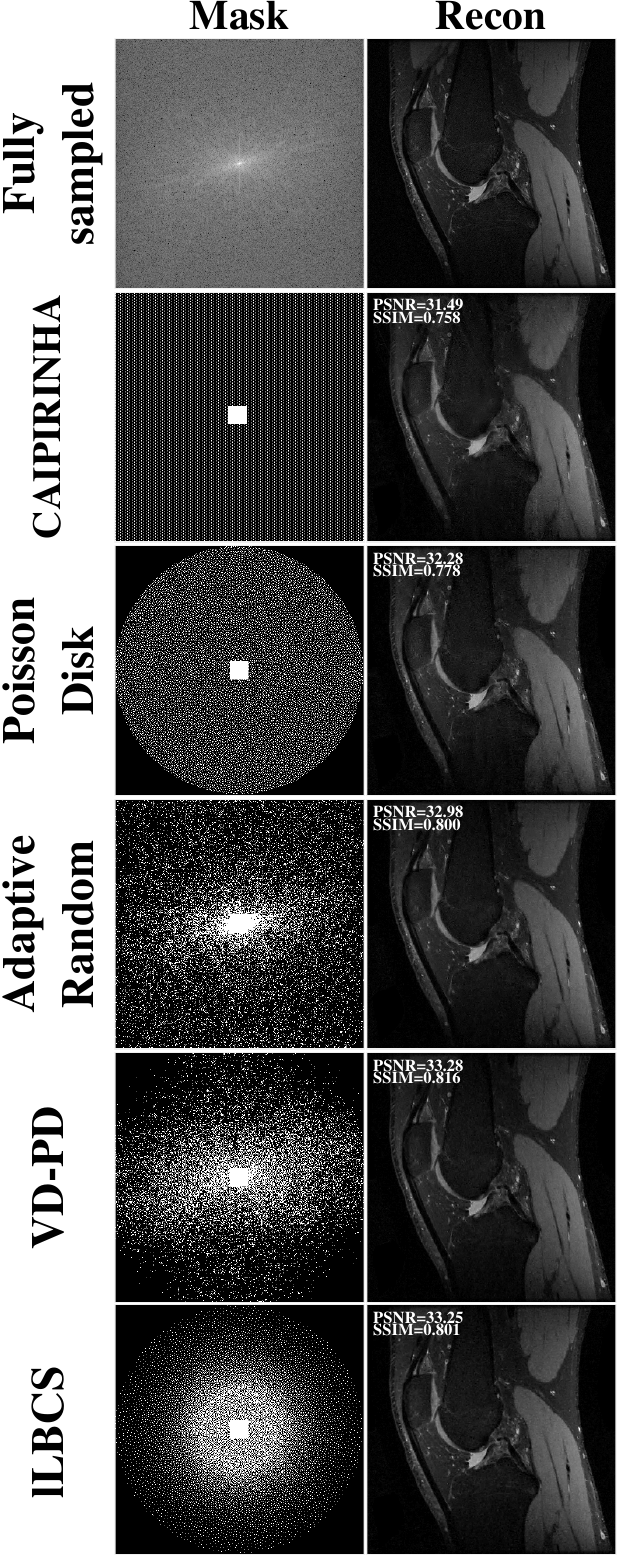}
    \caption{Optimized masks and example reconstructions of knee images under SENSE decoding at 6-fold acceleration. }
    \label{fig:recons_3D}
\end{figure}

In this section we demonstrate the effectiveness of lLBCS in providing masks for 3D MRI by comparing it to common 3D MRI subsampling masks: Controlled Aliasing in Parallel Imaging Results in Higher Acceleration (CAIPIRINHA) \citep{breuer2006controlled}, Poisson disc (PD) sampling \citep{bridson2007fast} and its variable density variant (VD-PD) \citep{vasanawala2011practical}, and adaptive random sampling method \citep{knoll2011adapted}. For brevity, we used only  the SENSE algorithm as the multi-coil reconstruction algorithm and for all masks we took the central 24 $\times$ 24 region as calibration region to estimate the coil sensitivities. As seen in Figure \ref{fig:recons_3D}, the mask given by lLBCS provides superior image quality compared to the benchmarks and comparable result to more recent VD-PD sampling whose polynomial decay parameter is optimized by a grid search (LB-VD).

\begin{figure}[!t]
    \centering
    \begin{subfigure}{0.48\linewidth}
        \centering
        \includegraphics[width=0.8\linewidth]{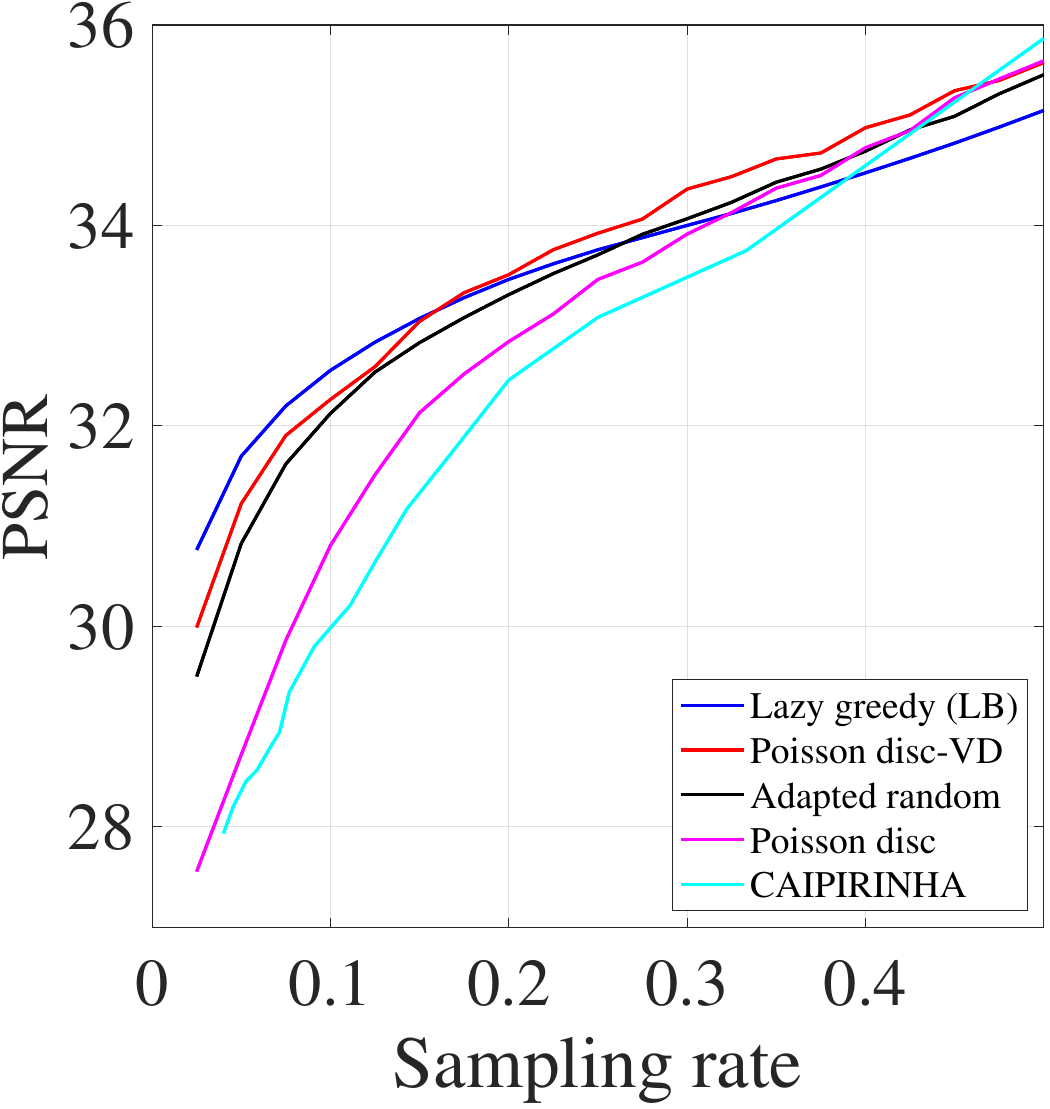}
    \end{subfigure}
    \hfill
    \begin{subfigure}{0.48\linewidth}
        \centering
        \includegraphics[width=0.8\linewidth]{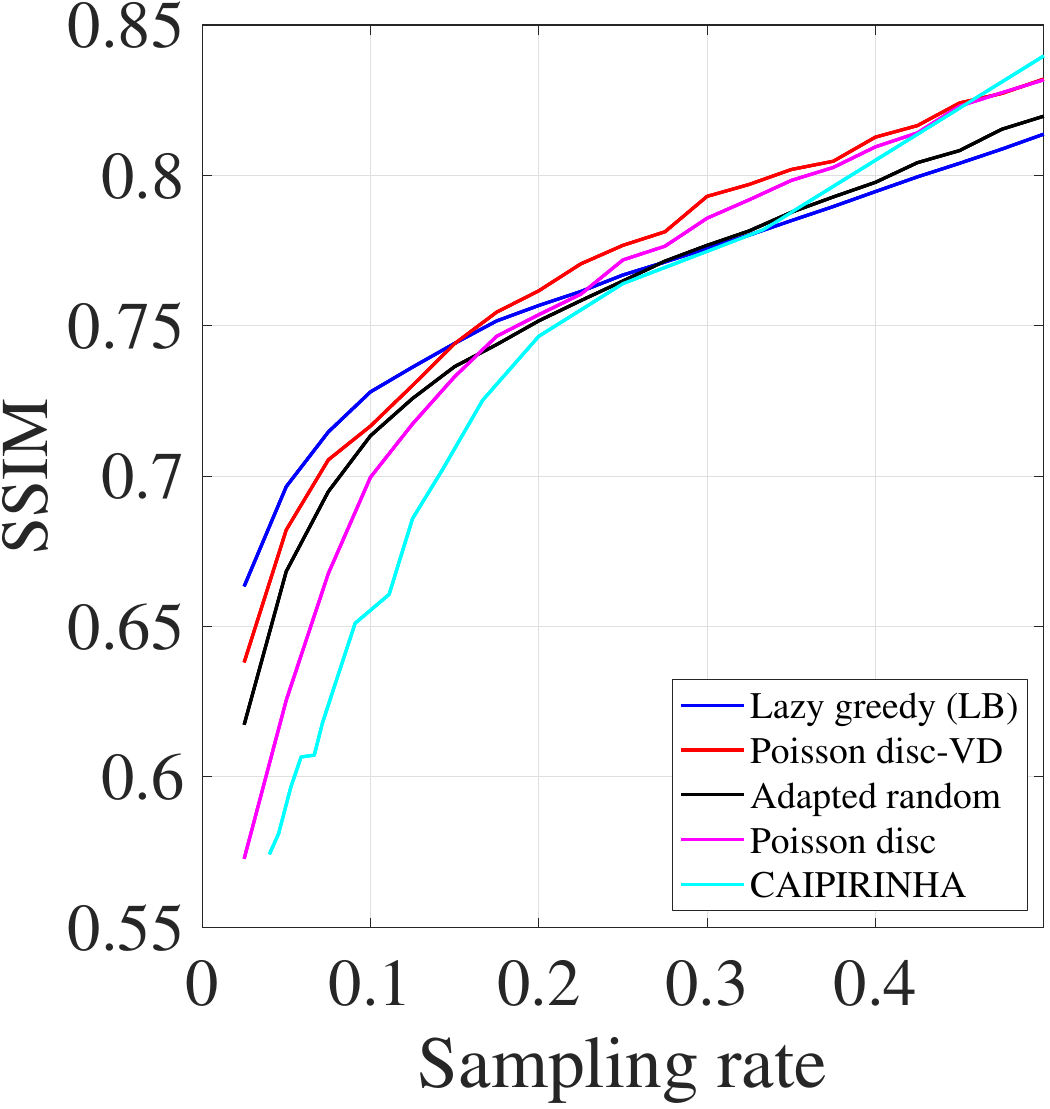}
    \end{subfigure}
    \caption{PSNR and SSIM performance of various masks averaged over 120 testing 3D volumes. The lLBCS model was trained using PSNR.}\label{fig:3D_performance}
\end{figure}

Although this trend holds for low sampling rates, Figure \ref{fig:3D_performance} shows that the baselines catch up around $15\%$ sampling and then outperform lLBCS. The trend is similar on both PSNR and SSIM evaluation, and several observations can be made. While our method does not uniformly outperform other baselines like in other cases, it remains competitive in the region of interest for such methods, namely low sampling rates. Indeed, as acquiring these 3D volumes required 40 minutes of scanning \citep{sawyer2013creation}, one aims at achieving gains by high accelerations. For instance, six-fold acceleration has been convincingly investigated in a prospective study for 3D knee MRI by \citet{fritz2016six}.

\begin{figure}[!t]
    \centering
    \includegraphics[width=0.6\linewidth]{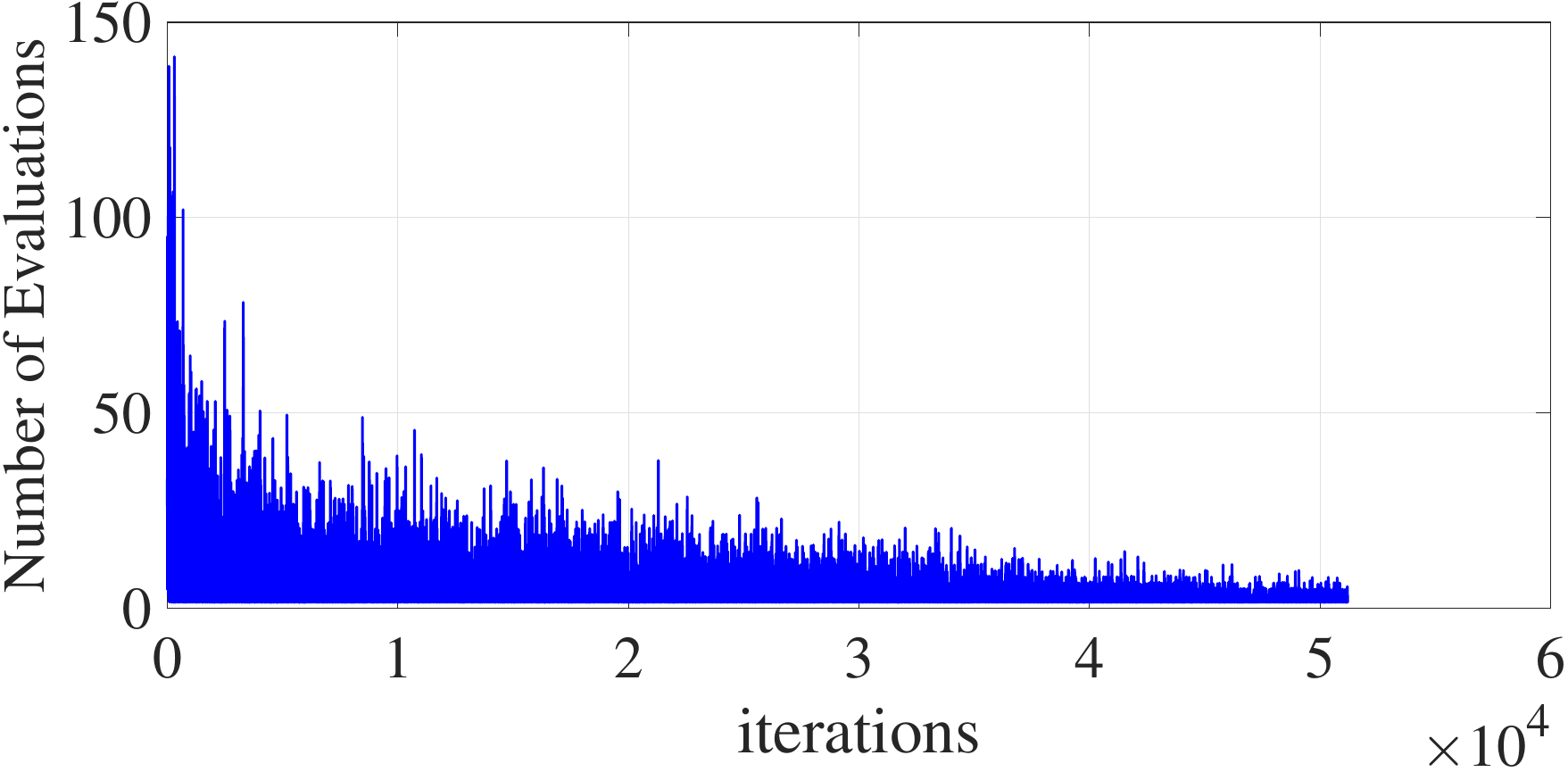}
    \caption{Number of lazy evaluations per round of the lazy LBCS algorithm, until $50\%$ sampling rate ($50624$ locations acquired, initialization with $24\times 24$ centermost frequencies).}\label{fig:lazy_eval}
\end{figure}

Figure \ref{fig:lazy_eval} shows the number of evaluations done per round of lLBCS, i.e. the number of times where the list of upper bounds $\rho(\cdot)$ is updated before adding an element to the mask $\omega$ in Algorithm \ref{alg:llbcs}. Two observations can be made: first, after the initial, expensive, step of evaluation, there are remarkably few updates in lLBCS, averaging at $2.644$ per round, whereas the regular LBCS algorithm would have 76\,368.5, which amounts to the enormous speedup of a factor 28\,884. This testifies to the efficiency of the cost reduction achieved by lLBCS. However, and this is our second observation, it seems that the number of lazy evaluations are steadily decreasing throughout the iterations, and we hypothesize that the decrease of performance observed in Figure \ref{fig:3D_performance} can be attributed to the list of upper bounds getting increasingly unreliable as the number of iterations increase.

This suggests avenues to further improve the performance of lLBCS. The simplest approach would consist in periodically re-computing the list of upper bounds, to more accurately track the potentially useful sampling locations. Although this step would not be necessary for \textit{submodular} functions, it could benefit our objective, which does not exactly satisfy submodularity. Alternatively, one could also re-evaluate a couple of elements from the list randomly at each iteration, spreading the computational load over a larger number of iterations.

\section{Discussion and Conclusion}\label{s:lbcs_discussion}
In this chapter, we presented two algorithmic contribution that broaden the scale of applications of LBCS \citep{gozcu2019rethinking} beyond its original limits, by successfully scaling to settings where the number of candidates grows drastically compared to 2D Cartesian MRI, to large dataset and to methods that present long-running times, as in multicoil and 3D MRI.


Adapting these algorithms to non-Cartesian settings such as radial sampling would be of great interest to clinicians, as radial sampling has seen a large success in body MRI due to the resistance to motion that it offers, enabling free breathing imaging \citep{coppo2015free,feng2016xd}. Although golden angle sampling has been widely used \citep{feng2014golden}, we anticipate that adjusting the distribution of radial spokes more closely to the underlying data distribution would improve the overall quality of reconstructed images. We anticipate however that this would involve challenges in simulating a non-Cartesian acquisition and quantifying the improvement of adding new measurements to an image, but recent deep learning reconstruction methods could provide a solution to this issue \citep{ramzi2022nc}.

Both sLBCS and lLBCS could certainly be improved by endowing them with additional heuristics. sLBCS would likely benefit from candidate lines being drawn from a probability favoring low frequency locations rather than fully at random. lLBCS could benefit form periodically re-computing the list of upper bounds $\rho$. Recent works such as \citet{zibetti2020fast} also showed the benefits of heuristics for mask design. For instance, they bias the selection of candidates based on the current line-wise NMSE and avoid selected elements at complex-conjugated position in order to accelerate the optimization procedure.

This should however not be the main point that one takes away from this chapter. Our work illustrates the great benefit of moving from model-based, model-driven approaches for mask design to learning-based, data-driven ones: trying to maximize a performance metric on data (data-driven approach), rather than some abstract mathematical structure (model-driven approach) such as coherence enables to more efficiently use the limited available resources. Similarly, moving from model-based to learning-based approaches allows to remove the constraints on the shape of the masks that can be explored, and our results show that this greatly benefits the reconstruction quality. Stated differently, our results suggest that model-based and/or model-driven approaches \textit{limit} the performance of CS methods applied to MRI, through their underlying model.


\section*{Bibliographic note}
For the lazy LBCS work\footnote{G{\"o}zc{\"u}, B., Sanchez, T., and Cevher, V. (2019). Rethinking sampling in parallel MRI: A data-driven approach. In \textit{27th European Signal Processing Conference (EUSIPCO)}.}, the work is mostly due to Baran G{\"o}zc{\"u}. The author of this dissertation contributed to the formalization of the lazy LBCS algorithm and to part of the 2D multi-coil experiments. Convinced of the potential of LBCS, Volkan Cevher provided the drive to further validate and extend the method.

For the stochastic LBCS work\footnote{Sanchez, T., G{\"o}zc{\"u}, B., van Heeswijk, R. B., Eftekhari, A., Il{\i}cak, E., \c{C}ukur, T., and Cevher, V. (2020a). Scalable learning-based sampling optimization for compressive dynamic MRI. In \textit{ICASSP 2020 - 2020 IEEE International Conference on Acoustics, Speech and Signal Processing (ICASSP)}, pages 8584–8588.}, Ruud van Heeswijk acquired the cardiac data, Efe Il{\i}cak and Tolga \c{C}ukur provided the vocal tract data. Baran G{\"o}zc{\"u} was mostly involved in supervising the project. Armin Eftekhari provided Propositions \ref{prop:1} and \ref{prop:2}, discussed in the previous chapter.

\cleardoublepage
\chapter{On the benefits of deep RL in accelerated MRI sampling} \label{ch:rl_mri}

In parallel to the refinements to the Learning-based Compressive Sampling (LBCS) \citep{gozcu2018learning} approach that we studied in the previous chapter, the progress of deep learning (DL) applied to reconstruction motivated several researchers to turn their attention to the problem of optimizing sampling patterns in a learning-based, data-driven fashion.

In this Chapter, we wish to focus on the state-of-the-art (SotA) approaches that leverage deep reinforcement learning (RL) to perform adaptive mask design in a sequential fashion \citep{bakker2020experimental,pineda2020active}. These approaches are particularly relevant to us for two main reasons.
\begin{enumerate}
    \item These two SotA approaches can be viewed as very natural extensions of (s)LBCS, which can be interpreted as a very basic RL policy. 
          This makes it both easy and insightful to compare LBCS to these methods, as it will provide a baseline to quantify the benefits brought by moving from LBCS to deep RL.
    \item More sophisticated RL policies promise to address two fundamental limitations of LBCS: they can perform adaptive sampling and incorporate long-term planning in their decision.
\end{enumerate}

However, the SotA approaches of \citet{pineda2020active} and \citet{bakker2020experimental} draw seemingly conflicting conclusions from their results. The work of \citet{pineda2020active} seems to indicate that long term planning could be the most important component in deep RL, as their results show that a non-adaptive, long term planning policy model trained on the dataset can perform as well as an adaptive, long-term planning policy. On the contrary, the contribution of \citet{bakker2020experimental} highlights the importance of adaptivity, as a greedy policy, that does not do long-term planning is found to closely match policies that do long term planning.

Our results synthesize this apparent conflict: \textit{We observe that sLBCS, a simple, easy-to-train method that does not rely on deep RL, does not attempt long term planning and is by definition not adaptive can perform as well as the state-of-the-art approaches of \citet{pineda2020active,bakker2020experimental}.}

This trend can be consistently observed on the fastMRI single-coil knee dataset \citep{zbontarFastMRIOpenDataset2019}, where we carry out evaluations across a variety of settings, whether full scale images, cropping the field of view, various mask designs used for training and various architectures. We observe that such small changes in the experimental pipeline can easily lead to different ordering of the RL methods and sLBCS in terms of performance or an invalidation of the conclusions drawn by the results. We refer to such occurrences as \textit{reversals}.
It is possible then to choose an experimental setting and performance metrics to support one's desired conclusion: that RL outperforms LBCS, that LBCS outperforms RL or that their difference is not significant.

Together with the observation that current SotA RL methods only add marginal value over the simple baselines at best, the work in this chapter highlights the need for further discussions in the community about standardized metrics, strong baselines, and careful design of experimental pipelines to evaluate MRI sampling policies fairly\footnote{The work in this chapter is based on the preprint: Sanchez, T.$^{*}$, Krawczuk, I.$^{*}$ and Cevher, V. (2021). On the benefits of deep RL in accelerated MRI sampling. \textit{Available at \url{https://openreview.net/pdf?id=fRb9LBWUo56}}. $^{*}$ denotes equal contribution.}.

\section{Reinforcement Learning to optimize sampling trajectories}

Before introducing how RL can be used to optimize sampling for MRI, let us briefly contextualize the works of \citet{jin2019self}, \citet{pineda2020active} and \citet{bakker2020experimental}.
\citet{jin2019self} were the first to attempt at using an RL-based approach to mask design, trying to train jointly a reconstruction model and a policy model through self-supervised learning. The approach is based on the AlphaGo model \citep{silver2017mastering}, a model that successfully learned to play go through self-play and that eventually ended-up the beating the world champion of Go. In this approach, the policy model is trained to learn the predictions done by a Monte-Carlo Tree Search (MCTS) \citep{coulom2006efficient}.

\citet{pineda2020active} leverage Double Deep Q-Networks (DDQN) \citep{van2016deep}, a state-of-the-art deep RL approach that is particularly suited to the problem at hand and that has recently shown to be a reliable and stable method. They train two variants of the model, one to perform non-adaptive sampling (Dataset-specific, DS-DDQN), and the other that is adaptive (Subject-specific, SS-DDQN). Their results show that DS-DDQN outperforms SS-DDQN in nearly all cases.

Finally, \citet{bakker2020experimental} took a different approach, leveraging the policy gradient theorem to enable the training of the policy network \citep[ch. 13]{sutton2018reinforcement}.  They both a greedy and a non-greedy (long horizon) policy, and find that, surprisingly, ``\textit{a simple greedy approximation of the objective leads to solutions nearly on-par with the more general non-greedy approach.}'' 

We summarize the features of the different models in Table \ref{tab:axes}, and for the reasons discussed above, we consider the method of \cite{pineda2020active} to be represented by their non-adaptive DS-DDQN, and the method of \citet{bakker2020experimental} to be represented by their greedy policy model.

\begin{table}[t]
    \centering
    \begin{tabular}{l|cc}
        \toprule
        \textbf{Sampling policy}                       & \textbf{Adaptive}      & \textbf{Long horizon}  \\
        \midrule
        LBCS \citep{gozcu2018learning}                 & \xmark                 & \xmark                 \\
        AlphaGo \citep{jin2019self}                    & \cmark                 & \cmark                 \\
        DDQN \citep{pineda2020active}                  & \xmark {\tiny(\cmark)} & \cmark                 \\
        Policy Gradient \citep{bakker2020experimental} & \cmark                 & \xmark {\tiny(\cmark)} \\
        \bottomrule
    \end{tabular}
    \vspace{1mm}
    \captionof{table}{Methods that will be considered in the paper. \cite{bakker2020experimental} considered fixed vs. adaptive and greedy vs. non-greedy, \cite{pineda2020active} considered data specific vs subject specific policies and also compared against greedy methods.}\label{tab:axes}
\end{table}

\subsection{A short RL primer}
Assume that we have an agent in an unknown environment, and that the agent can interact with the environment and potentially obtain rewards based on its actions. In reinforcement learning, we would like the agent to learn to maximize its cumulative reward from its interaction with the environment. This means that the agent should learn a strategy that allows it to adapt to its environment in order to maximize its reward. In Go, this could mean train the agent to win as many games as possible, or in MRI, this could amount to selecting the next line to acquire to maximize the reconstruction quality.

\begin{figure}[!ht]
    \centering
    \includegraphics[width=0.45\linewidth]{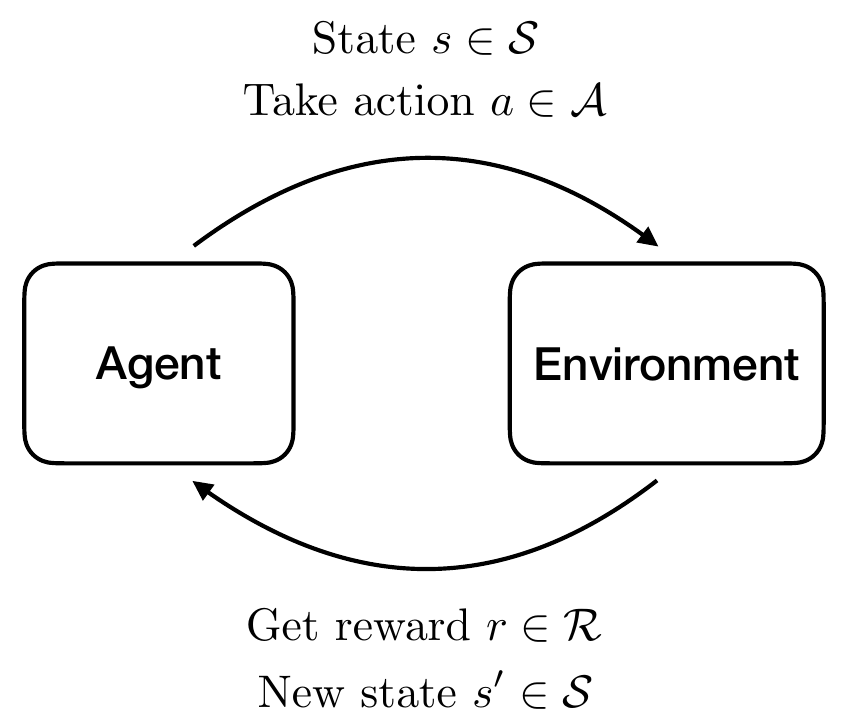}
    \caption{Illustration of an agent interacting with its environment.}
\end{figure}

Formally speaking, in RL, an agent acts in an \textit{environment} with an underlying model that may be known or not. The agent will stay in one of the many states $s \in \mathcal{S}$ of the environment, and take an action $a \in \mathcal{A}$, which allow it to move from its current state to another state $s'$. As a result of taking the action $a$, the agent receives a reward $r\in \mathcal{R}$ as feedback.

This process is repeated several times, and produces a sequence of states, actions and rewards. This results in a tuple
$$(S_1,A_1,R_2, S_2, \ldots, S_T)$$
where $T$ is the total number of steps taken. The states, actions and rewards are capitalized here as they refer to random variables. This tuple is referred to as an \textit{episode}. During an episode, the agent interacts with the environment in order to learn a better strategy. This strategy is described as a policy $\pi(s)$, which is a probability distribution over which actions taken in a state $s$ are likely to get the maximal cumulative reward. In order to quantify this, each state is associated with a value function $V_\pi(s)$, which predicts the expected amount of future reward that we would receive by acting following the current policy $\pi$. In RL, one generally aims at learning the policy and/or the value function.

\textbf{Modeling the environment.} In order to do that, we need to further describe the environment by a model. A model usually relies on two quantities, the \textit{transition probability} $P$ and the reward $R$. Starting in state $s$ and taking action $a$, the transition probability describe how likely it is that we end up moving to state $s'$ and obtaining reward $r$ as a result, namely
$$P(s',r|s,a) = \mathbb{P}[S_{t+1}=s', R_{t+1}=r | S_t = s, A_t = a]$$
where $\mathbb{P}$ describes here a probability. The reward function is defined as the expected reward for a given state action pair
$$R(s,a) = \mathbb{E}[R_{t+1}|S_t = s, A_t = a] = \sum_{r\in\mathcal{R}} r \sum_{s' \in \mathcal{S}} P(s',r|s,a)$$

\textbf{Value function and state-action function.} The value function aims at predicted the expected amount of future reward, that is commonly referred to as the \textit{return} and defined as
$$G_t \triangleq R_{t+1} + \gamma R_{t+2} + \ldots \gamma^{T-t-1} R_T = \sum_{k=t}^T \gamma^{k-1} R_{t+k}$$
where $\gamma \in [0,1]$ is called the \textit{discounting factor}.  The idea of the discounting factor is to penalize rewards in the future as a way to nudge the agent towards \textit{immediate} benefit. Note that the return $G_t$ always sums every future reward until the end of the episode, at time $T$.

We can then use the return to define formally the value function
\begin{equation}
    V_\pi(s) \triangleq \mathbb{E}_\pi[G_t|S_t =s]\label{eq:v_def}
\end{equation}
which means that the value of the state $s$ is the expected return given that we currently are in the state $S_t=s$. A related important quantity is the \textit{state-action function}
\begin{equation}
    Q_\pi(s,a) \triangleq \mathbb{E}_\pi [G_t|S_t=s, A_t=a]  \label{eq:q_def}
\end{equation}
which is similar to the value function except that it considers a state-action pair $(s,a)$. The two can be related as
\begin{equation}
    V_\pi(s) = \sum_{a\in \mathcal{A}} Q_\pi(s,a)\pi(a|s)\label{eq:value_to_q}
\end{equation}


\textbf{Markov Decision Processes.} Most RL problems can be expressed as Markov Decision Processes (MDPs). This implies in particular that the current state $S_t$ and action $A_t$ contain all the necessary information to future decisions, i.e.
$$\mathbb{P}(S_{t+1}|S_t,A_t) = \mathbb{P}(S_{t+1}|S_1, A_1 \ldots, S_t, A_t).$$
In other words, this means that the future $S_{t+1}$ is conditionally independent of the past ($S_1,A_1\ldots, S_{t-1},A_{t-1}$) given the present $S_t$, $A_t$. An MDP consists of five elements that we already encountered, namely $\mathcal{M} = \langle \mathcal{S}, \mathcal{A}, P, R, \gamma \rangle$. We have a set of states, a set of actions, the transition probability, the reward function and the discount factor. This description comprises everything that we need to tackle a RL problem.

\subsection{The special case of MRI}
The problem of optimizing the sampling mask of Cartesian MRI fits the description of RL well, but is not exactly described by an MDP. If we look to learn a sampling pattern on an image $\vx$, we can start from an empty mask and gradually make observations, trying to learn which are the best locations to acquire. It is natural to view the current mask $\omega_t$ as the current state $s_t$, and interpret the action $a_t$ as the next location that we seek to acquire. The reward is easily quantified as the improvement in reconstruction obtained by adding to the sampling mask the location corresponding to $a_t$:
\begin{equation}
    r(s_t,a_t) = \eta(\rx, \hat{\vx}_{\theta; \omega_{t+1}}) - \eta(\rx, \hat{\vx}_{\theta; \omega_{t}})\label{eq:reward_mri}
\end{equation}
where $\omega_{t+1} = \omega_t \cup a_t$ and $\hat{x}_{\theta; \omega_{t+1}} = \hat{x}_{\theta}(\mP_{\omega_{t+1}}\mF \vx)$\footnote{Note that we slightly abuse notation here and use the action $a_t$ to also refer to its corresponding sampling location.}. Recall also that $\eta(\cdot,\cdot): \mathbb{C}^P \times \mathbb{C}^P \to \mathbb{R}$ is a performance metric.
In the sequel, to avoid unnecessary clutter, we refer to the reconstruction at state $s_t$ simply as $\hat{\vx}_{t}$.

Note also that while the RL framework allows for transition \textit{probabilities}, the case of MRI, they are deterministic: we know in what state we will end up should we take an action.

The main difference lies in the fact that although the reward is well defined, it depends on the unobserved underlying ground truth $\vx$. As a result, our problem falls into the category of \textit{Partially Observable} Markov Decision Processes (POMDP) \citep{monahan1982state}. To take this into account, we need to introduce the set of observations $\mathcal{O}$ and the conditional observation function $O(s'| a,o)$. This leads us to define a POMDP as the 7-element tuple $\langle \mathcal{S},  \mathcal{O}, \mathcal{A}, P, O, R, \gamma \rangle$, where $\mathcal{S}$ is the set of all ground truths and mask $\{\langle \vx, \omega \rangle$\}, $\mathcal{O}$ is the set of all observations $\{\langle \hat{\vx}, \omega\rangle\}$, $\mathcal{A}$ is the set of all actions, corresponding typically to all columns to be acquired. The transition probability function is deterministic as discussed above, and the conditional observation probability will also be deterministic, as taking action $a_t$ from state $s_t$ will result in observing the tuple $\langle \hat{\vx}_{t+1}, \omega_{t+1} = \omega_t \cup a_t \rangle$. Finally, we have defined the reward function above, and the discount factor $\gamma$ can be taken as a parameter.

\subsection{Double Deep Q-Networks}
Until now, we have defined the fundamental building blocks of RL and the environment corresponding to the problem of learning sampling policies. We turn now to practical methods to learn sampling policies\footnote{This is a big jump forward, but the interested reader should read the excellent introduction to reinforcement learning \citep{weng2018a}.}.

The state of possible state-action pairs in RL is very large, and in most modern RL applications, it is simply impossible to explore it in an exhaustive manner. Similarly, storing the value for each state-action pair is impractical, and in practice, one will choose to represent the state-value function with a parametric model $Q_\phi(s,a)$ (with parameters $\phi$). Of course, deep neural networks have turned out to be prime candidates for such models, and led to the famous Deep Q-Networks (DQN) \citep{mnih2015human}. In Q-learning, the Q-network $Q_\phi(s,a)$ is learned by minimizing the objective
$$\mathcal{L}(\theta) = \mathbb{E}_{(s',r,a,s)}\left[\big(\overbrace{r + \gamma \max_{a'} Q_\phi(s',a')}^{\text{Target } Y^Q} - Q_\phi(s,a)big)^2\right].$$
One aims at solving this problem because $r + \gamma \max_{a'} Q_\phi(s',a')$ provides a more reliable estimate to the future return than $Q_\phi(s,a)$. Indeed, it can be proven that the optimal policy satisfies $Q_*(s,a) = r + \gamma_a Q_*(s',a')$ given any tuple (s,a,r,s'), and thus the loss quantifies how far one is from optimality.

However, this objective makes the training unstable and slow, the sequence of iterates $(s,a,r,s')$ from a single episode are heavily correlated, and because the target $Y_Q$ depends on $Q_\phi$ that is updated each iteration. As a result, \citet{mnih2015human} proposed two mechanisms to alleviate these issues. Experience replay is the idea that instead of computing updates directly on the sequence of iterates from the current episodes, the tuples $(s,a,r,s')$ are stored in a buffer from which a sample is drawn at random to perform the update. The second mechanism is periodic updating, where the network used in the target has its weights frozen and only periodically updated.

While these mechanisms allowed to fix some large issues of Q-learning, DQN was found to still suffer from overestimating the value of state-action pairs, and \citet{van2016deep} proposed to use Double-Q learning as a way to address this issue. As a result, the objective looks like
$$\mathcal{L}(\theta) = \mathbb{E}_{(s',r,a,s)\sim U(D)}[(\overbrace{r + \gamma \max_{a'} Q_{\phi^-}(s',a')}^{\text{Target } Y^{\text{DoubleQ}}} - Q_\phi(s,a))^2].$$
The changes here are the ones from DQN: $U(D)$ is a uniform distribution over the replay memory, and $\phi^-$ refer to the weights of the frozen network. The main difference however is to be found in the target $Y^{\text{DoubleQ}}$. Double Q-learning is based on the observation that in the Q-learning target $Y^Q$, the same values are used to select an action and to evaluate making it more likely to select overestimated values \citep{hasselt2010double}. This can be made explicit as
$$Y^Q = r + \gamma Q_\phi(s', \argmax_{a'} Q_\phi(s',a'))$$
In double Q-learning, two values of weights $\phi$ and $\phi'$ are used to decouple the action selection and evaluation, resulting in the target
$$Y^{\text{DoubleQ}} = r + \gamma Q_\phi(s', \argmax_{a'} Q_{\phi'}(s',a')).$$

After the $Q$-function has been trained, inference on new environments is done by following greedily the policy prescribed by the $Q$-function, namely
$$a_t = \argmax_a Q_\phi(s_t,a) \text{~for~}t=1,\ldots,T.$$


This is the model that \citet{pineda2020active} used in their article. They propose two variants of the model, a subject-specific DDQN (SS-DDQN) as well as a dataset-specific DDQN. The difference lies in the input given to the $Q$-function. In the SS-DDQN, the neural network will take as input $\hat{\vx}_t$ and $\omega_t$ and output a $|\mathcal{A}|$ dimensional vector, with the probability to take each action $a \in \mathcal{A}$ as the next move. On the contrary, the DS-DDQN will only take as input $\omega_t$, and as a result, the policy will only depend on the current mask and will \textit{not} adapt to the current subject.

\subsection{Policy Gradient}
\citet{bakker2020experimental} took a different approach to the problem, called policy gradient. Contrarily to Q-learning, this approach aims at directly learning a policy, without first learning a value function. The policy itself is parametrized as a function $\pi_\phi(s): \mathcal{S}\to \mathcal{A}$ that associates to each state the probability of taking an action $a \in \mathcal{A}$. It aims finding the set of parameters $\phi$ that maximize
$$\max_\phi \left\{\mathcal{J}(\phi) = V_{\pi_\phi}[s_1] = \sum_a \pi_\phi(a|s_1)Q_{\pi_\phi}(s_1,a_1)\right\}.$$
In this case, as the policy is parametrized by $\phi$ and not $Q$ as we previously had, the function $Q_{\pi_\phi}(s,a)$ must be \textit{computed} rather than simply evaluated on the input. This is achieved by using the relation connecting the value function to the state-action function and the definition of the state-action function (cf. Equations \ref{eq:value_to_q} and \ref{eq:q_def}).

The policy gradient theorem aims at providing an analytical expression for $\nabla_\phi \mathcal{J}(\phi)$, in order to optimize $\phi$ using a gradient-based approach. We will directly use its result and refer the interested reader to \citet[p.324ff]{sutton2018reinforcement}. The policy gradient theorem yields
\begin{equation}
    \nabla_\phi \mathcal{J}(\phi)\propto \mathbb{E}_{\pi_\phi}\left[ G_t \nabla \log \pi_\phi(A_t|S_t)\right].\label{eq:policy_gradient}
\end{equation}
This result is significant because it allows to estimate the gradient of the policy with an \textit{expectation} over the return and the gradient of the logarithm of policy. The expectation means that the gradient of the policy is amenable to a Monte-Carlo approximation, where one takes samples of the policy to approximate the expectation.

This is the basis used by \citet{bakker2020experimental}, but the result in their case is slightly modified to have
\begin{equation}
    \nabla_\phi \mathcal{J}(\phi)\propto \mathbb{E}_{\pi_\phi,\vx_0}\left[ \sum_{t=1}^{T} \nabla \log \pi_\phi(A_t|S_t) \sum_{t'=t}^{T}\left( \gamma^{t'-t} \left(r(S_{t'},A_{t'})- b(S_{t'})\right)\right)\right] \label{eq:bakker_grad1}
\end{equation}
where the expectation over $\vx_0$ is the expectation over the initial states, and $b(S_{t'})$ is a baseline that is aimed at reducing the high variance of the gradient estimator \citep[pp. 329-331]{sutton2018reinforcement}.

The expression in Equation \ref{eq:bakker_grad1} is not directly computable as it still contains an expectation over the policy, and $A_t$, $S_t$ depend on it. \citet{bakker2020experimental} propose both a \textit{greedy} and a \textit{non-greedy} approach to compute a Monte-Carlo (MC) estimate. They read
\begin{equation}
    \resizebox*{.9\linewidth}{!}{$\displaystyle
        \nabla_{\phi} J(\phi) \approx \frac{1}{q-1} \mathbb{E}_{\vx \sim \mathcal{D}} \sum_{i=1}^{q} \sum_{t=1}^{T}\left[\nabla_{\phi} \log \pi_{\phi}\left(\vx_{t}\right) \sum_{t^{\prime}=t}^{T} \gamma^{t^{\prime}-t}\left(r_{i, t^{\prime}}-\frac{1}{q} \sum_{j=1}^{q} r_{j, t^{\prime}}\right)\right] \text { (Non-greedy), }$}
\end{equation}
\begin{equation}
    \nabla_{\phi} J(\phi) \approx \frac{1}{q-1} \mathbb{E}_{\boldsymbol{x} \sim \mathcal{D}} \sum_{i=1}^{q} \sum_{t=1}^{T}\left[\nabla_{\phi} \log \pi_{\phi}\left(\vx_{t}\right)\left(r_{i, t}-\frac{1}{q} \sum_{j=1}^{q} r_{j, t}\right)\right] \text { (Greedy)}
\end{equation}
where $r_{i,t}$ denotes the reward for the $i$-th (out of $q$) MC sample at time step $t$. We see here that the non-greedy approach requires an additional sum in the expectation: for each time step $t$, a full rollout must be computed until the last acquisition step in order to do a single gradient update. On the contrary, a greedy approach is much faster, as it there is no sum inside the expectation, except the averaging over the $q$ different MC samples taken.

Note that, in the non-greedy case, \citet{bakker2020experimental} only sample multiple trajectories at the initial state, as sampling multiple ones at each state would lead to a combinatorial explosion. After training, inference on new environments is done by simply sampling from the obtained policy, i.e. $p(a|s) = \pi_\phi(a|s)$.


\subsection{LBCS as a simple RL policy}
It should be clear now to the reader that LBCS can easily be viewed as a simple greedy policy. Instead of training a sophisticated deep neural network, it simply relies on training a non-adaptive policy that does
$$\pi(a|s_t) = \mathbbm{1}_{a=a_t} \text{, where~} a_t = \argmax_a \left\{r(s_t,a) = \mathbb{E}_{\vx \sim p(\rvx)}\left[\eta(\vx, \hat{\vx}_{\theta; \omega_{t+1}}) - \eta(\vx, \hat{\vx}_{\theta; \omega_{t}})\right]\right\}.$$
where the policy is deterministic given the current state, and does not adapt to the current patient a gs the choice is based on the expectation over the training data. In this case, $\pi(a|s_t)$ really becomes $\pi(a|\omega_t)$. The inference is then simply done by following the policy, i.e. $\pi(a|\omega_t) = \mathbbm{1}_{a=a_t}$ for all $\vx \sim p(\rvx)$.


\subsection{The questions at hand}
After this primer into RL that introduced the different existing methods for RL-based mask design, we come back again to our initial question: What component of RL does matter most? How does patient-adaptivity help? What is the contribution in long-term planning? Are there other components of the MR pipeline that affect the performance of the method more than the policy?

As the works of \citet{pineda2020active} and \citet{bakker2020experimental} do not compare to each other, and work in very different settings, it is not possible to answer these questions directly from their respective works.

We chose to use LBCS \citep{gozcu2018learning} as starting point as it is clear that it implements a learning-based, data-driven policy that is \textit{not} adaptive \textit{nor} long-term planning. This method will provide the baseline from which we can quantify the improvements brought by adaptivity and long-term planning.

Secondly, as the methods of \citet{bakker2020experimental} and \citet{pineda2020active} are preprocessed fairly differently before being used to train RL policies, we wish to understand how these policies are sensitive to changes in parameters such as the training of the reconstruction algorithm, or changes in the image field of view, or the way that k-space is undersampled. In particular, this will help us understand whether the seemingly contradictory conclusions originate from different processing of the data, or whether they reflect something more fundamental. In the sequel, we will first define the processing pipeline precisely, before carrying thorough evaluations of these methods.

\section{The MRI data processing pipeline}\label{sec:pipeline}
\begin{figure}[!ht]
    \centering
    \includegraphics[width=\linewidth]{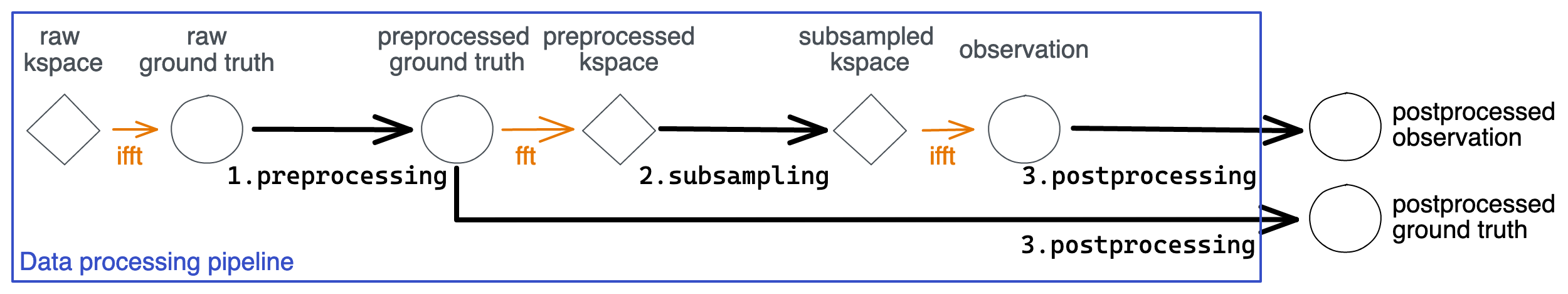}
    \caption{Illustration of the data processing pipeline of MRI subsampling. Diamonds represent data in Fourier domain (k-space) and circles represent data in image domain. The postprocessed observation and ground truth are the data that are subsequently used for training the reconstruction and policy models. The pipeline features three main blocks, namely \textit{preprocessing}, \textit{subsampling} and \textit{postprocessing}.}
    \label{fig:data_processing}
\end{figure}

In this section, we will step through the stages of the MRI data processing pipeline as shown in \Cref{fig:data_processing}. Such considerations are not reflected in the mathematical model representing the acquisition as presented in Equation \ref{eq:acquisition}. Recall that we consider an inverse problem, where we seek to recover a signal $\vx \in \mathbb{C}^P$ from partial observations $\vy \in \mathbb{C}^N$, $N\ll P$ obtained by subsampling a Fourier transform $\mF\in \mathbb{C}^{P\times P}$:
\begin{equation}
    \y_\omega = \Po\mF\x + \boldsymbol{\epsilon}, \label{eq:acquisition2}
\end{equation}
where $\boldsymbol{\epsilon} \in \mathbb{C}^N$ is the noise, $\omega \subseteq [P]:=\{1,\ldots, P\}$ is the index of sampling locations, with cardinality $N$, and $\Po$ is a diagonal matrix such that $(\mP_\omega)_{ii}=1 \text{ if } i \in \omega, 0 \text{ otherwise}$. This problem is inherently ill-posed, due to $N \ll P$, and we first construct an estimate $\hat{\x}_{\omega;\theta}=\ft(\y_\omega,\omega)$ of the original signal $\x$, where $\vf_\theta$ is a reconstruction method parametrized by $\theta$.

We turn now to the description of the practical pipeline, which is independent of the sampling method and simply operates on a given mask. We will also highlight a few common caveats.

As we will show in \Cref{s:ablations_results}, seemingly trivial changes at even a single of these stages can affect the results of the evaluation and lead to changes in the ordering of the performance of different policies, an observation that we will refer to as \textit{reversal}. We already stress that the main reason for such reversals to occur is that the RL policies generally do not generally bring a large or consistent improvement over sLBCS.
We mark sections where we observed changes leading to reversals with $^*$  and sections here they only shifted the results but do not lead to reversals with \textit{italics}.

\textbf{Data sources and storage.} In all cases, the ground truth signal $\vx$ is initially acquired as a complex signal in k-space, generally using multiple coils. The fastMRI dataset \citep{zbontarFastMRIOpenDataset2019} provides this raw multicoil data, as well as a simulated single-coil k-space which we use throughout this work.

\textbf{Preprocessing$^*$.}
Due to computational constraints, it is common to resize the images by cropping and/or resizing or use magnitude images over the raw data. Cropping and resizing changes the ground truth distribution, which  as seen in \Cref{tab:bakker_LH_auc} can also lead to reversals in the final results.


\begin{extremark}[Caveat \#1]
    Converting the raw ground truth to or starting from a magnitude induces a conjugate symmetry in Fourier space that is not present in real data, a distortion that makes our modeling less faithful to the physical model.
\end{extremark}

\textbf{Sampling.} While in the real world the data is actually acquired by sampling k-space (prospective sampling), in practice, the acquisition is generally simulated by retrospectively undersampling fully sampled Cartesian data following \Eqref{eq:acquisition}.

When training reconstructor and sampling method separately, training data are generally constructed with random masks that sample a certain fraction of center frequencies, and then the rest from a random distribution.
There has been considerable variation in defining the parameters of these distributions, but to this day, no systematic study of their effect on the reconstruction quality have been carried out.

\textbf{\emph{Post processing and reconstruction.}}
After sampling, the observation will be processed according to the implementation details of the reconstruction algorithm (e.g., normalized or standardized) and the reconstruction is computed.

Typical normalization, as used for instance in \cite{pineda2020active}, consist of dividing every input by a fixed value equal to the average energy of the dataset. \cite{bakker2020experimental} standardized their data and then clamped them to have standardized data in the range $[-6,6]$.

We found that standardization used by \cite{bakker2020experimental} was necessary only when paired with non-residual models, which tracks with observations in the literature that residual networks can work without normalization, although usually this is studied for network internal normalization which we also leverage (see e.g. \cite{zhangFixupInitializationResidual2018} which removes normalization entirely by using a proper initialization).

The results of Appendix \ref{app:bakker_mismatch} illustrate the impact of postprocessing.

\begin{extremark}[Caveat \#2]
    Normalization should occur consistently between ground truth and observation data, as failing to do this can lead to inconsistent values in the reconstruction, especially in parts where data consistency occurs. In \cite{bakker2020experimental}, the authors normalize observations and ground truth using their respective statistics and denormalize the reconstructed image using the \textit{ground truth} statistics, which is not compatible with the use of data consistency in the reconstructor. If one wishes to get a realistic estimate of the performance at deployment, it is also advisable to use only statistics available at test time (i.e. observation statistics, not ground truth statistics).
\end{extremark}

\textbf{Evaluation metrics$^*$.}
Finally, to judge the results, the most common metrics are peak signal-to-noise ratio (PSNR), mean squared error (MSE) and structural similarity index metric (SSIM).

All metrics tend to promote smooth images when used as a training loss \citep{muckleyStateoftheArtMachineLearning2020}, but it is widely known that MSE and PSNR focus on low-frequencies/high-energy components, which was also noted by \citet{bakker2020experimental}. However, while MSE and PSNR only operate on image differences and treats every pixel independently, SSIM is computed from local averages \citep{wang2004image,zbontarFastMRIOpenDataset2019}.

The metric is then reported as a curve plotted against sampling rate \citep{zhang2019reducing}, its reciprocal, the acceleration rate \citep{pineda2020active} or aggregated by computing the average performance, the performance at end of sampling, or the area under curve (AUC).

As discussed throughout the experiments, the choice of metric and its aggregation can greatly impact the conclusions drawn from results.

\begin{remark}[Area under curve (AUC) computation] We detail here the computation used to summarize performance curves with the AUC, used in Tables \ref{tab:bakker_LH_auc} and \ref{tab:pineda_auc}.

    The area under curve is a numerical integration of the performance curve of the form $\{r_t, \ell(\vx_i, \hat\vx_\theta(\vy_{\omega_t,i}))\}_{t=t_0}^T$ that relates the sampling rate or acceleration factor at the $t$-th step to its performance. We used the \texttt{sklearn} implementation, that implements it using a trapezoidal rule. We compute an individual AUC for each test sample and then aggregate the resulting set $\{\text{AUC}_i\}_{i=1}^{n_{\text{test}}}$ by computing its empirical mean and variance.

    This measure is susceptible to changes in reparation of the area under the curve, and this is the reason that the result changes when moving from sampling rate to acceleration factor (1/sampling rate). Indeed, when representing acceleration factors, in the case of \citet{pineda2020active} (cf. Table \ref{tab:pineda_auc}) $20$ out of $332$ sampling locations cover $90\%$ of the acceleration factor plot, biasing the AUC towards attributing most of its weight the high acceleration factors.
\end{remark}

\section{Re-examining deep RL for MRI sampling}
\label{s:re_examining}
Our initial impetus is the observation that the non-adaptive oracle used in \cite{bakker2020experimental} is highly reminiscent of the LBCS method \cite{gozcu2018learning}, which has been used as a strong baseline in the literature \citep{jin2019self,sanchez2020uncertaintydriven}.
To our surprise, evaluating these methods against the greedy RL method in the data processing pipeline used in \citet{zhang2019reducing} resulted in the fixed method not only coming close to the non-adaptive oracle, but actually closely matching the RL method (see \Cref{tab:bakker_LH_auc}).
Wanting to perform a fair comparison as well as to assess the impact of the variability in the pipelines used across the literature, we closely replicated the pipeline of \cite{bakker2020experimental}\footnote{We thank the authors for providing the original checkpoints and general help and responsiveness during this replication. Our results did not show the \textit{exact} same numbers as \citet{bakker2020experimental}, as we did not use the same train-val-test split and dataset undersampling. We discuss this further in Appendix \ref{app:bakker_pipeline_ablation}.} as well as  an extensive ablation study to understand this reversal and identify its source.

\subsection{Experimental setting}
In the sequel, the term \textit{setting} then refers to a particular choice of preprocessing, subsampling and postprocessing of the data, independently of the data source (we always use the same dataset) and the model or sampling policy used.
Preprocessing and postprocessing are relevant throughout the training of the reconstruction and the policy models, whereas changes in the sampling masks only directly affect the pretraining of the reconstructor, as the training the policy is done by successive rollouts using the policy model itself.\\

\textbf{Dataset and preprocessing.} Like the original paper of \cite{bakker2020experimental}, we used the fastMRI \citep{zbontarFastMRIOpenDataset2019} single-coil knee dataset for the experiments. We slightly modify follow their preprocessing (using complex data instead of magnitude data, different data normalization) that do not affect the relative ordering of the different methods. We provide ablations and a detailed discussion of these changes in \Cref{app:bakker_mismatch}.

Specifically, they \textit{crop} the data to $128\times 128$, which we refer to as (\texttt{c})).
\citet{zhang2019reducing} instead \textit{resizes} the data to $128\times128$ which we replicate by first cropping to $256\times256$ and then resizing them to $128\times128$.
This preprocessing results in images with different fields of view. We refer to this preprocessing as (\texttt{c+r}) and integrate it to our ablations.
We also evaluate horizontal (\texttt{h}) sampling masks, used in \cite{gozcu2018learning, jin2019self} in addition to the vertical (\texttt{v}) sampling masks used by \citet{bakker2020experimental,zhang2019reducing} and \cite{pineda2020active}.

The deep reconstructors used in \citet{pineda2020active,bakker2020experimental} pre-trained by randomly sampling a mask from a set of distributions with different parameters. We ablated over two pretraining regimes which we  abbreviate as \texttt{b} and \texttt{z}, respectively. We describe them in more detail in \Cref{ap:implementation}

\textbf{Reconstruction models.}
We ablate over the two reconstruction models used in the RL SotA works of \citet{pineda2020active} and \citet{bakker2020experimental}, namely a cascade of residual networks (\textit{cResNet}) architecture from \citet{zhang2019reducing}, used in \citet{pineda2020active} and the \textit{UNet} baseline provided in the fastMRI dataset, used in \citet{bakker2020experimental}. The training details and masks used for the training of the reconstruction models are described in \Cref{ap:implementation}.

For the UNet, like \citet{bakker2020experimental} we take a single channel for magnitude only data as input, while for the cResNet we take the 2 complex channels as inputs. All models are trained using $\ell_1$ loss and directly output the reconstruction without a final nonlinearity. The UNet has 837\,635 parameters in total, while the cResNet is larger with 1\,093\,479 parameters.

\textbf{Sampling methods}
In each ablation setting we compared the method of \citet{bakker2020experimental} to the following baselines and oracles (citations at the end of each item refer to prior works that also evaluate them):
\vspace{-2mm}
\begin{itemize}
    \item Random sampling (\textbf{Random}): Acquire a fixed proportion of low-frequency lines in Fourier and then randomly sample the remaining lines  \citep{jin2019self,pineda2020active,bakker2020experimental}.\\[-.5cm]
    \item Low-to-high frequencies (\textbf{LtH}): select k-space lines from low-to-high frequencies lines \citep{zhang2019reducing, pineda2020active,jin2019self}.
    \item (Stochastic) Learning-based Compressive Sampling (\textbf{LBCS}) \citep{gozcu2018learning, sanchez2019scalable}: This method trains a non-adaptive, greedy sampling policy that selects as a measurement candidate in each acquisition step the column that leads to the greatest average improvement over a sample from the training dataset. We use the stochastic version that scales better to large dataset and images \citep{jin2019self}.
    \item Non-adaptive Oracle (\textbf{NA Oracle}) \citep{bakker2020experimental}: This oracle is computed by training and evaluating LBCS directly on the test set, and can serve to illustrate the benefit of adaptivity in greedy methods. This is the instance of a non-adaptive, greedy sampling method \citep{bakker2020experimental}.
\end{itemize}

For the training of the policy models of \citet{bakker2020experimental}, we use the parameters of the greedy model in their paper, which we refer to simply as \textbf{RL} in the sequel.
We excluded the non-greedy version, as \citet{bakker2020experimental} notice that the performance of the non-greedy model with discount factor $\gamma=0.9$ is always close to one standard deviation of the greedy model, but significantly more computationally demanding.

We studied both the short and long horizon sampling regimes but only report on the long horizon for conciseness, with the short horizon results summarized in \Cref{app:bakker_pipeline_ablation}. Except for the deterministic LtH and NA Oracle, we report the performance of each method averaged on three runs/separately trained RL policies, along with the standard deviation.


\newcommand{\scrop}{(\texttt{c})}
\newcommand{\scropres}{(\texttt{c+r})}
\newcommand{\svert}{(\texttt{v})}
\newcommand{\shor}{(\texttt{h})}
\newcommand{\sbrate}{(\texttt{b})}
\newcommand{\szrate}{(\texttt{z})}

Summarizing, we start in the setting of \citet{bakker2020experimental} except for using complex data instead of magnitude data and then ablate over:
\textbf{i)} cropped \scrop{} vs cropped+resized \scropres, \textbf{ii)} vertical \svert{} vs horizontal \shor{} Cartesian sampling, \textbf{iii)} reconstruction with a pretrained UNet  \citep{ronneberger2015u,bakker2020experimental} or a cResNet \citep{zhang2019reducing} and \textbf{iv)} the training regime for the reconstructor proposed in \citet{bakker2020experimental} \sbrate{} or in \citet{zhang2019reducing} \szrate.
The base setting is referred to as \texttt{cvb}, for cropped (preprocessing) + vertical (mask) + Bakker et al. (sampling parameters) using a both a UNet and a cResNet reconstructor.

\subsection{Main results on \texorpdfstring{\citet{bakker2020experimental}}{Bakker et al. (2020)}}
\label{s:ablations_results}
We first evaluated the method of \citet{bakker2020experimental} on two settings, namely \texttt{cvb} and \texttt{c+rhz}, and report it with two different aggregations. \Cref{tab:bakker_LH_at_25} shows the SSIM of the policies at the final sampling rate considered ($25\%$), as used in as used in \citet{bakker2020experimental}, whereas \Cref{tab:bakker_LH_auc} has the SSIM aggregated over the whole acquisition trajectory by computing its AUC\footnote{A more complete ablation, using the setting \texttt{cvz} to ablate over the impact of different mask parameters, \texttt{c+rvz} to study the impact of different field of views and mask orientation is carried out in \Cref{app:bakker_ablation_full}.}.

Several observations can be made from these tables regarding \textit{reversals}. Focusing first on \Cref{tab:bakker_LH_at_25} only, we can see that although the first column, corresponding to the processing done by \citet{bakker2020experimental}, supports that RL outperforms LBCS, changing architecture (second column) already invalidates these results, with LBCS matching RL. But the change is even more significant by changing \textit{settings}. By moving from images where the field of view is cropped and undersampling occurs vertically (1st and 2nd column) to a different field of view with horizontal undersampling (3rd and 4th column), the relative ordering of LBCS and RL is \textit{reversed}, and LBCS shows the best performance.

However, if we include \Cref{tab:bakker_LH_auc} into the picture, then even the previous conclusions do not hold: on the \texttt{c+rhz} setting (3rd and 4th columns), RL matches or outperforms LBCS. \textit{Conclusions drawn from the results can be reversed or invalidated by using a different way to aggregate the results or by changing the setting on which the evaluation is carried out.}

Although contradicting conclusions can be drawn by looking at individual results, a consistent trend from Tables \ref{tab:bakker_LH_at_25} and \ref{tab:bakker_LH_auc}, as well as the ablations in \Cref{app:bakker_ablation_full}, is that \textit{the return on investment (ROI) of adopting RL over LBCS is generally marginal compared to what changes in the modeling pipeline can bring.}

We see for instance that improving the reconstruction architecture yields much more significant quantitative gains. Moving from a UNet to a cResNet brings an order of magnitude more improvement (around $0.01$ SSIM difference) than what RL brings over LBCS (around $0.0015$ at best).

We see also in \Cref{app:bakker_ablation_full} that the distribution of masks used to pretrain the reconstruction algorithm has a similar impact over performance. The \texttt{b} masks of \citet{bakker2020experimental} are discretely distributed from sampling rates $12.5\%$ to $25\%$, which matches exactly the short horizon experiment range. On the contrary the \texttt{z} masks of \citet{zhang2019reducing} are distributed from $7\%$ to $37.5\%$, whereas the long horizon experiment spans a range from $3\%$ to $25\%$ sampling rate. As a result, we observe that \texttt{b} masks consistently perform better than \texttt{z} in the short horizon experiment, whereas the opposite holds in the long horizon experiments.

As an example, in the long horizon case, using \texttt{z} masks over \texttt{b} masks brings a consistent improvement of $\sim 0.004$ in the SSIM AUC, whereas RL improves over LBCS by $\sim 0.0006$ in the SSIM AUC.

\begin{extremark}[Main conclusion]
    These results suggest that steps such as matching the pretraining and evaluation mask distributions and using strong reconstructors has a significantly larger influence and consistent effect on the performance than moving from a greedy, fixed policy (LBCS) to a greedy, adaptive one (RL).
\end{extremark}

\textbf{Additional observations.} We also see that the performance of LBCS always remains close to the NA Oracle, testifying to the generalization ability of the fixed LBCS masks as indicated by theory. Other reversals can be observed in both cases when changing the data processing for both the SSIM at $25\%$ and the AUC evaluations.

To conclude, note that the large gap in performance between the \texttt{cvb} and the \texttt{c+rhz} settings is not due to a significant difference in performance of the reconstruction methods, but rather originates from the difference in field of view. This can be seen by comparing the SSIM value of the observations (before reconstruction) using the deterministic LtH policy in both cases: in the \texttt{cvb} setting, one gets an AUC of $0.4989$ for the observation SSIM, whereas in the \texttt{c+rhz} setting, the AUC is $0.6534$.

\begin{figure}[!h]
    \begin{center}
        \resizebox{\linewidth}{!}{\begin{tabular}{lcccc}
    \toprule
    
   \multirow{2}{*}{\textbf{Policy}}& \multicolumn{2}{c}{\texttt{cropped, vert., Bakker}} & \multicolumn{2}{c}{\texttt{cropped+resized, horiz., Zhang}}\\
   \cmidrule(l){2-3} \cmidrule(l){4-5}
   & UNet  & cResNet & UNet  & cResNet \\
   \midrule
   \textbf{Random}  &  $0.5249\pm 0.0001$  & $0.5432\pm 0.0004$ &$0.6567\pm 0.0003$&$0.6725\pm 0.0006$ \\
   \textbf{LtH}  & $0.5832$                & $0.6197$ & $0.7325$ &$0.7714$ \\
   \textbf{LBCS}  & $0.6294\pm0.0009$      & $\mathbf{0.6417\pm 0.0011}$& $\mathbf{0.7768\pm 0.0000}$ &$\mathbf{0.7941 \pm 0.0000}$\\
   \textbf{RL} & $\mathbf{0.6298\pm 0.0002}$        & $\mathbf{0.6415\pm 0.0002}$ &$0.7761 \pm 0.0000$&$0.7935\pm 0.0001$\\
   \midrule
   \textbf{NA Oracle}  & $0.6301$          & $0.6428$          & $0.7771$    &$0.7942$\\
   
   \bottomrule
   \end{tabular}
   
   }
        \captionof{table}{SSIM at $25\%$ on the test set, on knee data comparing two models in the long horizon setting of \citet{bakker2020experimental}, for the \texttt{cvb} (cropped, vertical, Bakker-type of masks) and \texttt{c+rhz} (cropped+resized, horizontal, Zhang-like masks) settings. A different aggregation of the results is shown on Table \ref{tab:bakker_LH_auc}.}\label{tab:bakker_LH_at_25}
    \end{center}

\end{figure}

\begin{figure}[!h]
    \begin{center}
        \resizebox{\linewidth}{!}{\begin{tabular}{lcccc}
    \toprule
    
   \multirow{2}{*}{\textbf{Policy}}& \multicolumn{2}{c}{\texttt{cropped, vert., Bakker}} & \multicolumn{2}{c}{\texttt{cropped+resized, horiz., Zhang}}\\
   \cmidrule(l){2-3} \cmidrule(l){4-5}
   & UNet  & cResNet & UNet  & cResNet \\
   \midrule
   \textbf{Random}  &  $0.4348\pm 0.0001$  & $0.4432\pm 0.0002$ &$0.5860\pm 0.0002$&$0.5885\pm 0.0002$ \\
   \textbf{LtH}  & $0.4849$                & $0.5107$&$0.6678$ &$0.6902$ \\
   \textbf{LBCS}  & $0.5134\pm0.0004$      & $\mathbf{0.5243\pm 0.0003}$& $\mathbf{0.7035\pm 0.0001}$ &$0.7096 \pm 0.0001$\\
   \textbf{RL} & $\mathbf{0.5142\pm 0.0001}$        & $\mathbf{0.5242\pm0 .0002}$ &$\mathbf{0.7035 \pm 0.0002}$&$\mathbf{0.7111\pm 0.0009}$\\
   \midrule
   \textbf{NA Oracle}  & $0.5140$          & $0.5247$          & $0.7038$    &$0.7099$\\
   
   \bottomrule
   \end{tabular}
   
   }
        \captionof{table}{AUC on the test set using SSIM, on knee data comparing two models in the long horizon setting of \citet{bakker2020experimental}, for the \texttt{cvb} (cropped, vertical, Bakker-type of masks) and \texttt{c+rhz} (cropped+resized, horizontal, Zhang-like masks) settings. A different aggregation of the results is shown on Table \ref{tab:bakker_LH_at_25}.}\label{tab:bakker_LH_auc}
    \end{center}

\end{figure}

\subsection{Main results on \texorpdfstring{\cite{pineda2020active}}{Pineda et al. (2020)}}
\label{s:long_range}

Given these results, we can observe that if there is indeed a benefit to patient-adaptive greedy policies, it is highly sensitive to the parameters used, and the improvement over the non-adaptive greedy baseline is marginal.
One might wonder whether there will be a significant gain from adaptive RL policies if they are trained to perform long term planning on a longer horizon? To investigate this, we replicated the second SotA RL method described in  \cite{pineda2020active}.

\textbf{Experimental setting.} We used the pretrained checkpoints from \citet{pineda2020active}\footnote{The models can be found at \url{https://facebookresearch.github.io/active-mri-acquisition/misc.html}}, and compared to their Subject-Specific DDQN (SS-DDQN) as well as their Dataset-Specific DDQN (DS-DDQN).

This work uses a small discount factor of $\gamma=0.5$ and operates on the full fastMRI image of size $640\times332$, meaning it is both trained for a longer time horizon as well as having a lot more leeway for decision making.

We replicated the \textit{extreme} setting which starts with $2$ center frequencies ($0.6\%$ undersampling rate or $166\times$ acceleration) and is evaluated up to $100$ frequencies ($30.1\%$ undersampling rate or $3.32\times$ acceleration)\footnote{We thank the authors for providing us with the original scores and general responsiveness and helpfulness during this replication.}.
As can be seen in \Cref{tab:pineda_auc}, the LBCS mask is very close to the performance of both the data and sample specific DDQN policies provided by \cite{pineda2020active}.

\begin{figure}[!ht]
    \centering
    \begin{subfigure}[b]{0.49\textwidth}
        \centering
        \includegraphics[width=\linewidth]{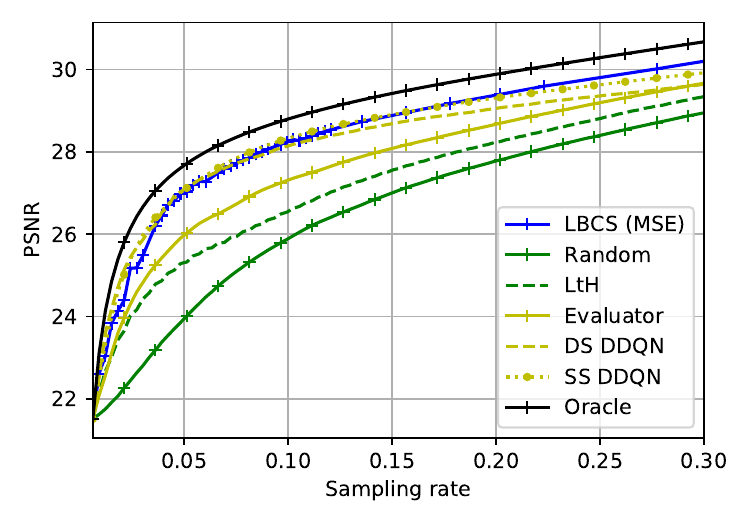}
        \caption{PSNR vs sampling rate}
        \label{fig:psnr_vs_sampling}
    \end{subfigure}
    \hfill
    \begin{subfigure}[b]{0.49\textwidth}
        \centering
        \includegraphics[width=\textwidth]{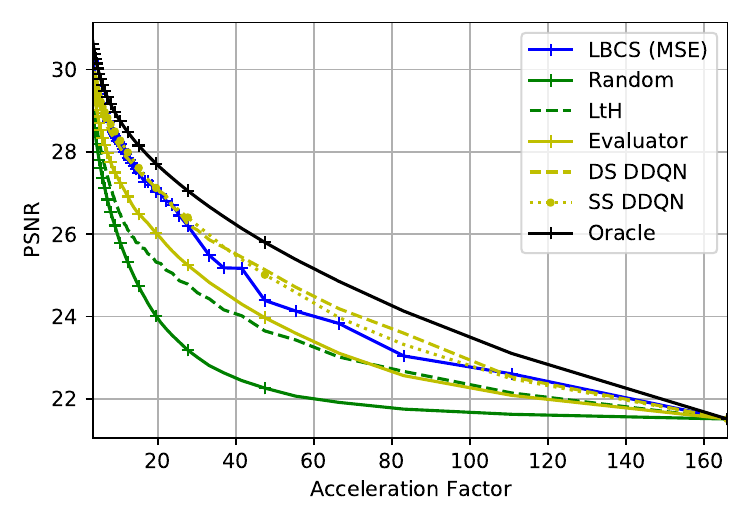}
        \caption{PSNR vs acceleration rate}
        \label{fig:psnr_acceleration_rate}
    \end{subfigure}
    \caption{PSNR performance plot on the test set, using the knee dataset with the processing from \cite{pineda2020active}. The plots feature two ways to report the same result, which are also displayed in Table \ref{tab:pineda_auc}. The SSIM performance plot can be found in \Cref{fig:pin_three_ssim} in the Appendix.}
    \label{fig:pin_three}
\end{figure}

\textbf{Results.} We see on \Cref{tab:pineda_auc} and \Cref{fig:pin_three}  other instances of reversals. On \Cref{tab:pineda_auc}, for a given metric, different ways of aggregating the results give rise to \textit{reversals} in interpretation. This is also the case when moving from SSIM to PSNR. Although there is a clear performance gain over Random and LtH masks, the gain from LBCS to DDQN is not consistent.

This discrepancy can be explained by the fact that, as displayed on \Cref{fig:pin_three}, the acceleration factor puts more emphasis on the sub $5\%$ range (i.e. below 17 lines) where LBCS has not yet caught up on DDQN. This results in the AUC weighting disproportionately high acceleration factors: $\sim 90\%$ of the plot and of the corresponding AUC consists of $20$ lines out of the $100$ acquired. Similarly, considering the final sampling rate ignores the performance of the method throughout the acquisition procedure.

We also performed a more detailed comparison of the masks given by the policies, using a subset of the first 200 test set images, keeping the order fixed in across methods, which can be found in \Cref{app:adaptivity}. They can be summarized as follows: \textbf{i)} the adaptivity of SS-DDQN mainly affects the ordering of frequencies, a large section of acquired frequencies being shared across samples and in similar regions to the LBCS mask at the final sampling rate, and \textbf{ii)} it confirms that the adaptive masks have a small edge only until about a sampling rate of $5\%$, after which LBCS catches up and overtakes the RL policy.

\begin{table}[h]
    \centering
    \resizebox{\linewidth}{!}{
\begin{tabular}{lcccccc}
 \toprule
\multirow{2}{*}{\textbf{Policy}}& \multicolumn{3}{c}{\textbf{SSIM}}& \multicolumn{3}{c}{\textbf{PSNR}}\\
\cmidrule(l){2-4}\cmidrule(l){5-7}
& Samp. rate & Acc. factor & Final rate & Samp. rate & Acc. factor & Final rate\\
\midrule
\textbf{Random} & 0.5801&0.4497&0.6723 &  26.489 &22.327&28.962\\
\textbf{LtH} & 0.5636 &0.4506&0.6686& 27.169&23.133&29.360\\
\textbf{LBCS} & 0.6079 &0.4787&\textbf{0.6886}& \textbf{28.491}&23.799&\textbf{ 30.211}\\
\textbf{DS-DDQN} & 0.6101&\textbf{0.4797}&0.6855&28.240&\textbf{23.978}&29.652 \\
\textbf{SS-DDQN} & \textbf{0.6139}&\textbf{0.4797}& 0.6882&28.424 &23.918&29.929\\
\midrule
\textbf{Adaptive Oracle} & 0.6341 &0.4910&0.7131& 29.013&24.498&30.683\\

\bottomrule
\end{tabular}}
    \captionof{table}{AUC on the test set when calculated against \textit{sampling rate} and \textit{acceleration} factor ($1/\text{sampling rate}$), as well as performance at the final sampling rate ($100$ lines acquired out of $332$) on the knee dataset, using the processing of \cite{pineda2020active}.}\label{tab:pineda_auc}
\end{table}

\section{Discussion}

As we have seen in \Cref{s:ablations_results} and \Cref{s:long_range}, the fixed mask of LBCS is competitive with or outperforms both the greedy policy gradient methods of \citet{bakker2020experimental} and the non-greedy DDQN methods of \citet{pineda2020active}.
Our ablations in \Cref{s:ablations_results} and \Cref{app:bakker_ablation_full,app:lbcs}, as well as the example of \Cref{fig:pin_three} highlight that the benefit of current RL methods over the fixed baseline, if it exists at all, is so small that a change of field of view, architecture or mask distribution (cf. Tables \ref{tab:bakker_LH_auc} and \ref{tab:bakker_auc_ssim_full}) can lead to a reversal in the relative performance ordering.

Note also that the computational resources required to train a deep RL model are much more significant than what is required to obtain a mask through sLBCS. sLBCS does not require to train a deep model and as a result does not need to perform any backpropagation. In the setting of \citet{bakker2020experimental}, this results in being able to train the mask using $20$ GB of RAM on a A-100 GPU in slightly less than $20$ minutes. With the same computational budget, the greedy deep RL method of \citep{bakker2020experimental}, which is trained significantly more quickly than longer horizon policies, takes roughly $25$ minutes for a single epoch out of $50$. However, both methods could be accelerated by optimizing the data processing computations, which take a large part of the computation time in both cases. In the setting of \citet{pineda2020active}, the comparison is even more striking. In communication with the authors, we learned that it took them more than $20$ days to train either of the DDQN models, while the training of sLBCS takes 30 minutes when distributed over 5 A-100 GPUs. This points to two main advantages of this testifies to the main advantage of sLBCS over RL-based approaches. Being a fixed policy, sLBCS does not require to do multiple iterations in order to converge, but is a one-shot approach. In addition, sLBCS can be \textit{heavily parallelized}, while deep RL-based approaches are inherently sequential, and result in a slow training procedure.

\Cref{tab:pineda_auc} confirms the expectation that methods trained on SSIM can underperform with respect to PSNR and vice versa. In general agreement with the literature, we see that using several metrics is important to capture these trade-offs, as reporting only one particular metric might hide underperformance in the other.

There have been various ways of reporting the metrics, ranging from showing sampling curves as in \Cref{fig:ssim_vs_sampling}, reporting the area under curve (AUC) tables or simply reporting the metric at the final sampling rate.
We see that reporting only the summarizing statistic can be deceitful and give relatively little information on the overall performance of the method.
For instance, using only the last sampling rate, one could not distinguish between a method that performs well at all sampling rates leading to the final one, and one that has overall mediocre results and quickly improves on the performance at the end.

For this reason, we consider sampling curves the gold standard for assessing the quality of accelerated MRI sampling methods, as a good performance at all sampling rates coupled with some stopping criterion could mean that MRI scans could be stopped when sufficient information has been acquired, further increasing their efficiency.
If one wishes to report a single number, we recommend using AUC unless only the final performance is of interest.

There is also the question of reporting acceleration factors or sampling rates, which put focus on very different regimes of sampling: acceleration factors, as shown in \cite{pineda2020active} focus on extreme undersampling rates. 
While is true that the high acceleration regimes are where it is most desirable to make improvements, results on sampling rates have the advantage of uniformly distributing the performance throughout a range of interest instead of a single acquisition determining a third of the metric as in \Cref{fig:psnr_acceleration_rate}.

Alternatives to using acceleration factors could consist in displaying sampling rates over a region of interest or computing the AUC by explicitly assigning more weight to low sampling rates instead of doing it implicitly and nonlinearly through acceleration factors.

Finally, a close study of the masks given by the SS-DDQN in \Cref{fig:ssddqn_lbcs}, the edge of the SS-DDQN over the LBCS mask in \Cref{fig:lbcs_vs_ssddqn} and the comparison of the LBCS and SSDQN edge over a fixed heuristic in \Cref{fig:lbcs_vs_lth,fig:ssddqn_vs_lth} reveals that the frequencies selected by the adaptive policy are very similar to those selected by the LBCS mask and that the variability between samples is concentrated at very early sampling rates.

\section{Relation to other works}
While we focused strictly on RL trained sampling policies leveraging pre-trained reconstructors, there are other paradigms for accelerated MRI sampling present in the literature.

The first line is that of using end-to-end training of masks using stochastic relaxations of the sampling mask to be able to differentiate it and optimize the reconstructor and sampling method jointly.  \citet{bahadir2019learning, huijben2020learning} directly optimize for a single sampling rate and sampling mask. More recent work extend this technique for jointly training an adaptive policy, doing away with the pretraining and making more efficient use of the capacity of the reconstructor  \citep{yin2021end,van2021active}. Another line of joint training relies on self-supervised learning using Monte-Carlo Tree Search, as was done in Deepmind's AlphaGo \citep{silver2017mastering,jin2019self}

An open question regarding this line of research is whether the joint training enables to achieve a better sampling policy or simply serves as a curriculum and a way to specialize the sampling policy and reconstructor onto each other. To our knowledge, neither  \citet{van2021active} nor \citet{yin2021end} investigated this. 
An interesting experiment could be to train a LBCS sampling mask on the co-trained reconstructor to see whether it can recover the sampling performance of the co-trained sampling network.

More similar to the spirit of our work, the recent study of \citet{shimron2021subtle} tackles subtle biases induced by the use of stored data and improper processing, \citet{edupuganti_uncertainty_2020} investigated specifically uncertainty methods for MRI reconstruction.

For works focusing on RL, \citet{deepRL2018} performed extensive ablation studies showing the sensitivity of RL methods to even minute variations in parameters.  \citet{Engstrom2020Implementation,ingredientsdeeppolicy} showed that the improvements of SotA RL methods can be traced to the exploitation of a subset of implementation details and algorithmic improvements and that even simple algorithms leveraging this subset can achieve SotA performance, similar to our work. This showcases that adapting to the data distribution seems to be sufficient to reach SotA in accelerated MRI sampling.

\section{Conclusion}
Taken together, our observations lead us to conclude the apparent conflict between the works of \citet{bakker2020experimental} and \citet{pineda2020active} is simply because at least \emph{in their current state}, neither of the RL SotA methods do not offer significant benefits over fixed, greedily trained masks.
Since greedy algorithms tend to perform near optimal in settings with submodularity \citep{krause2014submodular}, we conjecture a similar structure might be present in problems like the MRI sampling problem. Determining whether this conjecture holds or whether the lack of added value originates from specific RL algorithm and their training remains to be determined and should be the focus for any researcher set on applying RL to such a problem.

\textbf{Recommendations.} Our results also enable us to provide practical advice, summarized below. We provide a more extensive discussion of these statements in Appendix \ref{app:conclusions}.\vspace{-.1cm}
\begin{itemize}
    \item Focus on improvements in the reconstructor architecture, mask distribution and algorithms used for training the reconstructor.\\[-4mm]
    \item Compare against strong baselines, such as LBCS.\\[-4mm]
    \item Show sampling curves and use AUC to aggregate your results instead of performance at the final sampling rate.\\[-4mm]
    \item Be mindful about preprocessing settings when evaluating a policy model. We recommend using the cropped+vertical setting with the data normalization implemented by \citet{zbontarFastMRIOpenDataset2019}.
\end{itemize}\vspace{-.1cm}

We also want to emphasize that conducting the experiments in this paper would have been impossible if the RL methods had not been exemplary in terms of openness and reproducibility. Without access to the checkpoints and code, and without the authors' responsiveness we would not have been able to reproduce both works and add the missing baseline. Despite the theoretical guarantees of LBCS, we were surprised that it matched and sometimes simply outperformed more sophisticated methods.
We therefore do not view our work as criticism of these works but rather as an extension and a synthesis, and urge any future work to follow their lead in publishing codes, checkpoints and data.

\section*{Bibliographic note}
The idea of investigating  the \textit{limitations} of current deep RL methods originated from discussions with Igor Krawczuk. He also contributed the experiments on the setting of \citet{pineda2020active}.

\cleardoublepage
\chapter{Uncertainty driven adaptive sampling via GANs} \label{ch:gans}
In the previous chapter, we considered methods to optimize sampling based on deep reinforcement learning, and showed that in their current state, they fail to deliver on their promises of improving performance by doing long-term planning and adapting to the patient.

In this chapter, we take a step back from the question of designing the best sampling policy, and primarily aim at exploring how conditional Generative Adversarial Networks (GANs) \citep{goodfellow2014generative,mirza2014conditional} can be used to model inverse problems in a Bayesian fashion. We are particularly interested on how they can be used to learn  the posterior $p(\rvx|\yo)$, namely the distribution of ground truth images given observations.

In the context of inverse problems, GANs have mainly been used either as reconstruction models \citep{yang_dagan_2018,chen2022ai} or as generative priors \citep{bora2017compressed,jalal2021robust}, as discussed in Section \ref{ss:generative}. However, the work of \citet{adler2018deep} showed that GANs trained to approximate the posterior can learn to act as reconstructors and simultaneously provide uncertainty quantification (UQ).

We argue that there is even more that they can do when trained to approximate $p(\rvx|\yo)$. In this work, we show how conditional GANs naturally provide a criterion for adaptive sampling. Indeed, by sequentially observation the location with the largest posterior variance in the measurement domain, the GAN provides an adaptive sampling policy \textit{without ever being trained for it}.  In addition, our GAN-based sampling policy can easily be adapted to provide sampling in other measurement domains, contrarily to competing methods that are designed with Fourier sampling explicitly in mind \citep{zhang2019reducing}.

In particular, we show that when considering image domain sampling, our GAN-based policy can strongly outperform the non-adaptive greedy policy from sLBCS. However, in Fourier domain, the GAN-based policy does not match the performance of sLBCS. Indeed, in image domain, it seems that there is a greater variability in the location of the structure of interest, and as a result, adaptivity becomes ever more important. On the other side, in the previous chapter, we saw that adaptivity does not necessarily bring much in the Fourier domain. Nonetheless, we provide a more comprehensive explanation to the underperformance of our policy in Fourier, rooted in the concept of the \textit{information horizon} used to make a decision at each step. We show that our model uses \textit{less} information than LBCS to design its policy, and argue that this is the reason leading to its inferior performance.
However, when compared to models that use the same amount of information to inform their policy, such as \citet{zhang2019reducing}, our model shines and largely outperforms its competition.

Overall, this work proposes an all-in-one approach for reconstruction, uncertainty quantification (UQ) and sampling, and shows the importance of incorporating relevant \textit{information} into the policy for it to achieve a strong empirical performance.

The plan of this chapter is the following. In Section \ref{s:GANs_posterior}, we will first provide an introduction about GANs, and show how they can be used to model the posterior of inverse problems (Section \ref{ss:gan_posterior_modeling}). We will then show in Section \ref{ss:gan_sampling} how this posterior can be used to perform adaptive sampling; this is our main contribution. We will then briefly discuss the paper of \citet{zhang2019reducing} in Section \ref{ss:gan_zhang}, as they are the most conceptually similar work to ours, before moving to our experiments in Section \ref{sec:gan_experiment}. There, we validate that using the maximum posterior variance as a policy, GANs can be used to perform adaptive sampling in Fourier as well as image domains. We then briefly review relevant literature (Section \ref{ss:gans_rel}), discuss our results (Section \ref{sec:gans_discussion}) before concluding (Section  \ref{sec:gans_conclusion})\footnote{The work in this chapter is based on the following preprint and workshop paper:\\
    Sanchez, T., Krawczuk I., Sun, Z. and Cevher V. (2019). Closed loop deep Bayesian inversion: Uncertainty driven acquisition for fast MRI. \textit{Preprint available at} \url{https://openreview.net/pdf?id=BJlPOlBKDB}.\\
    Sanchez, T., Krawczuk, I., Sun, Z. and Cevher V. (2020). Uncertainty-Driven Adaptive Sampling via GANs. In \textit{NeurIPS 2020 Workshop on Deep Learning and Inverse Problems}. Available at \url{https://openreview.net/pdf?id=lWLYCQmtvW}.}.

\section{Generative adversarial networks (GANs) for posterior modeling}\label{s:GANs_posterior}

\subsection{Generative adversarial networks (GANs)}
We will begin by describing what GANs are and how they work. GANs are \textit{generative models}: given training data $\{\vx_1,\ldots, \vx_m\} \sim p(\rvx)$, they aim at learning to generate new samples that look like they have been generated by the distribution $p(\rvx)$. Several deep learning approaches have been explored to tackle the problem of generation, which we will discuss in \Cref{ss:gans_rel}, and GANs have shined in particular by their ability to handle multi-modal outputs \citep{goodfellow2016nips}.

GANs tackle the problem of generation as a two-player game, where two networks have opposite goals. The \textit{generator} tries to generate candidate samples that resemble the true distribution $p(\rvx)$, whereas the \textit{discriminator} aims at distinguishing real samples from generated ones. The generator network learns to map data from a latent space to the data distribution, and is described as $\vg_\theta: \mathbb{R}^L \to \mathbb{R}^P$. The discriminator performs a classification task, receiving a sample and outputting the probability that it has to be real, and so is described as  $d_\phi: \mathbb{R}^P \to \mathbb{R}$.

The two-player game can then be formally framed as \citep{goodfellow2014generative}:
\begin{equation}
    \min_\theta \max_\phi \mathbb{E}_{\rvx \sim p(\rvx)}\left[ \log \left(d_\phi(\rvx)\right)\right] + \mathbb{E}_{\rvz \sim p(\rvz)}\left[\log\left(1 - d_\phi\left(\vg_\theta\left(\rvz\right)\right)\right)\right]\label{eq:gan_objective}
\end{equation}
where $\rvz \sim p(\rvz)$ is a random variable in the latent space follow a simple distribution such as a multivariate Gaussian $p(\rvz) = \mathcal{N}(\rvz;\mathbf{0},\mI)$. The objective is simply a cross-entropy term for binary classification. The model is called \textit{adversarial} because $\vg_\theta$ and $d_\phi$ aim at achieving opposite goals, they are adversaries in the game. 

In practice, the models are trained by minimizing \Eqref{eq:gan_objective} in an alternating fashion, where $d_\phi$ is optimized for $k$ steps and then $\vg_\theta$ is optimized for a single step. This is due to the fact that the maximization over $\phi$ happens in the inner problem, and in theory, this would imply that each time that $\vg_\theta$ change, $d_\phi$ should be trained until convergence. As this would incur prohibitive costs, the optimization is done over $k$ steps only.

GANs have had tremendous practical success, and manage to generate photorealistic images \citep{karras2019style} or transfer artistic styles \citep{isola2017image} among other things. But their application are extremely widespread \citep{goodfellow2016nips,pan2019recent,kazeminia2020gans}.

\subsection{Conditional GANs}
However, in the present case, we are not interested in \textit{unconditional} generation, where the model directly learns to generate samples from the data distribution $p(\rvx)$. We rather want to be able to approximate \textit{conditional} distributions of the form $p(\rvx|\vy)$, where we want to condition the data distribution on some specific observation $\vy$.

The first conditional GANs aimed at doing class-conditional generation \citep{mirza2014conditional}, where the model should learn to generate images from the data distribution that are consistent with the class given as input. For instance, if the model aimed at learning to represent cats and dogs, conditional generation would be the ability to integrate class information in the generative model in order to obtain samples from different classes on demand, i.e. have a generator of the form $\vg_\theta(\vz, c)$, where $c$ is the label of the class of interest.

However, in the case of MRI and inverse problems, we are more interested in image-conditional generation, where the generator is conditioned on a partial observation, and aims at generating a full image that is both consistent with the partial observation and likely to originate from the data distribution $p(\rvx)$. This was explored later in tasks such as multimodal image-to-image translation, where a model aims at translating a same image between two domain (e.g. between a sketch and a realistic picture) \citep{zhu2017multimodal}. Particularly relevant to us is the work of \citet{adler2018deep}, which explicitly shows that learning a conditional Wasserstein GANs (WGANs) \citep{arjovsky2017wasserstein} in an inverse problem amounts to approximating the posterior $p(\rvx|\rvy)$. We discuss this work in greater depth as they were, until recently \citep{kovachki2020conditional}, the only work that formally connected conditional WGANs to learning the posterior in the context of inverse problems.
Similarly, \citep{belghazi2019learning} show that a conditional GAN can serve to learn an exponential amount of conditionals, although without explicitly connecting it to learning a posterior distribution.

\subsubsection{Conditional Wasserstein GANs}
The Wasserstein distance, also known as the Earth-Mover distance, is defined as a distance between two probability distributions $p$ and $q$
\begin{equation}
    \mathcal{W}(p,q) \triangleq \inf_{\gamma \in \Pi(p,q)}\mathbb{E}_{(\rvx,\rvy) \sim \gamma} \left[\|\rvx-\rvy\|\right]\label{eq:w1}
\end{equation}
where $\Pi(p,q)$ denotes the set of all joint distributions $\gamma(\rvx,\rvy)$ that respectively marginalize to $p$ and $q$. In Wasserstein GANs (WGANs), one aims to find the distribution $\vg_\theta$ that minimizes the Wasserstein distance with the reference distribution $p$. Although \Eqref{eq:w1} does not look like a two-player game at first glance, it can be expressed in a minmax form by leveraging the Kantorovich-Rubinstein duality  \citep[Remark 6.5 on p.~95]{villani2008optimal}, which states that
\begin{equation}
    \mathcal{W}(p, q) = \sup_{d: \|d\|_L \leq 1} \mathbb{E}_{\rvx\sim p}\left[d(\rvx)\right] -  \mathbb{E}_{\rvz\sim q}[d(\rvz)]\label{eq:kr}
\end{equation}
where $\|d\|_L \leq 1$ refers to functions with a Lipschitz constant smaller or equal to $1$. This results in the objective
\begin{equation}
    \min_\theta \mathcal{W}(p(\rvx),\vg_\theta) = \min_\theta \max_{\phi: \|d_\phi\|_L \leq 1} \mathbb{E}_{\rvx\sim p(\rvx)}\left[d_\phi(\rvx)\right] -  \mathbb{E}_{\rvz\sim p(\rvz)}[d_\phi(\vg_\theta(\rvz))]\label{eq:wgan_obj}
\end{equation}
where similarly to Equation \ref{eq:gan_objective}, we aim at training a generator to produce samples that resemble $p(\rvx)$, with the discriminator output being large for \textit{real} samples and small for \textit{generated} ones. Note that there are important differences in the objective, as we see that \Eqref{eq:wgan_obj} has no logarithms in the objective, and imposes a constraint on the Lipschitz constant of the discriminator. This has been enforced various ways, but the most common approach relies on regularizing the objective with a term called gradient penalty \citep{gulrajani2017improved} that penalizes large Lipschitz constants. One of the original motivations of WGANs is to reduce the mode collapse behavior observed in GANs \citep{arjovsky2017wasserstein}. Mode collapse is an issue where a GAN does not learn all modes of the distribution, but remains stuck on a single mode.

\subsection{Conditional GANs for posterior modeling}\label{ss:gan_posterior_modeling}
Most significant to this present work is however the extension of \Eqref{eq:wgan_obj} to an image conditional setting and the connection with a Bayesian modeling approach. In their work, \citet{adler2018deep} show that we can learn the posterior distribution $p(\rvx|\rvy)$ using WGANs, and using Kantorovich-Rubinstein duality, they show that the problem of learning the posterior by minimizing
$$\min_\theta \mathbb{E}_{\rvy \sim p(\rvy)}\left[\mathcal{W}(p(\rvx|\rvy),\vg_\theta(\cdot,\rvy))\right]$$
can be framed as
\begin{equation}
    \min_\theta \max_{\phi:\|d_\phi\|_L\leq 1} \mathbb{E}_{\substack{(\rvx,\rvy) \sim p(\rvx,\rvy) \\ \rvz \sim p(\rvz)}} \left[ d_\phi(\rvx,\rvy)  - d_\phi(\vg_\theta(\rvz,\rvy),\rvy)\right]\label{eq:wgan_objective}
\end{equation}
where $p(\rvx,\rvy)$ is the joint distribution of reference images and observations. We see that the main difference with \Eqref{eq:wgan_obj} lies in the fact that the generator and discriminator both take two inputs instead of one, and that we use a pair of samples from the joint distribution in order to learn the posterior. Note that it is easy to obtain samples from the joint distribution, as given a ground truth image $\rvx$, the corresponding observation can be obtained by applying the forward model of \Eqref{eq:acquisition} to it, with any mask $\omega$.

Using such a model applied to Computed Tomography (CT), \citet{adler2018deep} show that the learned model can perform both reconstruction and UQ, by taking the empirical mean and variance using samples $\{\vg_\theta(\vz_i,\vy)\}_{i=1}^{n_s}$.

\subsubsection{Deep Bayesian Inversion}
In their work, \citet{adler2018deep} take an additional step from Equation \ref{eq:wgan_obj}, in order to further prevent the issue of  \textit{mode collapse}. To tackle the issue \citet{adler2018deep}, propose to use a conditional mini-batch discriminator approach, inspired from \citet{karras2017progressive,salimans2016improved}. Their idea is to have the discriminator distinguish between pairs of conditional samples that can contain the original image or generated ones. Their objective becomes:
\begin{equation}
    \begin{split}
        \min_\theta \max_{\phi:\|d_\phi\|_L\leq 1} \mathbb{E}_{\substack{(\rvx,\rvy) \sim p(\rvx,\rvy) \\ \rvz_1, \rvz_2 \sim p(\rvz)}} \bigg[ &\frac{1}{2}\bigg(d_\phi\big(\big(\rvx,\vg_\theta(\rvz_2,\rvy)\big),\rvy\big)+d_\phi\big(\big(\vg_\theta(\rvz_1,\rvy),\rvx\big),\rvy\big)\bigg)\\
            &  - d_\phi\big(\big(\vg_\theta(\rvz_1,\rvy),\vg_\theta(\rvz_2,\rvy)\big),\rvy\big)\bigg]
    \end{split}
    \label{eq:ao_objective}
\end{equation}

\subsubsection{Neural Conditioner \citep{belghazi2019learning}}
Although not framed as a Bayesian approach, the work of \citet{belghazi2019learning} shows how to use a regular conditional GAN to learn an exponential amount of conditional distributions, which in the bottom line is the same task as learning the posterior distribution for various conditionals $\rvy$. As they are learning conditionals, their generator is called a \textit{neural conditioner}. Contrarily to \citet{adler2018deep}, \citet{belghazi2019learning} focus on the problem of inpainting, where an image $\rvx \in \mathbb{R}^P$ is divided between \textit{available} features indexed by the mask $\omega_a$ and \textit{requested} features that are not observed, and are indexed with the mask $\omega_r$. Their task then is to train a model to approximate the distribution $p(\vx_{\omega_r} | \vx_{\omega_a})$. This corresponds to an inverse problem where the observation is described as $\rvy = \rvx_{\omega_a} = \mP_{\omega_a} \rvx$. In the sequel, for compactness, we write $\mP_{\omega_a} \rvx$ as $a \cdot \rvx$. Overall, their objective reads
\begin{equation}
    \resizebox*{.9\linewidth}{!}{$\displaystyle \min_\theta \max_\phi \mathbb{E}_{\rvx, a, r}\left[ \log \left(d_\phi(r \cdot \rvx, a\cdot \rvx, a, r)\right)\right] + \mathbb{E}_{\rvx,a,r,\rvz}\left[\log\left(1 - d_\phi\left(\vg_\theta\left(a\cdot \rvx, a, r, \rvz\right) \cdot r, a\cdot \rvx, a, r\right)\right)\right]$}.\label{eq:nc_objective}
\end{equation}
Here, the discriminator and generator embed directly the mask information $a$ and $r$, and as in the conditional model of \Eqref{eq:ao_objective}, the generator takes the observation $a \cdot \rvx$ and the noise $\rvz$, and the discriminator takes as input both the candidate sample $r \cdot \rvx$ or $r \cdot \vg_\theta\left(a\cdot \rvx, a, r, \rvz\right)$ along with the observation $a \cdot \rvx$.

\subsection{Learning the posterior or learning moments}\label{ss:learning_moments}
If one is primarily interested in using the posterior to compute its mean and variance for reconstruction and uncertainty quantification, then alternative approaches exist that rely on supervised learning instead of adversarial training. Indeed, adversarial training can often be unstable, whereas training in a supervised fashion is often easier.

The idea, used in particular by \citet{zhang2019reducing} for MRI and proposed initially in \citet{kendall2017uncertainties}, assumes that the data is distributed normally around a mean $\boldsymbol{\mu}\in \mathbb{R}^P$ and a diagonal covariance $\text{diag}(\boldsymbol{\sigma})$, $\boldsymbol{\sigma} \in \mathbb{R}^P$, $\sigma_i \geq 0$. As discussed in Section \ref{ss:bayesian_recon}, minimizing the reconstruction error $\|\vy_\omega-\mA_\omega \hat{\vx}\|_2^2$ can be viewed as maximizing the likelihood of a Gaussian distribution with unit diagonal covariance. However, it is possible to model the uncertainty at a given pixel as a Gaussian centered at the reconstruction mean $\hat{\vx}_\theta$ with variance $\hat{\boldsymbol{\sigma}}^2_\theta$. Both the reconstruction mean and variance can be output of two heads of a model with shared parameters, i.e. $(\hat{\vx}_\theta, \hat{\boldsymbol{\sigma}}_\theta^2) = \vf_\theta(\vy_\omega)$. And they can be learned through minimizing the conditional negative log-likelihood (NLL)
\begin{equation}
    \min_\theta \ell_{\text{NLL}}(\theta) = \min_\theta \frac{1}{mP} \sum_{i=1}^m \sum_{j=1}^P \frac{(\vx_{i,j} - \hat{\vx}_{\theta,i,j})^2}{2\hat{\boldsymbol{\sigma}}_{\theta,i,j}^2} + \frac{1}{2} \log\left(2\pi\hat{\boldsymbol{\sigma}}_{\theta,i,j}^2\right)\label{eq:zhang_objective}
\end{equation}
where $\vx_{i,j}$ denotes the $j$-th pixel of the $i$-th data point for a training set of size $m$. Such an approach of aiming at solving $\min_\theta -\log(p_\theta(\rvx|\rvy))$ is referred to as a \textit{negative log-likelihood} in the machine learning (ML) literature, but in our context of inverse problems, \Eqref{eq:zhang_objective} amounts to assuming that the \textit{posterior} takes the shape of a Gaussian distribution, and that we learn its moments by minimizing the parameters of the neural network $\vf_\theta(\vy_\omega)$.

\begin{remark}
    There is here a considerable overlap with the terminology used in Bayesian Neural Networks \citep{kendall2017uncertainties}, where the aim is to learn a posterior distribution over \textit{weights},
    $$p(\theta|\rvx,\rvy) = \frac{p(\rvy|\rvx,\theta) p(\theta)}{p(\rvy|\rvx)}.$$
    We see that in our case, the notation is inverted compared to classical ML literature, as they denote the label or reference as $\rvy$ and the data as $\rvx$, whereas in our case, the reference image is denoted $\rvx$, and the data (i.e. the observation) is denoted as $\rvy$. This is because in standard ML, we do not consider an inverse problem where we aim at retrieving the data from partial observations obtained through a forward model, but rather assume that $\rvy$ contains some high-level structure (e.g. a class) that is hidden in some raw data $\rvx$.

    We see then that $p(\rvy|\rvx,\theta)$ in the ML notation corresponds to $p_\theta(\rvx|\rvy)$ in our notation, and serves in the ML case the role of likelihood, hence Equation \ref{eq:zhang_objective} describing $p_\theta(\rvx|\rvy)$ as a negative log-\textit{likelihood}. However, in an inverse problem, $p_\theta(\rvx|\rvy)$ plays the role of \textit{posterior}, as seen in Section \ref{ss:bayesian_recon}. We have
    $$p_\theta(\rvx|\rvy) = \frac{p(\rvy|\rvx)p(\rvx)}{p(\rvy)}.$$
    In the case of an inverse problem, $p(\rvy|\rvx)$ is the likelihood defined by the forward model of \Eqref{eq:acquisition}, $p(\rvx)$ the prior defined by the ground truth data and $p(\rvy)$ is the marginal. As computing probabilities from a discrete set of training data is difficult and marginalization is computationally prohibitive on such problems, \citet{zhang2019reducing} circumvent the problem by directly assuming that the posterior is a Gaussian distribution, and simply train a model to learn its moments from data. However, as we will see in the sequel, such modeling simplifications incur a cost, as this reductive assumption does not allow the posterior to model multi-modal distributions.
\end{remark}

\section{GAN based adaptive sampling}\label{ss:gan_sampling}
The initial motivation of this work was a belief that the benefits of generative methods have not been fully exploited in the Compressed Sensing (CS) setting. Although GANs have been used as priors \citep{bora2017compressed,narnhofer2019inverse}, as reconstructors \citep{yang_dagan_2018} and approximate posteriors for UQ \citep{adler2018deep}, their ability to naturally provide an adaptive policy has not been explored.

Although this connection between modeling the posterior and adaptive sampling was explored in the context of classical CS \citep{ji2008bayesian,seeger2010optimization}, this has not been exploited in the context of GANs, even though they provide a very flexible, learning-based model of the posterior distribution.

This is precisely out approach, which is in some sense very natural. Given an observation $\vy_\omega = \mP_\omega \mA \vx +\boldsymbol{\epsilon}$ and a trained conditional GAN $\vg_{\theta^*}(\cdot, \vy_\omega)$ approximating the posterior $p(\rvx|\vy_\omega)$, we can compute the estimated conditional pixel-wise variance in the transformed domain $\text{Var}[\mA\rvx|\vy_\omega]$ by taking several samples from the generator, transforming them to the observation domain $\mA$, and acquiring the location with the largest posterior variance, i.e.
\begin{equation}
    v_t = \argmax_{v: v\in [P]} \mathbf{P}_{v} \text{Var}[\mA \rvx | \rvy_{\omega_t}]. \label{eq:gudacs}
\end{equation}
Departing from the idea of explicitly training a policy to guide adaptive sampling \citep{jin2019self, zhang2019reducing,pineda2020active,bakker2020experimental}, our approach simply minimizes the variance of an approximate posterior. This turns out to yield a strong sampling policy \textit{despite the networks involved being trained solely for conditional generation}.

\begin{remark}
    In order to perform adaptive sampling using Equation \ref{eq:gudacs}, the model must fulfill some requirements. First of all, one must be able to \textit{sample} for the model, and secondly, sampling must be efficient.

    Although these requirements are readily filled by our approach, this is not the case of all models. In particular, this disqualifies methods that only learn the mean and pixel-wise variance of the posterior, by minimizing for instance the NLL (Equation \ref{eq:zhang_objective}). Such statistics do not capture the complex, multi-modal distributions that arise inverse problem, and sampling from the Gaussian distribution resulting of minimizing Equation \ref{eq:zhang_objective} will not work in practice, as the samples will not originate from the true posterior, which can be multi-modal, but rather from the Gaussian that simplifies the underlying distribution. We confirm in our experiments that sampling from the NLL, transforming the samples into Fourier and taking the maximum variance to guide the acquisition does not work.

    In addition, if one wishes to deploy such adaptive sampling in a clinical setting, the inference time must be shorter than the repetition time (TR), which is typically in the order of $10^2$-$10^3$ ms. While our approach and the one of \citet{zhang2019reducing} satisfy this constraint, this is not the case for all generative approaches. In particular, autoregressive models \citep{van2016pixel} or score-based sampling approaches suffer from long inference times \citep{song2021solving}.
\end{remark}

An important distinction between our current approach and the ones that were discussed previously is the \textit{information horizon} used by the different policies. The information horizon refers to how far ahead rollouts occur in order to train a given policy, and is described on Table \ref{tab:axes_2} for different policy models. While it is clear that policies that integrate long horizon planning leverage $N$-step information, other methods can make use of $0$-step or $1$-step information.

\begin{table}[ht]
    \centering
    \resizebox{\linewidth}{!}{\begin{tabular}{l|ccc}
            \toprule
            \textbf{Sampling policy}                       & \textbf{Adaptive}      & \textbf{Long horizon}  & \textbf{Information horizon} \\
            \midrule
            LBCS \citep{gozcu2018learning}                 & \xmark                 & \xmark                 & $1$-step                     \\
            AlphaGo \citep{jin2019self}                    & \cmark                 & \cmark                 & $N$-step                     \\
            DDQN \citep{pineda2020active}                  & \xmark {\tiny(\cmark)} & \cmark                 & $N$-step                     \\
            Policy Gradient \citep{bakker2020experimental} & \cmark                 & \xmark {\tiny(\cmark)} & $1$-step                     \\
            \midrule
            Evaluator \citep{zhang2019reducing}            & \cmark                 & \xmark                 & $0$-step                     \\
            GAN (Ours)                                     & \cmark                 & \xmark                 & $0$-step                     \\
            \bottomrule
        \end{tabular}}
    \vspace{1mm}
    \captionof{table}{Methods that will be considered in the paper. This table extends the perspective given in Table \ref{tab:axes}. Different methods make use of $0$/$1$/$N$-step information in their sampling policy.}\label{tab:axes_2}
\end{table}

The distinction lies in the fact that our GAN-based approach and the one of \citet{zhang2019reducing} make their choice based on the location with the maximum error at the current step, while LBCS or the Policy Gradient approach \citep{bakker2020experimental} integrate the actual error that remains when adding a location in the mask. Formally, given a test image $\vx$, greedily trained $1$-step policies aim at acquiring the location which yield to the best performance at the next step
\begin{equation}
    v_{t+1} = \argmax_{v\in[P]} \eta(\vx,\hat{\vx}_{\omega_{t}\cup v}).\label{eq:1step}
\end{equation}
Evaluating this for a given point requires \textit{computing a reconstruction for each candidate $v$}, in order to compute the \textit{actual} performance gain that would be achieved by adding $v$ to the mask $\omega_t$. $0$-step information methods rely instead on acquiring the location feature the largest \textit{current} error
\begin{equation}
    v_{t+1} = \argmax_{v\in[P]} \| \mP_v \mF\left(\vx- \hat{\vx}_{\omega_t}\right) \|.\label{eq:0step}
\end{equation}
Evaluating Equation \ref{eq:1step} requires only a \textit{single reconstruction}, and then selection is performed by masking in the Fourier space in order to find the location with the largest current error.

As we will see in the results, $0$-step methods do not outperform $1$-step methods, and we hypothesize that this is because they do not incorporate any feedback in the design of their policy, contrarily to $1$-step methods. $0$-step methods simply they pick the location with the largest current error, but do not incorporate the \textit{a posteriori} information on how good their action \textit{actually} was. This is mainly because $0$-step approaches use a heuristic that was not explicitly trained to be a policy, but to perform a surrogate task, such as estimating the variance of the posterior distribution in Fourier (in our case), or estimating the line-wise Fourier MSE \citep{zhang2019reducing}.

\section{Comparison with \citet{zhang2019reducing}}\label{ss:gan_zhang}
We have mentioned several times until now the conceptual similarities with the approach of \citet{zhang2019reducing} with our approach. We have shown in Section \ref{ss:learning_moments} that their approach is \textit{not} generative, but aims at learning the mean and variance of the posterior distribution by minimizing a NLL (Equation \ref{eq:zhang_objective}). These statistics do not allow to directly perform adaptive sampling, as we show in Appendix \ref{ss:app_brain}.

This is why, in order to perform adaptive sampling, \citet{zhang2019reducing} also train with their reconstructor an additional model, called an \textit{evaluator} $\ve_\phi(\hat{x},\omega)$, that tries to predict the line-wise error in Fourier domain. The role of the evaluator is not only to provide guidance for adaptive sampling, similarly to the variance in Fourier space in our case, but it also helps to refine the reconstruction, by adding an adversary that aims at discriminate between real reconstructed lines and reconstructed ones.

The idea of the evaluator is to provide an individual prediction of whether a line in k-space is likely to be real or reconstructed. Consider the $k$-th $v_k \in [P]$\footnote{For simplicity, we simply rewrite here the set of all pixels $\{1,\ldots,P\}$, as the argument is the same whether pixels or lines are considered.}, and a reconstruction $\hat{\vx}$. \citet{zhang2019reducing} define the $k$-th \textit{spectral map} as $\vm^k(\hat{\vx}) = \mF^{-1}(\mP_{v_k} \mF \hat{\vx})$, $\vm_i$ represents the inverse Fourier transform of a single k-space line, and has dimension $\vm_i \in \mathbb{C}^P$. The evaluator then aims at fitting target scores using the following kernel
\begin{equation}
    t(\hat{\vx},\vx)_k = \exp \left(-\gamma \| \vm^k(\hat{\x}) - \vm^k(\vx)\|_2^2\right)
\end{equation}
where $\gamma$ is a scalar hyperparameter. Note that $t(\hat{\vx},\vx) \in [0,1]$, where it is $1$ when the spectral maps of $\hat{\vx}$ and $\vx$ perfectly match. The evaluator $\ve_\phi(\hat{x},\omega)$ is then trained to minimize
\begin{equation}
    \min_\phi \ell_{\text{E}}^{\text{E}}(\phi) = \min_\phi \frac{1}{m} \sum_{i=1}^m\sum_{k=1}^P |e_\phi(\hat{\vx}_i,\omega)_k - t(\hat{\vx}_i,\vx_i)_k |^2.\label{eq:zhang_e}
\end{equation}
The intuition is simple, the evaluator tries to match the prediction of the kernel $t$. The evaluator is also added to the objective of the reconstruction method, contributing as
\begin{equation}
    \ell_{\text{E}}^{\text{R}}(\theta) = \frac{1}{m} \sum_{i=1}^m \sum_{k=1}^P |e_\phi(\hat{\vx}_{\theta,i},\omega)_k - 1 |^2
\end{equation}
where we see that it acts as an adversary: the evaluator is trained to predict how closely a reconstructed $\hat{\vx}$ matches a reference image $\vx$, while the reconstructor aims at reconstructing images that look realistic to the evaluator. The updated objective of Equation \ref{eq:zhang_objective} then becomes
\begin{equation}
    \min_\theta \ell_{\text{R}}(\theta) = \min_\theta \ell_{\text{NLL}}(\theta) + \beta \ell_{\text{E}}^{\text{R}}(\theta)\label{eq:zhang_r}
\end{equation}
The model is then trained by alternating updates between Equation \ref{eq:zhang_e} and Equation \ref{eq:zhang_r}. Although the model is inspired by the adversarial training employed in GANs, significant differences remains, especially regarding the learning of a distribution in a GAN against point estimates in \citet{zhang2019reducing}, and the construction of an evaluator that matches a kernel rather than some more general distance such as a KL-divergence \citep{goodfellow2014generative} or a Wasserstein distance \citep{arjovsky2017wasserstein}.

\section{Experiments}\label{sec:gan_experiment}
We first consider MRI adaptive sampling, and the move to image domain sampling on MNIST.
\subsection{MRI adaptive sampling}\label{ss:mri_gan}

\subsubsection{Experimental setting}
We carried out experiments on Fourier sampling with the same conditions as outlined in Section \ref{s:re_examining}

\textbf{Dataset.} We used again the fastMRI \citep{zbontarFastMRIOpenDataset2019} single-coil knee dataset for the experiments, using complex data and the same data normalization as previously. We cropped the data to $256\times 256$ before resizing them to $128\times 128$. We used horizontal sampling masks, and experimented among other things with the mask distribution of \citet{zhang2019reducing}. This corresponds to the \texttt{c+rhz} setting in the previous chapter.
Recall that for the mask distribution, the setting of \citet{zhang2019reducing} meaning the model was trained with 10 center frequencies always selected and between 3 and 37 randomly selected elements. The data were postprocessed to have the magnitude of the ground truth in the range $[0,1]$.

\textbf{Reconstruction models.}
For this MRI experiment, we used two types of reconstruction models. The first one is the c-ResNet of \citet{zhang2019reducing} that was trained to minimize the negative log-likelihood of \Eqref{eq:zhang_objective} along the adversarial contribution of the evaluator. The model was trained using Adam \citep{kingma2014adam} with $\beta=(0.5,0.999)$, $lr=6\cdot 10^{-4}$ over $100$ epochs. The learning rate was linearly decayed starting at $50$ epochs until it reached $0$ at the end of training.

The second type of model that we considered are different GAN versions. We observed that for Fourier-based conditional generation, the training of the Neural Conditioner (NC) \citep{belghazi2019learning} was generally unstable and generally did not lead to good performance, although it did very well in the image setting. We observed that the model proposed by \citet{adler2018deep} generally performed better than a standard conditional WGAN, and this is the one that we used in our experiments. We refer to it as the \textbf{AO} or \textbf{GAN} model. For the generator, we used the residual UNet (ResUNet) of \citet{belghazi2019learning}, and used also their ResNet-like architecture as discriminator. In the GAN case, the reconstruction was estimated by computing the empirical mean from a set of samples. In our experiments, we observed that merely $2$ samples managed to provide an informative mean and variance, although averaging on more samples led to an increase in overall image quality.

For the WGAN we use Adam \citep{kingma2014adam}, $\beta =(0,0.9)$, TTUR \citep{heusel2017gans} with learning rates of $lr_G=10^{-4}$ and $lr_D=3\cdot 10^{-4}$. The training is done over $100$ epochs, and the learning rate was halved at $50$ epochs. We used the gradient penalty \citep{gulrajani2017improved} with a weight of $10$, following  \citet{adler2018deep} along with a drift penalty with a weight of $10^{-3}$. The output of the generator was followed by a data-consistency layer \citep{schlemper2018deep, mardani2018neural}, which has the role of replacing the generated data with the observed ones at the last layer, and a sigmoid to keep the range of the data in $[0,1]$. Further details on the architecture and on the training are given in Appendix \ref{app:gan_implementation}.

\textbf{Sampling methods.}
Similarly to what we did in the previous Chapter (Section \ref{s:re_examining}), we evaluated the models on different policies, that we detail below.
\vspace{-2mm}
\begin{itemize}
    \item Random sampling (\textbf{Random}): Acquire a fixed proportion of low-frequency lines in Fourier and then randomly sample the remaining lines.\\[-.5cm]
    \item \textbf{Evaluator}: \citet{zhang2019reducing} proposed to train an evaluator that tries to estimate the current mean-squared error for each line in k-space. \\[-.5cm]
    \item (Stochastic) Learning-based Compressive Sampling (\textbf{LBCS}) \citep{gozcu2018learning, sanchez2019scalable}\\[-.5cm]
    \item Policy Gradient model (\textbf{RL}): The greedy, adaptive deep reinforcement learning method of \citet{bakker2020experimental}.\\[-.5cm]
    \item \textbf{GAN} (ours): We take samples from the conditional GAN posterior, transform them to Fourier and compute their empirical variance to guide the decision of what location to observe next.\\[-.5cm]
    \item \textbf{MSE oracle}: An adaptive oracle acquiring at each step the line with the largest mean squared error.\\[-.5cm]
\end{itemize}

For all the policies considered, we initialized the mask with $8$ low-frequency lines, and then carried out sequential sampling until a rate of $50\%$ sampling. Note however that the deep RL policy of \citet{bakker2020experimental} was trained only until $25\%$ sampling rate for computational reasons, and to provide a fair comparison, the aggregated results will be computed up to $25\%$ sampling.

\subsubsection{Results}
The results of our experiment are presented in Table \ref{tab:comp_knee} and Figure \ref{fig:knee_curve_ssim}.

Similar trends can be observed on PSNR (provided in Appendix \ref{fig:knee_curve_psnr}) and SSIM, where the $1$-step information policies (LBCS and RL) generally outperform the $0$-step information provided by the Evaluator and the GAN. On the GAN model, the GAN policy largely outperform the Evaluator, and on the reconstructor, the opposite is true, although the gap is smaller. However, the trend is \textit{not} the same if one goes until $50\%$ sampling, as shown on Figure \ref{fig:knee_curve_ssim}. Indeed, one sees that as the sampling horizon increases, the performance of the evaluator significantly worsens, whereas the GAN model matches the performance achieved by LBCS. Note that the Evaluator policy in this case is really the direct translation of the policy used in the case of \citet{zhang2019reducing}, but then reconstruction is carried out through the GAN. We see also that the RL policy is weaker on the reconstructor. This is due to the policy being trained on the GAN reconstruction model and then evaluated on the reconstructor. In addition, the results illustrate the inefficiency of sampling from the Gaussian NLL and transforming the samples into Fourier in order to sample: such a policy can perform even \textit{worse} than random sampling. 

We see also that while the MSE oracle provides an upper bound on the performance of our GAN policy, it is outperformed by LBCS. This further highlights the value of using $1$-step information horizon over $0$-step information horizon, as an adaptive $0$-step oracle can be outperformed by a $1$-step non-adaptive baseline.

We observe also interesting trade-offs. The evaluator shows a very good performance at low sampling rates ($<10\%$) matching and even outperforming $1$-step information methods, but its performance deteriorates at later stages. On the other hand, our GAN method is weaker at very early stages but stays much more consistent throughout the whole range of sampling rates.

We note also that using more than $2$ samples can boost the overall performance of the GAN reconstruction. For instance, using $5$ samples for the GAN reconstruction can boost the overall performance on the method: paired with the GAN policy, it reaches up to $31.43$ dB and $0.759$ SSIM, and on LBCS, it goes up to $32.10$ dB and $0.773$ PSNR. This amounts to an improvement of $0.8$ to $1$ dB and between $0.03$ to $0.04$ SSIM. However, averaging on increasingly many samples has a diminishing return, and in our experiments, a plateau is reached around $20$ samples.

We also tried to perform sampling from the Gaussian NLL of \citet{zhang2019reducing}, and as expected, the policy performed just like random sampling, as the NLL does not capture the complexity of the posterior distribution.

\begin{table}[!ht]
    \centering
    
\begin{tabular}{lcccc}
    \toprule
    \multirow{1}{*}{\textbf{Policy}}& \multicolumn{4}{c}{\textbf{Model}}\\
    & \multicolumn{2}{c}{GAN} & \multicolumn{2}{c}{Reconstructor}\\
    \cmidrule(r){2-3}\cmidrule(l){4-5}
    & \textsc{PSNR}& \textsc{SSIM}& \textsc{PSNR}& \textsc{SSIM}\\
    \midrule
    \textbf{Random}& $29.48$& $0.699$& $29.79$& $0.757$\\
    \textbf{Gaussian NLL}& $29.48$& $0.699$& $29.69$& $0.747$\\
    \textbf{Evaluator}& $30.03$& $0.710$& $30.77$& $0.765$ \\
    \textbf{GAN}& $30.67$& $0.726$ & $30.48$ & $0.762$\\
    \midrule
    \textbf{LBCS}& $\mathbf{31.06}$& $\mathbf{0.735}$& $\mathbf{31.41}$& $\mathbf{0.781}$\\
    \textbf{RL}& $30.88$& $0.731$ & $30.68$& $0.766$\\
    \midrule
    \textbf{MSE Oracle}& $30.98$& $0.734$ & $31.23$& $0.778$\\
    \bottomrule
\end{tabular}
    \caption{Average PSNR and SSIM test set AUC (one AUC per image) on knee data in the \texttt{c+rh} setting. The AUC is computed between the $8$ initial lines and $25\%$ sampling rate.}\label{tab:comp_knee}
\end{table}

\begin{figure}[!ht]
    \centering
    \includegraphics[width=.48\textwidth]{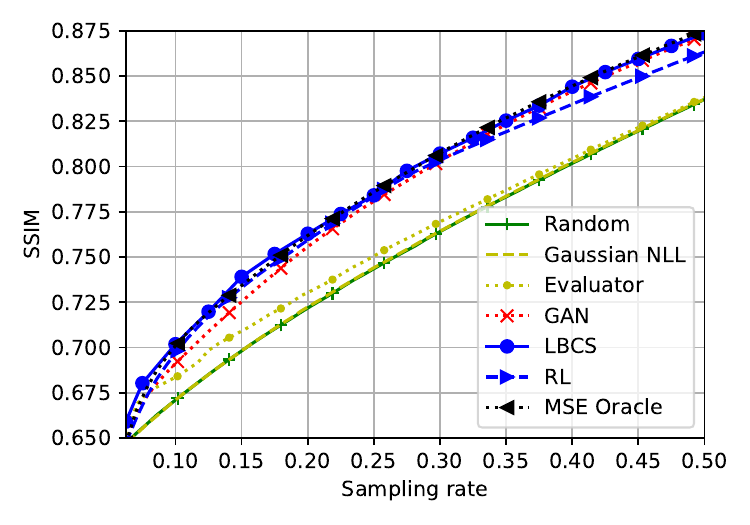}
    \includegraphics[width=.48\textwidth]{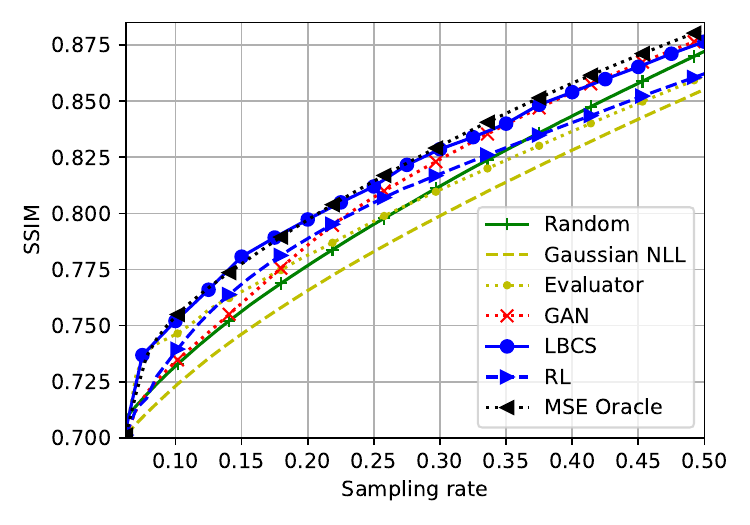}
    \caption{SSIM plots of results shown in Table \ref{tab:comp_knee}, showing the different performance of the policies. The left plot is the result of the evaluation on our GAN model, and the right plot is evaluated using the reconstructor of \citet{zhang2019reducing}. Although Table \ref{tab:comp_knee} computes the AUC until $25\%$, this table extends the evaluation until $50\%$ sampling rate.}\label{fig:knee_curve_ssim}
\end{figure}

\begin{figure}[!ht]
    \centering
    \includegraphics[width=0.5\textwidth]{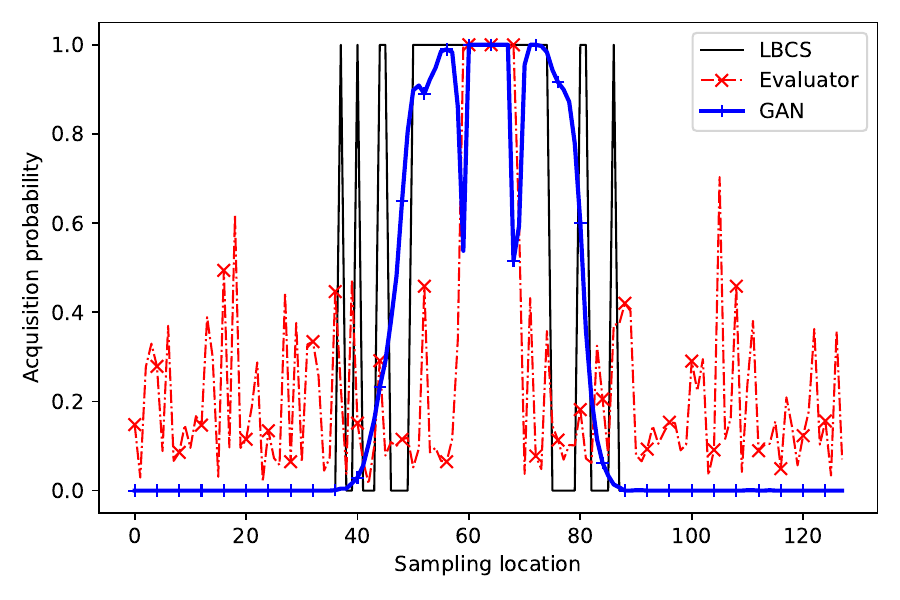}
    \caption{Acquisition probability of sampling locations by LBCS, our method and the evaluator of \citet{zhang2019reducing}, evaluated on the testing set at $25\%$ sampling rate, using the GAN model for reconstruction.}\label{fig:policy_dist}
\end{figure}

\textbf{Policy distribution.} Turning now to Figure \ref{fig:policy_dist}, we can see how different the policy looks through the testing set for LBCS, the evaluator and our GAN. The probabilities are computed as average of the masks that are observed at sampling rate $25\%$ of the testing set, and we see very different behaviors. First, as LBCS is a fixed policy, it is clear that the same locations are acquired for all images, and as a result, the policy is either $0$ or $1$. Then the GAN policy is quite concentrated around low frequencies, and spans a range that is consistent with the locations acquired by LBCS. On the contrary, the evaluator tends to acquire a lot of high frequencies, and it is hard to discern much structure in the acquisition.

Note that the GAN policy might suffer from acquiring less often some very central frequencies. It might fail to do so due to the uncertainty on these locations being quite small, although the error remains non-negligible.

\begin{figure}[!ht]
    \centering
    \includegraphics[width=.7\linewidth]{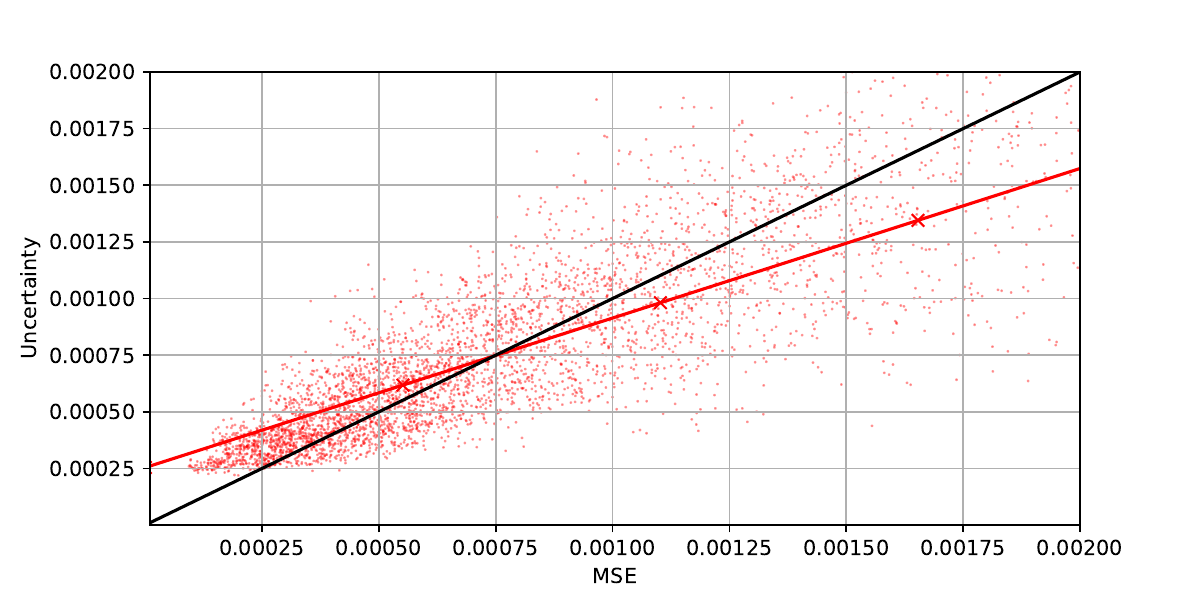}
    \caption{Uncertainty (empirical variance) against MSE for the GAN model. Each point represents a given test sample at a given sampling rate. For clarity, not all data points are represented. The red line is the result of performing a linear regression on the data points, and has $R^2=0.705$. The black line corresponds to $\text{Uncertainty}=\text{MSE}$}\label{fig:calibration}
\end{figure}

\textbf{Uncertainty calibration.} We might also wonder how good is the uncertainty (aggregated empirical variance) at predicting the mean squared error. This is shown for our GAN model on Figure \ref{fig:calibration}, where we see that overall estimation is quite good, but has some limitations still. At high errors, the model tends to be over-confident, and show a large variability in its uncertainty values, whereas at lower errors, it becomes slightly underconfident, although the uncertainty tends to have a lower variance.

However, note that this result is based on aggregating the uncertainty over the whole image, and might not reflect the ability of the model to predict good pixel-wise MSE. Nonetheless, the results suggest that the aggregated empirical variance of our model correlates well with MSE.

\begin{figure}[!ht]
    \centering
    \includegraphics[width=\textwidth]{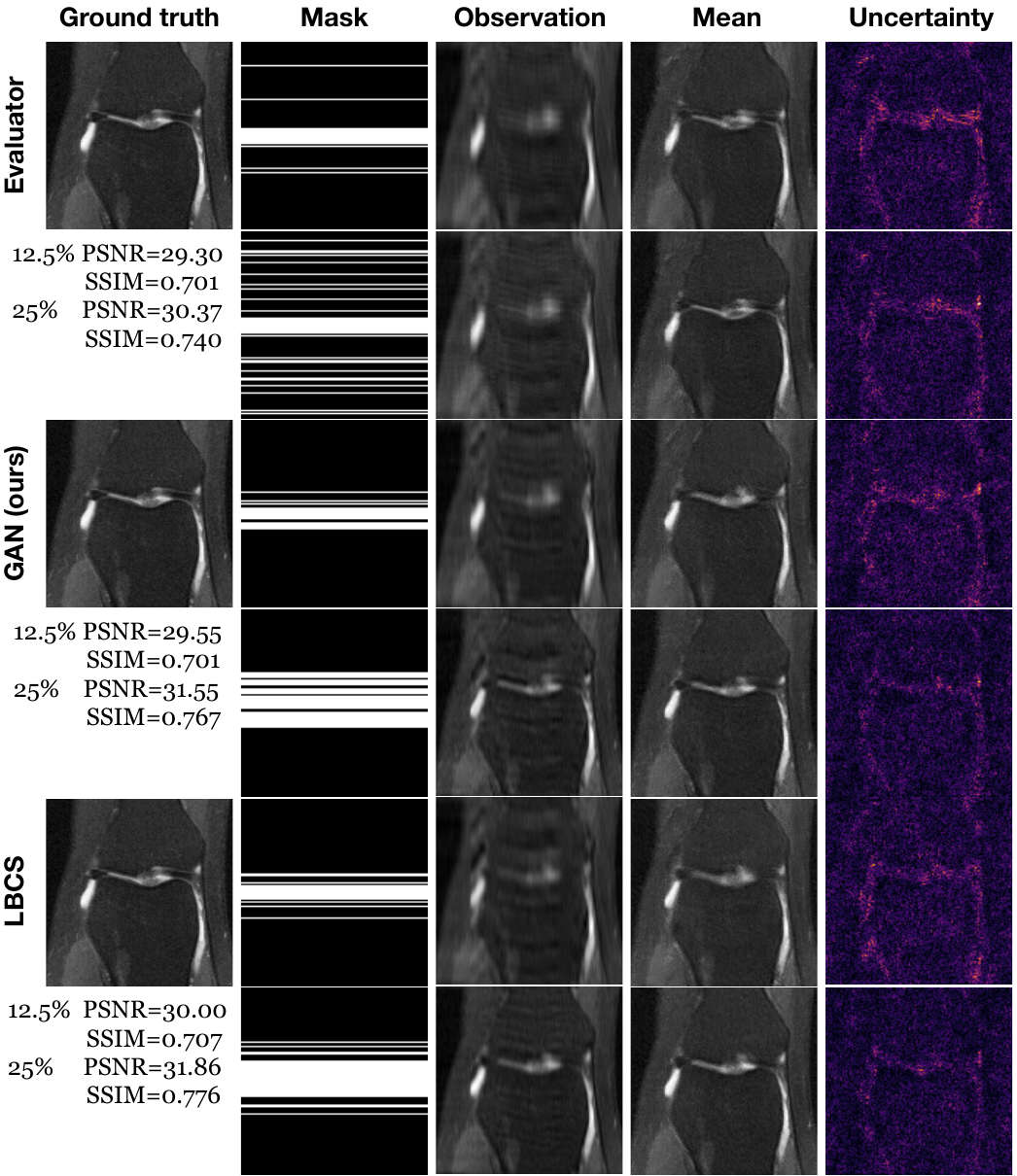}
    \caption{Visual illustration of the evaluator, GAN and LBCS policies evaluated at $12.5\%$ and $25\%$ on a slice of knee data, using the GAN model for reconstruction.}\label{fig:knee_visual}
\end{figure}

\subsection{Beyond Fourier sampling: image-domain conditional sampling}
Although our method was motivated by Fourier sampling, the work of \citet{belghazi2019learning} proposed a similar, GAN-based approach for inpainting tasks. This motivated us to consider the problem of adaptive sampling in image domain as well.

\subsubsection{Experimental setting}
We carried out our experiments on the MNIST dataset. The training set contains 60\,000 handwritten digits of size $28\times 28$. We held out 10\,000 digits for validation purposes. The test set consists of 10\,000 digits. We zero-padded all images to $32\times 32$ for convenience.

We evaluated the NC approach of \citet{belghazi2019learning} using their discriminator architecture, as well as the ResUNet and cResNet as two generators, and we compared it to a conditional Wasserstein GAN.

Similarly to the MRI case, the models were trained with gradient penalty \citep{gulrajani2017improved} and the generators are followed with a data consistency layer as well as a sigmoid. All models were trained using Adam \cite{kingma2014adam} with learning rate $10^{-4}$, $\beta=(0.5, 0.999)$ and weight decay $10^{-4}$. All models were trained for $300$ epochs. The images were undersampled using masks that were either drawn at random or drawn with a probability decaying away from the center.

The Neural Conditioner (NC) \cite{belghazi2019learning} generator and discriminator were updated every other round. In the WGAN case, we optimized the generator once, followed by four updates of the discriminator, as done in \cite{adler2018deep}.

\begin{remark}
    Similarly to \cite{belghazi2019learning}, we found gradient regularization of the discriminator to be essential. However, we obtained very poor training results using the authors proposed gradient penalty, and observed that the gradient penalty of \cite{gulrajani2017improved} led to a stable training.

    Also, \citet{belghazi2019learning} aim at controlling the encoder's Lipschitz constant by using spectral-normalization on the encoding part of the generator. However, we did not observe any effect of doing this, and did not retain it in our model.
\end{remark}

After training, we carried out several rounds of evaluation for different types of sampling as reported in Table \ref{tab:comp_mnist_im}. We found that $2$ samples were sufficient for the adaptive sampling with GAS to yield good results. We also observed that more samples $20$ to $100$ made little improvement in the sampling quality or image quality, except that it suppressed the noise of the empirical posterior mean and variance a bit more efficiently.

Regarding the stochastic LBCS approach \citep{sanchez2019scalable}, there are two parameters to be adjusted, namely the number of training samples to be used at each stage of the greedy mask optimization, and the size of candidate locations to be used. We set these to $64$ and $256$ respectively. We used the negative MSE as the performance metric.

In this experiment, as we have smaller images in image domain, we performed pixel-wise masking rather than line-wise masking like in MRI. This greatly increases the number of candidate locations that the policy can sample. We used for training a mixture of random masks and masks that sample more frequently locations near the center of the image.

Note also that the approach of \citet{zhang2019reducing} is \textit{not} applicable here, as it is not clear how to train their evaluator in image domain, due to their introduction of spectral maps. Also, even if it were adapted, their approach scales very poorly to pixel-wise sampling, as this would require the computation of a spectral map for each individual pixel, all of which are fed at once as input of the evaluator.

\subsubsection{Results}
\begin{figure}[!ht]
    \centering

    \includegraphics[valign=t,width=.9\linewidth]{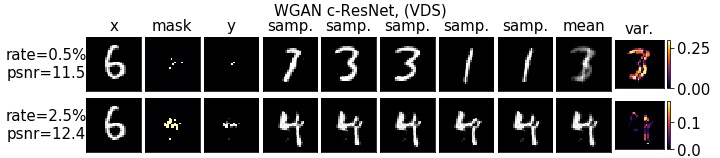}

    \includegraphics[width=.9\linewidth]{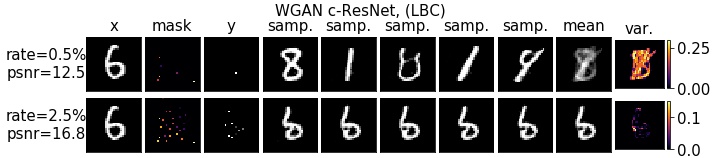}

    \includegraphics[width=.9\linewidth]{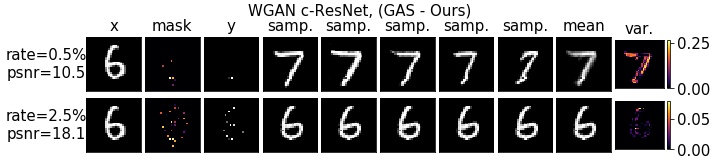}

    \caption{Image domain illustration of the WGAN cResNet sampling using three different methods. The columns contain respectively: the ground truth image, the mask (with brighter color meaning that the location has been selected recently), the observation, several samples and the empirical posterior mean and variances. The first row shows the result obtained with $0.5\%$ and $2.5\%$ sampling respectively. Here, VDS is our random policy, LBC is LBCS and GAS stands for Generative Adaptive Sampling and is our GAN-based policy.}\label{fig:mnist_image}
\end{figure}

Our main results are presented on Table \ref{tab:comp_mnist_im}, which provides the AUC for PSNR, SSIM, as well as the downstream classification accuracy. This was computed using a simple convolutional network that was pretrained on MNIST without any undersampling. A visual assessment of different policies is shown on Figure \ref{fig:mnist_image}.

Focusing first on Table \ref{tab:comp_mnist_im}, we see a very different picture than on the MRI case (Table \ref{tab:comp_knee}). Here, our $0$-step adaptive GAN policy very largely outperforms the $1$-step fixed LBCS policy across all metrics. This highlights a fundamentally different paradigm between image domain and Fourier domain sampling. In image domain, it seems that adaptivity matters much more than in Fourier. This makes intuitive sense, and is well illustrated in the case of MNIST, where the sampling pattern focuses on areas that are ambiguous, which can be very different for different images: this is illustrated more clearly on Figure \ref{fig:mnist_image2} in Appendix \ref{app:s_mnist}. This is also consistent with results on CIFAR10 shown in Appendix \ref{app:s_cifar}.

\begin{table}[!ht]
    \centering
    \begin{tabular}{lll|ccc}
        \toprule
        Policy & Model & Architecture & PSNR             & SSIM            & Accuracy        \\
        \midrule
        \multirow{2}{*}{Random}
               & NC    & ResUNet      & $17.10$          & $0.76$          & $0.67$          \\
               & WGAN  & c-ResNet     & $16.41$          & $0.76$          & $0.67$          \\
        \midrule
        \multirow{4}{*}{LBCS}
               & NC    & ResUNet      & $21.39$          & $0.90$          & $0.92$          \\
               & NC    & cResNet      & $20.33$          & $0.88$          & $0.93$          \\
               & WGAN  & ResUNet      & $20.29$          & $0.89$          & $0.90$          \\
               & WGAN  & cResNet      & $20.48$          & $0.89$          & $0.92$          \\

        \midrule
        \multirow{4}{*}{\begin{minipage}{1cm}GAN (Ours)\end{minipage}}
               & NC    & ResUNet      & $\mathbf{40.15}$ & $\mathbf{0.94}$ & $0.91$          \\
               & NC    & cResNet      & $31.47$          & $0.93$          & $\mathbf{0.93}$ \\
               & WGAN  & ResUNet      & $31.73$          & $0.92$          & $0.88$          \\
               & WGAN  & cResNet      & $\mathbf{35.81}$ & $\mathbf{0.96}$ & $\mathbf{0.95}$ \\

        \bottomrule
    \end{tabular}
    \caption{Average test set AUC (one AUC per image) on MNIST, in image domain.}\label{tab:comp_mnist_im}
\end{table}

It is surprising to see that, although the different generator architectures have roughly the same number of parameters, they can lead to very different performance when used as policies. The ResUNet performs significantly better with the NC and the cResNet shows a much better performance on the WGAN. However, there is not a significant difference on the performance with the LBCS policy. Although this is beyond the scope of this work, it would be interesting to form a better understanding of why such a discrepancy occurs.

Figure \ref{fig:mnist_image} features also an impressive ability of conditional GANs: although the model was only trained on pairs of data $(\vx,\vy)$ from a \textit{given} mode of the distribution, the model successfully produces varied conditional outcomes $\{\vx_i\}$ that span \textit{multiple modes} of the conditional distribution. Figure \ref{fig:mnist_image} also shows an issue that our GAN policy can encounter: sometimes, the model can be confidently wrong: at $0.5\%$ percent, the model thinks the digit should be a seven, which is incorrect\footnote{For other generative models such as variational autoencoders, it has been shown that these models can frequently be overconfident \citep{nalisnick2018deep}. Although evaluating the likelihood in GANs is an open problem, it is not unlikely that a similar behavior should be found in GANs.}. However, adding more samples allows it to converge to the correct digit. Note that in this case, it shows a level of uncertainty that is generally low, as opposed to the random policy (VDS) and LBCS, which feature a higher level of uncertainty.

\section{Related works}\label{ss:gans_rel}
In this work, we have investigated how GANs can be used for inverse problems to perform reconstruction, uncertainty quantification and adaptive sampling. A flurry of different approaches have been proposed to tackle these problems in the literature, and we briefly discuss a few particularly relevant works now.

Although we have used GANs as the generative model in this work, many other approaches have been explored in the context of inverse problems. Variational Autoencoders (VAEs) \citep{rezende2014stochastic,kingma2013auto} have been applied to inverse problems \citep{tonolini2019variational,zhang2021conditional} as well as the task of learning conditional distributions \citep{ivanov2018variational}. But other approaches such as autoregressive models
\citep{oord2016conditional,van2016pixel} or conditional normalizing flows \citep{papamakarios2019normalizing,sun2020deep,denker2021conditional} have also been explored. Lately, score-based approaches, based on the idea of learning to sample by learning the score of the distribution, have gained popularity \citep{ramzi2020denoising,song2021solving}.

Although all of these methods can provide reconstruction and uncertainty quantification, autoregressive models and score-based distribution suffer from slow inference times, limiting their potential use in a setting of adaptive acquisition in MRI, where the inference should take at most the same time as a readout. This can in principle be achieved by our model, which has an inference time in the order of milliseconds.

In the vein of the Neural Conditioner (NC) \citep{belghazi2019learning}, we mention the works of \citet{douglas2017universal,gautam2020masking} and a very recent energy based approach \citep{strauss2021arbitrary}. However, these approaches focus on missing data imputation or inpainting, and are not designed for settings where the acquisition does not happen in image domain. Other approaches train a conditional embedder produce a noise distribution which will subsequently enable conditional sampling \citep{whang2020approximate}, or motivate a conditional generator based on approximately triangular transport maps \citep{kovachki2020conditional}.

Generative adversarial networks have also been used in the context of inverse problems, and several works used unconditional GANs as priors \citep{bora2017compressed, anirudh2018unsupervised, latorre2019fast, narnhofer2019inverse,jalali2019solving, hussein2020image} and subsequently performed optimization in the latent space. Another approach has been to directly train GANs for reconstruction along additional losses in order to increase the quality of the reconstruction \citep{yang_dagan_2018}. In such settings, the noise channel of the generator is often discarded at test time as the aim is to obtain the best image quality.

There has also been some recent interest into quantifying the uncertainty in the reconstruction. This notably motivated the negative log-likelihood in \citet{zhang2019reducing}, where the level of uncertainty can be used as a criterion to decide when sampling can be stopped. A recent work, based on Stein's unbiased risk estimator, studied the calibration of the uncertainty for variational autoencoders \citep{edupuganti_uncertainty_2020}. Score-matching is also very promising for UQ, although it suffers from long inference times \citep{ramzi2020denoising,song2021solving}.

As discussed, \citet{zhang2019reducing} is the work more conceptually similar to ours. 
Their model aims at doing reconstruction and uncertainty quantification, while also providing adaptive sampling through their evaluator. It is also trained in an adversarial fashion, although it is not a GAN. However, their approach is \textit{not} generative, and only yields point estimates of the mean and the variance by minimizing a multivariate Gaussian negative log likelihood. Their approach supposes the pixel-wise uncertainty to be independent random variables normally distributed around the mean, an assumption not required in \citet{adler2018deep, belghazi2019learning}. This assumption also implies that sampling from this distribution to estimate the Fourier variance is not possible, as this would require access to the full covariance matrix, and our results confirm this.

We note also that the optimization of the sampling pattern is not restricted to MRI, but has also been applied on the standard compressed sensing setting, notably motivated by early work on Bayesian Compressive Sensing \citep{ji2008bayesian,wu2019deep,wu2019learning}.

Finally, our methodology relies on the idea of sequentially acquiring the most uncertain measurement at each round, a concept which has first been proposed and leveraged in the context of optimal experimental design \citep{lindley1956measure, chaloner1995bayesian, vanlier2012bayesian}. While until recently, most approaches in this realm relied on a statistical prior model as well as expensive MCMC sampling \citep{vanlier2012bayesian,pauwels2014bayesian}, but variational approaches have also been explored \citep{foster2019variational}. However, most of these approaches do not scale up to very high-dimensional problems. On the contrary, GANs are naturally fit to deal with high-dimensional problems, and provide with a natural way of empirically quantifying the uncertainty in this acquisition.

\section{Discussion}\label{sec:gans_discussion}

Our results show the potential of conditional GANs in tackling inverse problems, as they offer a good reconstruction, uncertainty quantification and adaptive sampling at once, by only training in a standard adversarial fashion. Nonetheless, several points are worth discussing

\textbf{Generically trained generative reconstruction method.} The current generative model was trained in a generic fashion and \emph{not} specifically to optimize the quality of masks designed through it. The ability for designing masks stems purely from training it as a rigorous Bayesian modeling of the continuum of inverse problems. This allows the posterior to be conditioned on incrementally collected information in a closed loop fashion. However, one could imagine ways to improve the quality of the policy, by explicitly training the model to not only provide good samples, but simulate a \textit{rollout} of samples. The discriminator could be then trained to distinguish not between samples but \textit{sequences} of samples.

Alternatively, our method could easily be incorporated in a reinforcement learning-based framework aimed at jointly training reconstruction and sampling such as \citet{jin2019self}. This could give the best of both worlds, giving principled uncertainty estimates to the RL sampler, moving beyond greedy sampling and possibly speeding up the training of the reconstruction method by focusing on regions with less reliable variance estimates instead of using masks sampled from parametric distributions as in this work.

\textbf{Limitations of the model.} While our approach and the one of \citet{belghazi2019learning} are successfully producing conditioned samples on unseen masks shapes, we observe that they nonetheless depend on the type of masks used throughout training. For instance, our method, being trained on pixel-wise undersampling for MNIST, would not be as good in the inpainting of large holes. Hand-crafting the type of masks that we want to condition on in the future remains an important step towards obtaining good generalization performance, and further work needs to be done to have a method able to truly condition on arbitrarily structured observations.

We also observed an expected trade-off in performance between models trained on a single type of masks and a single sampling rate and models trained on varying types of masks and sampling rates. In this sense, adaptive sampling comes at a cost: a small drop in performance compared to a method \textit{specialized} on a more restricted setting. Quantifying precisely this trade-off is beyond the scope of this work and should be further studied.

\textbf{Impact of data consistency.} We found the data-consistency layer to be a fundamental piece in all settings, as it forces the model to incorporate observed data into the output image, and the model is then much less prone to hallucinating unlikely features as more observations are being acquired. In addition, this ensures the fact that the input will never be discarded by the generator, and prevents the model from collapsing into a mode that is inconsistent with observations.

\textbf{Limitations of the posterior estimation.} We leveraged the approach of \citep{adler2018deep}, which is the first of its kind to allow to construct a posterior estimator from samples of the joint distribution. While it works well empirically, the authors did not provide any analysis or guarantees on how well the generator captures the tails of the posterior distribution. Our observations suggest that unusual images, that far away from the mean of the learned distribution, might not be accurately captured, i.e. the estimated variance is lower than expected. This could be due to the limited training data available, but might also an artifact of the cWGAN approach which tends to struggle with capturing weaker modes of their distributions. Specifically, the problematic examples might be an indication that while the loss shown in Equation \ref{eq:ao_objective} avoids mode collapse, there might still be some modes that are underrepresented.

\textbf{Adaptivity in image domain.} Contrarily to our conclusions in Chapter \ref{ch:rl_mri} and Section \ref{ss:mri_gan}, adaptive sampling seems to be greatly beneficial in image domain sampling. In particular, in a simple task such as MNIST, one can clearly see the benefit of adaptive policies, that tailor the sampling pattern to the relevant locations of the image (in the case of MNIST, the digit itself). This is also concurrent with our results on CIFAR10 (Appendix \ref{app:s_cifar}). This raises the question of predicting \textit{when} adaptive sampling would be beneficial over fixed sampling. Does it depend on the structure of the sampling domain itself, or would the distribution of images considered impact this as well? Does the adaptivity of the model depend on the distribution of masks used for training? All of these questions should be further investigated.

\section{Conclusion}\label{sec:gans_conclusion}

We presented a novel approach for reconstruction and adaptive sampling, rooted in Bayesian modeling. We showed that, in line with recent advances \citep{belghazi2019learning,adler2018deep}, GANs perform well at this task and yield useful posteriors for Fourier domain reconstruction as well as image domain, allowing both high quality reconstructions and derivation of a sampling heuristic. This natural sampling heuristic achieved surprisingly strong sampling results, without ever training to act as a policy, and is competitive with the state-of-the-art on the fronts of reconstruction, uncertainty quantification and adaptive sampling.

\section*{Bibliographic note}
Zhaodong Sun provided initial results in the work. Igor Krawczuk helped with the code, implemented and provided the results for the methods of \citet{zhang2019reducing} and \citet{bakker2020experimental}, and carried out the CIFAR10 experiments. Volkan Cevher provided the initial motivation to look into the Deep Bayesian Inversion approach of \citet{adler2018deep} which started this project.

\cleardoublepage
\cleardoublepage
\chapter{Conclusion and future works}\label{ch:conclusion}
\markboth{Conclusion and future works}{Conclusion and future works}

\section*{Summary of the thesis}

In this thesis, we studied the problem of optimizing sampling patterns for Cartesian MRI. More specifically, we considered the problem of optimizing sampling patterns in a sequential context, using learning-based, data-driven methods. Our work consists of both methodological novelties and insights into what impacts the performance of models, what design choices matter most and \textit{why} some methods generally perform worse than others.

In \textbf{Chapter \ref{ch:sampling}}, we brought a taxonomy of approaches that have been taken to tackle the problem of optimizing sampling patterns, which helped shed light on the development of approaches to optimize sampling for Cartesian MRI. More concretely, we proposed that similarly to reconstruction methods, the field of mask optimization has undergone a shift from \textit{model-based} to \textit{learning-based} approaches \citep{ravishankar2019image,doneva2020mathematical}, but that it has also shifted from \textit{model-driven} towards \textit{data-driven} criteria to evaluate the quality of sampling masks. It was naturally observed that moving toward learning-based, data-driven approaches enables the model to perform better \citep{chauffert2013variable,zijlstra2016evaluation}, a trend that was also consistently confirmed throughout our work. We also observed that within the realm of learning-based, data-driven methods, several trends arose, where researchers tackled the problem of learning to design masks as either a fixed, static problem or a sequential, dynamic approach. Following the sequential approach to mask design, our work focused on studying the problem of optimizing sampling on pre-trained reconstruction models. This constitutes the precise framework of our contributions.

Within this context, in \textbf{Chapter \ref{chap:lbcs}}, we first investigated algorithms to drastically improve the scaling of LBCS \citep{gozcu2018learning}.  We showed that ideas from submodular optimization \citep{krause2014submodular} can be leveraged to scale up LBCS by several orders of magnitude, and in some cases without degrading its performance. Our algorithms, sLBCS and lLBCS, have been shown to provide strong mask designs for a range of clinically relevant MRI settings, including multicoil dynamic MRI and 3D MRI. These settings were previously not accessible to LBCS, due to the prohibitive cost of evaluating all candidate locations on the full dataset. sLBCS tackled this limitation by proposing a stochastic sampling of both candidate locations and data points, and achieved results on par with LBCS, and even sometimes outperforming the method. This method is easily implemented and broadly applicable, and we used it as a strong baseline throughout the rest of the thesis. lLBCS was proposed to address the case of multicoil and 3D MRI, where LBCS cannot be evaluated, due to its prohibitive computational cost. By using lazy evaluations, lLBCS can scale up to these problems. lLBCS presents very strong performances at high undersampling rates, but we hypothesized that its list of upper bounds becomes gradually worse as more sampling locations are added to the mask, making its performance gradually worse than baselines. However, several options were proposed to further improve its performance and scalability, and should be investigated as future work.

Our work illustrates the benefit of a flexible learning-based, data-driven approach to the problem of optimizing sampling. Moreover, by focusing on the problem of optimizing sampling patterns on pre-trained reconstruction models, sLBCS and lLBCS integrate a future-proof dimension, as they will remain readily available to any reconstruction method that is applicable to a sequence of sampling rates.

In \textbf{Chapter \ref{ch:rl_mri}}, we aimed at resolving a seeming conflict between mask design methods that used deep reinforcement learning \citep{pineda2020active,bakker2020experimental}. These methods showed a potential to improve over sLBCS and lLBCS by integrating long-horizon sampling as well as patient-adaptive decisions. However, \citet{pineda2020active} and \citet{bakker2020experimental} reached seemingly contradictory conclusions. \citet{pineda2020active} suggested that long-term planning was the most important component, as their Dataset-Specific DDQN (non-adaptive) policy often matched the performance of their Subject-Specific DDQN (adaptive) policy. On the side of \citet{bakker2020experimental}, it was found that a \textit{greedily} trained adaptive deep RL policy could match the performance of policies trained with a long-horizon dependency. In other words, \textit{does the ability to plan multiple steps ahead matter most, as suggested by \citet{pineda2020active}, or is it rather the ability to adapt sampling to a given patient, as suggested by \citet{bakker2020experimental}?}

Our methodology was simply to consider a baseline policy that is greedily trained (no long-term planning) and non-adaptive, but that is trained to maximize the same reward function as considered by \citet{pineda2020active,bakker2020experimental}. This baseline could then serve as the reference point from which one could observe what brings most performance improvement. However, it became quickly clear that our previously proposed algorithm, sLBCS, could exactly fulfill this role, and it was then natural to use it as the comparison baseline.  Exhaustive experiments allowed us to bring elements of answers to this important question, albeit in a surprising way: it seems indeed that \textit{neither long-term planning, nor adaptive sampling} are able to bring a strong return on investment over sLBCS.

Exploring this question led us to consider the different processing choices that were done in the works of \citet{pineda2020active} and \citet{bakker2020experimental}, such as the choices of architecture, how the models were pretrained, the metrics used for reporting and how results were aggregated and presented. Surprisingly again, we showed that \textit{inconsistencies in reporting could easily lead to incoherent results}, and that subtle changes in the processing or the reporting pipeline could lead to cases where the SotA deep RL methods of \citet{pineda2020active,bakker2020experimental} were outperformed by sLBCS.

Moreover, the return on investment of moving from sLBCS to a deep RL policy model was at best marginal compared to other changes such as the architecture or the mask distribution used to pretrain it. Indeed, our results show very clearly that the benefit of using a deep RL policy over sLBCS is very limited compared to the improvements achieved by improving the reconstruction architecture or using a distribution of sampling masks tailored to the acquisition horizon considered. However, there is a clear benefit of moving from hand-crafted heuristics such as random sampling and low-pass acquisition policies to a \textit{learning-based}, \textit{data-driven} such as sLBCS.

These observations led us to wonder whether these results are due to current limitations of deep RL policies or if they describe some more fundamental limitations of the problem. We believe the second option to be more plausible, in light of more recent works such as \citet{bakker2021learning}, where in their case, the best performing adaptive policies \textit{explicitly learn} to become non-adaptive. This also opens some further questions: could we design even simpler \textit{learning-based}, \textit{data-driven} approaches that retain the benefits of sLBCS, while being even easier to train? Can we also find new paradigms of acquisition that would allow to further improve over the \textit{learning-based}, \textit{data-driven} paradigm?

Finally, in \textbf{Chapter \ref{ch:gans}}, we proposed a novel approach to sampling, based on Bayesian modeling. Inspired by previous works to model inverse problems in a Bayesian fashion using GANs \citep{adler2018deep}, we showed that in the context of MRI, this Bayesian modeling allows us to not only perform reconstruction and uncertainty quantification, but also provides a \textit{natural} way to perform adaptive sampling. Indeed, we show that acquiring the location with the largest posterior variance provided a good, natural policy, without the model ever being trained explicitly to perform adaptive sampling. Our GAN-based approach was however outperformed by sLBCS and RL-based policies, but did outperform the approach of \citet{zhang2019reducing}. We postulated that the reason for this is that our GAN approach and the one of \citet{zhang2019reducing} rely on $0$-step information, i.e. base their decision on the current error, rather than receiving feedback about how their decision \textit{actually} performed, which we describe as $1$- or $N$-step information.

Nonetheless, this last work provides a promising all-in-one approach to Bayesian modeling of inverse problems, and would deserve to be investigated in greater depth. GAN-based approach have generally achieved strong performances in MRI reconstruction \citep{chen2022ai}, and seem like a promising approach to provide an integrated approach to MRI reconstruction, uncertainty quantification and sampling. But as our methodology is tied to learning the posterior rather than GANs themselves, we anticipate that novel contributions to Bayesian modeling would also benefit from these observations, further giving credit to learning the full posterior distribution rather than the pixel-wise mean and variance.

\section*{Insights}
\begin{figure}[!t]
    \centering
    \includegraphics[width=\linewidth]{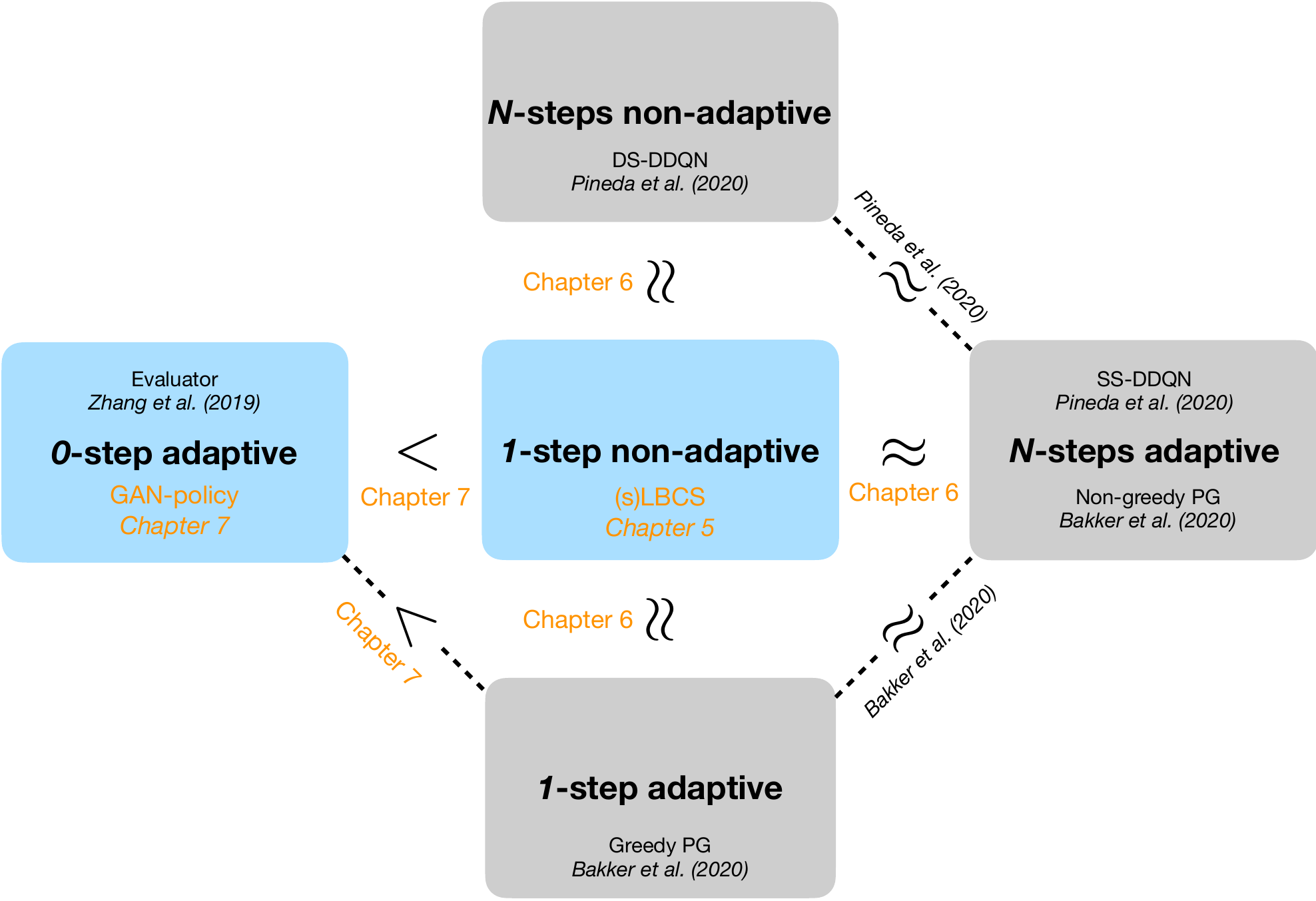}
    \caption{Summary of the contributions of this thesis relating to the design of a mask optimization algorithm. }\label{fig:summarize_contributions}
\end{figure}

In addition to the methodological contributions, our work provides insights into how sampling policies can be designed efficiently. Figure \ref{fig:summarize_contributions} shows the connections that we established in this thesis. We used the common notions of fixed (non-adaptive) and adaptive policies and refined the distinction between greedy and long-horizon policies to include the \textit{information horizon} accessible to the policy, distinguishing between $0$-step and $1$-step greedy policies. The first are driven by the error at the current reconstruction, whereas the second integrates feedback about the \textit{actual} effect on reconstruction of adding a sampling location to the mask.

We found that this is the most significant distinction between the existing methods, as in Chapter 6, our results suggest that at least in their current state, neither long-horizon planning ($N$-steps information horizon) nor adaptivity allow significant and consistent improvements over (s)LBCS, a $1$-step non-adaptive policy. The distinction between 0-step and 1-step methods was explored in Chapter 7, where we showed that even an adaptive oracle that acquires the line with the largest current mean-squared error is outperformed by a non-adaptive policy that leverages $1$-step information.

Overall, our experiments suggest that the most important structure that a policy should include in order to achieve near state-of-the-art performance is $1$-step information, i.e. taking into account the \textit{real} impact on performance of acquiring a new location. Further refinements, at least in the current state, seem to bring at most very modest performance improvements, challenging the common trend to design ever-more complex models.

\newpage

\section*{Future works}
Nevertheless, even if our thesis brings methodological contributions and insights to the problem of learning sampling masks for Cartesian MRI, many important questions still remain open.

This thesis does \textit{not} provide a definitive answer to whether the low return on investment of deep RL methods applied to MRI results from current limitation or if it arises from an intrinsic structure of the problem. It is not impossible that further advances in deep learning would allow to efficiently train complex models, and to achieve significant performance gain over methods such as (s)LBCS. Similarly, although the results of this thesis provide evidence toward the validity of the conclusions of Figure \ref{fig:summarize_contributions}, it is not impossible either that a deeper connection would be able to explain these observations more adequately. As a result, further work, both at the theoretical and practical levels would be necessary in order to understand what really matters in sampling mask optimization.

Moreover, our results focus on \textit{pre-trained reconstruction models} for sequential acquisitions, and it is not clear how \textit{jointly} training the reconstruction and the policy models might affect the picture given in Figure \ref{fig:summarize_contributions}. Indeed, joint training is challenging as it adds variance to the training procedure, and it is difficult to estimate its impact.
Although a few works considered this problem \citep{jin2019self,van2021active,yin2021end}, it remains to be understood whether joint training mainly results in stronger reconstruction models by enabling them to focus on a more restricted set of masks, in stronger policies by co-designing them with the reconstruction model, or if both, or none, of these conclusions hold. Answering these questions would be an important step towards designing end-to-end, data-driven MRI pipelines.



A crucial interrogation relates however to the clinical value of optimized sampling masks for MRI sampling. It is known  that there are some significant artifacts introduced by prospective undersampling, but their effect has not been sufficient studied in the case of deep learning-based MRI reconstruction \citep{yu2022validation}. However, there have not been any studies that consider the distortion of a prospective acquisition when using an optimized sampling. Are there ways to design masks that are robust to the distortions of prospective sampling and can retain good performance? Although robustness is a growing concern in MRI reconstruction \citep{antun2020instabilities,johnson2021evaluation,darestani2021measuring}, most works only consider retrospective undersampling.

Additionally, the world of MR imaging is not restricted to Cartesian imaging, and many exciting recent works also aim at optimizing sampling trajectories for non-Cartesian acquisitions \citep{lazarus2019sparkling,weiss2019pilot,wang2021b,chaithya2022benchmarking}. Non-Cartesian acquisitions can be in practice much faster than Cartesian trajectories, and although they suffer from different limitations, they are definitely a topic that would be worth further investigating.

Beyond these more conceptual directions, many technical questions need also to be addressed for the field of MRI sampling optimization to really mature into a clinically relevant application. We should clarify what really matters at different levels, first in terms of metrics: what kind of metrics do we want to use? Where does the reconstruction quality matters? In terms of the problems considered themselves: should we care about sequential reconstruction or optimize a fixed mask? Do we want to jointly train reconstruction methods and sampling policies? How can we ensure that a trained policy would be robust in a prospective acquisition?

These questions leave many avenues of research open, and promises exciting discoveries in the years to come.

\cleardoublepage

\addtocontents{toc}{\vspace{\normalbaselineskip}}
\cleardoublepage
\bookmarksetup{startatroot}

\appendix
\chapter{Appendix for Chapter \ref{ch:rl_mri}}

\section{Implementation details}\label{ap:implementation}
\textbf{Pretraining regimes.} The deep reconstructors used in \cite{pineda2020active,bakker2020experimental} are pre-trained by randomly sampling a mask from a set of distributions with different parameters. \cite{bakker2020experimental} uses a discrete set of sampling rates at ($25\%$,$25\%$,$25\%$, $16.7\%$,$16.7\%$, and $12.5\%$) with ($25\%$,$16.7\%$,$12.5\%$, $16.7\%$,$12.5\%$,$12.5\%$) of the total selected frequencies being allocated to center frequencies and the rest sampled uniformly,while  \cite{zhang2019reducing} uses a distribution  where $10$ center frequencies are always acquired, and between $0$ and $38$ additional frequency lines are acquired following a uniform distribution (as a reminder, the total number of possible sampling locations is $128$ lines or columns).

All UNet models of the ablations of the \citet{bakker2020experimental} setting were trained following the hyperparameters described in the paper, i.e. using Adam \citep{kingma2014adam} and training to 50 epochs, otherwise using the hyperparameters from \citet{zbontarFastMRIOpenDataset2019} (except for the $10\times$ learning rate drop 40 epochs, we instead kept the initial $10^{-3}$ rate throughout).
We used a batch size of $32$.

For the cResNet models used in the ablations we used Adam as an optimizer with learning rate $10^{-3}$,betas of $(0.9,999)$, trained for $50$ epochs and a batch size of $32$. The architecture is the same as in \citet{zhang2019reducing} except only outputting the mean of the image (no uncertainty channel), and using $72$ channels for the 3 residual blocks, with $18,36,72$ channels in the encoder and $72,36,18$ channels in the decoder. In the comparison with \citet{pineda2020active}, we used the checkpoint provided at \url{https://facebookresearch.github.io/active-mri-acquisition/misc.html} by the authors.

\textbf{Computational hardware.} We performed all of your experiments on a DGX-2 server using $A100$ GPUs. On this machine the reconstruction model of \cite{pineda2020active} fits into GPU memory with an effective batchsize of $50$ (we use subbatching to enable arbitrary batch sizes) and it took $\sim 3$ hours to train each of the reported LBCS masks. Meanwhile, masks for the comparisons of \cite{bakker2020experimental} could be trained in $\sim 20$ minutes, while training the RL policy took on the order of days as reported in the authors original paper.
While one might be able to obtain a certain speedup using a more optimized or parallel RL algorithm, a key bottleneck is the sequential nature of the optimization.

\clearpage
\FloatBarrier

\section{Further experiments and results on Bakker's setting}
\subsection{Data processing differences}\label{app:bakker_pipeline_ablation}\label{app:bakker_mismatch}
In section \ref{s:re_examining}, we mentioned that we slightly deviated from the setting of \cite{bakker2020experimental}. We now detail these changes and show that they do not change the \textit{relative} performance of the different methods.
\begin{enumerate}
    \item \textbf{Train-test split:} we used a different train-test split than \cite{bakker2020experimental}, as we randomly split $10\%$ of the training set as a test set, and used the fastMRI validation set for test. We used the $50\%$ more central slices, resulting in $15599$ training slices, $1743$ validation slices and $3564$ test slices.
    \item \textbf{Complex data undersampling:} \cite{bakker2020experimental} use magnitude ground truth images as the reference that is undersampled. We discussed in \Cref{sec:pipeline} the issue with this approach. We chose to use complex preprocessing of data, followed by taking magnitude of the observation obtained \textit{after} undersampling.
    \item \textbf{Data range: }PSNR and SSIM need to be provided with a maximal data range in their computation: \cite{bakker2020experimental} used the maximal intensity in the ground truth  \textit{volume}, while we used the maximal intensity of each ground truth \textit{slice} or image.
    \item \textbf{Data standardization:} \cite{bakker2020experimental} used a unique data standardization, where observations and ground truth data are standardized using their respective statistics, but denormalized, after reconstruction, using only the ground truth statistics. While this ensures a more closely matching data range, this introduces a mismatching data range that biases models leveraging data consistency, used in most state-of-the-art models. We performed \textit{matched} data normalization and denormalization using the observation statistics for the observation, and the ground truth ones for the ground truth.
\end{enumerate}

The third and fourth changes are related to postprocessing, and do \textit{not} require retraining a model. We see on the three first columns of Table \ref{tab:bakker_pipeline_ssim_auc_sh} that these changes do not alter the relative ordering of the methods. However, in absolute numbers, the worse performing method in the initial setting outperforms the best method the best performing one after these changes (highlighted in italic on Table \ref{tab:bakker_pipeline_ssim_auc_sh}). While comparing these numbers would be a mistake, this highlights the impact of subtle postprocessing changes and the need for care in comparing methods, especially across different papers.

\textbf{Ablation over Magnitude vs Complex data.} We first illustrate the impact of doing the appropriate preprocessing by evaluating the model of \cite{bakker2020experimental} using their data processing pipeline against a pipeline with complex preprocessing. The SSIM performance AUC is reported in Table \ref{tab:bakker_pipeline_ssim_auc_sh}.

This would correspond to a simulation of what would happen if the model of \cite{bakker2020experimental} was to be used in deployment, as one would receive undersampled complex observations that are only then transformed to magnitude images. While this does not induce any \textit{reversal} in this case, we see that in the matched setting, the gap between LBCS and RL shrinks. We see also again that training and evaluating on the matched setting leads to larger performance improvements than the ones obtained by training sophisticated policies.


\begin{figure}[!ht]
    \begin{center}
        \resizebox{\textwidth}{!}{\begin{tabular}{lccc|cc}
\toprule

\multirow{2}{*}{\textbf{Policy}} & \textbf{Bakker's } & \textbf{Individual } & \textbf{Matched } & \textbf{Mismatched} & \textbf{Complex preprocessing} \\
& \textbf{setting} & \textbf{data range}& \textbf{standardization}& \textbf{setting}&  \textbf{(matched - ours)}\\
\midrule
\textbf{Random (SH)} & \textit{0.6251} & 0.5964 & 0.5798 & 0.5595 & 0.5608 \\
\textbf{LtH (SH)} & 0.6073 & 0.5762 & 0.5650 & 0.5617 & 0.5719 \\
\textbf{LBCS (SH)} & 0.6375 & 0.6099 & 0.5928 & 0.5684 & 0.5732 \\
\textbf{RL (SH)} & \textbf{0.6388} & \textbf{0.6112} & \textit{\textbf{0.5941}} & \textbf{0.5698} & \textbf{0.5738} \\
\midrule
\textbf{NA Oracle (SH)} & 0.6383 & 0.6107 & 0.5937 & 0.5717 & 0.5738\\
\bottomrule
\end{tabular}}
        \captionof{table}{AUC on the test set, using SSIM, on various data processing on Bakker's knee setting. The mismatched setting refers to the evaluation of the model trained in the setting of \cite{bakker2020experimental}, but evaluated complex data preprocessing instead of magnitude data preprocessing. This makes the reconstruction and policy models to be out of the distribution. A model trained on the complex preprocessing is reported in the complex preprocessing (matched - ours) column.}\label{tab:bakker_pipeline_ssim_auc_sh}
    \end{center}
\end{figure}

\subsection{Exact reproduction of Bakker et al.}
In our experiments, we did not use the exact same train-val-test split and dataset undersampling as \citet{bakker2020experimental}. We re-evaluated our model in the exact setting of \citet{bakker2020experimental}, as illustrated on the first column of Table \ref{tab:bakker_pipeline_ssim_auc_sh}. On Table \ref{tab:bakker_pipeline_ssim_at_25_sh}, we present the results at $25\%$ sampling against the ones reported in the paper of \citet{bakker2020experimental}.

\begin{figure}[!ht]
    \begin{center}
        \resizebox{0.6\textwidth}{!}{\begin{tabular}{lccc}
    \toprule
\textbf{Policy (SH)} & \textbf{Our code} & \textbf{Bakker} & \textbf{Difference}\\
\midrule
\textbf{Random} & 0.6974 & 0.6948 & \textbf{+0.0026}\\
\textbf{Equi (one)} & 0.7078 & 0.7049  & \textbf{+0.0029}\\
\textbf{Equi (two)} & 0.7090 & 0.7064  & \textbf{+0.0026}\\
\textbf{NA Oracle} & 0.7235 & 0.7213 & \textbf{+0.0022}\\
\textbf{LBCS} & 0.7222 & --  & --\\
\textbf{RL (Greedy)} & 0.7250 & 0.7230 & \textbf{+0.0020}\\
\textbf{RL $\mathbf{\gamma=0.9}$} & 0.7250 & 0.7232 & \textbf{+0.0018}\\
\bottomrule
\end{tabular}}
        \captionof{table}{SSIM at $25\%$ on the test set used by \citet{bakker2020experimental} and subsampling the dataset by a factor $2$, in order to replicate their setting as close as possible. We compare the numbers that we obtain to the ones reported in the article of \citet{bakker2020experimental}.}\label{tab:bakker_pipeline_ssim_at_25_sh}
    \end{center}
\end{figure}

We see that the performance difference is in the same order of $0.0018$ to $0.0029$, and that the relative ordering of methods is preserved. We conclude from these results that our results are overall consistent with the results displayed in the paper of \citet{bakker2020experimental}, although this is not an \textit{exact} replication.


\subsection{Further results in the Bakker experiment}\label{app:bakker_ablation_full}
We provide the results of our full ablation study on the components described in section \ref{s:re_examining}. We report observations on the \texttt{cvb}, \texttt{cvz}, \texttt{c+rvz} and \texttt{c+rhz} settings. The summary of the full trajectory with an AUC is presented on Table \ref{tab:bakker_auc_ssim_full} and the performance at the end of sampling is shown in Table \ref{tab:bakker_at_25_ssim_full}.

It is difficult to consistently establish a trend for when reversals will happen, but several important observations can be made from the results.

First of all, by comparing Tables \ref{tab:bakker_auc_ssim_full} and Table \ref{tab:bakker_at_25_ssim_full}, we see that reversals can happen by considering one way of reporting over another (see \texttt{c+rvz} cResNet in both tables). This highlights again the impact of the way results are reported.

The effect of the masks used for pretraining the reconstructor is also interesting. Comparing the \texttt{cvb} and \texttt{cvz} results, once can consistently see the following trend. In the short horizon setting (LHS of the tables), using \texttt{cvb} leads to consistent improvements over \texttt{cvz}. In the long horizon setting, the opposite is true. Recall that the \texttt{b} masks are discretely distributed from sampling rates $12.5\%$ to $25\%$, which matches the short horizon experiment range. The \texttt{z} masks span a continuous range, from roughly $7\%$ to $37.5\%$, whereas the long horizon experiment spans a range from $3\%$ to $25\%$ sampling rate. These results suggest that matching the pretraining regime to the regime on which evaluation will be carried out has a significant influence on the performance of the sampling policy, an observation which has, to our knowledge, never been quantified before.

\textbf{PSNR evaluation.} We also provide a PSNR evaluation for the models trained on SSIM in Tables \ref{tab:bakker_auc_psnr_full} and \ref{tab:bakker_at_25_psnr_full}. There does not seem a clear trend or correlation between the policy's performance on SSIM and PSNR. We observe however the same kind of dynamics in the settings, where cropped + resized data naturally have a higher PSNR than cropped ones, and cResNet improves the reconstruction quality of UNet.

\begin{figure}[!ht]
    \begin{center}
        \resizebox{0.95\textwidth}{!}{\begin{tabular}{lcccc|cccc}
\toprule
\multirow{3}{*}{\textbf{Policy}} & \multicolumn{4}{c|}{\textbf{Short Horizon}} & \multicolumn{4}{c}{\textbf{Long Horizon}} \\
\cmidrule(l){2-5}\cmidrule(l){6-9}
 & \multicolumn{2}{c}{\texttt{cvb}} & \multicolumn{2}{c|}{\texttt{c+rhz}} & \multicolumn{2}{c}{\texttt{cvb}} & \multicolumn{2}{c}{\texttt{c+rhz}} \\
  \cmidrule(l){2-3}\cmidrule(l){4-5}\cmidrule(l){6-7}\cmidrule(l){8-9}
 & \multicolumn{1}{c}{\textbf{UNet}} & \multicolumn{1}{c}{\textbf{CResNet}} & \multicolumn{1}{c}{\textbf{UNet}} & \multicolumn{1}{c|}{\textbf{CResNet}} & \multicolumn{1}{c}{\textbf{UNet}} & \multicolumn{1}{c}{\textbf{CResNet}} & \multicolumn{1}{c}{\textbf{UNet}} & \multicolumn{1}{c}{\textbf{CResNet}} \\
\midrule
\textbf{Random}    & {0.5608}            & {0.5659}  & {0.7292} & {0.7338}  &  0.4348 & 0.4432&0.5860& 0.5885\\
\textbf{LtH}         & {0.5719}            & 0.5764         & 0.7309      & 0.7412   & 0.4849   & 0.5107&0.6678 &0.6902      \\
\textbf{LBCS}        &{0.5732}            & 0.5828           & \textbf{0.7430} & \textbf{0.7526} & 0.5134& \textbf{0.5243}&\textbf{0.7035} &0.7096 \\
\textbf{RL }         & \textbf{0.5738}            & \textbf{0.5830}   & \textbf{0.7430} & 0.7524 &\textbf{0.5142} & \textbf{0.5242} &\textbf{0.7035}&\textbf{0.7111} \\
\midrule
\textbf{NA Oracle}          & \multicolumn{1}{l}{0.5738} & {0.5832}  & 0.7430          & 0.7527& 0.5140& 0.5247  & 0.7038 &0.7099\\[2mm]
\bottomrule
\toprule
\multirow{3}{*}{\textbf{Policy}} & \multicolumn{4}{c|}{\textbf{Short Horizon}} & \multicolumn{4}{c}{\textbf{Long Horizon}} \\
\cmidrule(l){2-5}\cmidrule(l){6-9}
 & \multicolumn{2}{c}{\texttt{cvz}} & \multicolumn{2}{c|}{\texttt{c+rvz}} & \multicolumn{2}{c}{\texttt{cvz}} & \multicolumn{2}{c}{\texttt{c+rvz}} \\
 \cmidrule(l){2-3}\cmidrule(l){4-5}\cmidrule(l){6-7}\cmidrule(l){8-9}
 & \multicolumn{1}{c}{\textbf{UNet}} & \multicolumn{1}{c}{\textbf{CResNet}} & \multicolumn{1}{c}{\textbf{UNet}} & \multicolumn{1}{c|}{\textbf{CResNet}} & \multicolumn{1}{c}{\textbf{UNet}} & \multicolumn{1}{c}{\textbf{CResNet}} & \multicolumn{1}{c}{\textbf{UNet}} & \multicolumn{1}{c}{\textbf{CResNet}} \\
 \midrule
\textbf{Random } & {0.5580} & 0.5661 & 0.6851 & 0.6984 & 0.4483 & 0.4621 & 0.5252 & 0.5276 \\
\textbf{LtH } & {0.5663} & 0.5759 & {0.6739} & {0.7028} & 0.5122 & 0.5218 & 0.6292 & 0.6444 \\
\textbf{LBCS } & {0.5712} & {0.5822} & {0.7034} & 0.7235 & 0.5174 & 0.5328 & 0.6568 & 0.6712 \\
\textbf{RL } & \textbf{0.5717} & \textbf{0.5829} & \textbf{0.7042} & \textbf{0.7239} & \textbf{0.5183} & \textbf{0.5334} & \textbf{0.6582} & \textbf{0.6733} \\
\midrule
\textbf{NA Oracle } & 0.5718 & 0.5835 & 0.7038 & 0.7236 & 0.5180 & 0.5336 & 0.6568 & 0.6717\\
\bottomrule
\end{tabular}}
        \captionof{table}{AUC on the test set, using SSIM, for the full ablation study using the model of \cite{bakker2020experimental}. The short and long horizon results are \textit{not} comparable with each other, as AUCs are integrated on the whole range of sampling rates. The top right part of the table (long horizon) replicated the results of table \ref{tab:bakker_LH_auc}, excluding the standard deviation for legibility. The rest of the ablation were \textit{not} averaged on several seeds due for computational reasons. Recall that \texttt{cvb} stands for cropped, vertical lines, Bakker-like mask distribution, \texttt{c+rhz} stands for cropped then resized, horizontal lines and Zhang-like masks, \texttt{cvz} stands for cropped, vertical lines and Zhang-like masks and \texttt{c+rvz} stands for cropped then resized, vertical lines and Zhang-like masks.}\label{tab:bakker_auc_ssim_full}
    \end{center}
\end{figure}

\begin{figure}[!ht]
    \begin{center}
        \resizebox{0.95\textwidth}{!}{\begin{tabular}{lcccc|cccc}
\toprule
\multirow{3}{*}{\textbf{Policy}} & \multicolumn{4}{c|}{\textbf{Short Horizon}} & \multicolumn{4}{c}{\textbf{Long Horizon}} \\
\cmidrule(l){2-5}\cmidrule(l){6-9}
 & \multicolumn{2}{c}{\texttt{cvb}} & \multicolumn{2}{c|}{\texttt{c+rhz}} & \multicolumn{2}{c}{\texttt{cvb}} & \multicolumn{2}{c}{\texttt{c+rhz}} \\
 \cmidrule(l){2-3}\cmidrule(l){4-5}\cmidrule(l){6-7}\cmidrule(l){8-9}
 & \textbf{UNet} & \textbf{CResNet} & \textbf{UNet} & \textbf{CResNet} & \textbf{UNet} & \textbf{CResNet} & \textbf{UNet} & \textbf{CResNet} \\
 \midrule
\textbf{Random} & {0.607} & 0.6141 & 0.7565 & 0.7667 & 0.5249 & 0.5432 & 0.6567 & 0.6567 \\
\textbf{LtH } & {0.6267} & 0.6313 & {0.7602} & {0.7786} & 0.5832 & 0.6197 & 0.7325 & 0.7714 \\
\textbf{LBCS } & {0.6288} & {0.6413} & \textbf{0.7751} & \textbf{0.7923} & 0.6294 & \textbf{0.6417} & \textbf{0.7768} & \textbf{0.7941} \\
\textbf{RL } & \textbf{0.6298} & \textbf{0.6417} & {0.7745} & \textbf{0.7921} & \textbf{0.6298} & \textbf{0.6415} & {0.7761} & {0.7935} \\
\midrule
\textbf{NA Oracle } & 0.6301 & 0.6421 & 0.7751 & 0.7926 & 0.6301 & 0.6428 & 0.7771 & 0.7942\\[2mm]
\bottomrule
\toprule
\multirow{3}{*}{\textbf{Policy}} & \multicolumn{4}{c|}{\textbf{Short Horizon}} & \multicolumn{4}{c}{\textbf{Long Horizon}} \\
\cmidrule(l){2-5}\cmidrule(l){6-9}
 & \multicolumn{2}{c}{\texttt{cvz}} & \multicolumn{2}{c|}{\texttt{c+rvz}} & \multicolumn{2}{c}{\texttt{cvz}} & \multicolumn{2}{c}{\texttt{c+rvz}} \\
 \cmidrule(l){2-3}\cmidrule(l){4-5}\cmidrule(l){6-7}\cmidrule(l){8-9}
 & \textbf{UNet} & \textbf{CResNet} & \textbf{UNet} & \textbf{CResNet} & \textbf{UNet} & \textbf{CResNet} & \textbf{UNet} & \textbf{CResNet} \\
 \midrule
\textbf{Random } & {0.6053} & 0.6148 & 0.7201 & 0.7401 & 0.5502 & 0.5693 & 0.6191 & 0.6359 \\
\textbf{LtH } & {0.619} & 0.6309 & {0.7052} & {0.7487} & 0.6190 & 0.6309 & 0.7052 & 0.7487 \\
\textbf{LBCS } & {0.6282} & {0.6413} & {0.7411} & {0.7710} & 0.6293 & \textbf{0.6451} & {0.7453} & \textbf{0.7746} \\
\textbf{RL } & \textbf{0.6288} & \textbf{0.6416} & \textbf{0.7417} & \textbf{0.7714} & \textbf{0.6304} & \textbf{0.6452} & \textbf{0.7470} & {0.7738} \\
\midrule
\textbf{NA Oracle } & 0.6292 & {0.643} & 0.741 & 0.7717 & {0.6304} & {0.6464} & 0.7457 & 0.7755\\
\bottomrule
\end{tabular}

}
        \captionof{table}{SSIM at 25\% sampling rate, using SSIM, with the model of \cite{bakker2020experimental}. This is the counterpart of the results shown in Table \ref{tab:bakker_auc_ssim_full}. Here, the results across short and long horizon are comparable.}\label{tab:bakker_at_25_ssim_full}
    \end{center}
\end{figure}

\clearpage
\FloatBarrier

\begin{figure}[!t]
    \begin{center}
        \resizebox{.95\textwidth}{!}{\begin{tabular}{lcccc|cccc}
\toprule
\multirow{3}{*}{\textbf{Policy}} & \multicolumn{4}{c|}{\textbf{Short Horizon}} & \multicolumn{4}{c}{\textbf{Long Horizon}} \\
\cmidrule(l){2-5}\cmidrule(l){6-9}
 & \multicolumn{2}{c}{\texttt{cvb}} & \multicolumn{2}{c|}{\texttt{c+rhz}} & \multicolumn{2}{c}{\texttt{cvb}} & \multicolumn{2}{c}{\texttt{c+rhz}} \\
  \cmidrule(l){2-3}\cmidrule(l){4-5}\cmidrule(l){6-7}\cmidrule(l){8-9}
 & \multicolumn{1}{c}{\textbf{UNet}} & \multicolumn{1}{c}{\textbf{CResNet}} & \multicolumn{1}{c}{\textbf{UNet}} & \multicolumn{1}{c|}{\textbf{CResNet}} & \multicolumn{1}{c}{\textbf{UNet}} & \multicolumn{1}{c}{\textbf{CResNet}} & \multicolumn{1}{c}{\textbf{UNet}} & \multicolumn{1}{c}{\textbf{CResNet}} \\
\midrule
\textbf{Random}    & {24.206}            & {24.180}  & {27.641} & {28.365}  &  20.802 & 20.836 &22.375& 22.371\\
\textbf{LtH}         & {24.486}            & 24.420         & 27.628      & 28.640   & 23.570   & 23.578&26.607 &27.317      \\
\textbf{LBCS}        &\textbf{24.493}            & \textbf{24.517}           & \textbf{28.254} & \textbf{29.154} & 23.574& {23.635}&\textbf{27.220} &\textbf{27.802} \\
\textbf{RL }         & {24.466}            & {24.506}   & {28.247} & 29.147 &\textbf{23.585} & \textbf{23.646} &{27.210}&{27.757} \\
\midrule
\textbf{NA Oracle}          & 24.510 & {24.521}  &28.257         & 29.152& 23.591 & 23.639  &27.222 &27.757\\[2mm]
\bottomrule
\toprule
\multirow{3}{*}{\textbf{Policy}} & \multicolumn{4}{c|}{\textbf{Short Horizon}} & \multicolumn{4}{c}{\textbf{Long Horizon}} \\
\cmidrule(l){2-5}\cmidrule(l){6-9}
 & \multicolumn{2}{c}{\texttt{cvz}} & \multicolumn{2}{c|}{\texttt{c+rvz}} & \multicolumn{2}{c}{\texttt{cvz}} & \multicolumn{2}{c}{\texttt{c+rvz}} \\
 \cmidrule(l){2-3}\cmidrule(l){4-5}\cmidrule(l){6-7}\cmidrule(l){8-9}
 & \multicolumn{1}{c}{\textbf{UNet}} & \multicolumn{1}{c}{\textbf{CResNet}} & \multicolumn{1}{c}{\textbf{UNet}} & \multicolumn{1}{c|}{\textbf{CResNet}} & \multicolumn{1}{c}{\textbf{UNet}} & \multicolumn{1}{c}{\textbf{CResNet}} & \multicolumn{1}{c}{\textbf{UNet}} & \multicolumn{1}{c}{\textbf{CResNet}} \\
 \midrule
\textbf{Random } & {23.978} & 24.184 & 25.845 & 26.949 & 20.967 & 21.437 & 20.428 & 20.625 \\
\textbf{LtH } & {24.108} & 24.396 & {25.641} & {27.048} & 23.441 & 23.703 & 24.487 & 25.510 \\
\textbf{LBCS } & \textbf{24.269} & {24.490} & {26.515} & \textbf{27.870} & \textbf{23.600} & 23.846 & \textbf{25.307} & \textbf{26.316} \\
\textbf{RL } & 24.226 & \textbf{24.498} & \textbf{26.530} & {27.861} & {23.581} & \textbf{23.858} & {25.240} & {26.246} \\
\midrule
\textbf{NA Oracle } & 24.271 & 24.513 & 26.528 & 27.856 & 23.603 & 23.848 & 25.309 & 26.250\\
\bottomrule
\end{tabular}}
        \captionof{table}{AUC on the test set, using PSNR, for the full ablation study using the model of \citet{bakker2020experimental}. The short and long horizon results are \textit{not} comparable with each other, as AUCs are integrated on the whole range of sampling rates. The rest ablation were \textit{not} averaged on several seeds for computational reasons. Recall that \texttt{cvb} stands for cropped, vertical lines, Bakker-like mask distribution, \texttt{c+rhz} stands for cropped then resized, horizontal lines and Zhang-like masks, \texttt{cvz} stands for cropped, vertical lines and Zhang-like masks and \texttt{c+rvz} stands for cropped then resized, vertical lines and Zhang-like masks. }\label{tab:bakker_auc_psnr_full}
    \end{center}
\end{figure}

\begin{figure}
    \begin{center}
        \resizebox{.95\textwidth}{!}{\begin{tabular}{lcccc|cccc}
\toprule
\multirow{3}{*}{\textbf{Policy}} & \multicolumn{4}{c|}{\textbf{Short Horizon}} & \multicolumn{4}{c}{\textbf{Long Horizon}} \\
\cmidrule(l){2-5}\cmidrule(l){6-9}
 & \multicolumn{2}{c}{\texttt{cvb}} & \multicolumn{2}{c|}{\texttt{c+rhz}} & \multicolumn{2}{c}{\texttt{cvb}} & \multicolumn{2}{c}{\texttt{c+rhz}} \\
 \cmidrule(l){2-3}\cmidrule(l){4-5}\cmidrule(l){6-7}\cmidrule(l){8-9}
 & \textbf{UNet} & \textbf{CResNet} & \textbf{UNet} & \textbf{CResNet} & \textbf{UNet} & \textbf{CResNet} & \textbf{UNet} & \textbf{CResNet} \\
 \midrule
\textbf{Random} & {24.583} & 24.604 & 28.241 & 29.113 & 21.520 & 21.684 & 23.547 & 23.843 \\
\textbf{LtH } & {25.056} & 24.984 & {28.176} & {29.567} & 25.056 & 24.984 & 28.176 & 29.567 \\
\textbf{LBCS } & \textbf{25.068} & \textbf{25.117} & \textbf{29.088} & {30.214} & 25.043 & {25.077} & \textbf{29.226} & \textbf{30.234} \\
\textbf{RL } & {25.032} & {25.108} & {29.050} & \textbf{30.218} & \textbf{25.071} & \textbf{25.101} & {29.194} & {30.188} \\
\midrule
\textbf{NA Oracle } & 25.095 & 25.120 & 29.090 & 30.233 & 25.111 & 25.111 & 29.211 & 30.229\\[2mm]
\bottomrule
\toprule
\multirow{3}{*}{\textbf{Policy}} & \multicolumn{4}{c|}{\textbf{Short Horizon}} & \multicolumn{4}{c}{\textbf{Long Horizon}} \\
\cmidrule(l){2-5}\cmidrule(l){6-9}
 & \multicolumn{2}{c}{\texttt{cvz}} & \multicolumn{2}{c|}{\texttt{c+rvz}} & \multicolumn{2}{c}{\texttt{cvz}} & \multicolumn{2}{c}{\texttt{c+rvz}} \\
 \cmidrule(l){2-3}\cmidrule(l){4-5}\cmidrule(l){6-7}\cmidrule(l){8-9}
 & \textbf{UNet} & \textbf{CResNet} & \textbf{UNet} & \textbf{CResNet} & \textbf{UNet} & \textbf{CResNet} & \textbf{UNet} & \textbf{CResNet} \\
 \midrule
\textbf{Random } & {24.369} & 24.623 & 26.657 & 27.957 & 21.791 & 22.497 & 21.854 & 22.506 \\
\textbf{LtH } & {24.551} & 24.961 & {26.223} & {28.143} & 24.551 & 24.961 & 26.223 & 28.143 \\
\textbf{LBCS } & \textbf{24.817} & {25.089} & {27.439} & \textbf{29.069} & 24.850 & \textbf{25.084} & {27.518} & \textbf{29.145} \\
\textbf{RL } & {24.765} & \textbf{25.102} & \textbf{27.451} &  \textbf{29.069} &\textbf{25.102} & {24.807} & \textbf{27.583} & {29.041} \\
\midrule
\textbf{NA Oracle } & 24.813 & {25.123} & 27.429 & 29.043 & {24.837} & {25.101} & 27.446 & 29.072\\
\bottomrule
\end{tabular}

}
        \captionof{table}{PSNR at 25\% sampling rate, using PSNR, with the model of \citet{bakker2020experimental}. This is the counterpart of the results shown in Table \ref{tab:bakker_auc_psnr_full}, where the acronyms used are explicited. Here, the results across short and long horizon are comparable. }\label{tab:bakker_at_25_psnr_full}
    \end{center}
\end{figure}

\clearpage
\FloatBarrier

\section{LBCS vs long-range adaptivity}\label{app:adaptivity}


Using the setting of \citet{pineda2020active}, we studied qualitatively how the different policies yield different mask designs.  As can be seen in \cref{fig:ssddqn_lbcs}a., d. and e, LBCS puts more emphasis on the center frequencies, but acquires similar sections of k-space as the SS-DDQN.
It also creates a more symmetric mask, which is in line with \cite{pineda2020active} observations that SSIM creates more asymmetric masks.
More interestingly, as can be seen in \cref{fig:ssddqn_lbcs}c., most of the variation is concentrated at the early sampling rates (left of the plot) with the std and especially the coefficient of variation ($\frac{\sigma}{\mu}$) decaying towards zero in most locations.
This implies that while SS-DDQN is indeed adapting to each image individually, this mainly affects ordering early on and after 20 -40  samples ($6-12\%$ sampling rate) LBCS starts to catch up.
This is also supported by two observations:

\begin{enumerate}
    \item LBCS underperforms SSDQN by a larger margin than on the full evaluation, which we interpret as the mask being good \emph{on average} and with a smaller sample there is a higher chance of individual suboptimality.
    \item the per-image difference between LBCS and SSDQN grows in favor of SS-DDQN until about 40 samples where LBCS starts to slowly recover. The distribution of the difference however becomes much wider, implying there are images where LBCS performs wildly different from SS-DDQN.
\end{enumerate}

\begin{figure}[!ht]
    \centering
    \begin{subfigure}[b]{0.3\textwidth}
        \centering
        \includegraphics[width=\linewidth]{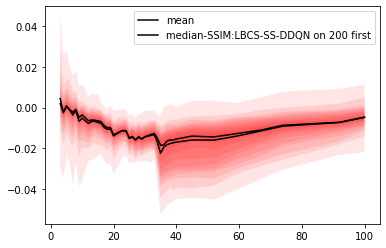}
        \caption{LBCS-SS-DDQN}
        \label{fig:lbcs_vs_ssddqn}
    \end{subfigure}
    \hfill
    \begin{subfigure}[b]{0.3\textwidth}
        \centering
        \includegraphics[width=\textwidth]{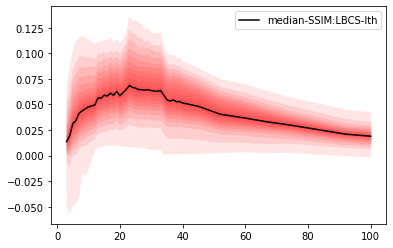}
        \caption{LBCS-LtH}
        \label{fig:lbcs_vs_lth}
    \end{subfigure}
    \hfill
    \begin{subfigure}[b]{0.3\textwidth}
        \centering
        \includegraphics[width=\textwidth]{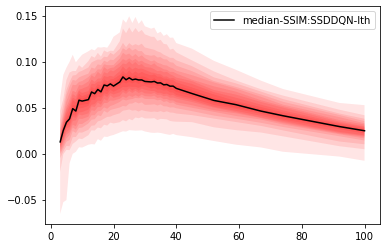}
        \caption{SS-DDQN-LtH}
        \label{fig:ssddqn_vs_lth}
    \end{subfigure}
    \caption{Per Image distribution of SSIM differences between sampling methods across the sampling process. Each shade corresponds to a $10$ percent region, with the lightest shade indicating max and min regions.}
\end{figure}

\begin{figure}
    \centering
    \includegraphics[width=\linewidth]{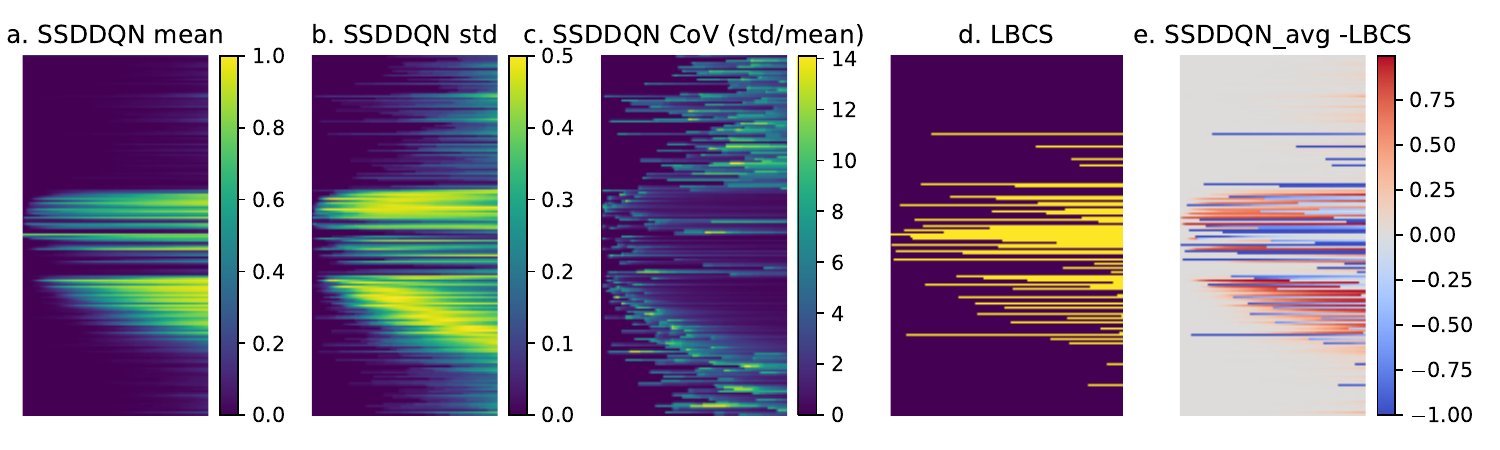}
    \caption{SS-DDQN variability and comparison with the LBCS mask. CoV stands for Coefficient of Variation here.}
    \label{fig:ssddqn_lbcs}
    \vspace{-.4cm}
\end{figure}

\begin{figure}[!ht]
    \centering
    \begin{subfigure}[b]{0.49\textwidth}
        \centering
        \includegraphics[width=\linewidth]{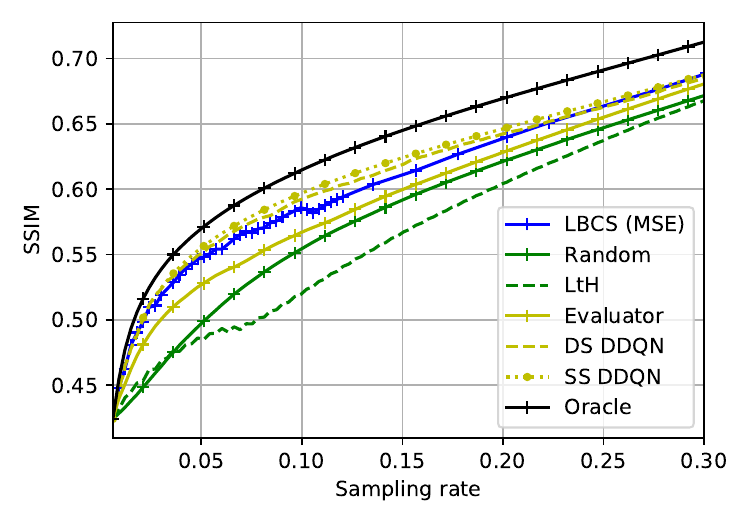}
        \caption{SSIM vs sampling rate}
        \label{fig:ssim_vs_sampling}
    \end{subfigure}
    \hfill
    \begin{subfigure}[b]{0.49\textwidth}
        \centering
        \includegraphics[width=\textwidth]{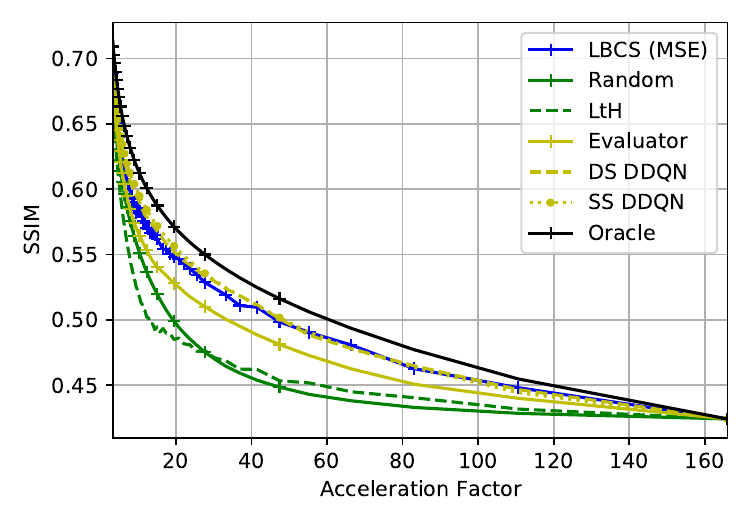}
        \caption{SSIM vs acceleration rate}
        \label{fig:ssim_acceleration_rate}
    \end{subfigure}
    \vspace{-.2cm}
    \caption{Two ways to report the same result, SSIM version of \cref{fig:pin_three}}
    \vspace{-.4cm}
    \label{fig:pin_three_ssim}
\end{figure}

\clearpage
\FloatBarrier

\section{LBCS complexity and parameters}
\label{app:lbcs}

\Cref{table:LBCS_args} summarizes the hyperparameters for sLBCS used and \Cref{table:comparison} gives an overview of the computational complexity.

\begin{figure}[!ht]
    \centering
    \begin{subfigure}[b]{.49\textwidth}
        \centering
        \includegraphics[width=\linewidth]{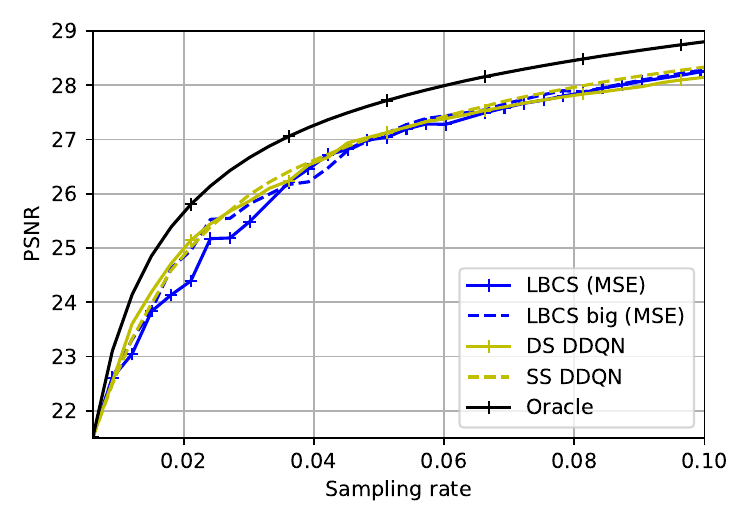}
        \caption{PSNR vs sampling rate}
        \label{fig:psnr_big_lbcs}
    \end{subfigure}
    \begin{subfigure}[b]{.49\textwidth}
        \centering
        \includegraphics[width=\textwidth]{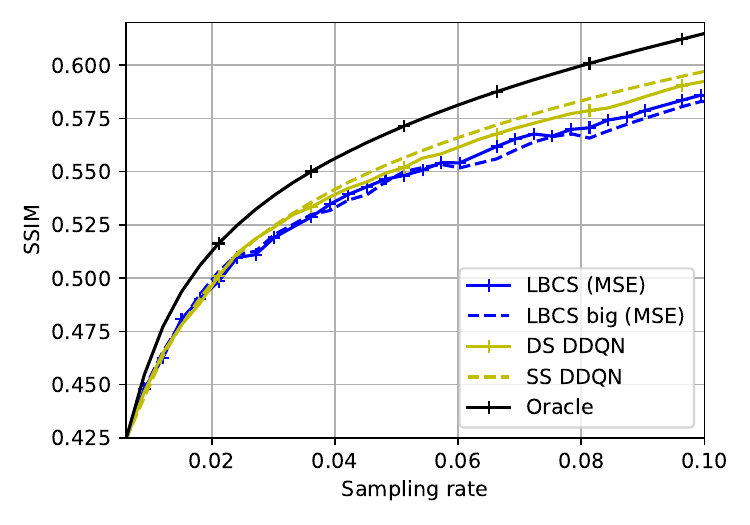}
        \caption{SSIM vs sampling rate}
        \label{fig:ssim_big_lbcs}
    \end{subfigure}
    \vspace{-.2cm}
    \caption{PSNR and SSIM of LBCS with larger batch size (LBCS - big) trained on MSE, zoomed to the region where the smaller batch size LBCS underperformed. Note that it fully matches or outperforms both versions of DDQN on PSNR.}
    \vspace{-.4cm}
    \label{fig:lbcs_big}
\end{figure}

As we previously discussed, sLBCS is almost completely parallelizable, which leads to the stark runtime differences noted in \Cref{ap:implementation}. As can be seen when comparing \Cref{fig:pin_three} and \Cref{fig:lbcs_big}, once a certain data batch size $l$ and sampling candidate set cardinality $k$ is reached, LBCS performance saturates, although it continues to benefit especially in the low sampling regime.
This means that one could very feasibly reduce $l$ for the comparison with \citet{bakker2020experimental}, we attempted to use a larger batchsize to explore what would be the result of scaling sLBCS up.

\begin{table}[!ht]
    \centering
    \resizebox{\linewidth}{!}{\begin{tabular}{l c c c}
            \toprule
            \centering \textbf{Method}               & \textbf{Forward}                           & \textbf{Backward} & \textbf{Total}       \\ [0.5ex]
            \midrule
            \citep{bakker2020experimental}           & $q(8)E(50)B(16)H(n_r+n_p)=6'400H(n_r+n_p)$ & $6400Hn_p$        & $6'400H( n_r+2 n_p)$ \\
            LBCS for  \citet{bakker2020experimental} & $n(H)k(128)l(256)n_r=32'768Hn_r$           & 0                 & $32'768Hn_r$         \\
            \midrule
            \citep{pineda2020active}                 & $5e6(n_r+n_p)$                             & $5e6(n_p)$        & $5e6(n_r+2n_p)$      \\
            LBCS for \citet{pineda2020active}        & $n(100)k(64)l(20)n_r=128e3n_r$             & 0                 & $128e3n_r$           \\
            \bottomrule
        \end{tabular}}
    \caption{Approximate computational cost of the compared methods. Note that at test time, LBCS is basically free while the RL policies will still need to be deployed.
        $n_r,n_p$ are the parameter counts of the reconstruction and sampling policies respectively}
    \label{table:comparison}
\end{table}

\begin{table}[!ht]
    \centering
    \resizebox{\linewidth}{!}{\begin{tabular}{l c c c c}
            \toprule
            \centering \multirow{2}{*}{\textbf{Setting}}          & Num. lines               & Max. cardinality & Candidate    & Data batch \\
                                                                  & $\vert \mathcal{S}\vert$ & $N$              & set size $k$ & size $l$   \\
            \midrule
            Short horizon \newline \citep{bakker2020experimental} & 128                      & 16               & 128          & 256        \\
            Long horizon \citep{bakker2020experimental}           & 128                      & 28               & 128          & 256        \\
            \citep{pineda2020active}                              & 332                      & 100              & 64           & 20         \\
            \citep{pineda2020active} "big"                        & 332                      & 48               & 200          & 256        \\
            \bottomrule
        \end{tabular}}
    \caption{The hyperparameters used for the stochastic LBCS masks throughout comparisons}
    \label{table:LBCS_args}
\end{table}

\clearpage
\FloatBarrier
\section{Visual comparisons}

\begin{figure}[ht]
    \centering
    \includegraphics[width=\linewidth]{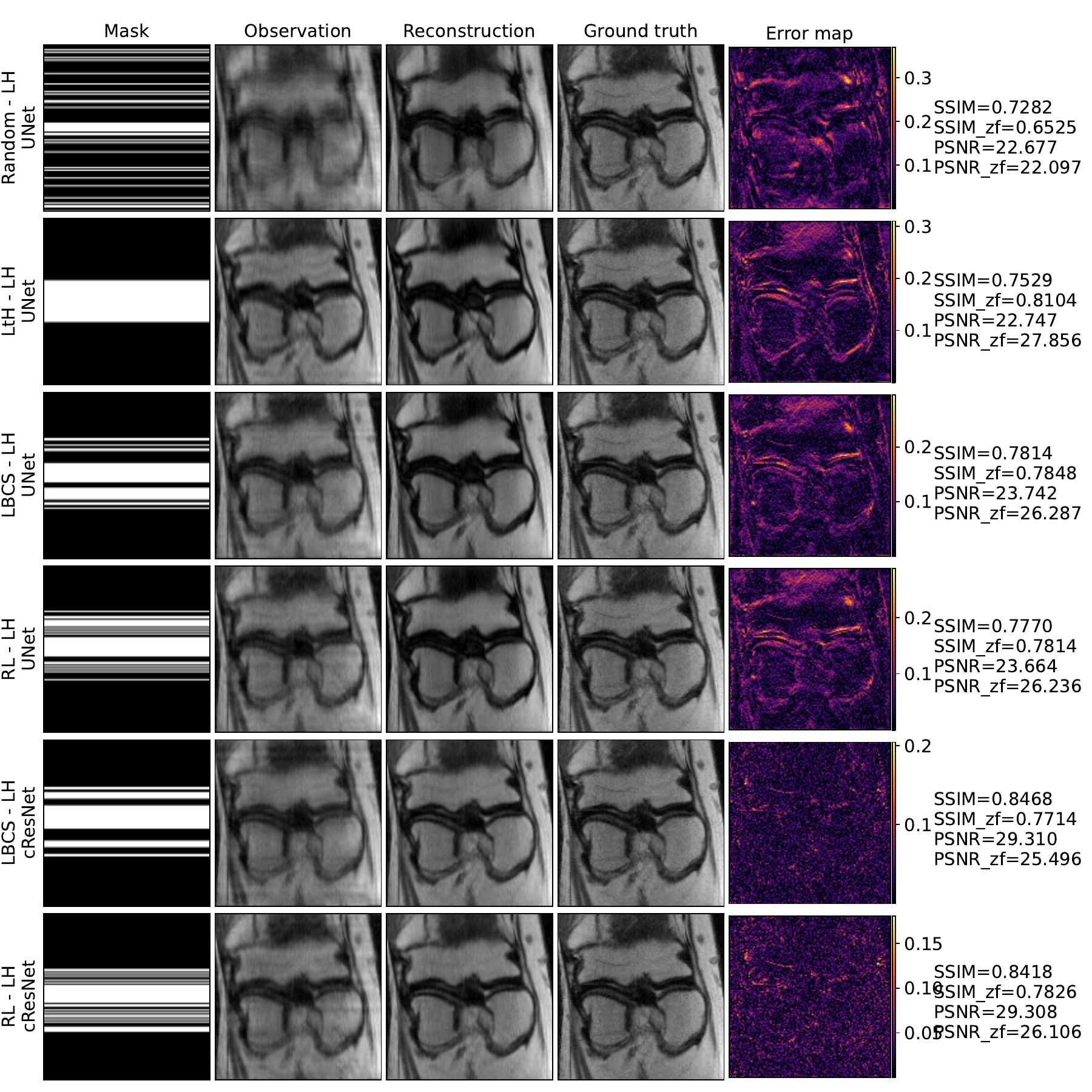}
    \caption{Visualization of masks, observations, reconstructions and ground truths and error maps ($|\text{reconstruction}-\text{ground truth}|$) at $25\%$ sampling rate, for different policies (Random, LtH, LBCS, RL). The data are processed according to the \texttt{c+rhz} setting, i.e. cropped then resized images, horizontal undersampling and Zhang-type distribution of masks. The SSIM and PSNR values are given on the right, and here, \textit{zf} refers to zero-filled, and is the SSIM/PSNR taken between the observation and the ground truth.}
    \label{fig:visualize_25}
    \vspace{-.4cm}
\end{figure}

\begin{figure}[ht]
    \centering
    \includegraphics[width=\linewidth]{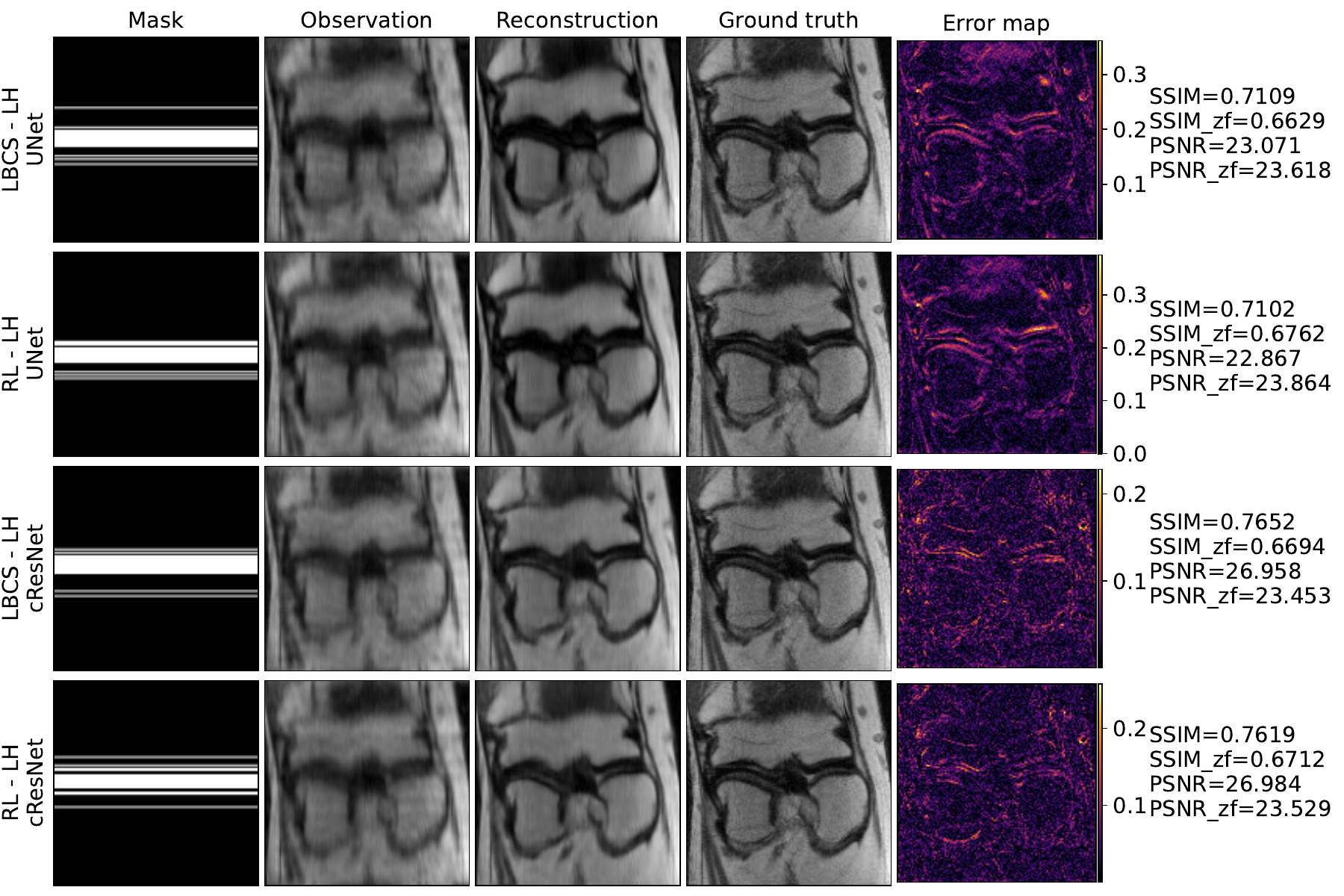}
    \caption{Additional visualization for the image displayed in Figure \ref{fig:visualize_25}. The results feature $12.5\%$ sampling rate.}
    \label{fig:visualize_12}
    \vspace{-.4cm}
\end{figure}

\begin{figure}[!ht]
    \begin{subfigure}[b]{0.6\textwidth}
        \centering
        \includegraphics[width=\linewidth]{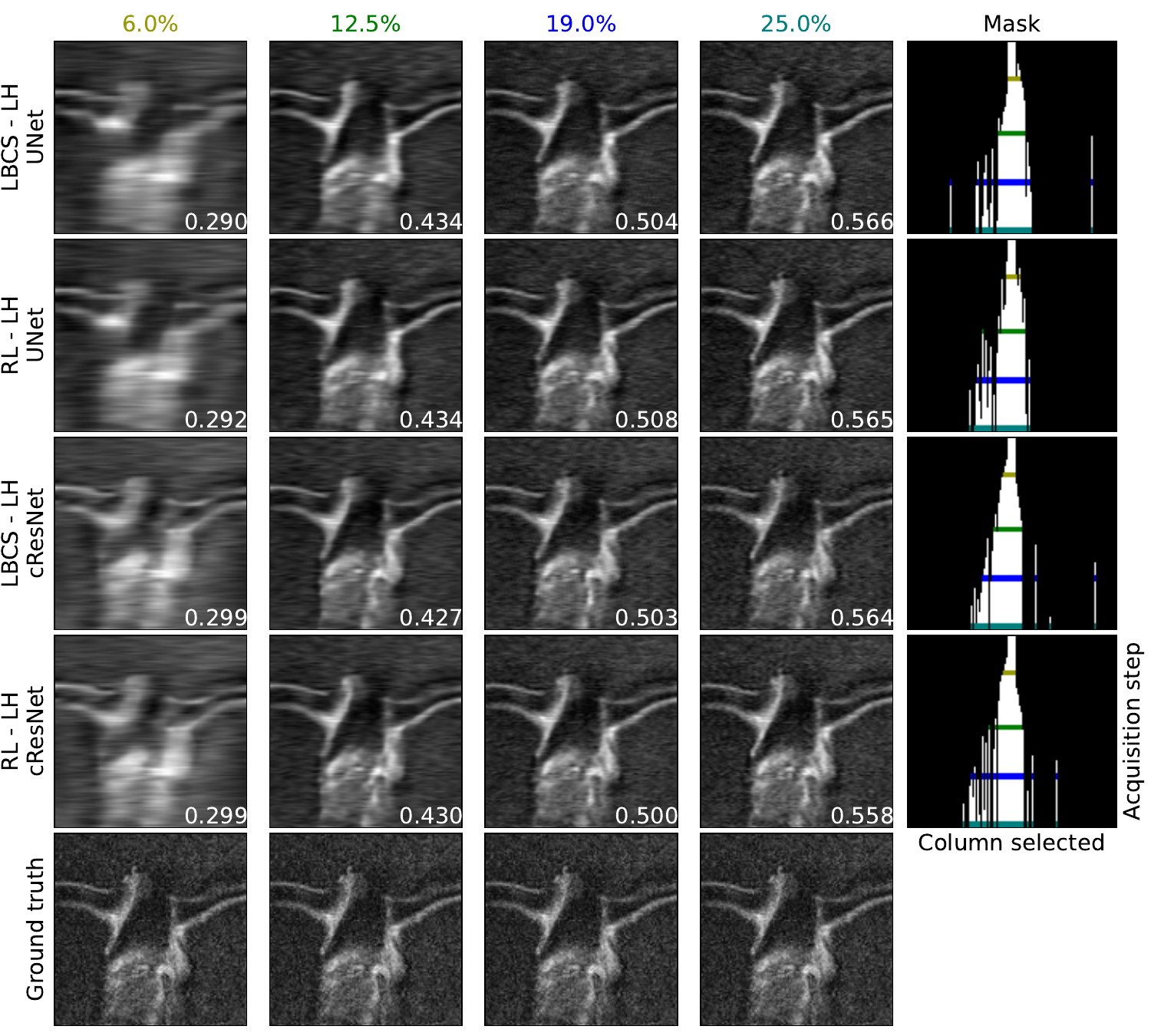}
        \caption{}\label{fig:recon_plot_a}
    \end{subfigure}
    \hfill
    \begin{subfigure}[b]{0.6\textwidth}
        \centering
        \includegraphics[width=\linewidth]{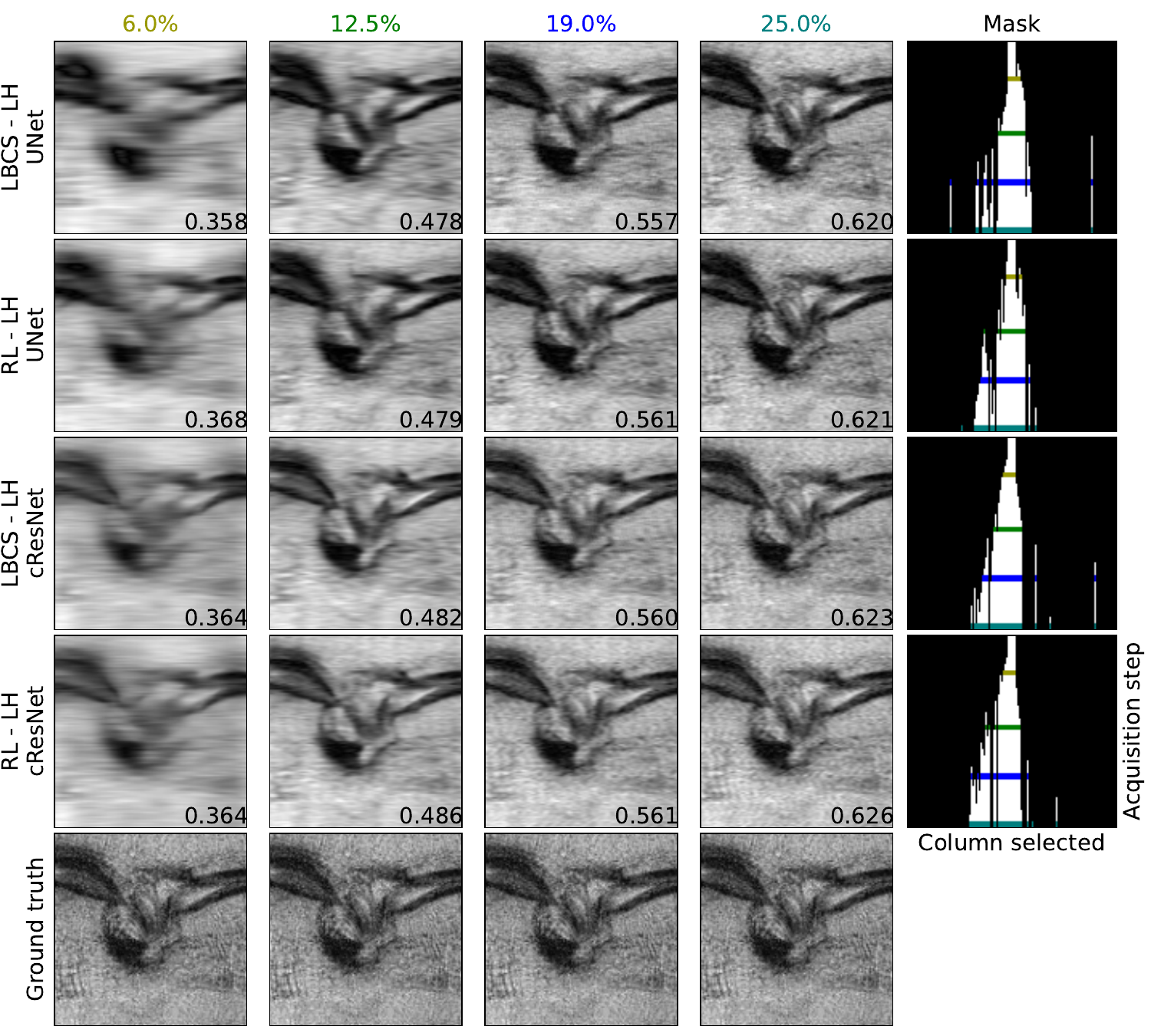}
        \caption{}\label{fig:recon_plot_b}
    \end{subfigure}
    \centering
    \caption{Visualization of reconstructed images at different sampling rates ($6\%$, $12.5\%$, $19\%$ and $25\%$) for two sampling policies (LBCS and RL) and two reconstruction algorithms (UNet and cResNet). The data are processed according to the \texttt{cvb} setting, i.e. cropped images, vertical undersampling and Bakker-type distribution of masks. The last row shows the ground truth (repeated), and each reconstruction has the corresponding SSIM displayed at the bottom right of the image. The rightmost column display the columns acquired during sampling (in white) as a function of the acquisition steps: starting on top with only center frequencies and progressively adding more and more lines to the sampling mask. The top plot (a) displays a proton density, fat saturated (PDFS) image, and the bottom plot displays a proton density (PD) image \citep{zbontarFastMRIOpenDataset2019}.}
    \label{fig:recon_plot}
    \vspace{-.4cm}
\end{figure}

In this section, we present a visual comparison over a selected set of models, policies, settings and images. We present a visual evaluation of the models and policies at sampling rates $25\%$ and $12.5\%$ on Figures \ref{fig:visualize_25} and \ref{fig:visualize_12}. In addition, we present a more exhaustive set of reconstruction, at various sampling rates, on Figure \ref{fig:recon_plot}, where we display both types of sequences that were used to generate the data, namely proton density (PD) and proton density, fat saturated (PDFS) of images \citep{zbontarFastMRIOpenDataset2019}.

Focusing first on Figures \ref{fig:visualize_25} and \ref{fig:visualize_12}, we can observe that random and low-to-high sampling lie significantly behind the performance of LBCS and RL. This is the reason why they are not included in the rest of the figures. We can see on the observation that random sampling tends to miss important structures, and results in a severely aliased observation. On the contrary LtH, that focuses on low frequencies, obtains a good quality observation, but fails to yield an improvement after reconstruction, and generally loses out on higher frequency details, yielding a poorer performance especially on edges. In the rest of the comparison, it is hard to notice any significant visual difference between the images obtained by LBCS and the RL method of \citet{bakker2020experimental}. It is however clear that in the \texttt{c+rhz} setting, the cResNet yields a significantly better and sharper reconstruction at $25\%$ sampling, compared to the UNet. This is also confirmed by the results in Table \ref{tab:bakker_at_25_ssim_full}. On this particular image, this trend is also observed at $12.5\%$ sampling rate on Figure \ref{fig:visualize_12}, but this is \textit{not} a consistent trend, as this is not highlighted by the AUC computation of Table \ref{tab:bakker_auc_ssim_full}.

Turning now to Figure \ref{fig:recon_plot}, it is interesting first to discuss the policies obtained by LBCS and RL respectively. The LBCS policy is fixed for the reconstruction algorithm, so the first and third rows of Figures \ref{fig:recon_plot_a} and \ref{fig:recon_plot_b} will each feature the same policy. The adaptive RL policy, on the second and fourth rows seems to have central backbone of common frequency, but varies more around higher frequencies, as observed in the Figure 5 of the appendix of \citet{bakker2020experimental}. However, in both cases, these differences have very little quantitative impact, and the same is true visually: there is no clear visual difference between the reconstructed images at either stage.

While this is not an exhaustive visual investigation, and does not directly assess the suitability of the different policies for various downstream tasks, this lack of visual difference could suggest that RL might not bring a significant improvement over simpler techniques such as LBCS in such cases. However, the question remains open in the case where the sampling policy would be tailored directly for the downstream task, rather than optimized for reconstruction quality, but this falls beyond the scope of the current work.

\clearpage
\FloatBarrier
\section{Detail on the practical recommendations}\label{app:conclusions}
In this appendix, we discuss the recommendations issued in the conclusion, and provide the supporting evidence from our results.

\textbf{Focus on improvements in the reconstructor architecture, mask distribution and algorithms used for training the reconstructor.}
Our results and ablations on the setting of \citet{bakker2020experimental} consistently show that the improvements obtained by changing the reconstructor architecture or the mask distribution are orders of magnitude more impactful than moving from LBCS to RL. This is supported the results of both Section \ref{s:ablations_results} and Appendix \ref{app:bakker_ablation_full}, where we see that moving from a UNet to a cResNet typically brings a significantly larger improvement (typically $0.01$ SSIM) than what RL brings over LBCS in the best case (at most $0.0015$ SSIM).This trend is also seen in Appendix \ref{app:bakker_ablation_full} when moving from \texttt{cvb} in the short horizon setting to \texttt{cvz} in the long horizon setting. We refer the reader to this section for more details.

\textbf{Compare against strong baselines, such as LBCS.} This point is established throughout our paper, where all results illustrate that, at best, RL methods bring moderate improvement over LBCS. This improvement often comes at the cost of prohibitive computational expense, even on higher end DGX-2 servers\footnote{The authors of \citep{pineda2020active} confirmed to us that it took more than 20 days to train their model.}, while training LBCS required at most a couple of hours (cf. Appendix \ref{ap:implementation}).

\textbf{Show sampling curves and use AUC to aggregate your results instead of performance at the final sampling rate.}
There is no consensus on how results should be aggregated from sampling curves. \citet{bakker2020experimental,van2021active} reported performance at the end of the acquisition, and \citet{pineda2020active} reported AUC curves computed on the acceleration factor. \citet{gozcu2018learning,yin2021end} reported performance at selected sampling rates, and for other works such as \citet{jin2019self}, it is not clear how results were aggregated.

We believe that reporting the AUC on sampling rates, computed on the whole range of acquisition steps allows to most meaningfully quantify the performance of a policy on its whole trajectory. It does not require to select a sampling rate at which the result should be evaluated, and using sampling rates as opposed to acceleration factor allows to equally weight the contribution of each acquired line.

Table \ref{tab:pineda_auc} and Figure \ref{fig:pin_three} compellingly illustrate that reporting the policy at a given sampling rate is not representative of its performance throughout the acquisition procedure.

\textbf{Be mindful about preprocessing settings when evaluating a policy model. We recommend using the cropped+vertical setting with the data normalization implemented by \citet{zbontarFastMRIOpenDataset2019}.} We discussed in \Cref{sec:pipeline} that \citet{bakker2020experimental} used a data normalization that is, among other things, incompatible with data consistency, a commonly used building block for cascading networks \citep{schlemper2018deep,zhang2019reducing}. This can be prevented by using a normalization based on observation rather than ground truth statistics, as implemented in \citet{zbontarFastMRIOpenDataset2019}.

Regarding the experimental setting, vertical masks have ubiquitously used on the fastMRI dataset \citep{zbontarFastMRIOpenDataset2019,huijben2020learning,bakker2020experimental,pineda2020active} and cropping has been the most common preprocessing to alleviate the computational burden of the large images ($640\times368$) used in the dataset \citep{bakker2020experimental,Huijben2020Deep,yin2021end}. Evaluating models on cropped data with vertical masks will then facilitate reproducibility among different works. We would additionally recommend to researchers to evaluate their models on the cropped+resized in addition to the cropped only setting, as the images display a significantly different field of view (cf. Figures \ref{fig:visualize_25} and \ref{fig:recon_plot}).

\cleardoublepage
\chapter{Appendix for Chapter \ref{ch:gans}}

\section{Implementation details}\label{app:gan_implementation}

\subsection{Knee experiments}
The GAN experiment was carried out using a residual UNet (ResUNet) \citet{belghazi2019learning}, using the architecture described in  Table \ref{tab:resunet} below. The discriminator architecture is detailed in Table \ref{tab:ncresnet}. They consisted of respectively $1.2$ million and $19.5$ million parameters.

We also used the cascade of residual networks (cResNet) \citet{zhang2019reducing} as our reconstruction baselines, along with their evaluator. One block of the architecture is given in Table \ref{tab:cresnet}. The network was composed of $3$ such blocks. The evaluator architecture is detailed in \ref{tab:evaluator}. They amount to $29.5$ million and $97.7$ million parameters respectively. These numbers were obtained from the usage of \citet{zhang2019reducing} by \citet{pineda2020active}.

\begin{table}[!ht]
    \centering
    \resizebox{.35\textwidth}{!}{\begin{tabular}{lc}
            \toprule
            Type                       & Output size               \\
            \midrule
            Input                      & $6 \times 128 \times 128$ \\
            $\text{ResBlock}_1(3,2,1)$ & $64 \times 64 \times 64$  \\
            $\text{ResBlock}_2(3,2,1)$ & $128 \times 32 \times 32$ \\
            $\text{ResBlock}_3(3,2,1)$ & $256 \times 16 \times 16$ \\
            $\text{ResBlock}_3(3,2,1)$ & $512 \times 8 \times 8$   \\
            $\text{ResBlock}_3(3,2,1)$ & $1024 \times 4 \times 4$  \\
            Global average pooling     & $1024 \times 1 \times 1$  \\
            $\text{Conv}(1,1,0)$       & $1 \times 1 \times 1$     \\
            \bottomrule
        \end{tabular}}

    \caption{Discriminator ResNet. This was inspired by the discriminator of \cite{belghazi2019learning}. The model used ReLU activations, and consisted of $19.5$ million parameters.}
    \label{tab:ncresnet}
\end{table}

\begin{table}[!ht]
    \centering
    \resizebox{.75\textwidth}{!}{\begin{tabular}{llcc}
            \toprule
                                     & Type                                              & Output size                & Comments                                          \\
            \midrule
            Input                    & --                                                & $3 \times 128 \times 128$  &                                                   \\
            \multirow{5}{*}{Encoder} & $\text{ResBlock}_1(3,1,1)$                        & $64 \times 128 \times 128$ &                                                   \\
                                     & Avg Pool +$\text{ResBlock}_2(3,1,1)$              & $64 \times 64 \times 64$   &                                                   \\
                                     & Avg Pool +$\text{ResBlock}_3(3,1,1)$              & $64 \times 32 \times 32$   &                                                   \\
                                     & Avg Pool +$\text{ResBlock}_4(3,1,1)$              & $64 \times 16 \times 16$   &                                                   \\
                                     & Avg Pool +$\text{ResBlock}_5(3,1,1)$              & $64 \times 8 \times 8$     &                                                   \\[2mm]
            Bottleneck               & Avg Pool +$\text{ResBlock}_6(3,1,1)$              & $64 \times 4 \times 4$     &                                                   \\[2mm]
            \multirow{5}{*}{Decoder} 
                                     & $\text{Upsample}_1+\text{ResBlock}_7(3,1,1)$      & $64 \times 8 \times 8$     & $\text{Cat}[\text{Upsample}_1,\text{ResBlock}_5]$ \\
                                     & $\text{Upsample}_2+\text{ResBlock}_8(3,1,1)$      & $64 \times 16 \times 16$   & $\text{Cat}[\text{Upsample}_2,\text{ResBlock}_4]$ \\
                                     & $\text{Upsample}_3+\text{ResBlock}_9(3,1,1)$      & $64 \times 32 \times 32$   & $\text{Cat}[\text{TrConv}_3,\text{ResBlock}_3]$   \\
                                     & $\text{Upsample}_4+\text{ResBlock}_{10}(3,1,1)$   & $64 \times 64 \times 64$   & $\text{Cat}[\text{TrConv}_4,\text{ResBlock}_2]$   \\
                                     & $\text{Upsample}_5+\text{ResBlock}_{11}(3,1,1)$   & $64 \times 128 \times 128$ & $\text{Cat}[\text{TrConv}_5,\text{ResBlock}_1]$   \\[2mm]
                                     & $\text{Conv}(1,1,0) + \text{Sigmoid} + \text{DC}$ & $2 \times 128 \times 128$  &                                                   \\
            \bottomrule
        \end{tabular}}

    \caption{ResUNet architecture used in the experiments, inspired by the architecture in \cite{belghazi2019learning}. The activation used throughout is a LeakyReLu with slope $0.2$. We had $n_{\text{channels,in}}$ containing $2$ channels for the observation, plus $1$ channel for the noise.  $\text{ResBlock}(3,1,1)$ denotes a residual block with a $3\times 3$ kernel size, stride $1$ and padding $1$. This results in a model with $1.2$ million parameters. Bilinear upsampling is used. }
    \label{tab:resunet}
\end{table}

\begin{table}[!ht]
    \centering
    \resizebox{.75\textwidth}{!}{\begin{tabular}{llcc}
            \toprule
                                        & Type                                                & Output size                & Comments                          \\
            \midrule
            Input                       & --                                                  & $2 \times 128 \times 128$  &                                   \\
            \multirow{3}{*}{Encoder}    & $\text{Conv}_1(3,2,1)$                              & $32 \times 64 \times 64$   &                                   \\
                                        & $\text{Conv}_2(3,2,1)$                              & $64 \times 32 \times 32$   &                                   \\
                                        & $\text{Conv}_3(3,2,1)$                              & $128 \times 16 \times 16$  &                                   \\[2mm]
            \multirow{3}{*}{Bottleneck} & $\text{ResBlock}_1(3,1,1)$                          & $128 \times 16 \times 16$  & Skip-add from module $i-1$        \\
                                        & $\text{ResBlock}_i(3,1,1)$                          & $128 \times 16 \times 16$  & $i$ blocks (total of 6 ResBlocks) \\
                                        & $\text{ResBlock}_6(3,1,1)$                          & $128 \times 16 \times 16$  & Skip to module $i+1$              \\[2mm]
            \multirow{3}{*}{Decoder}    & $\text{TrConv}_1(4,2,1)$                            & $64 \times 32 \times 32$   &                                   \\
                                        & $\text{TrConv}_2(4,2,1)$                            & $32 \times 64 \times 64$   &                                   \\
                                        & $\text{TrConv}_3(4,2,1)$                            & $16 \times 128 \times 128$ &                                   \\[2mm]
                                        & $\text{Conv}(1,1,0) + \text{(Sigmoid)} + \text{DC}$ & $3 \times 128 \times 128$  & Sigmoid only at the last block    \\
            \bottomrule
        \end{tabular}}

    \caption{One block of cResNet, used in \cite{zhang2019reducing}. The model used ReLu activations along with instance normalization.  In the paper, we used 3 blocks with $6$ residual blocks in the Bottleneck layer. We had $n_{\text{channels,in}}$ containing $2$ channel for the observation.  $n_{\text{channels,out}}=3$: $2$ channels for the reconstruction, and $1$ channel for the variance. The entire model consisted of three such blocks, for a total of $29.5$ million parameters.}
    \label{tab:cresnet}
\end{table}

\begin{table}[!ht]
    \centering
    \resizebox{.75\textwidth}{!}{\begin{tabular}{llcc}
            \toprule
                                 & Type                   & Output size                 & Comments                 \\
            \midrule
            Input spectral maps  & $-$                    & $128 \times 128 \times 128$ &                          \\
            Input mask embedding & $\text{Conv}(1,1,1)$   & $6 \times 1 \times 1$       & Replicate \& concatenate \\
            Input tensor         & $-$                    & $134 \times 128 \times 128$ &                          \\
                                 & $\text{Conv}(4,2,1)_1$ & $256 \times 64 \times 64$   &                          \\
            Evaluator            & $\text{Conv}(4,2,1)_2$ & $512 \times 32 \times 32$   &                          \\
                                 & $\text{Conv}(4,2,1)_3$ & $1024 \times 16 \times 16$  &                          \\
                                 & $\text{Conv}(4,2,1)_4$ & $2048 \times 8 \times 8$    &                          \\
                                 & GAP                    & $1024 \times 1 \times 1$    &                          \\
            Output vector        & Conv $(1,1,0)$         & $128 \times 1 \times 1$     &                          \\
            \bottomrule
        \end{tabular}}
    \caption{Evaluator architecture \citep{zhang2019reducing}. First, spectral maps are created and concatenated with an embedded mask. Then, the data is processed through convolutional layers, using $\text{LeakyReLU}(0.2)$ and instance normalization. This results in a total of $97.7$ million parameters.}\label{tab:evaluator}
\end{table}

\clearpage
\FloatBarrier
\section{Extended results}
We first provide additional knee results in Section \ref{ss:app_knee}, then evaluate our GAN model on a small MRI dataset of brain images in Section \ref{ss:app_brain} and finish with additional MNIST evaluations in Section \ref{app:s_mnist}.

\subsection{Additional MRI knee results}\label{ss:app_knee}
\begin{figure}[!ht]
    \centering
    \includegraphics[width=.48\textwidth]{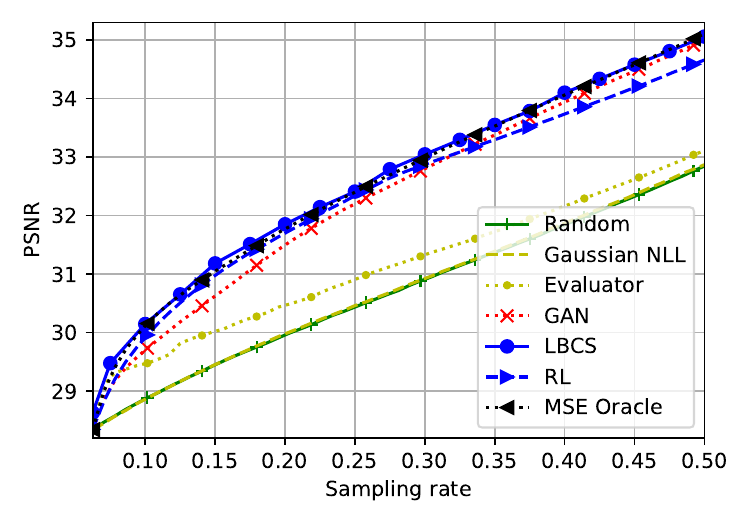}
    \includegraphics[width=.48\textwidth]{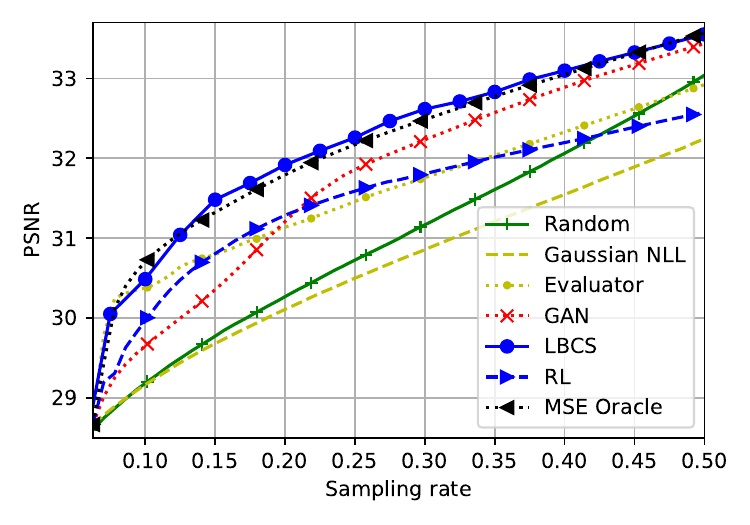}
    \caption{PSNR plots of results shown in Table \ref{tab:comp_knee}, showing the different performance of the sampling methods. The left plot is the result of the evaluation on our GAN model, and the right plot is evaluated using the reconstructor of \citet{zhang2019reducing}. Although Table \ref{tab:comp_knee} computes the AUC until $25\%$, this table extends the evaluation until $50\%$ sampling rate.}\label{fig:knee_curve_psnr}
\end{figure}

In Figure \ref{fig:knee_curve_psnr}, we find the PSNR counterpart of the SSIM plot of Figure \ref{fig:knee_curve_ssim}. It is interesting to see that there are noticeably different behaviors between the two models, and the results are overall consistent with the trends seen on the SSIM plot of Figure \ref{fig:knee_curve_ssim}. Note also that the scales are different in each plot.

\subsection{Brain experiment}\label{ss:app_brain}

\subsubsection{Experimental setting}
The dataset used in the first three experiments (subsections) below consists of a proprietary dataset of 2D T1-weighted brain scans. In our experiments, we use 100 slices of sizes $256\times 256$ from five such subjects, $20$ per subject. Three subjects ($60$ slices) were used for training the network, two subjects ($30$ slices) for testing. The data were then massively augmented with both rigid transformations and elastic deformations to counter overfitting as our dataset is very small, following the recommendations of \citet{ronneberger2015u,schlemper2017deep}. Namely, we apply both rigid transformations and elastic deformations. Specifically, at training time, each image was dynamically augmented with a randomly applied translation of $\pm 6$ pixels along $x$- and $y$-axes, rotations of $[0, 2\pi)$, reflection along the $x$-axis with $50\%$ probability. We also apply elastic deformations using the implementation in \citep{simard2003best}with $\alpha \in [0, 40]$ and $\sigma \in [5, 8]$.

Data from individual coils was processed via a complex linear combination, where coil sensitivities were estimated from an $8\times 8$ central calibration region of Fourier space \citep{bydder2002combination}. The acquisition used a field of view (FOV) of $220\times 220$ mm and a resolution of $0.9\times 0.7 $mm. The slice thickness was $4.0$mm. The imaging protocol comprised a flip angle of $70\deg$, a TR/TE of $250.0/2.46$ ms, with a scan time of 2 minutes and 10 seconds.

The training conditions for the Brain experiment were the same as the Knee experiment.

\subsubsection{Results}
We observe on Tables \ref{tab:comp_brain_psnr} and \ref{tab:comp_brain_ssim} results that are consistent with the ones on the knee dataset: LBCS outperforms both the Evaluator and the GAN, but it is interesting to observe that, in this setting, the GAN performance is quite close to the one of LBCS, especially in terms of SSIM. Note that in this context, the policy of \citet{bakker2020experimental} tends to perform worse than previously.

\begin{figure}[h]
\begin{minipage}[c]{0.48\textwidth}
\begin{center}
    \begin{tabular}{lcc}
     \toprule
    \multirow{1}{*}{\textbf{Policy}}& \multicolumn{2}{c}{\textbf{Model}}\\
    \cmidrule{2-3} & GAN & Recon\\
    \midrule
    \textbf{Gaussian NLL}  & - & $25.37$\\
    \textbf{Evaluator} & $32.19$& $29.95$\\
    \textbf{GAN} & $33.77$& $31.18$\\
    \midrule
    \textbf{LBCS} & $\mathbf{34.03}$& $\mathbf{32.03}$\\
    \textbf{RL} & $31.45$& $28.23$ \\
    \bottomrule
    \end{tabular}
    \captionof{table}{Average PSNR test set AUC (one AUC per image) for the brain experiment.}\label{tab:comp_brain_psnr} 
\end{center}
\end{minipage}
\hfill
\begin{minipage}[c]{0.48\textwidth}
\begin{center}
     \begin{tabular}{lcc}
     \toprule
    \multirow{1}{*}{\textbf{Policy}}& \multicolumn{2}{c}{\textbf{Model}}\\
    \cmidrule{2-3} &GAN & Recon\\
    \midrule
    \textbf{Gaussian NLL} & - & $0.72$\\
    \textbf{Evaluator} & $0.83$& $0.85$\\
    \textbf{GAN} & $\mathbf{0.88}$ & $\mathbf{0.87}$\\
    \midrule
    \textbf{LBCS} &  $\mathbf{0.88}$& $0.84$\\
    \textbf{RL} &$0.80$ & $0.80$\\
    \bottomrule
    \end{tabular}
    \captionof{table}{Average SSIM set AUC (one AUC per image) for the brain experiment.}\label{tab:comp_brain_ssim} 
\end{center}
\end{minipage}
\end{figure}

\clearpage
\FloatBarrier

\subsection{Additional MNIST results}\label{app:s_mnist}
\begin{figure}[!ht]
    \centering
    \includegraphics[valign=t,width=.8\linewidth]{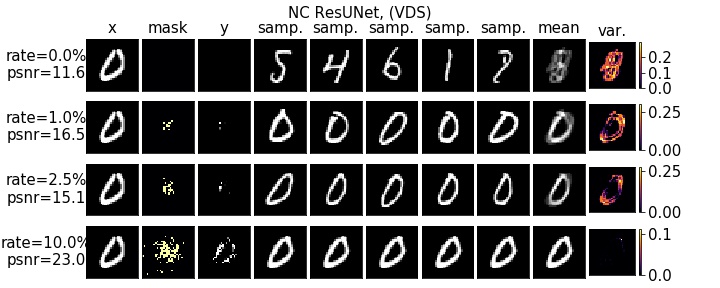}

    \includegraphics[valign=t,width=.8\linewidth]{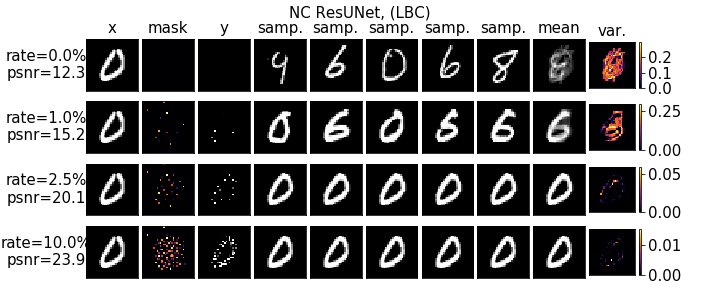}

    \includegraphics[width=.8\linewidth]{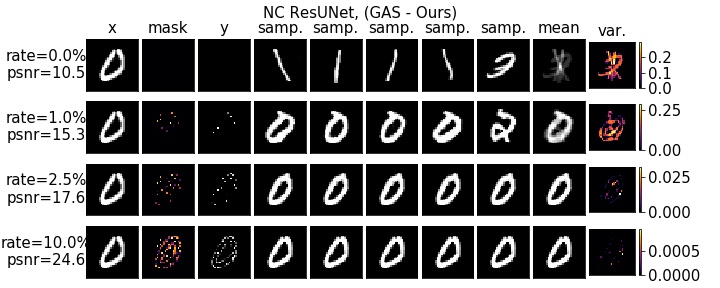}
    \caption{Additional Image domain results for the MNIST experiment, featuring different sampling rates, using a NC ResUNet model for conditional sampling.}\label{fig:mnist_image2}
\end{figure}

\begin{figure}[!ht]
    \centering
    \includegraphics[valign=t,width=.5\linewidth]{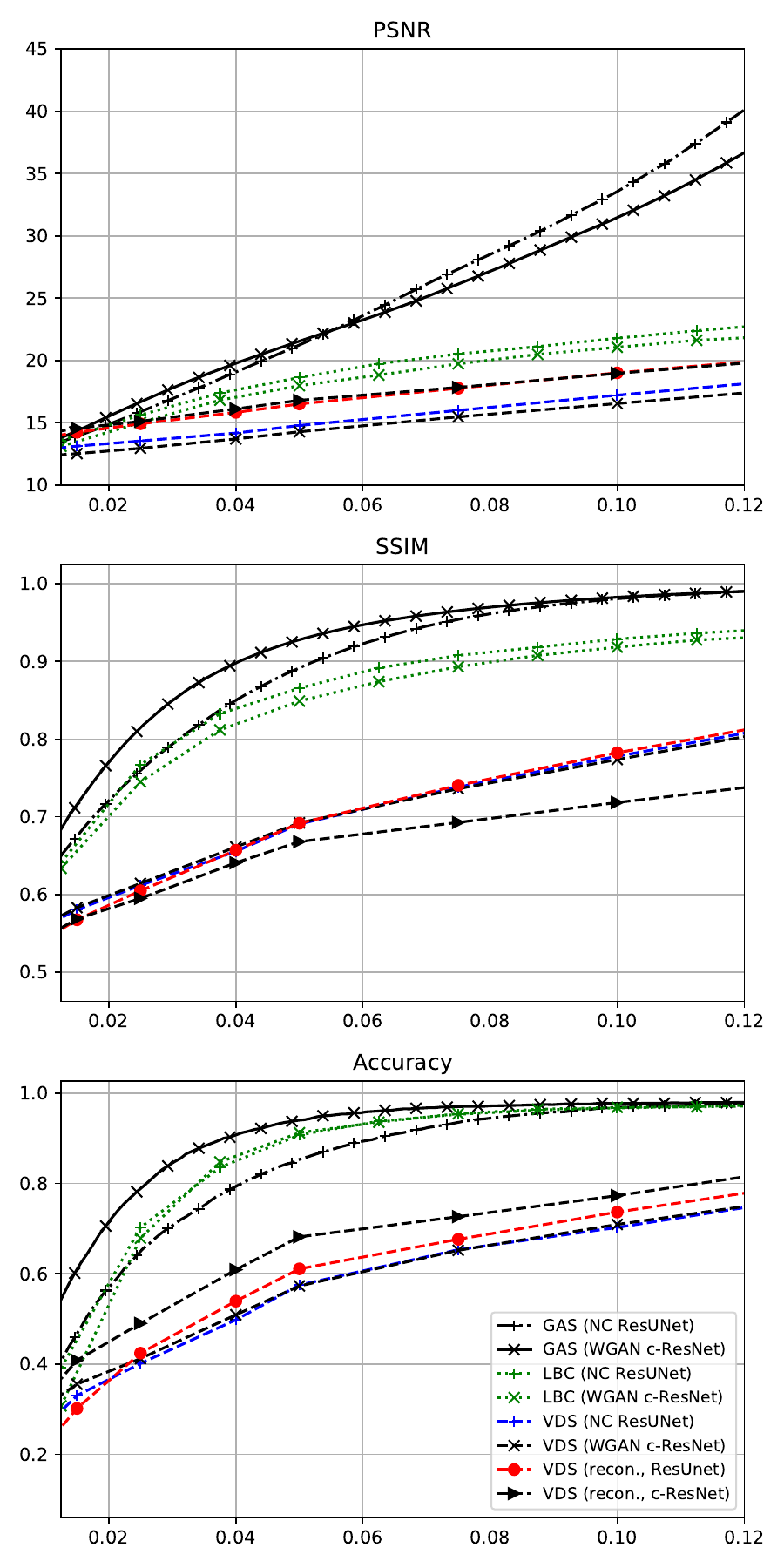}
    \caption{Image domain plots for the MNIST experiment, showing PSNR, SSIM and Accuracy for some selected models. These results are presented compactly in Table \ref{tab:comp_mnist_im}.}\label{fig:plt_mnist_image}
\end{figure}

In Figure \ref{fig:mnist_image2}, we provide additional reconstruction at different sampling rates for the different policies considered. Here  GAS refers to generative adaptive sampling and corresponds to our GAN policy. We also provide the PSNR, SSIM and Accuracy against sampling rate on Figure \ref{fig:plt_mnist_image}, which are the counterpart to Table \ref{tab:comp_mnist_im}. We see in particular that the strong performance of the GAN policy is due to the PSNR and SSIM reaching very high values around $5\%$ sampling, where the sampling pattern has really acquired almost all pixels relevant to the digit. As there is almost no background, the error decreases very quickly. We see nonetheless that in every case, the adaptive policy manages to bring a noticeable improvement over LBCS, which was not the case in Fourier image.

\clearpage
\FloatBarrier

\subsection{CIFAR10 experiment}\label{app:s_cifar}
Finally, we provide results on CIFAR10, for which the pixelwise subsampling task is much more challenging. We chose to use grayscale versions of the images.

\textbf{Experiment setting.} For CIFAR10, we modified the training procedures slightly. For the WGAN we use ExtraAdam~\citep{gidel2018variational}, TTUR~\citep{heusel2017gans} with learning rates of $lr_G=10^{-4}$ and $lr_D=3\cdot 10^{-4}$, one sided gradient penalty~\citep{petzka2018on} and betas of $(0,0.9)$. Number of epochs was set to $200$ with the learning rates being halved every $50$ steps. Batchsize was set to $128$ and we used 8 GPUs in total, yielding an effective batch size of $1024$.
For the NC we retain batch size $128$ and the Adam settings, simply moving to 8 GPUs and increasing training time to 200 epochs.

\begin{figure}[!ht]
    \centering
    \includegraphics[valign=t,width=0.8\linewidth]{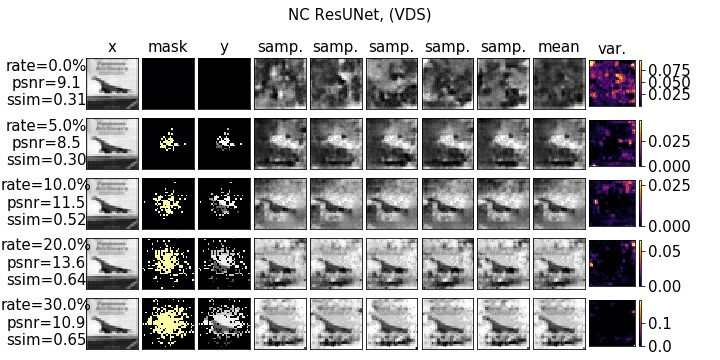}
    \includegraphics[valign=t,width=0.8\linewidth]{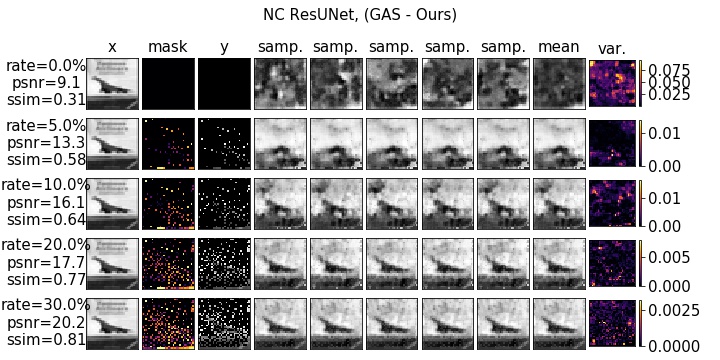}
    \caption{CIFAR10 results, featuring different sampling rates, using a NC ResUNet model for conditional sampling. Observe that while the model saw only masks going from $0.5\%$ to $20\%$ during training, it generalizes well up to $30\%$, gaining $2.5$dB in the case of GAS.  GAS stands for Generative Adaptive Sampling (GAN policy), and VDS refers to variable-density sampling policies.}\label{fig:image_cifar10_1}
\end{figure}

\begin{figure}[!ht]
    \centering
    \includegraphics[valign=t,width=0.8\linewidth]{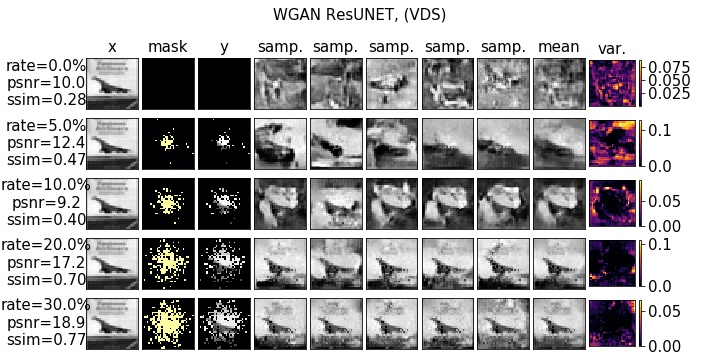}
    \includegraphics[valign=t,width=0.8\linewidth]{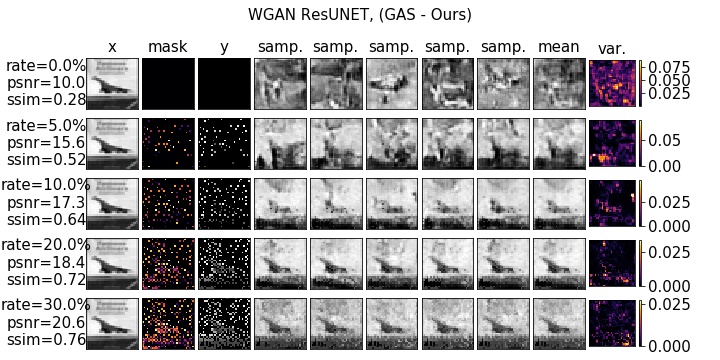}
    \caption{CIFAR10 results, featuring different sampling rates, using a WGAN ResUNet model for conditional sampling. Observe that while the model saw only masks going from $0.5\%$ to $20\%$ during training, it generalizes well up to $30\%$.  GAS stands for Generative Adaptive Sampling (GAN policy), and VDS refers to variable-density sampling policies.}\label{fig:image_cifar10_2}
\end{figure}

\begin{figure}[!ht]
    \centering
    \includegraphics[valign=t,width=0.5\linewidth]{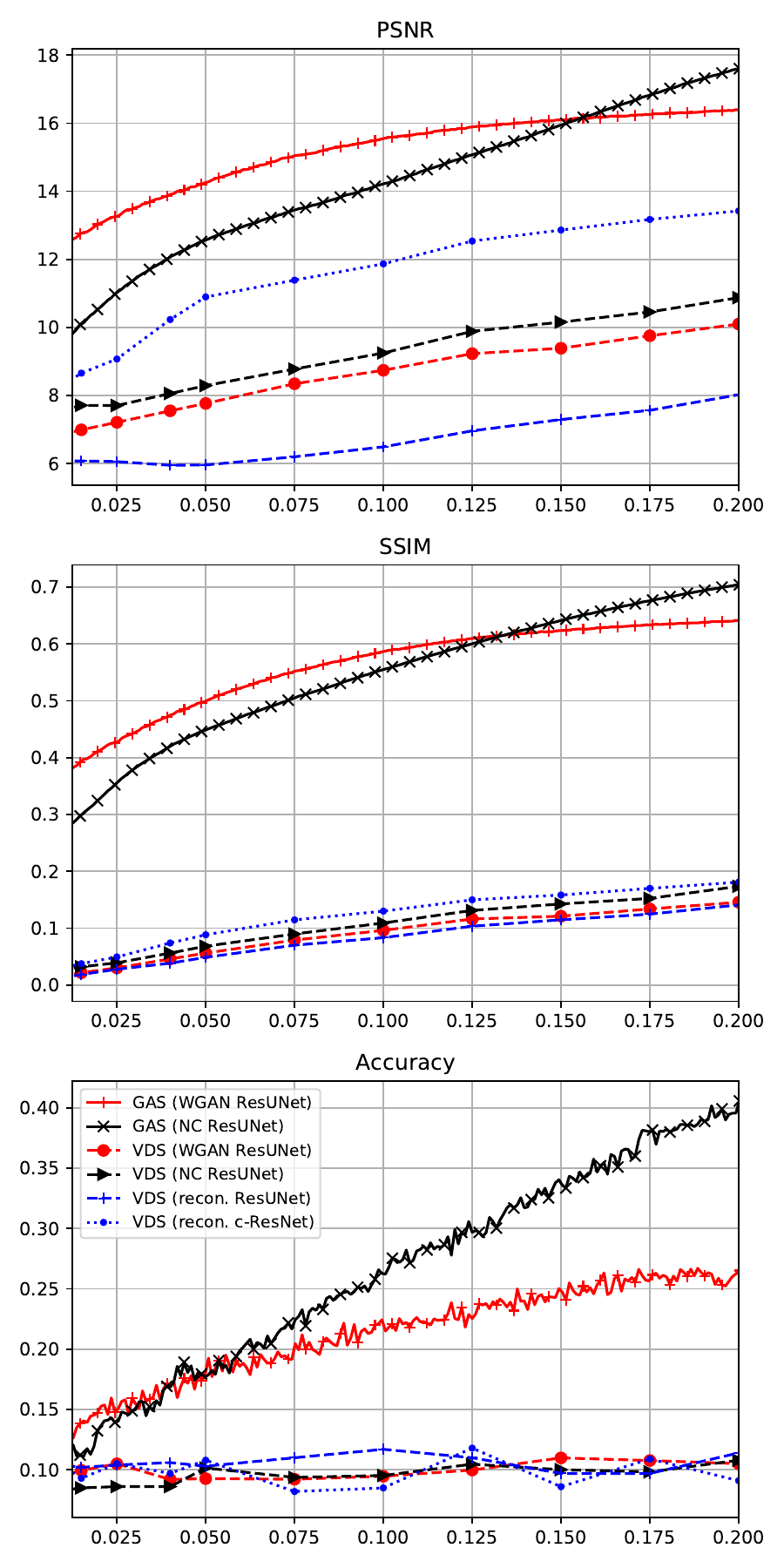}
    \caption{Image domain plots, showing PSNR, SSIM and Accuracy for some selected models. GAS stands for Generative Adaptive Sampling (GAN policy), and VDS refers to variable-density sampling policies.}\label{fig:plt_cifar10}
\end{figure}

\textbf{Results.} \Cref{fig:image_cifar10_1} and \cref{fig:image_cifar10_2} showcase GAS results using a WGAN and NC respectively. We evaluate VDS (variable-density sampling) and GAS (generative adaptive sampling, GAN policy) on a subset of the test set of CIFAR10 composed of $1000$ images and report the results on Table \ref{tab:comp_cifar}. The results are detailed in the plots of Figure \ref{fig:plt_cifar10}.

\begin{table}[!ht]
    \centering
    \resizebox{0.8\textwidth}{!}{\begin{tabular}{lll|ccc}
            \toprule
            Algorithm            & Model  & Architecture & PSNR [dB]$~(\uparrow)$   & SSIM $(\uparrow)$        & Accuracy  $ [\%]~(\uparrow)$ \\
            \midrule
            \multirow{2}{*}{VDS}
                                 & NC     & ResUNet      & $9.32\pm 0.70$           & $0.11\pm0.03$            & $0.10\pm 0.17$               \\
                                 & WGAN   & ResUNet      & $8.72\pm 0.62$           & $0.09\pm 0.03$           & $0.10\pm 0.17$               \\
            \midrule
            \multirow{2}{*}{VDS} & Recon. & ResUNet      & $6.73\pm 0.51$           & $0.09\pm 0.02$           & $0.11\pm0.17$                \\
                                 & Recon. & c-ResNet     & $ 11.79 \pm 0.72$        & $0.13 \pm 0.03$          & $0.10\pm0.16$                \\
            \midrule
            \multirow{2}{*}{\begin{minipage}{1cm}GAS (Ours)\end{minipage}}
                                 & NC     & ResUNet      & $14.36 \pm 1.89$         & $0.55\pm 0.08$           & $\mathbf{0.27 \pm 0.22}$     \\
                                 & WGAN   & ResUNet      & $\mathbf{15.26\pm 2.34}$ & $\mathbf{0.57 \pm 0.09}$ & $0.21 \pm 0.18$              \\
            \bottomrule
        \end{tabular}}
    \caption{Average test set AUC (one AUC per image) with standard deviation on CIFAR10, in image domain. This was computed on a subset of the test set.}\label{tab:comp_cifar}
\end{table}


\backmatter
\cleardoublepage
\phantomsection
\addcontentsline{toc}{chapter}{Bibliography}
\bibliographystyle{apalike}
\bibliography{bibliography}

\end{document}